\documentclass{article}
\usepackage{arxiv}

\usepackage[utf8]{inputenc} 
\usepackage[T1]{fontenc}    

\usepackage{array, textcomp, url, comment, tabularx} 
\usepackage{graphicx, pdflscape}

\usepackage[table,xcdraw,dvipsnames]{xcolor}

\usepackage{amsmath, amssymb, amsthm, amsfonts, mathtools, stmaryrd, pifont} 
\usepackage{algorithm, algorithmicx}
\usepackage[noEnd=true, 
            commentColor=black!70, 
            italicComments=true]{algpseudocodex}
\usepackage{enumitem, booktabs, multirow, multicol}
\usepackage[compress]{cite}
\usepackage{xstring}
\usepackage{nicefrac}       

\newcommand{\hollowstar}{\text{\ding{73}}}
\newcommand{\filledplus}[1][black]{%
  \mathrel{\vcenter{\hbox{%
    \tikz[x=1.5ex,y=1.5ex]{
      \draw[line width=0.4pt, draw=#1, fill=white]
        (0,0.4) -- (0,0.1) -- (-0.3,0.1) -- (-0.3,-0.1) -- (0,-0.1)
        -- (0,-0.4) -- (0.3,-0.4) -- (0.3,-0.1) -- (0.6,-0.1)
        -- (0.6,0.1) -- (0.3,0.1) -- (0.3,0.4) -- cycle;
    }
  }}}%
}

\usepackage{etoolbox}
\AtBeginEnvironment{table}{\footnotesize}
\AtBeginEnvironment{tabular}{\footnotesize}
\AtBeginEnvironment{figure}{\footnotesize}

\usepackage[caption=false,listofformat=empty]{subfig}

\usepackage{tikz}
\usetikzlibrary{%
    shapes.geometric, decorations.pathmorphing,
    arrows, arrows.meta,
    backgrounds, positioning,shadows,
    calc, matrix, graphs}

\newcommand{\withOrcid}[2]{%
    \begingroup%
    \hypersetup{urlcolor=black}%
    \ifstrempty{#2}{#1}{\href{https://orcid.org/#2}{%
    {#1}\includegraphics[scale=0.06]{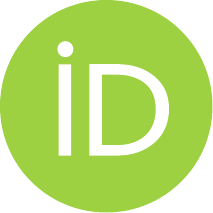}}}%
    \endgroup%
}

\usepackage{hyperref}
\hypersetup{
 colorlinks=true,
 linkcolor=blue,
 citecolor=blue,
 filecolor=magenta, 
 urlcolor=blue,
 pdfauthor = {Cruz, J.M. and Talbi, E-G.},
}

\algnewcommand\algorithmicdata{\textbf{Data:}}
\algnewcommand\algorithmicinports{\textbf{InPorts:}}
\algnewcommand\algorithmicoutports{\textbf{OutPorts:}}
\algnewcommand\algorithmicvars{\textbf{Variables:}}
\makeatletter
\renewcommand{\ALG@beginalgorithmic}{\small}
\algnewcommand\InPorts{%
	\algpx@endCodeCommand%
	\ifnumcomp{0}{<}{\FSSize{algpx@startNewCodeBoxQueue}}{\setbool{algpx@setNorth}{true}}{}%
	\algpx@drawInItem{\algorithmicinports}%
}
\algnewcommand\OutPorts{%
	\algpx@endCodeCommand%
	\ifnumcomp{0}{<}{\FSSize{algpx@startNewCodeBoxQueue}}{\setbool{algpx@setNorth}{true}}{}%
	\algpx@drawInItem{\algorithmicoutports}%
}
\algnewcommand\Vars{%
	\algpx@endCodeCommand%
	\ifnumcomp{0}{<}{\FSSize{algpx@startNewCodeBoxQueue}}{\setbool{algpx@setNorth}{true}}{}%
	\algpx@drawInItem{\algorithmicvars}%
}
\algnewcommand\Data{%
	\algpx@endCodeCommand%
	\ifnumcomp{0}{<}{\FSSize{algpx@startNewCodeBoxQueue}}{\setbool{algpx@setNorth}{true}}{}%
	\algpx@drawInItem{\algorithmicdata}%
}
\renewcommand{\ALG@name}{Algorithm}
\newcommand{\algorithmname}{\ALG@name}
\patchcmd{\fnum@algorithm}{\thealgorithm}{\thealgorithm:}{}{}
\makeatother

\newcommand{\ie}{i.e., }
\newcommand{\eg}{e.g., }
\newcommand{\etal}{\emph{et al.} }

\newcommand{\R}{\mathbb{R}}
\newcommand{\X}{\mathfrak{X}}
\newcommand{\Z}{\mathbb{Z}}
\newcommand{\N}{\mathbb{N}}

\newcommand{\nha}[1]{%
    \IfStrEqCase{#1}{%
        {long}{Neuromorphic Optimiser}%
        {short}{NeurOptimiser}%
    }[neurOptimiser]
}
\newcommand{\nhu}[1]{%
    \IfStrEqCase{#1}{%
        {long}{Neuromorphic Heuristic Unit}%
        {short}{NHU}%
    }[NHU]
}
\MakeRobust{\nha}
\MakeRobust{\nhu}

\definecolor{mutedOrange}{RGB}{244, 162, 97}  
\definecolor{softPurple}{RGB}{162, 162, 208} 
\definecolor{pastelYellow}{RGB}{233, 196, 106} 
\definecolor{lightGrayBlue}{RGB}{209, 209, 233} 
\definecolor{softBlueGreen}{RGB}{128, 206, 215} 

\definecolor{inportColor}{RGB}{85, 170, 255}  
\definecolor{outportColor}{RGB}{255, 85, 85}  

\definecolor{varColor}{RGB}{167, 201, 87} 
\definecolor{var_continuous_col}{RGB}{97, 155, 138} 
\definecolor{var_output_col}{RGB}{77, 136, 255} 
\definecolor{var_index_col}{RGB}{74, 144, 226} 
\definecolor{textdark}{RGB}{69, 69, 69} 
\definecolor{mutedGreen}{rgb}{0.2, 0.6, 0.2}
\definecolor{forestGreen}{rgb}{0.13, 0.55, 0.13}

\definecolor{pastelCoral}{HTML}{F28B82}       
\definecolor{pastelBlue}{HTML}{AECBFA}        
\definecolor{mintGreen}{HTML}{C8E6C9}         
\definecolor{lavenderBlue}{HTML}{D7DFFB}      
\definecolor{champagne}{HTML}{FDF6E3}         
\definecolor{darkGray}{HTML}{555555}          
\definecolor{arrowRed}{HTML}{E57373}          
\definecolor{arrowBlue}{HTML}{64B5F6}         

\tikzstyle{perturbation} = [
    rectangle, drop shadow, minimum width=2.5cm, minimum height=2.5cm, text centered, rounded corners=5pt, draw=black, top color=mintGreen!10, bottom color=mintGreen!60
    ]
\tikzstyle{selection} = [
    rectangle, drop shadow, minimum width=2.5cm, minimum height=2.5cm, text centered, rounded corners=5pt, draw=black, top color=mutedOrange!10, bottom color=mutedOrange!40
    ]
\tikzstyle{spiking} = [
    rectangle, drop shadow, minimum width=2.5cm, minimum height=2cm, text centered, rounded corners=5pt, draw=black, top color=champagne!10, bottom color=champagne!80
    ]
\tikzstyle{receiver} = [
    rectangle, drop shadow, minimum width=2.5cm, minimum height=2.5cm, text centered, rounded corners=5pt, draw=black, top color=lavenderBlue!10, bottom color=lavenderBlue!80
    ]
\tikzstyle{sender} = [
    rectangle, drop shadow, minimum width=2.5cm, minimum height=2.5cm, text centered, rounded corners=5pt, draw=black, top color=lavenderBlue!10, bottom color=lavenderBlue!80
    ]

\tikzstyle{main_process} = [
    rectangle, drop shadow, minimum width=3cm, minimum height=3cm, text centered, rounded corners=5pt, draw=black, top color=mintGreen!10, bottom color=mintGreen!60
    ]
\tikzstyle{layer} = [
    rectangle, drop shadow, minimum width=1.5cm, minimum height=10cm, text centered, rounded corners=5pt, draw=black, top color=champagne!10, bottom color=champagne!80
    ]
\tikzstyle{aux_process_3} = [
    rectangle, drop shadow, minimum width=2cm, minimum height=3cm, text centered, rounded corners=5pt, draw=black, top color=lavenderBlue!10, bottom color=lavenderBlue!80
    ]
\tikzstyle{aux_process_4} = [
    rectangle, drop shadow, minimum width=2cm, minimum height=3cm, text centered, rounded corners=5pt, draw=black, top color=lavenderBlue!10, bottom color=lavenderBlue!80
    ]

\tikzstyle{inPort} = [
    rectangle, minimum width=0.7cm, minimum height=0.6cm, text centered, draw=black, fill=pastelCoral!70, rounded corners=5pt, opacity=0.9
    ]
\tikzstyle{outPort} = [
    rectangle, minimum width=0.7cm, minimum height=0.6cm, text centered, draw=black, fill=pastelBlue!70, rounded corners=5pt, opacity=0.9
    ]
\tikzstyle{var_funct} = [rectangle, minimum width=0.7cm, minimum height=0.6cm, text centered, draw=black, fill=softPurple!40, rounded corners=5pt, opacity=0.9]
\tikzstyle{var_index} = [rectangle, minimum width=0.7cm, minimum height=0.6cm, text centered, draw=black, fill=varColor!40, rounded corners=5pt, opacity=0.9]
\tikzstyle{arrow} = [darkGray,thick,->,>=latex]
\tikzstyle{varconnect} = [<->, dashed, thick, >=latex]

\tikzstyle{outport} = [rectangle, minimum width=0.7cm, minimum height=0.6cm, text centered, draw=black, fill=outportColor!40, rounded corners=5pt, opacity=0.9]
\tikzstyle{inport} = [rectangle, minimum width=0.7cm, minimum height=0.6cm, text centered, draw=black, fill=inportColor!40, rounded corners=5pt, opacity=0.9]
\tikzstyle{var_funct} = [rectangle, minimum width=0.7cm, minimum height=0.6cm, text centered, draw=black, fill=softPurple!40, rounded corners=5pt, opacity=0.9]
\tikzstyle{var_index} = [rectangle, minimum width=0.7cm, minimum height=0.6cm, text centered, draw=black, fill=varColor!40, rounded corners=5pt, opacity=0.9]

\newcommand{\ERT}{\ensuremath{\mathrm{ERT}}}

\providecommand{\algorithmA}{Neuropt-Hyb}
\providecommand{\algorithmB}{Neuropt-Izh}
\providecommand{\algorithmC}{Neuropt-Lin}
\providecommand{\algorithmD}{RANDOMSEARCH}
\newcommand{\ntables}{12}
\providecommand{\algDtables}{\StrLeft{RANDOMSEARCH auger noiseless}{\ntables}}
\providecommand{\algCtables}{\StrLeft{Neuropt-Hyb}{\ntables}}
\providecommand{\algBtables}{\StrLeft{Neuropt-Izh}{\ntables}}
\providecommand{\algAtables}{\StrLeft{Neuropt-Lin}{\ntables}}

\newcommand{\cocoversion}{{\scriptsize\sffamily{}\color{Gray}Data produced with COCO v2.7.3}}

\providecommand{\bbobecdfcaptionallgroups}[1]{
Empirical cumulative distribution of simulated (bootstrapped) runtimes, measured in number of $f$-evaluations, divided by dimension (\# f-evals/dimension) for the $51$ targets $10^{[-8..2]}$ for all function groups and all dimensions. 
}

\definecolor{NavyBlue}{HTML}{000080}
\definecolor{Magenta}{HTML}{FF00FF}
\definecolor{Orange}{HTML}{FFA500}
\definecolor{CornflowerBlue}{HTML}{6495ED}
\definecolor{YellowGreen}{HTML}{9ACD32}
\definecolor{Gray}{HTML}{BEBEBE}
\definecolor{Yellow}{HTML}{FFFF00}
\definecolor{ForestGreen}{HTML}{228B22}
\definecolor{Lavender}{HTML}{FFC0CB}
\definecolor{SkyBlue}{HTML}{87CEEB}
\definecolor{Goldenrod}{HTML}{DDF700}
\definecolor{VioletRed}{HTML}{D02090}
\definecolor{LimeGreen}{HTML}{32CD32}

\newcommand{\stripzero}[1]{%
  \ifnum#1<10 %
    \number#1%
  \else
    #1%
  \fi
}

\newcommand{\getmodelchar}[1]{%
    \IfStrEq{#1}{Fix}{Fixed}{%
    \IfStrEq{#1}{Rand}{Random}{%
    \IfStrEq{#1}{Dir}{Directional}{%
    \IfStrEq{#1}{DErand}{DE/current-to-rand/1}{Unknown}}}}%
}
\newcommand{\getmodelcharShort}[1]{%
    \IfStrEq{#1}{Fix}{Fix.}{%
    \IfStrEq{#1}{Rand}{Rand.}{%
    \IfStrEq{#1}{Dir}{Dir.}{%
    \IfStrEq{#1}{DErand}{DE/...}{Unknown}}}}%
}
\newcommand{\getdim}[1]{%
  \ifnum#1=2 2%
  \else\ifnum#1=3 3%
  \else\ifnum#1=5 5%
  \else\ifnum#1=10 10%
  \else\ifnum#1=20 20%
  \else #1%
  \fi\fi\fi\fi\fi
}
    
\newcommand{\getidvariant}[1]{%
    \IfStrEq{#1}{00}{Fixed as \texttt{spk\_cond}}{%
    \IfStrEq{#1}{01}{$L_2$ as \texttt{spk\_cond}}{Unknown}}%
}
\robustify{\getidvariant}

\def\sectionname{Section}

\begin{document}
\frenchspacing

\title{NeurOptimisation: The Spiking Way to Evolve}

\author{%
    \withOrcid{%
    Jorge M.~Cruz-Duarte}{0000-0003-4494-7864}\\
    University of Lille,\\ CNRS, Inria, Centrale Lille,\\ UMR 9189 CRIStAL, F-59000 Lille, France\\
    \texttt{jorge.cruz-duarte@univ-lille.fr}\\
    \And%
    \withOrcid{%
    El-Ghazali Talbi}{0000-0003-4549-1010}\\
    University of Lille,\\ CNRS, Inria, Centrale Lille,\\ UMR 9189 CRIStAL, F-59000 Lille, France\\
    \texttt{el-ghazali.talbi@univ-lille.fr}
}


\maketitle

\begin{abstract}

The increasing energy footprint of artificial intelligence systems urges alternative computational models that are both efficient and scalable. Neuromorphic Computing (NC) addresses this challenge by empowering event-driven algorithms that operate with minimal power requirements through biologically inspired spiking dynamics.
We present the \nha{short}, a fully spike-based optimisation framework that materialises the neuromorphic-based metaheuristic paradigm through a decentralised NC system. The proposed approach comprises a population of \nhu{long}s (\nhu{short}s), each combining spiking neuron dynamics with spike-triggered perturbation heuristics to evolve candidate solutions asynchronously. 
The \nha{short}'s coordination arises through native spiking mechanisms that support activity propagation, local information sharing, and global state updates without external orchestration. 
We implement this framework on Intel's Lava platform, targeting the Loihi 2 chip, and evaluate it on the noiseless BBOB suite up to 40 dimensions.
We deploy several \nha{short}s using different configurations, mainly considering dynamic systems such as linear and Izhikevich models for spiking neural dynamics, and fixed and Differential Evolution mutation rules for spike-triggered heuristics.
Although these configurations are implemented as a proof of concept, we document and outline further extensions and improvements to the framework implementation.
Results show that the proposed approach exhibits structured population dynamics, consistent convergence, and milliwatt-level power feasibility.
They also position spike-native MHs as a viable path toward real-time, low-energy, and decentralised optimisation.

\smallskip
\textit{This work has been submitted to the IEEE Transactions on Evolutionary Computation for possible publication. Copyright may be transferred without notice, after which this version may no longer be accessible.}

\end{abstract}


\section{Introduction}
\lettrine{N}{euromorphic} Computing (NC) has emerged as a promising paradigm to address the increasing demands of intelligent systems for low energy consumption, scalability, and real-time responsiveness \cite{li2023electromagnetic-009}. This approach, grounded in the biological principles of spiking neural activity, proposes a model of computation based on asynchronous, event-driven processes and collocated memory and processing. Particularly, Spiking Neural Networks (SNNs), as the third generation of neural models, operate with sparse and localised computation, encouraging the implementation of distributed architectures without central synchronisation \cite{Sanaullah2023snns-review, Nunes2022snns-review}. These properties are particularly valuable in embedded and edge scenarios, where energy and latency constraints prevail. Platforms such as Loihi, TrueNorth, SpiNNaker, and BrainScaleS represent this shift from synchronous von Neumann architectures to biologically inspired substrates \cite{ivanov2022neuromorphic-1fa}. Hence, SNNs are computationally plausible and hardware-aligned, cobblestoning a new path for sustainable, flexible, and efficient artificial intelligence.

Parallel to this technological advance, heuristic-based algorithms, also known as Metaheuristics (MHs), have proven strong capabilities in addressing complex optimisation problems without requiring gradient information. Evolutionary Algorithms (EAs), Particle Swarm Optimisation (PSO), and Differential Evolution (DE), among others, have provided robust solutions in domains where traditional techniques fail \cite{zhan2022survey-90b, liu2024large-scale-db6}. These methods exploit population-based strategies and adaptive search mechanisms to explore challenging domains. Still, their deployment remains computationally intensive and energy demanding when executed on conventional processors. 

The inherent parallelism and locality of most MHs suggest a strong and appealing synergy with NC architectures. These two fields are now mature, and it is only a matter of time before they meet. 
Thus, the integration of both paradigms will yield a new generation of MHs, whose operators and coordination emerge from spike-based dynamics. 
Recently, the concept of Neuromorphic-based Metaheuristics (Nheuristics) was introduced as a formal algorithm class that embeds heuristic search mechanisms within spiking neural dynamics \cite{talbi2025nheuristics}. 
Nheuristics set the foundations for defining a unified design space that covers spiking neuron models, solution encoding, learning mechanisms, and communication topologies. 

This work introduces a \nha{long} framework, from now on referred to as \nha{short}. This is a general-purpose optimisation framework founded on the Nheuristics paradigm through fully spike-driven computation. The \nha{short} architecture consists of multiple \nhu{long}s (\nhu{}s), each embedding heuristic operators within the spike-triggered neuron dynamics. These units evolve candidate solutions asynchronously, embedding heuristic operators within biologically inspired or abstract state transitions.
Additionally, we design decentralised coordination mechanisms such as a tensor contraction layer for spike propagation, a neighbour manager for local information exchange, and a high-level selector for global decision-making. 
We demonstrate the flexibility of the approach by implementing both two-dimensional linear dynamic and Izhikevich (nonlinear) neuron models with multiple spike-triggered rules, including DE-based spike-triggered heuristics.
Moreover, we provide an open-source implementation using Intel's Lava NC framework \cite{lava2023}, targeting the Loihi 2 chip, and validate the approach on the BBOB test suite from \cite{hansen2021coco}.
Results demonstrate structured search dynamics, consistent convergence behaviour, and sub-watt power estimates under realistic configurations. 


\section{Related Work}

\noindent%
The integration of NC, particularly SNNs, with MH optimisation has rendered notable but fragmentary progress. The existing systems reported primarily target specific combinatorial problems by embedding spiking dynamics within custom architectures, often yielding efficient but rigid solutions.

Several contributions follow this direction. 
Lin and Fan proposed a transient chaotic neural network based on ferroelectric memristors for the Travelling Salesperson Problem (TSP), achieving improved convergence via polarisation‐driven annealing in a handcrafted spiking design \cite{lin2022ferroelectric-ea6}. 
Kim \etal presented an NC-based Max‑Cut solver using organic memristors, where the architecture's tight coupling to problem structure yields promising embedded performance \cite{kim2023organic-c54}.

Other approaches have addressed Quadratic Unconstrained Binary Optimisation (QUBO) problems using spike-based computation.
Lele \etal designed a swarm of spiking solvers for QUBO on RRAM hardware, achieving energy-efficiency operation via compute-in-memory co‑design \cite{lele2023neuromorphic-e98}. 
Similarly, Pierro \etal developed a spiking simulated annealing algorithm on Loihi 2, reporting real-time execution and energy savings \cite{pierro2024solving-cc2}.
Notwithstanding, both implementations remain structurally bound to the QUBO problem class.

Beyond problem-specific designs, several research efforts have studied and explored the broader integration of SNNs and optimisation from a generalisation perspective. 
Roy \etal surveyed SNNs as key components in the development of neuromorphic intelligence, focusing on biologically plausible learning rules, including spike-timing-dependent plasticity, surrogate gradients, and hardware landscape \cite{roy2019towards-276}. Withal, their analysis remains within classification and inference tasks, bypassing search and optimisation. 
Besides, Shen \etal reviewed evolutionary spiking neural networks, which utilise evolutionary principles to train deep spiking architectures, often achieving performance comparable to that of conventional methods \cite{shen2024evolutionary-24f}. 
Nevertheless, their work centred around neural architectural search without implementing spike-driven heuristics as optimisation operators.
Schuman \etal introduced the Evolutionary Optimisation for Neuromorphic Systems (EONS) framework, which evolves SNN topology and hyperparameters for deployment on NC simulators \cite{Schuman2020}. A later variant extended the method to edge computing tasks using hybrid simulation on \(\mu\)Caspian hardware \cite{schuman2021real}. While these approaches support hardware-aware adaptation, the evolutionary logic remains CPU-based and offline.

Complementary strategies have explored the application of swarm intelligence in conjunction with spiking-based computation. 
Fang \etal proposed a model integrating SNNs with swarm-based mechanisms for continuous and combinatorial problems, although limited to software simulations sans specific NC hardware alignment \cite{Fang2019sosfe}. 
Similarly, Sasaki and Nakano presented a spiking variant of PSO using deterministic oscillators, delivering theoretical guarantees and competitive performance \cite{Sasaki2023osnn}. However, their model requires manual parameter tuning and lacks hardware implementation.
Javanshir \etal provided an extensive hardware and algorithmic survey, where evolutionary strategies are acknowledged, including encoding schemes and implementations, but without proposing a spike-based MH framework \cite{Javanshir2022}. 
More recently, Snyder \etal developed a Bayesian optimiser, LavaBO, and related asynchronous coordination extensions within Intel's Lava NC framework \cite{Snyder2023neuromorphic,Snyder2024asynchronous,Snyder2024parallelized}.
Although the spike activity in these implementations remains a secondary interface, CPU-based Bayesian methods drive the optimisation process.

From these studies, two critical observations emerge. First, there is a decentralised effort to develop energy-efficient MHs by exploiting the architectural affordances of NC substrates. Second, while SNNs are well suited to parallel and low-power execution, existing applications in optimisation remain rigid and problem-specific. The structural compatibility between NC and MH design has yet to yield a general, reusable framework in which both coordination and search arise natively from spike-driven dynamics.

\section{Foundations of Neuromorphic Optimisation} \label{Sec:Foundations}

\noindent%
This section presents the theoretical foundations of the \nha{short} framework, commencing with core concepts in optimisation and heuristic search, and leading to the formulation of its neuromorphic, spike-driven components.

\subsection{Optimisation Problem and Heuristics}\label{Sec:Found:OptHeu}

\noindent%
The proposed approach is based on the following key concepts, which establish the terms used to describe the approach in the remainder of this section. First of all, we employ the minimisation problem \( (\mathfrak{X}, f, \text{min}) \) definition such as
\begin{equation}\label{Eq:Min_Problem}
    \textstyle\pmb{x}_* = \operatorname{argmin}_{\pmb{x}\in\mathbb{R}^d}\left\{ f(\pmb{x})\right\},
\end{equation}
where \( \pmb{x}_{*} \in \mathfrak{X} \) is the optimal solution within the feasible domain  \( \mathfrak{X}\subseteq\mathbb{R}^d  \) that minimises the objective function \( f:\mathbb{R}^d\mapsto\mathbb{R}  \), \ie  \( f\left(\pmb{x}_{*}\right) \leq f(\pmb{x}),\, \forall\ \pmb{x}\, \in \mathfrak{X}  \) \cite{Cruz-Duarte2021b}. 
This definition inherently incorporates constraints in the feasible domain specification. In contrast,
\begin{equation}\label{Eq:FeasibleDomain}
\mathfrak{X} = \left\{
    \pmb{x} \in \mathbb{R}^d
    \mid
    x_{l,j} \leq x_j \leq x_{u,j}, \, \forall\, j \in \mathbb{N}_d
    \right\},
\end{equation}
stands for a simple, feasible domain, defined by the lower and upper bound constraints \( \pmb{x}_l, \pmb{x}_u \in\mathbb{R}^d\). In this work, we deal with continuous minimisation problems, but the definition above can be easily extended to other domains.

Now, when addressing optimisation problems using heuristic-based algorithms, particularly those involving multiple probes, it is crucial to consider the concepts of population, neighbourhoods, and the best candidate solution. 
Therefore, a population \( \pmb{X} \) consists of a finite set of \( n \) candidate solutions associated to some units that evolve iteratively over time \(t\), \ie \( \pmb{X} = \{\pmb{x}_i\}_{i=1}^{n} \). 
Analogously, a neighbourhood is defined as \( \pmb{N} \subseteq \pmb{X} \). A common structure assumes that each candidate belongs to a neighbourhood of \( m \) neighbours, thus \( \pmb{N}_i \subseteq \pmb{X}, \, \forall i \in \mathbb{N}_n, \) with \( \#\pmb{N}_i = m < n \). Another relevant structure considers the historical trajectory of a given candidate, where a neighbourhood consists of its past states such that \( \pmb{N}_i^t = \{\pmb{x}_i^\tau\}_{\tau=1}^{t}. \)
With this, the best candidate solution from a given set is obtained via \( \arg\!\min_{\pmb{x} \in \pmb{A}} \{ f(\pmb{x}) \} \), since \( \pmb{A} \) stands for an arbitrary subset. For instance, we refer to \( \pmb{p}_i^t \) as the particular best candidates in the sets \( \pmb{N}_i^t \).
Plus, employing an auxiliary population \(\pmb{P}^t=\{\pmb{p}_i^t\}_{i=1}^n\), we determine the global best candidate solution \(\pmb{g}^t\) at time step \(t\), as well as more informed neighbourhoods, \ie \( \pmb{P}_{n,i}^t \subseteq \pmb{P}^t, \, \forall i \in \mathbb{N}_n\).

From the perspective of optimisation, these components naturally align with well-established heuristic concepts. A heuristic can be defined as a rule that generates or modifies candidate solutions for a given problem~\cite{Drake2020, talbi2009metaheuristics}. We follow a layered interpretation that distinguishes between Simple Heuristics (SHs) and Metaheuristics (MHs) based on their operational scope and domain dependency~\cite{Cruz-Duarte2020a}. SHs act directly on the problem domain and can be classified into three categories: generative, perturbative, and selective. 
Thus, an MH can be presented as,
\begin{equation}\label{Eq:MH-Def}
    \operatorname{MH} \triangleq\left\langle h_\text{init}, h_\text{sop}, h_\text{fin}\right\rangle=h_\text{fin}\left(h_\text{sop}\right) \circ h_\text{init},
\end{equation}
where $h_\text{sop}$ is the search operator, which usually comprises a perturbation \(h_\text{per}\) succeeded by a selection \(h_\text{sel}\) of candidates for the next step.
The others correspond to the initialisation \(h_\text{init}\) and finalisation \(h_\text{fin}\) heuristics, which are particular implementations of the generative and selective heuristics, respectively.
The main difference between \(h_\text{sel}\) and \(h_\text{fin}\) is the working domain, while the former interacts with the problem domain, the latter does it with the heuristic space; \ie low-level and high-level selectors.
Although \(h_\text{init}\) and \(h_\text{fin}\) are crucial for the overall process, $h_\text{sop}$ is the real signature of the well-known MHs in the literature. 
For example, consider DE, which has numerous variants in the literature \cite{Mohamed2021differential}. Most of these variant follow the notation DE/\emph{mutation}/\emph{differences}/\emph{crossover} that can be expressed as in \eqref{Eq:MH-Def} using
\(
    h_\text{sop} = (h_\text{sel,g}\circ h_\text{per,DC})\circ(h_\text{sel,d}\circ h_\text{per,DM})
\).
This search operator is chiefly composed of a differential mutation \(h_\text{per,DM}\) and a differential crossover \(h_\text{per,CR}\) corresponding to the DE/\emph{mutation}/\emph{differences}/\(\cdot\) and DE/\(\cdot\)/\(\cdot\)/\emph{crossover} parts.
To illustrate these, we selected DE/\emph{current-to-best}/1/\emph{bin}, such that
\begin{equation} \label{Eq:DE-mutation} 
    h_\text{per,DM}\{\pmb{x}_i^t\} \triangleq \displaystyle {\pmb{x}}_i^t + F (\pmb{g}^t - \pmb{v}_{i}^t)
        + F \left( \pmb{x}_{r_1}^t - \pmb{x}_{r_2}^t \right),
\end{equation}
\begin{equation} 
    h_\text{per,DC}\{\pmb{x}_i^t\} \triangleq \pmb{x}_i^{t+1}\odot  \mathcal{H}(p_\text{CR} - \pmb{r}) + \pmb{x}_i^{t} \odot \mathcal{H}^\text{c}(p_\text{CR} - \pmb{r}),
\end{equation}
where \(F\!\in\! [0, 2]\), \(r_1,r_2\!\sim\!\mathcal{U}(0,1),r_1\!\ne\! r_2\), and \(\pmb{r}\ni r_j\!\sim\!\mathcal{U}(0,1)\), and \(p_\text{CR}\!\in\![0,1]\). Plus, \(\odot\) is the Hadamard-Schur's product, and \(\mathcal{H},\mathcal{H}^\text{c}:\R^D\to \Z_2^D\) are the element-wise Heaviside function and its complement such that $\mathcal{H}(\pmb{y})+\mathcal{H}^\text{c}(\pmb{y})=\pmb{1}$, for an arbitrary $\pmb{y}\in\R^d$.
Moreover, \(h_\text{sel,d}\) and \(h_\text{sel,g}\) stand for the direct and the greedy selection \cite{Cruz-Duarte2021b}.

\subsection{Spiking Neuron Models as Adaptive Perturbation Units} \label{Sec:Found:SpikingNeuron}

\noindent%
Spiking neurons constitute the fundamental computational element in our proposal. They enable dynamic, state-dependent perturbations through biologically grounded membrane potential dynamics \cite{Nunes2022snns-review}. These models simulate the evolution of neuronal states, driven by external inputs and internal synaptic activity, resulting in spike events when the membrane potential exceeds a specific threshold. The Leaky Integrate-and-Fire (LIF) model efficiently captures this behaviour, while more complex models like Izhikevich and Hodgkin–Huxley include ionic channel mechanisms and nonlinear adaptations \cite{Sanaullah2023snns-review, Yamazaki2022snns-review}. 
We consider the integration of different neuron models to encourage practitioners to implement the most suitable spiking dynamics for their strategies. 
Thus, each candidate solution \( \pmb{x} \) is represented by \(d\) independent spiking neurons that can share information via spike-triggered mechanisms.

\begin{figure}[!htp]
    \centering
    \resizebox{0.5\columnwidth}{!}{
        \begin{tikzpicture}[scale=1.0, node distance=0.1]
    \draw[thick, fill=pastelBlue!60, opacity=0.7] plot [smooth cycle, tension=0.5] 
    coordinates {
        (4,     -1.2)            
        (4.5,   0.8)          
        (5.5,   1.3) 
        (7.8,     1.5)            
        (6.8,   -0.8) 
        (5.0,   -0.9)
        };
    \node (U) at  (7.5, -0.8) {\( \mathfrak{V}\subseteq\mathbb{R}^p \)};
    
    \draw[thick, fill=pastelCoral!60, opacity=0.7] plot [smooth cycle, tension=1.5] 
    coordinates {
        (4, -1.2) 
        (4.3, -0.2) 
        (5.4, -0.2) 
        (6., -1.2) 
        (5.0, -1.6)
        };
    \node (U) at  (6.1, -1.4) {\( \mathcal{S} \)};

    \node (xt) at (1, 0.75) {} 
        node [left=of xt] {\( x_j^t \)};
    \node (xtp) at (1, -1.0) {} 
        node [left=of xtp] {\( x_j^{t+1} \)};
    \node (xtpp) at (1, 0.25) {} 
        node [left=of xtpp] {\( x_j^{t+2} \)};
    \node (vt) at (6.5, 0.75) {} 
        node [right=of vt, xshift=-2.5] {\( \pmb{v}_j^{t} \)};
    \node (vtp) at (4.3, -0.8) {} 
        node [right=of vtp, yshift=3, xshift=-5] {\( \pmb{v}_j^{t+1} \)};
    \node (vtpp) at (6.4, -0.64) {} 
        node [above=of vtpp, xshift=5, yshift=-5] {\( \pmb{v}_j^{t+2} \)};

    \draw[<->, >=latex, thick] (1, 2) -- (1, -2) node[pos=0, right, yshift=-1] {\( \mathfrak{X}_j\subset\mathfrak{X} \)};

    \fill (xt) circle (2pt);
    \fill (xtp) circle (2pt);
    \fill (xtpp) circle (2pt);
    \fill (vt) circle (2pt);
    \fill (vtp) circle (2pt);
    \fill (vtpp) circle (2pt);
    
    \draw[->, thick, >=latex, color=blue] (xt) 
        to[out=30,in=140] node[below] { $\mathcal{T}$} (vt);
    \draw[->, thick, >=latex, color=red] (vtp) 
        to[out=-135,in=-30] node[above=0.1] { $\mathcal{T}^{-1}$} (xtp);
        
    \draw[->, thick, >=latex, color=blue] (xtp) 
        to[out=45,in=140] node[below] { $\mathcal{T}$} (vtp);
    \draw[->, thick, >=latex, color=red] (vtpp) 
        to[out=140,in=30] node[pos=0.7, below=0.1, xshift=0] { $\mathcal{T}^{-1}$} (xtpp);
    
    \draw[thick, ->, >=latex] (vt) 
        to[out=180, in=80] node[pos=0.35, above, xshift=0] {\( h_d \)} (vtp);
    \draw[thick, ->, >=latex] (vtp) 
        to[out=-45, in=200] node[pos=0.21, below, xshift=5] {\( h_s \)} (vtpp);

    \coordinate (A) at (8.5,2);

    \coordinate (B) at ($(A) + (-65:2)$);
    \coordinate (C) at ($(A) + (25:1)$);

    \node[below] at (C) {\(s_j\)};
    
    \draw[thick, <->, >=latex] (C) -- (A) -- (B);
    
    \foreach \i in {0, 1, 2} {
        \coordinate (tp\i) at ($(A)!0.28*(\i+0.7)!(B)$);
        \fill (tp\i) circle (2pt);
    }
    \coordinate (spike) at ($(tp1) + (25:0.5)$);
    \draw[thick, ->, >=latex, color=forestGreen] (tp1) -- (spike);

    \node[right, yshift=2, color=textdark] at (tp0) {\(s_j^{t}\)};
    \node[right, yshift=2, color=forestGreen] at (spike) {\(s_j^{t+1}\)};
    \node[right, yshift=2, color=textdark] at (tp2) {\(s_j^{t+2}\)};

    \fill[textdark!50] (tp0) circle (1pt);
    \fill[forestGreen] (tp1) circle (1pt);
    \fill[textdark!50] (tp2) circle (1pt);

\begin{pgfonlayer}{background}
    \draw[dashed, thick, color=textdark!50] (vt) -- (tp0);
    \draw[dashed, thick, color=forestGreen!80] (vtp) -- (tp1);
    \draw[dashed, thick, color=textdark!50] (vtpp) -- (tp2);
\end{pgfonlayer}

\end{tikzpicture}
    }
    \caption{%
    Example of three search steps carried out internally by a spiking neuron in an evolutionary process. A position component \(x_j^t\) is mapped via \( \mathcal{T} \) to render \( \pmb{v}_j^t \). The blue- and red-shaded regions are the neuromorphic space \( \mathfrak{V} \) and the spiking condition set \(\mathcal{S}\), where the dynamic \( h_d \) and spike-triggered \( h_s \) rules reign. The floating axis \(s_j\) shows the associated spiking activity.
    }    
    \label{Fig:commutative-diagram}
\end{figure}
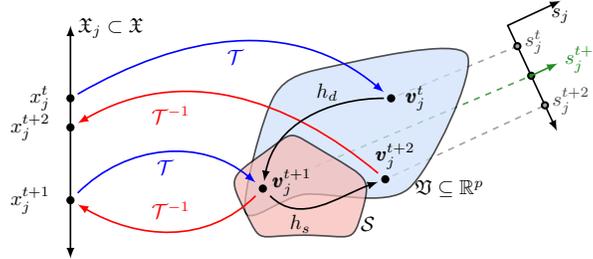

First, before describing the particular elements of this model, \figurename~\ref{Fig:commutative-diagram} presents an example of its behaviour on three representative search steps.
It features the \(j\)\textsuperscript{th} candidate solution component \(x_j^t\in\X_j\subset\X \) at the step \(t\), corresponding to a neuron, which evolves through spiking-driven state transitions represented with \(\pmb{v}_j^t\in\mathfrak{V}\subseteq\R^p\). This is possible due to the transformation \( \mathcal{T}:\X_j\leftrightarrow\mathfrak{V}\), where its corresponding \( \pmb{v}_j^t \) updates according to the rules \( h_d \) and \( h_s \). This procedure iterates as a new state arises via spike-triggered transition, which fires the spike \(s_j^{t+1}\), within the spiking condition set \( \mathcal{S} \).

We commence dissecting the process overviewed above, focusing solely on the neuromorphic space to establish the foundations of our proposal. A neuron is usually modelled throught its membrane potential \( v^t \), which accumulates inputs and evolves dynamically based on synaptic interactions \cite{Schuman2022}.
To describe this behaviour, we regard the LIF model due to its simplicity \cite{Sanaullah2023snns-review}.
It models the neuron's membrane potential charging and discharging dynamics as a leaky capacitor, which is numerically written using the Euler method, such as
\begin{equation}\label{Eq:LIF-NUM}
    v^{t+1} \gets \begin{cases}
    v^{t} -\textstyle\frac{\Delta t}{\tau_m}\left(v^{t}-v_{\text{rest}} + i_{\text{syn}}^{t}\right), &\text{if}\, v^t< v_{\text{th}},\\
    v_{\text{rest}},&\text{if}\, v^t\geq v_{\text{th}},
    \end{cases}
\end{equation}
since \( \tau_m \) stands for the membrane time constant, \( v_{\text{rest}} \) corresponds to the resting or equilibrium potential, and \( i_{\text{syn}}^t \) denotes the synaptic input, and \(\Delta t\) represents the numerical time step \cite{Yamazaki2022snns-review}. 
Therefore, when \( v^t \) reaches a threshold \( v_{\text{th}} \), a spike is fired, triggering an immediate reset.
From now on, we assume all variables are nondimensionalised. 

This numerical model is low-cost and easy to implement, but it falls short of emulating all the dynamic artefacts of neurons \cite{Nunes2022snns-review}. Notwithstanding, the LIF model sheds some light on how different rules can govern this internal dynamic, and how problems might require specific configurations for threshold, state-update dynamics, and spike-triggering conditions.

Moving forward with a richer-behaviour model, we regard the Izhikevich model, which is also computationally efficient and capable of reproducing the spiking and bursting artefacts of some kinds of neurons \cite{Yamazaki2022snns-review}.
This model comprises two differential equations that, similar to LIF, can be expressed in their numerical form as follows
\begin{equation}\label{Eq:Izhikevich-NUM}
    \begin{pmatrix}
        v^{t+1} \\ u^{t+1}    
    \end{pmatrix}
    \gets \begin{cases}
    h_d(v^t,u^t), &\text{if}\, v^t< v_{\text{th}},\\
    h_s(v^t,u^t),&\text{if}\, v^t\geq v_{\text{th}},
    \end{cases}
\end{equation}
with 
\begin{equation}\label{Eq:Izhikevich-NUM-heu}
\begin{aligned}
    \textstyle h_d\!\left(\begin{matrix}
        v^t\\ u^t
    \end{matrix}\right)&\!\triangleq\!\begin{pmatrix}
        \textstyle v^t + (\frac{1}{25} (v^t)^{2}\!+\!5 v^t\!+\!140-u^t\!+\!i_\text{syn}^t)\Delta t\\
        \textstyle u^t + a(b v^t-u^t)\Delta t
    \end{pmatrix},\\
    \textstyle h_s\!\left(\begin{matrix}
        \textstyle v^t\\  
        \textstyle u^t
    \end{matrix}\right)&\!\triangleq\!\begin{pmatrix}
        \textstyle c\\ 
        \textstyle u^t+d 
    \end{pmatrix}.
\end{aligned}
\end{equation}
Likewise, \( v^t \), \( v_\text{th}\), and \(i_\text{syn}^t\) stand for the membrane potential, threshold potential, and input current. In addition, \( u^t \) corresponds to the recovery variable, and \(a\), \(b\), \(c\), and \(d\) are control parameters \cite{Izhikevich_2003}. 

With these expressions, it is time to define the dynamical updating rule \(h_d\) and the spike-triggered rule \(h_s\), \ie  \( h_d, h_s:\mathbb{R}^p\mapsto\mathbb{R}^p \). The former accounts for the neuron's intrinsic dynamics, and the latter models the after-spike reset behaviour.
Now, we specify the internal neuromorphic state vector, such as
\(
    \pmb{v}_j\in\mathfrak{V}\subseteq\mathbb{R}^{p},
\) 
where \( p \ge 1\) is given by the spiking neuron model, evolving under a defined dynamic.
In this state, \(v_{1,j}\in\pmb{v}_j\) acts as an analogue to membrane potential, while \(v_{k,j},\,\forall k\in\mathbb{N}_p\backslash\{1\},\) are model-specific auxiliary components.

Therefore, we propose a generalised neuron model that applies perturbations within a transformed space, where rules or heuristics are dynamically adjusted based on spike-triggered updates.
The state variable change is given by,
\begin{equation}\label{Eq:GeneralModel}
    \textstyle \pmb{v}^{t+1} \gets
    \begin{cases}
        h_d(\pmb{v}_j^t), & \text{if}\; \Phi(\pmb{v}_j^t,t,\dots) \neq 1,\\
        h_s(\pmb{v}_j^t), & \text{otherwise},\\
    \end{cases}
\end{equation}
where
\begin{equation}\label{Eq:SpikingCondition}
    \textstyle\Phi(\pmb{v}_j^t,t,\dots) \triangleq \bigvee_k \varphi_k(\pmb{v}_j^t,t,\dots),
\end{equation}
since \(\Phi(\pmb{v}_j^t,t,\dots)\!:\!\mathfrak{V}\times\Z_{+}\times\dots\mapsto\Z_2\) corresponds to the spiking condition function depending on either the state variable or external information.
In this work, we consider two main spiking conditions to determine \(\Phi\) for simplicity.
The first \(\varphi_s(\pmb{v}_j^t) \) is directly related to the spiking condition set \(\mathcal{S}\), which could be part, at least partially, of \(\mathfrak{V}\), as \figurename~\ref{Fig:commutative-diagram} depicts. 
Hence, we can recursively specify that \(\mathcal{S}=\{\pmb{v}_j\in\mathfrak{V} \mid \varphi_s(\pmb{v}_j)=1\}\).
The simplest \(\varphi_s\) implementation stands for the thresholding described for the LIF and Izhikevich models, \ie
\begin{equation}\label{Eq:SimpleSpikingCondition}
    \varphi_s(\pmb{v}_j)=|v_{1,j}|\geq \vartheta_j,     
\end{equation}
where \(\vartheta_j\in\R_+\) is a threshold membrane potential.

Associated with the \(\pmb{v}_j\) changes, the spiking signal \(s_j\in\Z_2\)  represents the self-firing events that a neuron produces. 
Thus, the function \(\varphi_s\) can also be employed as an indicator function, such that the \(s_j\) generation can be defined as,
\begin{equation}\label{Eq:Spiking-Signal}
    s_j^t\gets\varphi_s(\pmb{v}_j^t).
\end{equation}
This firing event is complemented by presynaptic signals from neighbouring units, whose activity is mediated by the spike-driven communication mechanism established in the SNN. 

Additionally, the second condition, \(\varphi_a(\pmb{v}_j^t) \), corresponds to spikes induced by the environment, typically described as an activation signal \(a_j\in\mathbb {Z} _ 2 \) determined from spikes fired in the neighbourhood.
This indicates whether adjacent neurons have recently emitted action potentials.
Hence, the simplest definition for this function stands for \(\varphi_a(\pmb{v}_j^t,\dots)=a_j^t\).


Let us now instantiate the illustrative example in \figurename~\ref{Fig:commutative-diagram} with concrete data. The dynamic behaviour of spiking neurons is best understood through their phase portraits, where, for two-dimensional models, one can directly observe the evolution of the membrane potential and the recovery variable under diverse regimes. In this context, we implement the Izhikevich model using \eqref{Eq:GeneralModel} with \(h_d\) and \(h_s\) as defined in \eqref{Eq:Izhikevich-NUM-heu}, and a spiking condition \(\Phi\) and region \(\mathcal{S}\) determined by \(\varphi_s\) in \eqref{Eq:SimpleSpikingCondition}, setting \( \vartheta_j = 30 \). 
For the simulation, we use \(\Delta t = 0.01\) and \(T = 100\).

\figurename~\ref{Fig:Izh-UV} depicts the phase-space trajectories and their associated signals of several cortical, resonator, and thalamo-cortical neurons, generated with the Izhikevich model and using the parameters reported in \cite{Izhikevich_2003}. 
The corresponding time series of the membrane potential, the recovery variable, and the spike signal accompany each portrait.
From a dynamic systems perspective, \( h_d \) governs the evolution of \( \pmb{v} \), capturing the neuron's intrinsic dynamics. This corresponds to the trajectories observed following the vector field before reaching the spiking threshold. Once a spike occurs, the system experiences an instantaneous reset through \( h_s \), as indicated by the abrupt jumps where trajectories are reset to new states.

\newcommand{\threepw}{0.15\linewidth}
\begin{figure}[!htp]
    \centering%
    \subfloat[Regular Spiking\label{Fig:Izh-RS}]{%
        \includegraphics[width=\threepw, trim=9pt 5pt 5pt 6pt, clip]{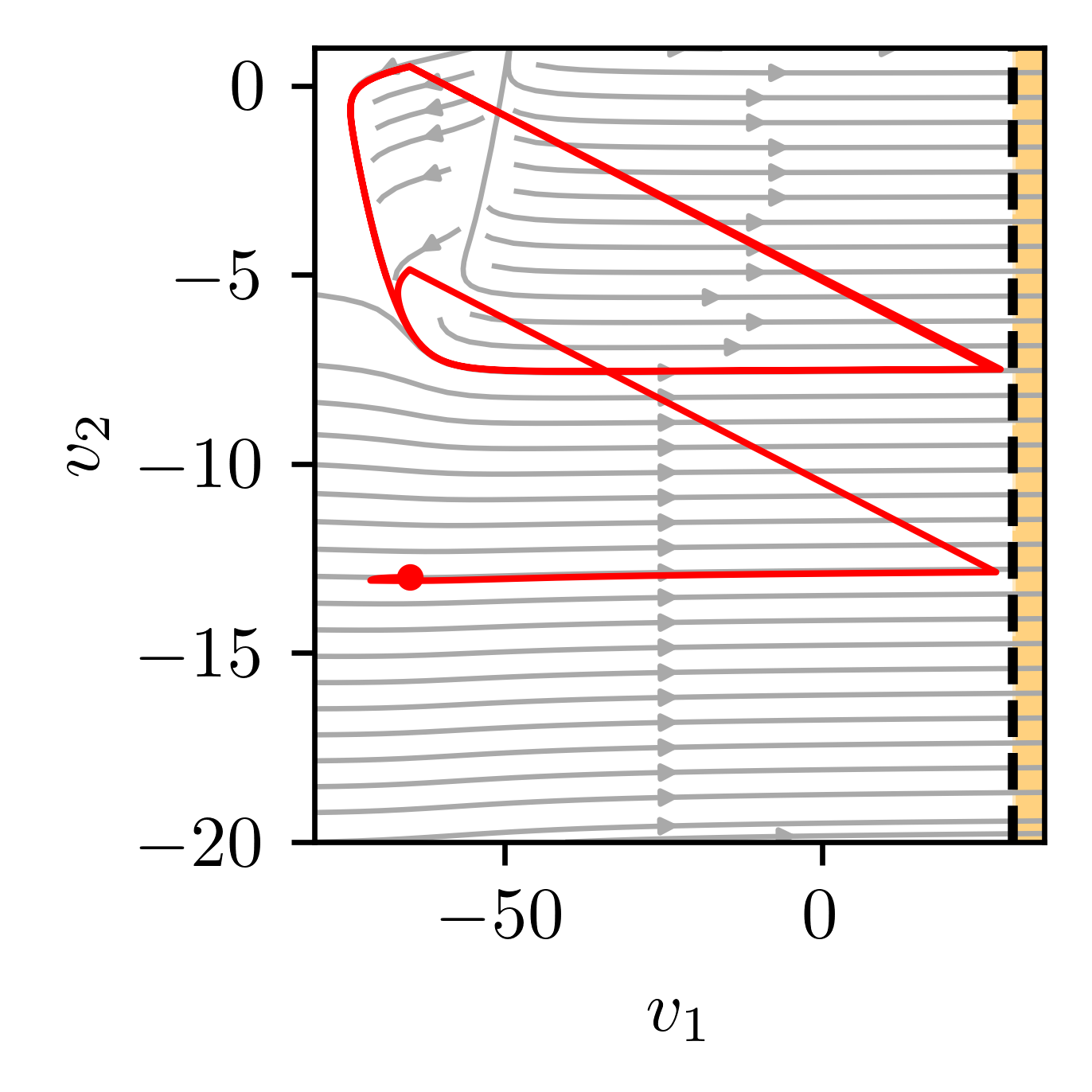}%
        \includegraphics[width=\threepw, trim=9pt 0pt 5pt 0pt, clip]{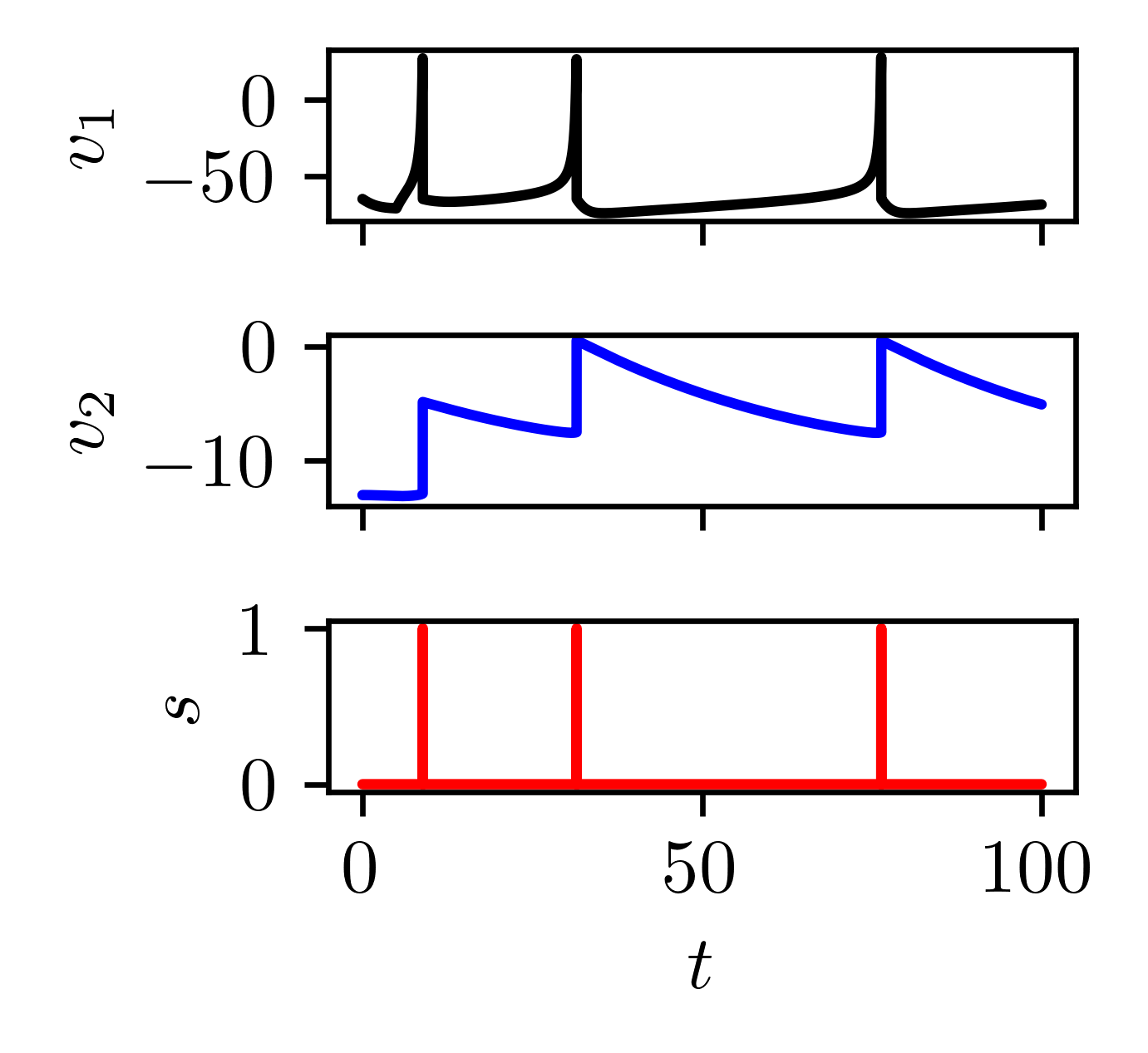}
    }
    \subfloat[Chattering\label{Fig:Izh-CH}]{%
        \includegraphics[width=\threepw, trim=9pt 5pt 5pt 6pt, clip]{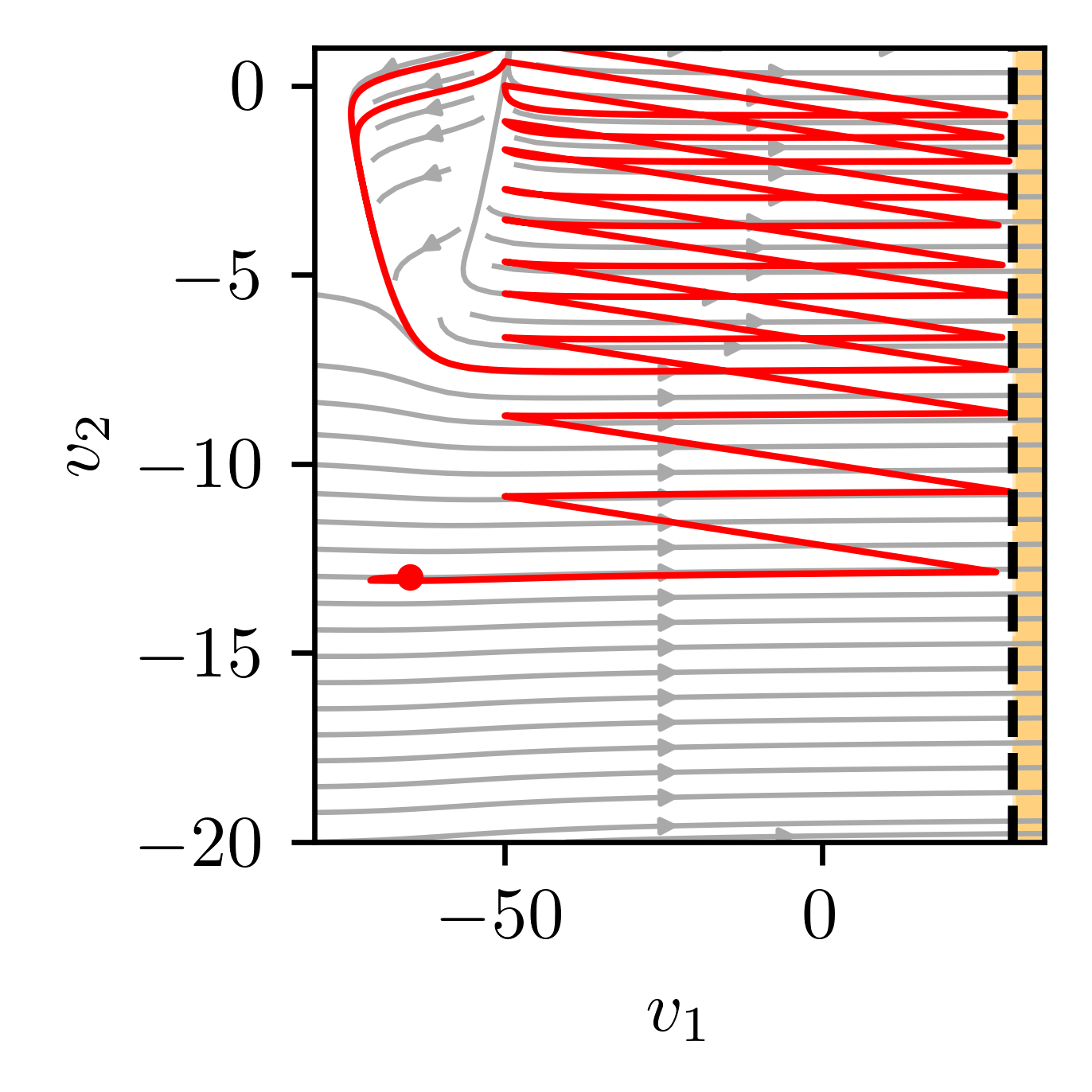}%
        \includegraphics[width=\threepw, trim=9pt 5pt 5pt 0pt, clip]{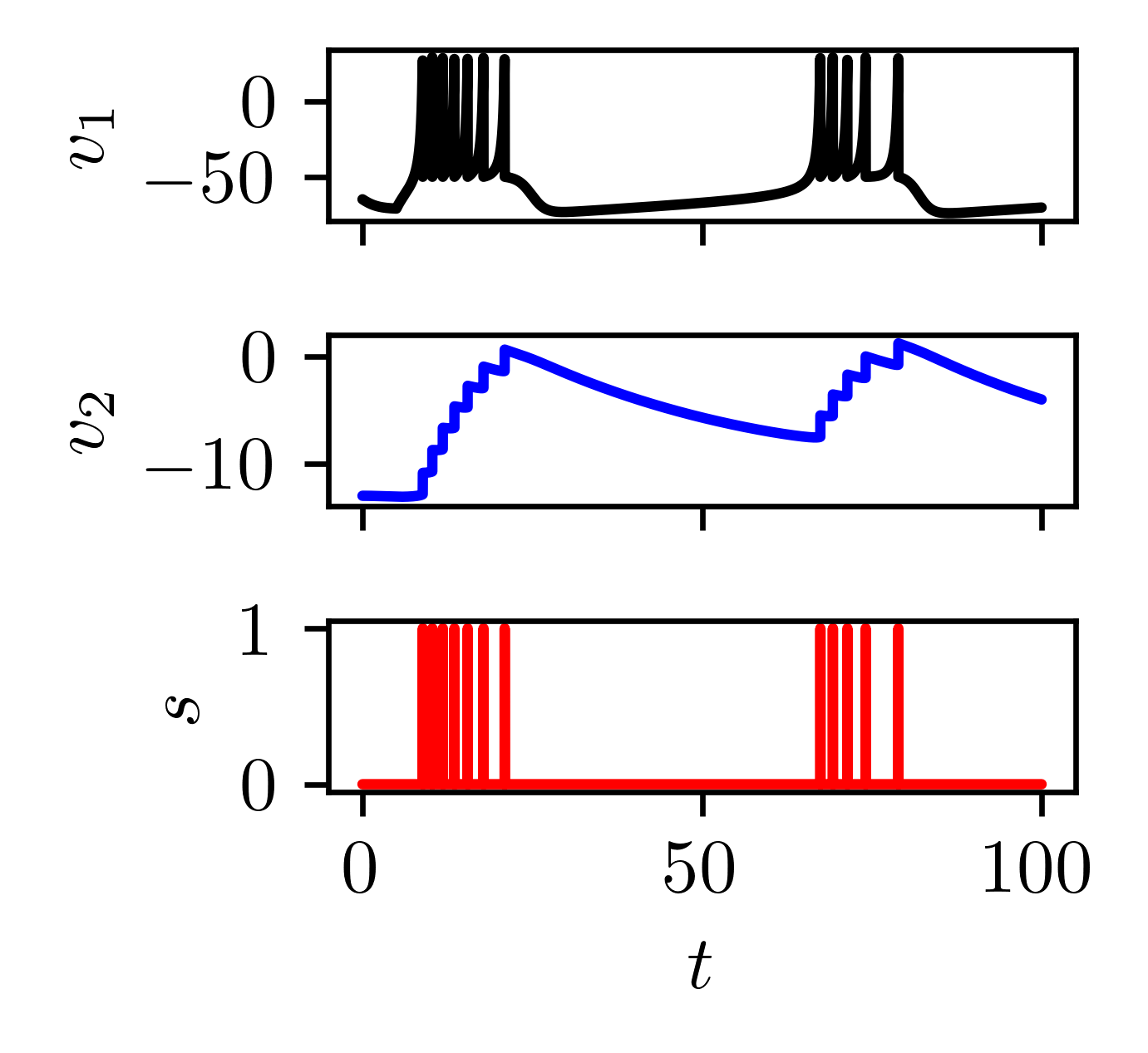}
    }\\[0pt]
    \subfloat[Resonator\label{Fig:Izh-RZ}]{%
        \includegraphics[width=\threepw, trim=9pt 5pt 5pt 6pt, clip]{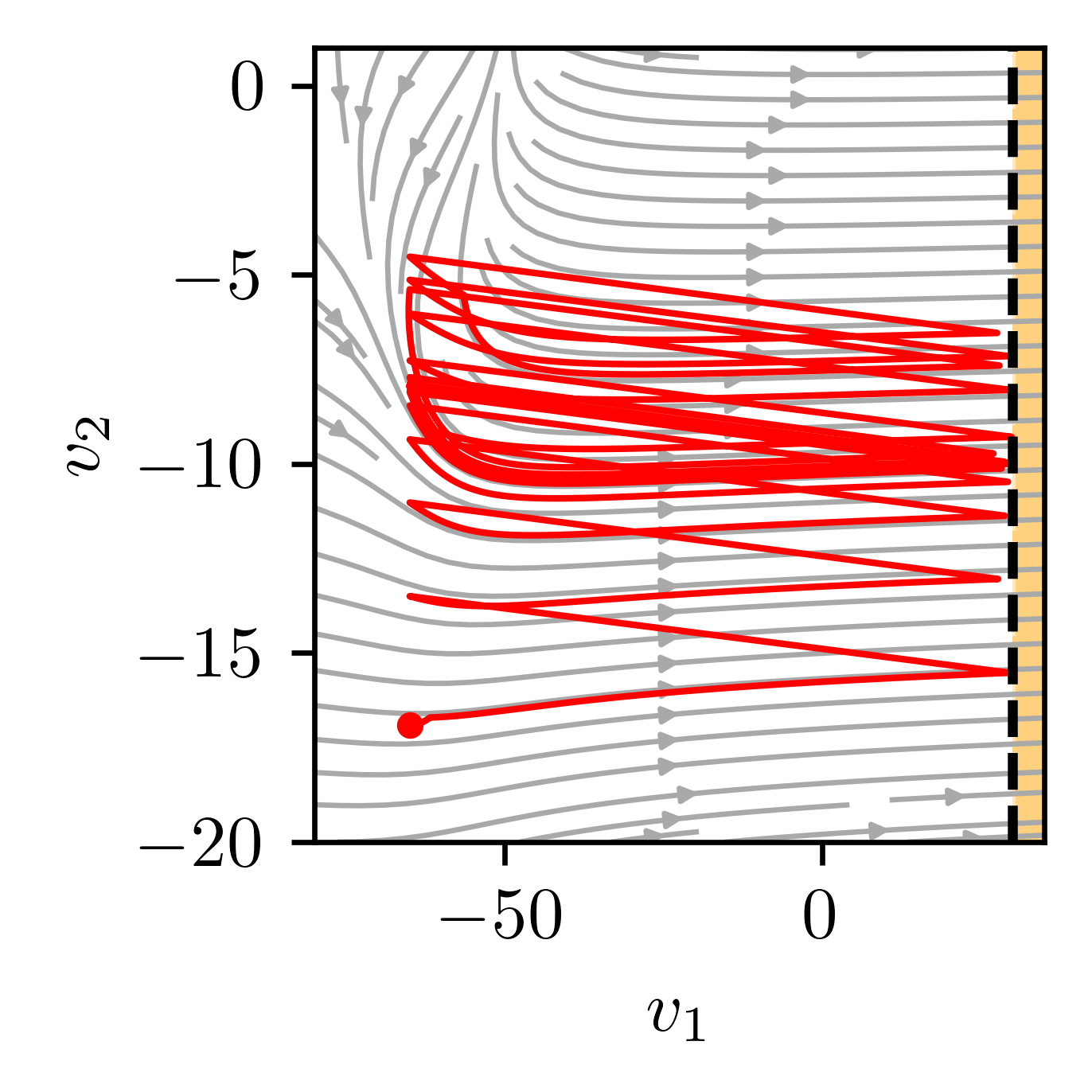}%
        \includegraphics[width=\threepw, trim=9pt 5pt 5pt 0pt, clip]{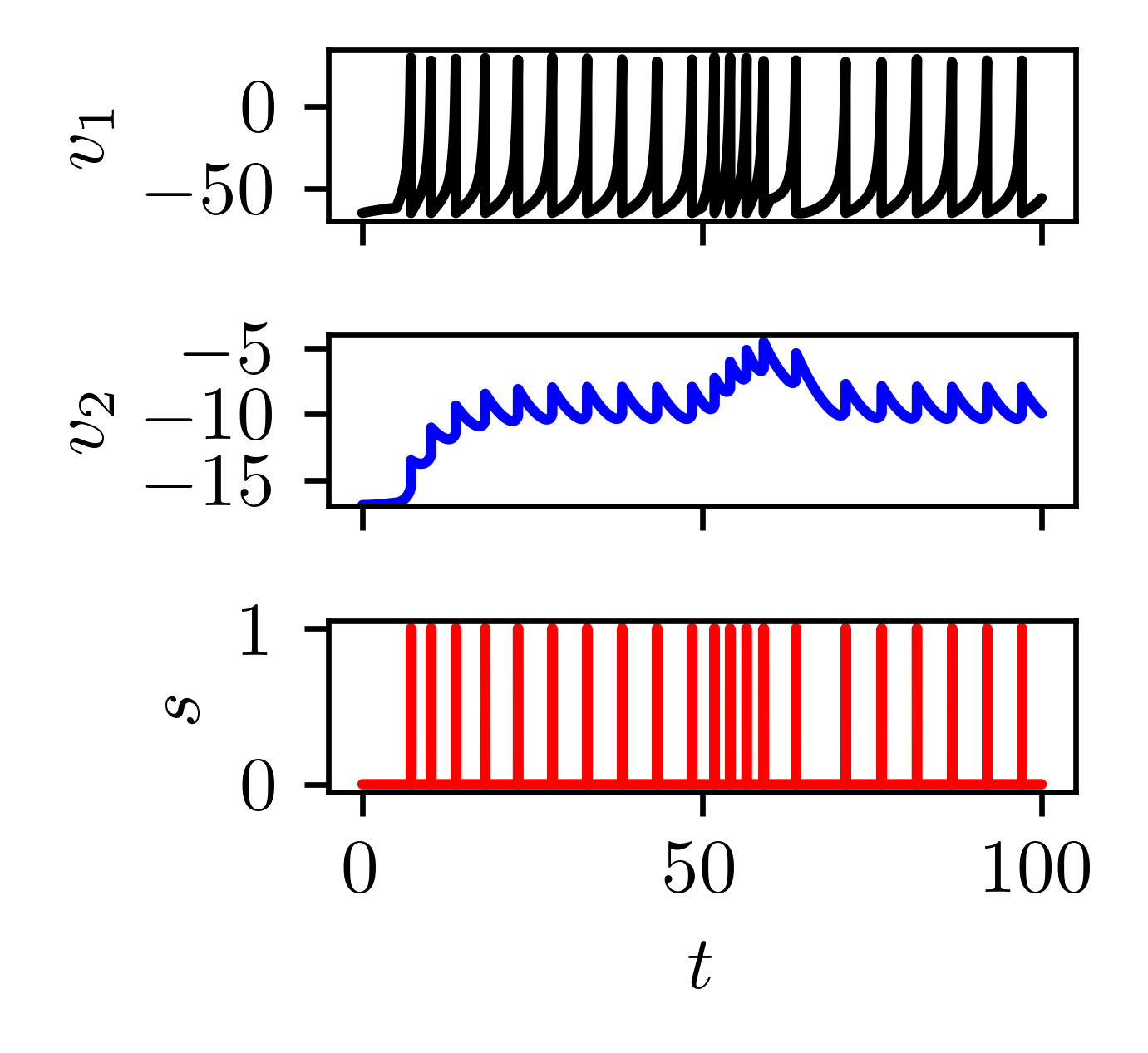}
    }
    \subfloat[Thalamo-Cortical\label{Fig:Izh-TC}]{%
        \includegraphics[width=\threepw, trim=9pt 5pt 5pt 6pt, clip]{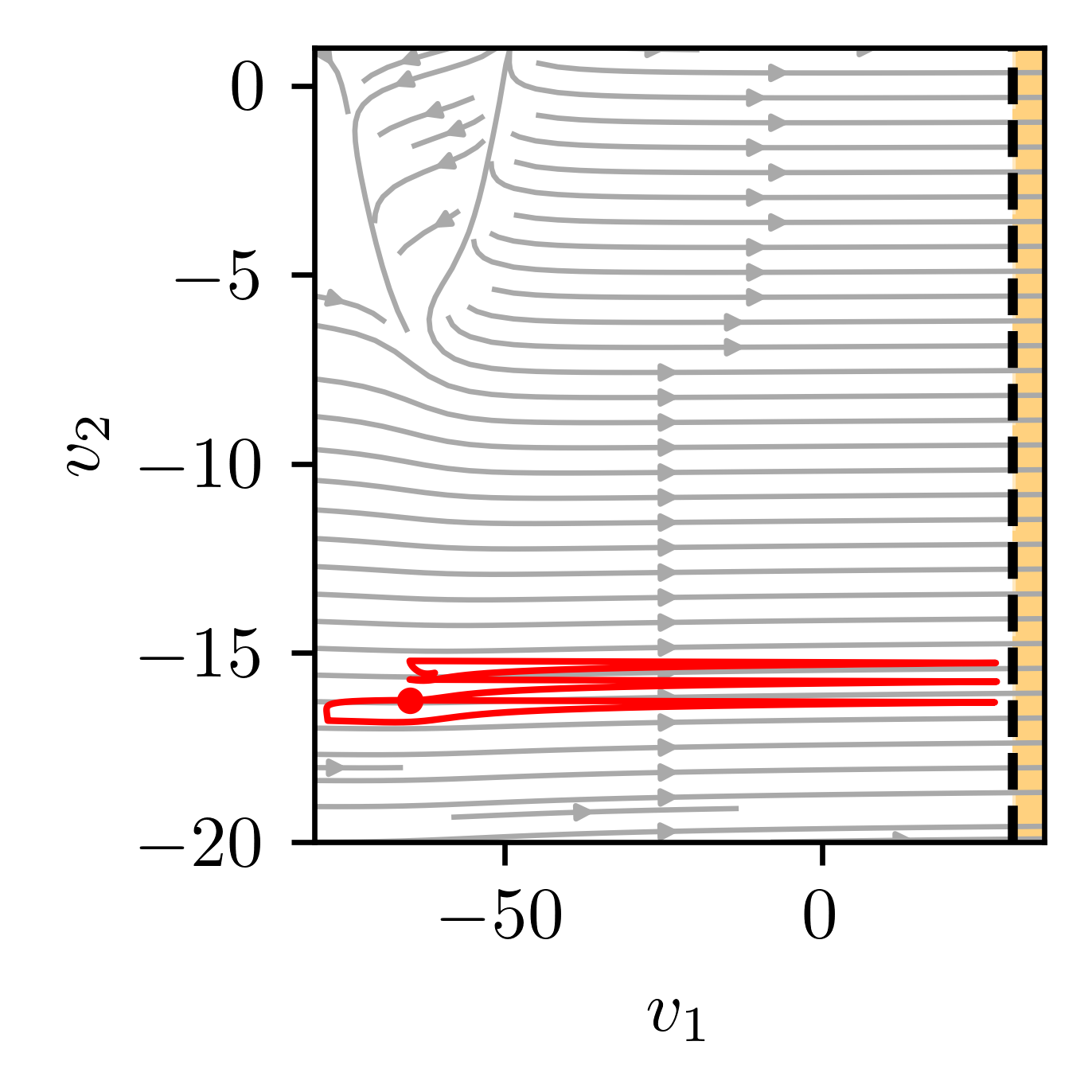}%
        \includegraphics[width=\threepw, trim=9pt 5pt 5pt 6pt, clip]{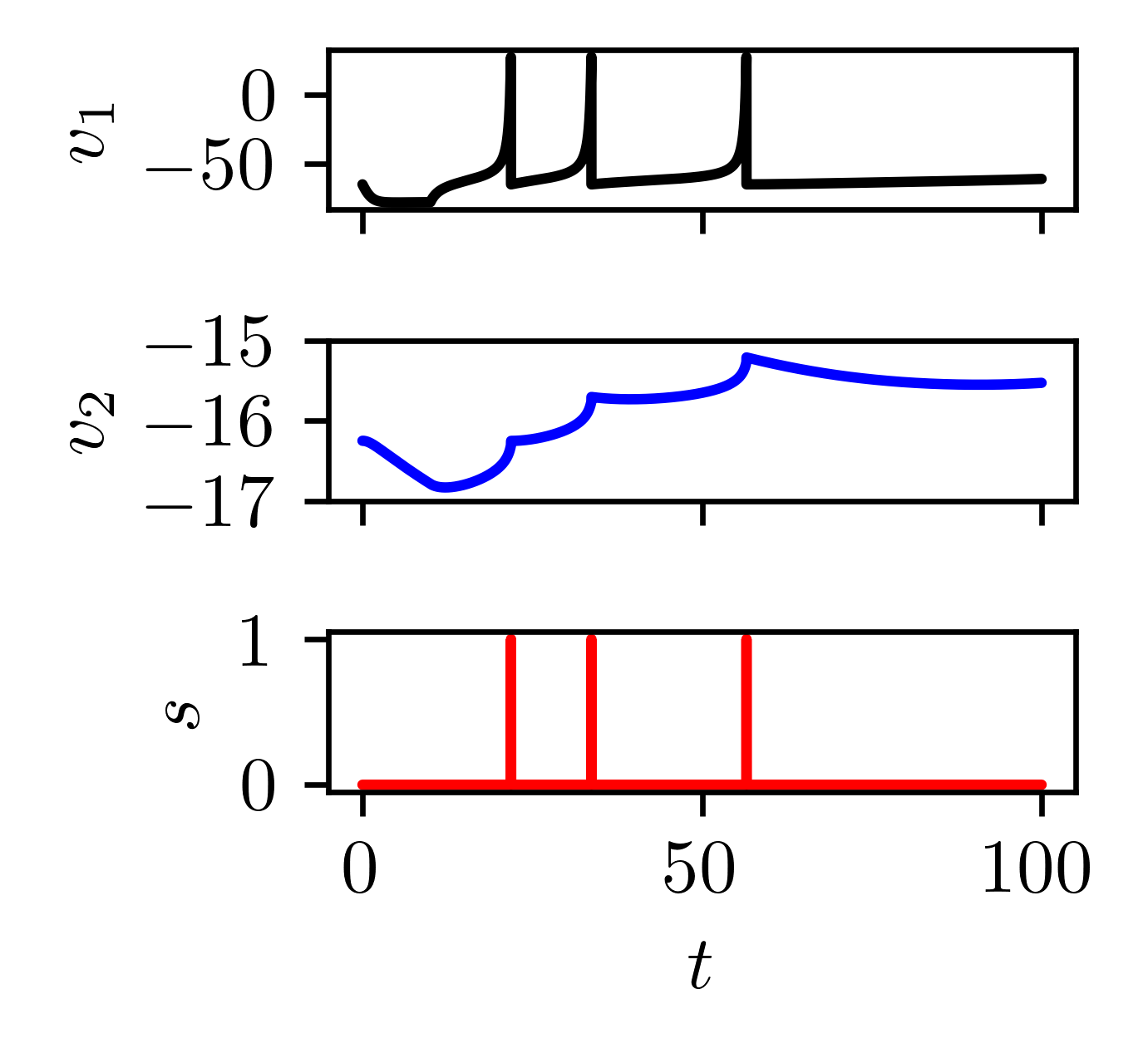}
    }\hfill
    \caption{Phase portraits (\(v_1\) vs. \(v_2\)) and time signals (\(v_1,v_2,s\) vs. \(t\)) of the four dynamical behaviours generated using the Izhikevich model and the parameters described in \cite{Izhikevich_2003}. The dashed black line and orangish-shaded region represent the thresholding condition \(\varphi_s\) and spiking region \(\mathcal{S}\).
    }
    \label{Fig:Izh-UV}
\end{figure}

To wrap up the idea mentioned above, let us consider a simpler model based on a two-dimensional dynamic system for a linear deterministic behaviour under \( h_d \) and stochastic spiking dynamics under \( h_s \), as shown below
\begin{equation}\label{Eq:2dsystem-diff}
\begin{aligned}
    h_d(\pmb{v}) &\triangleq \pmb{v} + \pmb{A}\pmb{v} \Delta t,\\ 
    h_s(\pmb{v}) &\triangleq \vartheta_1 + \vartheta_2 r_1 e^{\mathrm{i}2\pi r_2},\quad r_1,r_2\sim\mathcal{U}(0, 1).\\
\end{aligned}
\end{equation}
For this example, we choose a neuromorphic space \( \mathfrak{V} =[-2,2]^2 \) and a disc-shaped spiking subset \(\mathcal{S}\) defined through,
\begin{equation}\label{Eq:2dsystem-diff-pred}
    \varphi_s(\pmb{v})=\begin{cases}
        ||\pmb{v}||_2\leq \vartheta_\text{attractor},&\text{if } \operatorname{tr}(\pmb{A}) \leq0,\\
        ||\pmb{v}||_2\geq \vartheta_\text{repeller},&\text{if } \operatorname{tr}(\pmb{A}) >0,\\
    \end{cases}
\end{equation}
where \( \vartheta_\text{attractor} = 0.5\) and \( \vartheta_\text{repeller}=1.5 \) are thresholds for stable and unstable equilibrium points. 
Respectivelly, the pair \((\vartheta_1,\vartheta_2)\) for \(h_s\) are chosen \((0.5,1.0)\) and \((0.0,1.5)\).
We select these cases to exemplify how we can utilise the system's nature to design spiking cores by adapting thresholding mechanisms to the process of evolution. Plus, the linear coefficients of \( h_d \) and initial points were selected to display the state evolution under four different equilibrium points according to \cite{steven2024nonlinear}. For the simulation, we employ \(\Delta t = 0.01\) and \(T = 20\).

\begin{figure}[!htp]
    \centering%
    \subfloat[Stable Node\label{Fig:2ndSys:sink}]{%
        \includegraphics[width=\threepw, trim=9pt 5pt 5pt 6pt, clip]{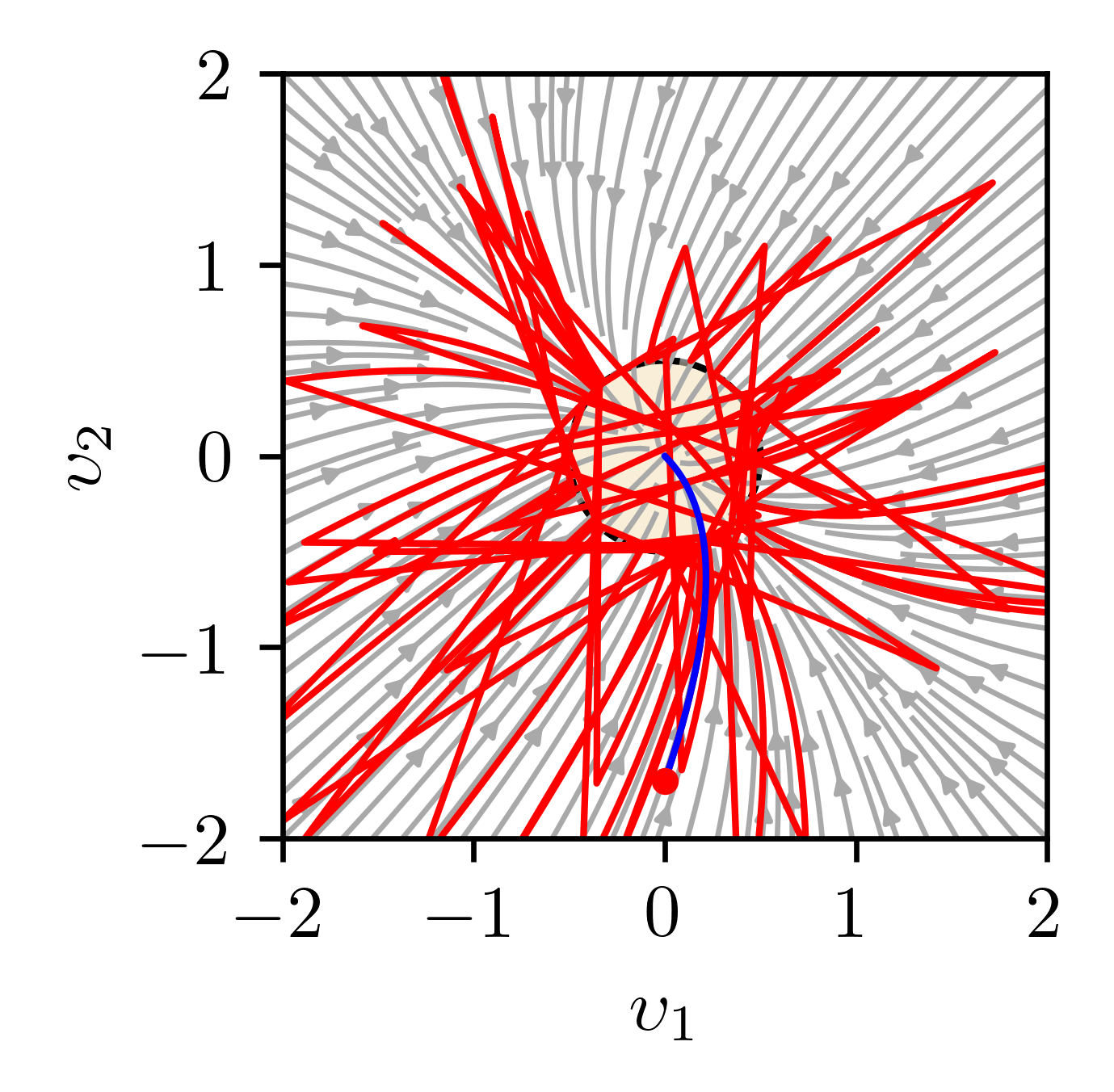}%
        \includegraphics[width=\threepw, trim=9pt 5pt 5pt 6pt, clip]{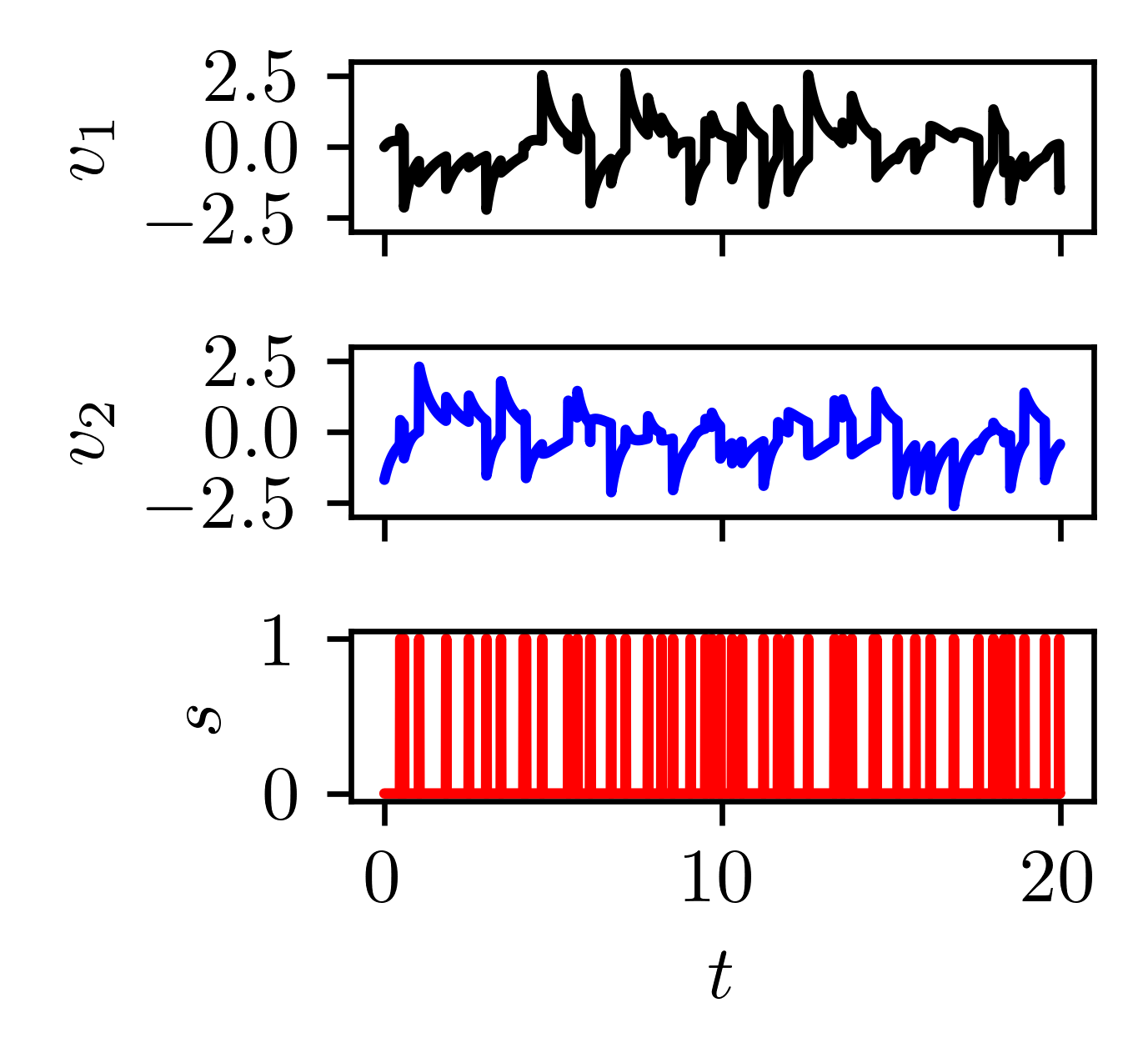}   
    }
    \subfloat[Stable Spiral\label{Fig:2ndSys:spiralsink}]{%
        \includegraphics[width=\threepw, trim=9pt 5pt 5pt 6pt, clip]{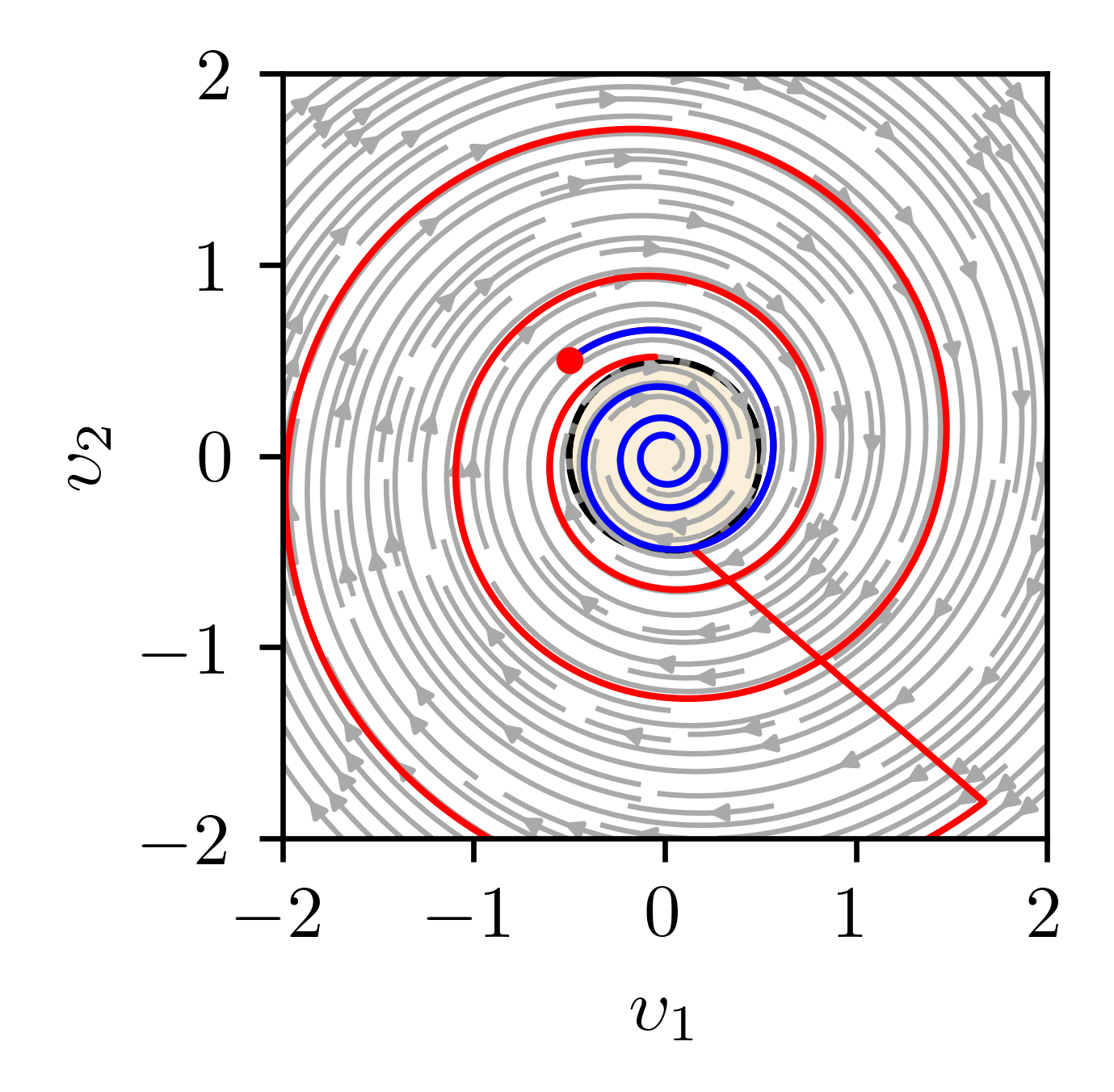}%
        \includegraphics[width=\threepw, trim=9pt 5pt 5pt 6pt, clip]{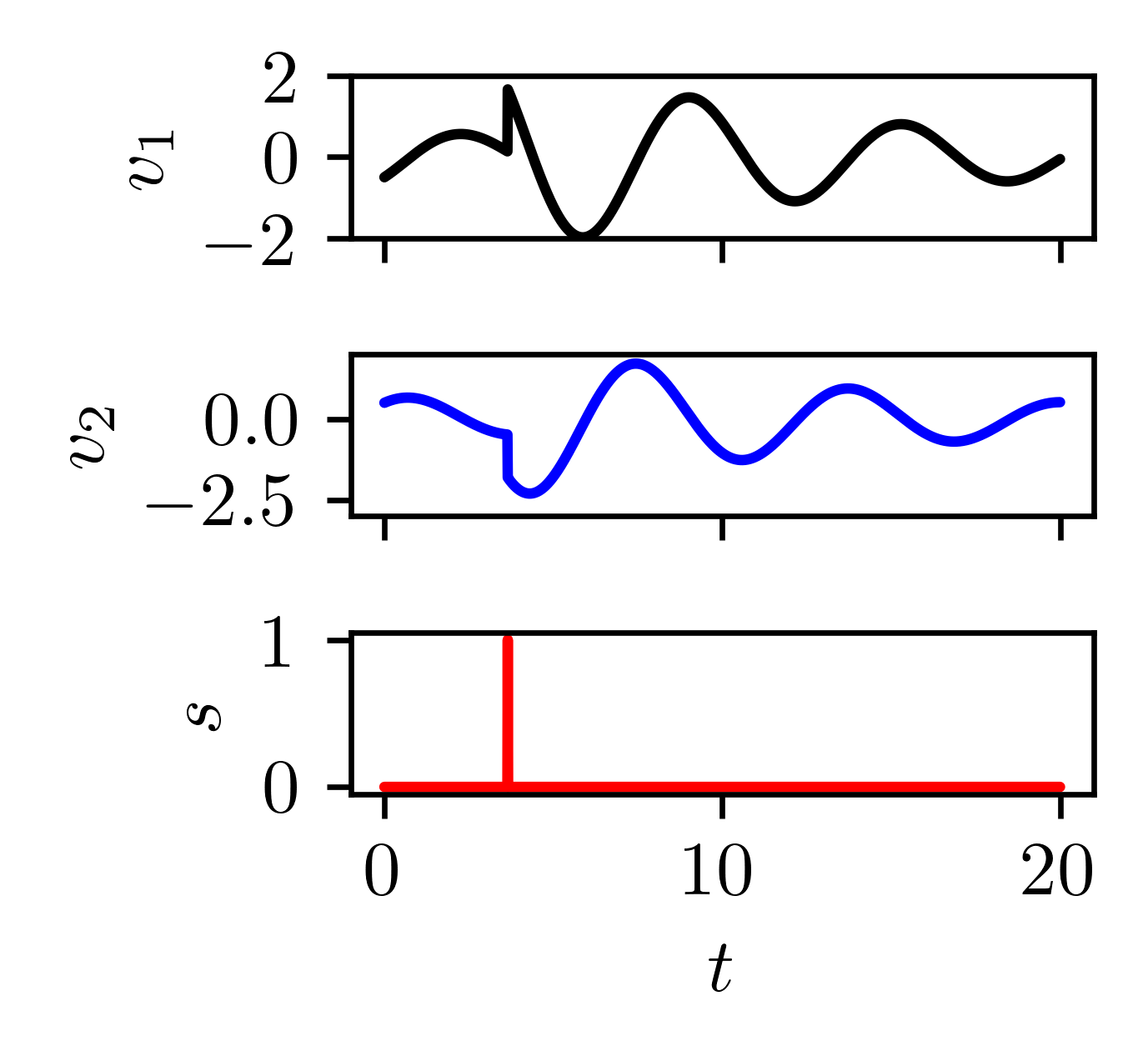}   
    }\\[0pt]
    \subfloat[Unstable Node\label{Fig:2ndSys:source}]{%
        \includegraphics[width=\threepw, trim=9pt 5pt 5pt 6pt, clip]{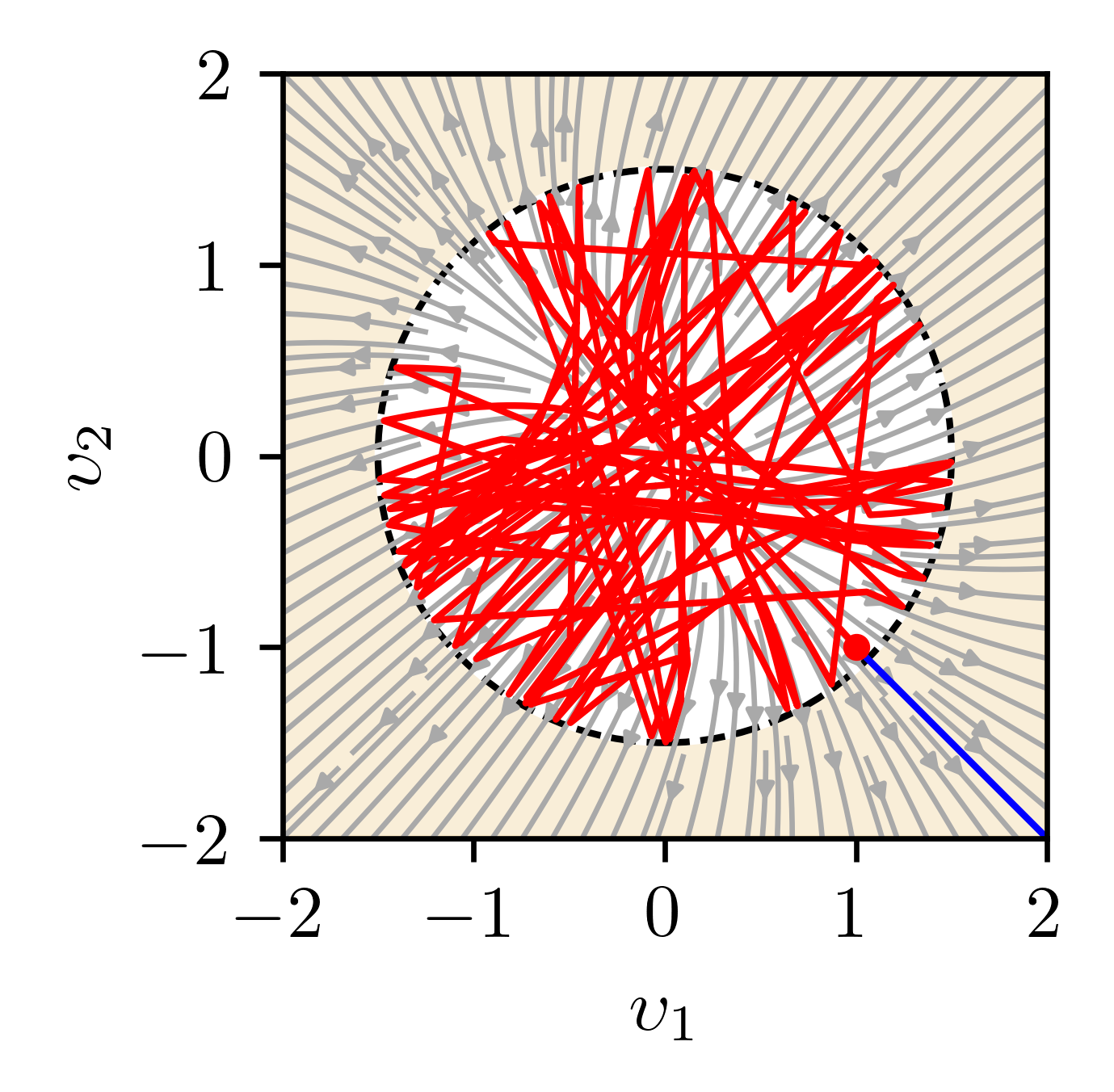}%
        \includegraphics[width=\threepw, trim=9pt 5pt 5pt 6pt, clip]{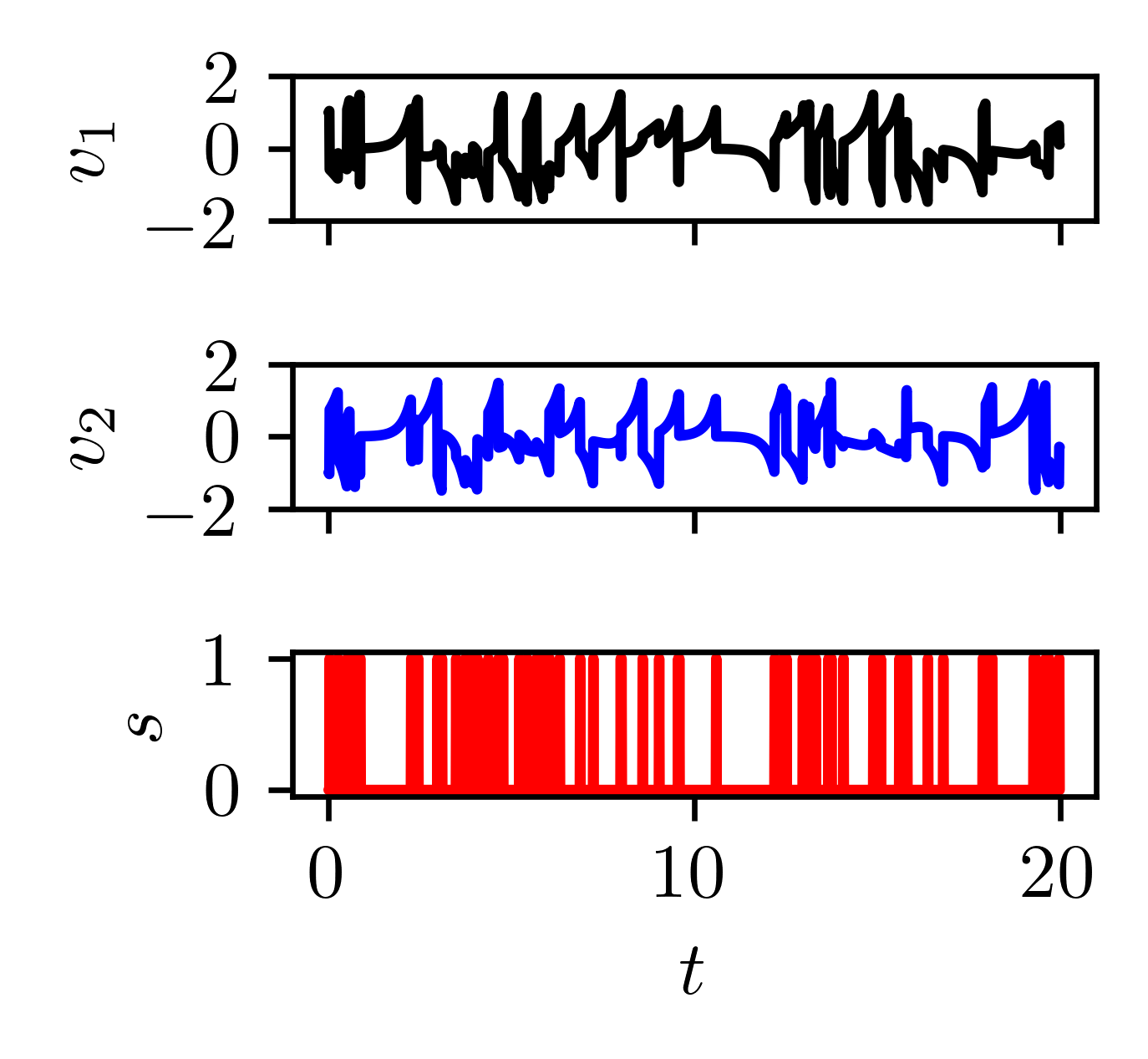}   
    }
    \subfloat[Unstable Spiral\label{Fig:2ndSys:spiralsource}]{%
        \includegraphics[width=\threepw, trim=9pt 5pt 5pt 6pt, clip]{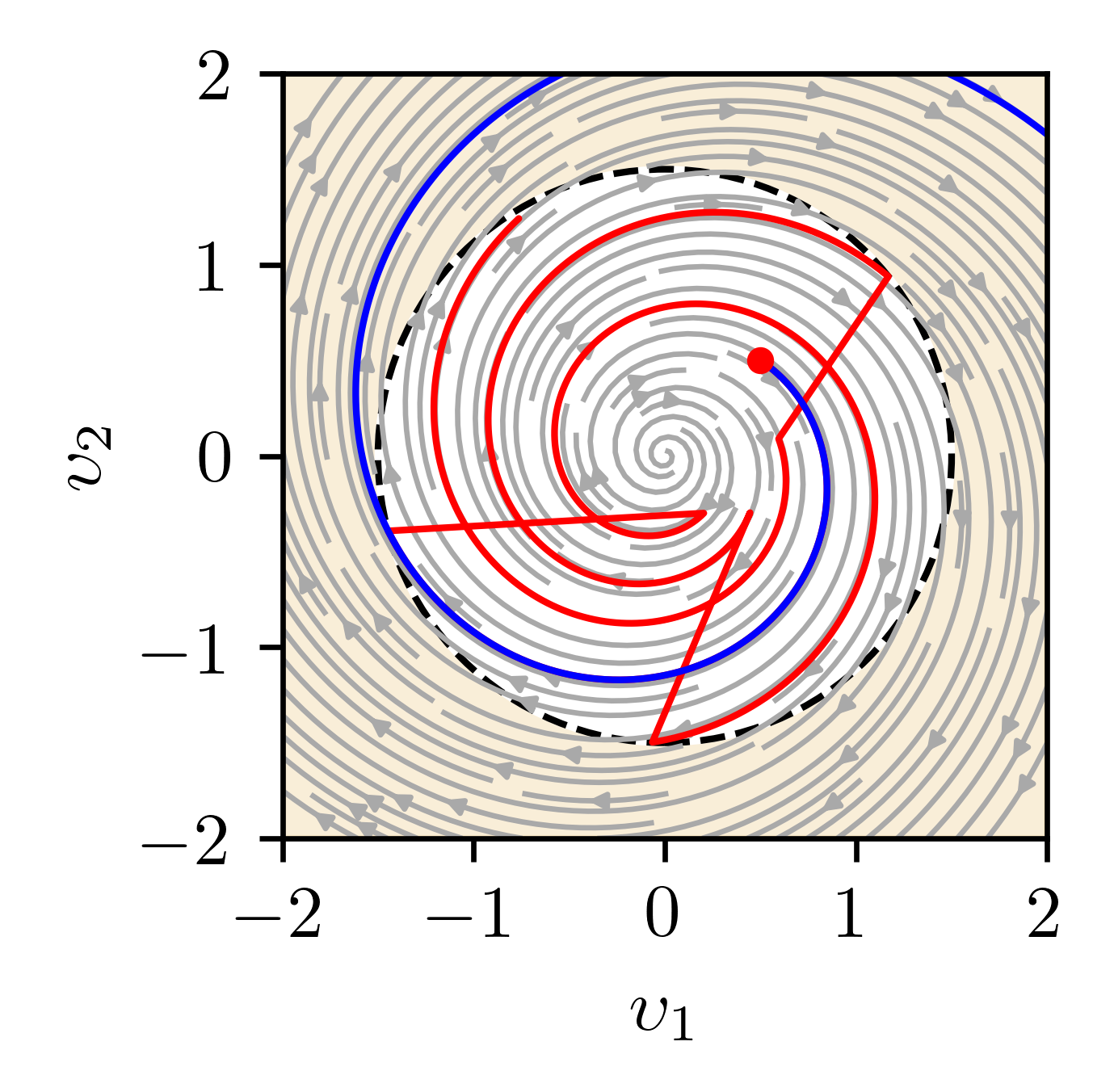}%
        \includegraphics[width=\threepw, trim=9pt 5pt 5pt 6pt, clip]{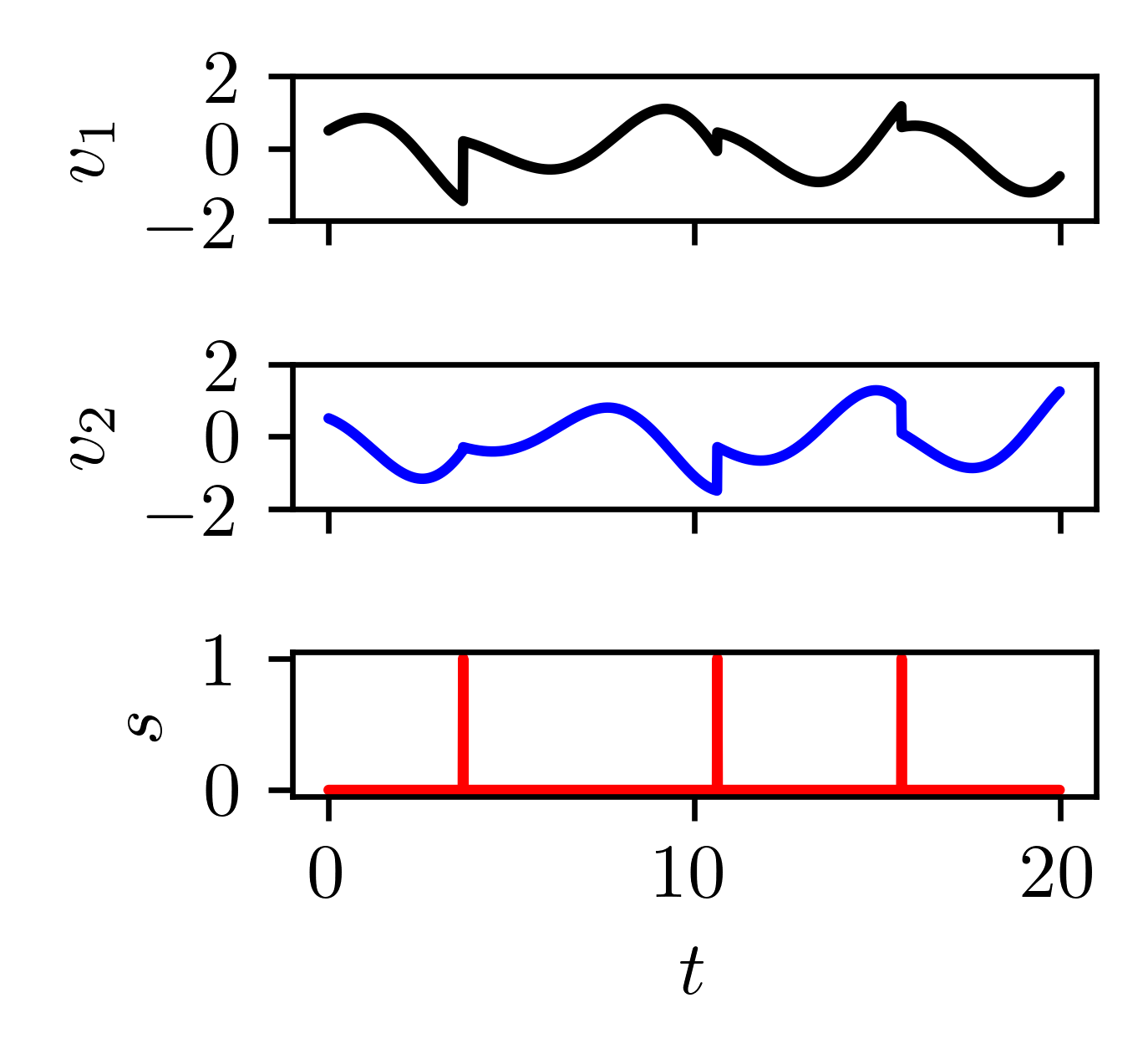}   
    }
    \caption{Phase portraits (\(v_1\) vs. \(v_2\)) and time signals (\(v_1, v_2, s\) vs. \(t\)) of four equilibrium points in a linear system. The dashed black circle and orangish-shaded region depict the thresholding condition \( \varphi_s \) and spiking region \(\mathcal{S}\). The red and blue lines show trajectories with and without spiking, respectively. 
    }\vspace{-1ex}
    \label{Fig:2ndSys}
\end{figure}

Similar to the previous portraits, \figurename~\ref{Fig:2ndSys} shows the phase-space trajectories and time signals for four illustrative systems. In this case, two trajectories (red and blue lines) per class start from a marked red point, but only the blue one changes when it enters the orangish-shaded region representing the spiking condition set. The first and second rows of \figurename~\ref{Fig:2ndSys} correspond to stable and unstable behaviours. Moreover, the first column depicts fast dynamics, \ie while trajectories in the stable node converge to the equilibrium, \(\varphi_{s}\) prevents them from reaching it, in the unstable case, trajectories diverge but are bounced back. The second column portrays spiral behaviours with slower evolution due to complex eigenvalues.

\subsection{Heuristic Interpretation of Spiking Dynamics}\label{Sec:Found:Heuristics}

\noindent%
\figurename{}~\ref{Fig:Izh-UV} and \ref{Fig:2ndSys} support the importance of state-dependent changes in shaping the behaviour of neuromorphic optimisation strategies. This formulation opens a pathway for integrating heuristic operations as rules that modify candidate solutions through spike-driven state transitions. So, designing such a spiking neuron requires specifying three principal components according to \eqref{Eq:GeneralModel}: the dynamic update rule \( h_d \), the spiking condition function \( \Phi \) to control \(\mathcal{S}\), and the spike-triggered rule \( h_s \). These components, which can be considered heuristics, must be carefully defined to shape the search dynamics, including the spiking conditions, which themselves act as selection criteria. In the following paragraphs, we describe these components from a heuristic perspective.

Foremost, spikes are fired when the neuron's internal state satisfies a condition \(\varphi_s\), which acts as a selection predicate for triggering transitions. The simplest case corresponds to the thresholding heuristic defined in \eqref{Eq:SimpleSpikingCondition}. Still, this principle can be extended to incorporate additional parameters reflecting internal dynamics or contextual information from the optimisation process. Some other representative conditions can be,
\begin{subequations}\label{Eq:SelfSpikingCond}
\begin{align}
    \varphi_s(\pmb{v}_j,\dots) &\triangleq ||\hat{\pmb{\alpha}}_\text{spk}\cdot\pmb{v}_j||_{q}>\vartheta_j, 
        \label{subEq:vp:lp} \\
    \varphi_s(\pmb{v}_j,\dots) &\triangleq ||\pmb{v}_j||_2 < \displaystyle {\varepsilon_\text{spk}}/({1+t}). 
        \label{subEq:vp:shrink}
\end{align} 
\end{subequations}
From these, the weighted Minkowski (\(\ell_q\)--norm) condition in \eqref{subEq:vp:lp} extends \eqref{Eq:SimpleSpikingCondition}, using the unit vector \(\hat{\pmb{\alpha}}_\text{spk}\in[0,1]^p\) to modulate each component's relevance.
Plus, the condition in \eqref{subEq:vp:shrink} implements a proximity criterion to the origin with a shrinking tolerance \(\varepsilon_\text{spk} > 0\), useful for systems with stable equilibrium points, \eg \figurename{}~\ref{Fig:2ndSys:sink} and \ref{Fig:2ndSys:spiralsink}.

Most of the parameters in \eqref{Eq:SelfSpikingCond} are either constants, deterministic sequences, or random variables. While the threshold \(\vartheta_j\) can be defined as a scalar, this static choice may ignore the potential for environment-aware modulation of spiking activity during the search process. To introduce flexibility and contextual awareness, we consider the following variants,
\begin{subequations}\label{subEq:vt}
\begin{align}%
    \vartheta_j &\triangleq \alpha_\text{thr}, 
        \label{subEq:vt:fix}
        \\ 
    \vartheta_j &\triangleq \alpha_\text{thr}|g_j - p_j|, 
        \label{subEq:vt:pg}
        \\ 
    \vartheta_j &\triangleq \alpha_\text{thr}|x_{\text{ref},j} - p_j|. 
        \label{subEq:vt:pgn}
\end{align}
\end{subequations}
From these, \(\alpha_\text{thr}>0\) is the base value that directly defines the threshold in the fixed case at \eqref{subEq:vt:fix}. 
The difference-based variant in \eqref{subEq:vt:pg}, inspired by \cite{Sasaki2023osnn}, uses the distance between the global \(p_j \in \pmb{p}\) and particular \(g_j \in \pmb{g}\) best components, whereas \eqref{subEq:vt:pgn} employs a target reference \(x_{\text{ref},j}\in\pmb{x}_\text{ref}\) instead of the global best component.


Subsequently, the heuristic \( h_d \) is defined by the evolution of a dynamic system chosen for modelling the spiking neuron. When the system implemented is discrete, this expression can be used directly. For example, two-dimensional strange attractors such as the Lozi and H\'enon maps~\cite{aslimani:hal-02499326,steven2024nonlinear}.
When the system is continuous, with the form given by \( \dot{\pmb{v}}_j = \mathfrak{g}(\pmb{v}_j) \), we must consider a numerical integration method for approximating the new state \(\pmb{v}_j^{t+1}\).
The simplest approximation method, such as the Euler method \cite{steven2024nonlinear}, is given by,
\begin{equation}\label{Eq:EulerApprox}
    h_d(\pmb{v}_j) \triangleq \pmb{v}_j + \mathfrak{g}(\pmb{v}_j) \Delta t,
\end{equation}
where the time step \(\Delta t>0\) must be specified and can be considered to control the detail level during the dynamic search around \(\mathfrak{V}\).
Two examples of these systems are discussed above, which are already in their numerical form in \eqref{Eq:Izhikevich-NUM-heu} and \eqref{Eq:2dsystem-diff}.
We avoid presenting them in their differential expression for brevity.
Still, their corresponding \(\mathfrak{g}\) forms are straightforward to infer from \eqref{Eq:Izhikevich-NUM-heu} and \eqref{Eq:2dsystem-diff}, using \eqref{Eq:EulerApprox}.
Notice that we focus on two-dimensional systems, but systems with one or more than two dimensions can also be implemented with ease, \eg LIF, Lorenz, and Hodgkin-Huxley \cite{Liu2022gamma-review, steven2024nonlinear}.
Naturally, other alternative numerical methods can be considered, \eg the 4\textsuperscript{th}-order Runge-Kutta algorithm \cite{steven2024nonlinear}.


Upon spiking, the spike-triggered heuristic \( h_s \) performs a transition over the internal state according to a perturbative operation. According to \sectionname~\ref{Sec:Found:OptHeu}, this rule can be formalised as \( \pmb{v}_j^{t+1} \gets h_s(\pmb{v}_j^t)\triangleq h_\text{per}\{\pmb{v}_j^t\} \), where \( h_\text{per} \) is chosen from a finite but massive set of heuristic operators existing in the literature. We consider four representative classes of perturbative rules, all of which are consistent with the original reset strategies from spiking neuron models \cite{Nunes2022snns-review} and the definition of simple heuristics~\cite{Cruz-Duarte2020a}.

The first class corresponds to a stochastic reset, related to a random search operation, defined as
\begin{equation}\label{Eq:hs:rand}
    h_s(\pmb{v}_j^t) \triangleq \pmb{\xi}_j^t
\end{equation}
since the state is sampled from the neuromorphic space \(\mathfrak{U}\) using any random distribution, \eg \(\pmb{\xi}_j^t \ni \xi_{j,k}^t\sim \mathcal{N}(0, 1)\).

Subsequently, the second class represents a reset toward a well-known state, similar to the reset mechanism of the LIF and Izhikevich models~\cite{Sanaullah2023snns-review}. We use the best visited state \( \bar{\pmb{v}}_{j}^{t} = \mathcal{T}\{p_j^t\} \), perturbated with slight noise \(\pmb{\epsilon}_j^t \ni \epsilon_{j,k}^t\sim \mathcal{N}(0, \sigma)\) to avoid hitting the exact position, such as,
\begin{equation} \label{Eq:hs:fixed}
    h_s(\pmb{v}_j^t) \triangleq \bar{\pmb{v}}_{j}^{t} + \pmb{\epsilon}_j^t,
\end{equation}
where \(\sigma >0\) corresponds to the standard deviation of the component-wise noise applied to \(\bar{\pmb{v}}_{j}^{t}\).

The third class defines a directional displacement from the current state toward a given good-quality state \(\pmb{v}_\text{ref}^t\), as shown
\begin{equation}\label{Eq:hs:dir}
    h_s(\pmb{v}_j^{t}) \triangleq \pmb{v}_j^t + \alpha_d (\pmb{v}_\text{ref}^t - \pmb{v}_j^t) + \pmb{\epsilon}_j^t,
\end{equation}
since \(\pmb{v}_\text{ref}^t\) can be the neuromorphic space representation via \(\mathcal{T}\) of either the particular best position, the global best position or a combination of both.
Plus, \(\alpha_d >0\) controls the step size and \(\pmb{\epsilon}_j^t\) introduces stochastic variability.

Lastly, the fourth class of spike-triggered heuristics is based on mutation strategies from DE~\cite{Mohamed2021differential}, which corresponds to the next step of the generalisation path previously described. 
Therefore, based on the perturbation heuristic in \eqref{Eq:DE-mutation} for the DE/\emph{current-to-best}/1 mutation, we rewrite this and include another in terms of the state variables, such as
\begin{subequations} \label{Eq:DEvariants}
\begin{align} 
    h_s(\pmb{v}_j^{t}) &\triangleq \textstyle\pmb{v}_j^t + F (\tilde{\pmb{v}}^t_j - \pmb{v}_{j}^t) + F ( \bar{\pmb{v}}_{r_{1},j}^t - \bar{\pmb{v}}_{r_{2},j}^t ), 
        \label{suEq:de:cu2be}
        \\ 
    h_s(\pmb{v}_j^{t}) &\triangleq \textstyle{\pmb{v}}_j^t + F (\bar{\pmb{v}}_{r_1,j}^t - \pmb{v}_{j}^t) + F ( \bar{\pmb{v}}_{r_{2},j}^t - \bar{\pmb{v}}_{r_{3},j}^t ), 
        \label{suEq:de:cu2ra}
\end{align}
\end{subequations}
These correspond to the \emph{current-to-best}, \emph{rand-to-best}, and \emph{current-to-rand}, respectively. Regard that \( \bar{\pmb{v}}_{r_k}^t \) are distinct state vectors randomly sampled and transformed from \( \pmb{P}_{n,i}^t \), and \(\tilde{\pmb{v}}_j^t\) denote the global best state representation using \(\mathcal{T}\). 

\subsection{Bidirectional Mapping}

\noindent%
The mapping between candidate positions and the neuromorphic space is crucial for implementing the proposed framework. A bidirectional transformation \(\mathcal{T}\), as shown in \figurename~\ref{Fig:commutative-diagram}, is hence required to relate the neuromorphic and problem spaces.
Therefore, the \(\mathcal{T}\) between \( x_j \) and \(  \pmb{v}_j \) can be defined as 
\( 
    \mathfrak{V}\ni\pmb{v}_j \triangleq \mathcal{T}\{x_j\}
    \Longleftrightarrow
    \mathcal{T}^{-1}\{x_j\} \triangleq
    x_j\in\pmb{x}\in\mathfrak{X} 
\).
A simple implementation of \(\mathcal{T}\) corresponds to the linear mapping,
\begin{equation} \label{Eq:ImplementedTransformation}
    \pmb{v}_j=\mathcal{T}(x_j)\triangleq
    \begin{pmatrix}
        \alpha (x_j - x_{\text{ref},j}),
        &
        2r-(1-x_{\text{ref},j})
    \end{pmatrix}^\intercal,
\end{equation}
where \(\alpha \in\R \) is a gain factor, \( {x}_{\text{ref},j}\in\pmb{x}_\text{ref}\in\X \) is a reference point, and \(r\sim\mathcal{U}(0,1)\) is a random value.
Notice that the first component is directly related to the candidate position element, while the second component introduces a stochastic perturbation into the encoding process. 
This random value must be retained throughout execution to ensure deterministic reversibility and consistent internal dynamics.

On the one hand, \(\alpha\) controls the encoding resolution because this factor scales the displacement from \( \pmb{x}_{\text{ref}} \).
Thus, small \(\alpha\) values promote smoother dynamics, whereas larger ones amplify sensitivity to positional deviations. 
For that reason, choosing \(\alpha\) correctly, depending on the problem's nature, boosts a rich and detailed search.
The most straightforward setup can be \(\alpha=1\), which works for most cases but should be considered for fine-tuning tasks.
On the other hand, \( \pmb{x}_{\text{ref}} \) opens a colourful palette of search behaviours across the problem domain, including focused local refinement, exploratory displacement, and collective coordination through neighbourhood-informed centring.
Similar to the previous essential components, we consider distinct references, for example,
\begin{subequations}\label{Eq:xref}
\begin{align}
    \pmb{x}_\text{ref} &\triangleq w_1\pmb{p}+w_2\pmb{g},  
        \label{subEq:xref:pg} \\
    \pmb{x}_\text{ref} &\triangleq \textstyle w_1\pmb{p}+w_2\pmb{g}+\sum_{k=1}^m w_{k+2} \pmb{p}_k, 
    \label{subEq:xref:pgn} 
\end{align}
\end{subequations}
where \(\pmb{p}_k\) denotes the \(k\)\textsuperscript{th} neighbour's best-known position, \(m\) is the number of neighbours, and \(\{w_k\}_{k=1}^m\) is a set of normalised weights.
For simplicity, we adopt uniform weighting, ensuring even contribution and preserving neutrality across influence sources \cite{Sasaki2023osnn}. 
Nevertheless, weighted reference models based on additional information, such as fitness values, can be considered a promising research avenue.

\section{The Neuromorphic Optimisation Framework} \label{Sec:Approach}


\subsection{Overall Architecture of NeurOptimiser}

\noindent%
\figurename~\ref{Fig:nha} depicts the general structure of the \nha{short}, which requires essential parameters before starting the procedure, such as the problem \( (\mathfrak{X}, f, \text{min})\), the number of \nhu{long}s (\nhu{}s) \(n\), and the adjacency matrices \(\pmb{W}_s,\pmb{W}_x\in\mathbb{Z}_{2}^{n\times n}\) for the SNN and information neighbourhoods. 
The \nha{short} is composed of $n$ \nhu{}s that process and refine candidate solutions. Plus, it comprises a Tensor Contraction Layer, a Neighbour Manager, and a High-Level Selector, which coordinate the transmission and reception of essential encoded information, such as spikes, candidate positions, and fitness values.
These components are conceived as asynchronous processes using Intel's Lava NC framework, targeting the Loihi 2 chip \cite{lava2023}.

\begin{figure}[!htp]
    \centering
    \resizebox{0.75\columnwidth}{!}{
        \input{figures/nheuristic_algorithm}%
    }
    \caption{\nha{long} (\nha{short}) architecture composed of \(n\) \nhu{long}s (\nhu{short}s) units, a Tensor Contraction Layer, a Neighbourhood Manager, and a High-Level Selector.}
    \label{Fig:nha}
\end{figure}

As an overview, the \nha{}'s process is described as follows. First, \nhu{}s produce their initial candidate solution, while other processes wait until they have information in their respective input ports.
After several steps, we can summarise the asynchronous behaviour of each process, such as:
\begin{itemize}
    \item A \nhu{} reads all its communication channels to process the information independently and then updates its current position and spiking signal.
    \item The Tensor Contraction Layer receives every spiking signal from \nhu{}s, processes them in a matrix structure, and then propagates the resulting activations to all \nhu{}s.
    \item The Neighbour Manager reads the particular positions and their fitness from \nhu{}s, processes them, and shares the information grouped by neighbourhoods. 
    \item The High-Level Selector receives the same information that the Neighbour Manager receives and selects the best candidate position. It then broadcasts this to all \nhu{}s.
\end{itemize}
Each process is detailed in this section's remaining paragraphs. 

\subsection{Internal Structure of a Neuromorphic Heuristic Unit}\label{Sec:Prop:NHU}

\begin{figure}[!htp]
    \centering
    \resizebox{0.6\columnwidth}{!}{
        \def\roti{-90}
\def\rotii{-90} 

\begin{tikzpicture}[rotate=90, node distance=1.5cm, scale=0.8, every node/.style={transform shape}, every shadow/.style={opacity=.8}]

\newcommand{\portshifttwo}{0.5cm} %
\newcommand{\portshiftthree}{0.75cm} %
\newcommand{\rightd}{1.5cm} %

\node (pnh) [perturbation] {\rotatebox{\rotii}{\parbox{1.5cm}{\centering Spiking Core}}};
\node (ssh) [spiking, above=of pnh, yshift=-0.2cm] {\rotatebox{\rotii}{\parbox{2cm}{\centering Spiking Handler}}};
\node (snh) [selection, below=of pnh, yshift=-0.2cm] {\rotatebox{\rotii}{Selector}};
\node (prh) [receiver, right=of pnh, xshift=0.0cm] {\rotatebox{\rotii}{\parbox{2cm}{\centering Receiver}}};
\node (psh) [sender, right=of snh, xshift=0.0cm] {\rotatebox{\rotii}{\parbox{2cm}{\centering Sender}}};

\node (ssh_s_out) [outPort] at ([xshift=0cm, yshift=0.4cm] ssh.west) {\rotatebox{\rotii}{$\pmb{S}$}};
\node (ssh_a_in) [inPort] at ([xshift=0cm, yshift=-0.4cm] ssh.west) {\rotatebox{\rotii}{$\pmb{A}$}};
\node (ssh_s_in) [inPort] at ([xshift=-0.5cm, xshift=0] ssh.south) {\rotatebox{\rotii}{$\pmb{s}$}};
\node (ssh_a_out) [outPort] at ([xshift=0.5cm, xshift=0] ssh.south) {\rotatebox{\rotii}{$\pmb{a}$}};

\node (pnh_g_in) [inPort] at ([yshift=\portshifttwo] pnh.west) {\rotatebox{\rotii}{$\pmb{g}$}};
\node (pnh_fg_in) [inPort] at ([yshift=-\portshifttwo] pnh.west) {\rotatebox{\rotii}{${f}_{{g}}$}};

\node (pnh_s_in) [inPort] at ([xshift=0.5cm] pnh.north) {\rotatebox{\rotii}{$\pmb{a}$}};
\node (pnh_s_out) [outPort] at ([xshift=-0.5cm] pnh.north) {\rotatebox{\rotii}{$\pmb{s}$}};

\node (pnh_x_out) [outPort] at ([yshift=0, xshift=-0.8cm] pnh.south) {\rotatebox{\rotii}{$\pmb{x}$}};
\node (pnh_p_in) [inPort] at ([yshift=0cm, xshift=0] pnh.south) {\rotatebox{\rotii}{$\pmb{p}$}};
\node (pnh_fp_in) [inPort] at ([yshift=0, xshift=0.8cm] pnh.south) {\rotatebox{\rotii}{$f_{\pmb{p}}$}};

\node (pnh_xn_in) [inPort] at ([xshift=0cm, yshift=\portshifttwo] pnh.east) {\rotatebox{\rotii}{$\pmb{P}_{n}$}};
\node (pnh_fxn_in) [inPort] at ([xshift=0cm, yshift=-\portshifttwo] pnh.east) {\rotatebox{\rotii}{$\pmb{f}_{{p}_{n}}$}};

\node (snh_x_in) [inPort] at ([xshift=-0.8cm] snh.north) {\rotatebox{\rotii}{$\pmb{x}$}};
\node (snh_p_out) [outPort] at ([yshift=\portshifttwo] snh.east) {\rotatebox{\rotii}{$\pmb{p}$}};
\node (snh_fp_out) [outPort] at ([yshift=-\portshifttwo] snh.east) {\rotatebox{\rotii}{$f_{\pmb{p}}$}};
\node (function) [var_funct] at (snh.west) {\rotatebox{\rotii}{$f(\pmb{x})$}};

\node (psh_p_in) [inPort] at ([xshift=0cm, yshift=\portshifttwo] psh.west) {\rotatebox{\rotii}{$\pmb{p}$}};
\node (psh_fp_in) [inPort] at ([xshift=0cm, yshift=-\portshifttwo] psh.west) {\rotatebox{\rotii}{$f_{\pmb{p}}$}};
\node (psh_p_out) [outPort] at ([yshift=0cm, xshift=-0.6cm] psh.south) {\rotatebox{\rotii}{$\pmb{P}$}};
\node (psh_fp_out) [outPort] at ([yshift=0cm, xshift=0.6cm] psh.south) {\rotatebox{\rotii}{$\pmb{f}_{p}$}};

\node (prh_p_in) [inPort] at ([yshift=0cm, xshift=-0.6cm] prh.north) {\rotatebox{\rotii}{$\mathcal{P}_{n}$}};
\node (prh_fp_in) [inPort] at ([yshift=0cm, xshift=0.6cm] prh.north) {\rotatebox{\rotii}{$\pmb{F}_{n}$}};
\node (prh_p_out) [outPort] at ([xshift=0cm, yshift=\portshifttwo] prh.west) {\rotatebox{\rotii}{$\pmb{P}_{n}$}};
\node (prh_fp_out) [outPort] at ([xshift=0cm, yshift=-\portshifttwo] prh.west) {\rotatebox{\rotii}{$\pmb{f}_{{p}_{n}}$}};

\node (ext_s_out) [outPort] at ([yshift=0cm, xshift=-\rightd] ssh_s_out.west) {\rotatebox{\rotii}{$\pmb{S}$}};
\node (ext_a_in) [inPort] at ([yshift=0cm, xshift=-\rightd] ssh_a_in.west) {\rotatebox{\rotii}{$\pmb{A}$}};
\node (ext_g_in) [inPort] at ([xshift=-\rightd] pnh_g_in.west) {\rotatebox{\rotii}{$\pmb{g}$}};
\node (ext_fg_in) [inPort] at ([xshift=-\rightd] pnh_fg_in.west) {\rotatebox{\rotii}{$f_{{g}}$}};
\node (ext_pn_in) [inPort] at ([yshift=0.2cm] prh_p_in.north |- ssh.north) {\rotatebox{\rotii}{$\mathcal{P}_{n}$}};
\node (ext_fpn_in) [inPort] at ([yshift=0.2cm] prh_fp_in.north |- ssh.north) {\rotatebox{\rotii}{$\pmb{F}_{n}$}};

\node (ext_p_out) [outPort] at ([yshift=-2cm, xshift=0cm] psh_p_out.south) {\rotatebox{\rotii}{$\pmb{P}$}};
\node (ext_fp_out) [outPort] at ([yshift=-2cm, xshift=0cm] psh_fp_out.south) {\rotatebox{\rotii}{$\pmb{f}_{{p}}$}};
\node (function_ext) [var_funct, xshift=0.0cm] at (ext_fg_in |- function){\rotatebox{\rotii}{$f(\pmb{x})$}};

\node (int_var) [var_index, below=of function_ext.west, anchor=west, yshift=-1cm] {\rotatebox{0}{\texttt{agent\_id}}};
\node (ssh_int_var) [var_index] at ([yshift=0, xshift=0cm] ssh.east) {\rotatebox{\rotii}{$i$}};
\node (prh_int_var) [var_index] at ([yshift=0, xshift=0cm] prh.south) {\rotatebox{\rotii}{$i$}};
\node (psh_int_var) [var_index] at ([yshift=0, xshift=0cm] psh.north) {\rotatebox{\rotii}{$i$}};

\draw [arrow, text=textdark] (function_ext.east) 
	-- node[midway, below, xshift=-0.0cm, yshift=-0.4cm, text=textdark] {{$f:\mathbb{R}^d\mapsto\mathbb{R}$}}
	(function.west);

\draw [arrow] (pnh_s_out.north) 
	-- node[midway, left, yshift=0cm, text=textdark] {$\pmb{s}\in\mathbb{Z}_2^{d}$} 
	(ssh_s_in.south);
\draw [arrow] (ssh_a_out.south) 
	-- node[midway, right, yshift=0cm, text=textdark] {$\pmb{a}\in\mathbb{Z}_2^{d}$} 
	(pnh_s_in.north);

\draw [arrow] (pnh_x_out.south) 
	-- node[midway, left, yshift=0cm, text=textdark] {$\pmb{x}\in\mathbb{R}^{d}$}  
	(snh_x_in.north);

\draw [arrow] (snh_p_out.east) 
	-| coordinate (p_mid) ++(.2cm, 1.3cm) 	
	-| node[pos=0.3, below, yshift=0.06cm, text=textdark] {$\pmb{p}\in\mathbb{R}^{d}$}
	(pnh_p_in.south);

\fill[darkGray] (p_mid) circle (2pt);

\draw [arrow] (snh_fp_out.east) 
	-| coordinate (fp_mid) ++(.35cm, 2.4cm)
	-| node[pos=0.25, above, xshift=0cm, text=textdark] {$f_{\pmb{p}}\in\mathbb{R}$}
	(pnh_fp_in.south);

\fill[darkGray] (fp_mid) circle (2pt);

\draw [arrow] (p_mid)
	|- 
	(psh_p_in.west);
	
\draw [arrow] (fp_mid) 
	|- 
	(psh_fp_in.west);

\draw [arrow] (prh_p_out.west) 
	-- node[midway, above, text=textdark, xshift=0.2cm] {\rotatebox{\roti}{{$\pmb{P}_n\in\mathbb{R}^{m\times d}$}}}
    (pnh_xn_in.east);
\draw [arrow] (prh_fp_out.west) 
	-- node[midway, below, text=textdark, xshift=0.2cm] {\rotatebox{\roti}{{$\pmb{f}_{p_n}\in\mathbb{R}^{m}$}}}
    (pnh_fxn_in.east);

\draw [arrow] (ext_a_in.east) 
	-- 
	node[midway, below, yshift=-0.2cm, xshift=-0.1cm, text=textdark] {$\pmb{A}\in\mathbb{Z}_2^{n\times d}$}
	(ssh_a_in.west);
\draw [arrow] (ssh_s_out.west) 
	--
	node[midway, above, yshift=0.2cm, text=textdark] {$\pmb{S}\in\mathbb{Z}_2^{n\times d}$} 
	(ext_s_out.east);
	
\draw [arrow] (ext_g_in.east) 
	--
	node[midway, above, yshift=0.3cm, text=textdark] {$\pmb{g}\in\mathbb{R}^{d}$} 
	(pnh_g_in.west);
	
\draw [arrow] (ext_fg_in.east) 
	--
	node[midway, below, yshift=-0.3cm, text=textdark] {$f_{{g}}\in\mathbb{R}$} 
	(pnh_fg_in.west);
	
\draw [arrow] (ext_pn_in.south) 
	-- 
	node[midway, left, text=textdark] {\rotatebox{\roti}{$\mathcal{P}_n\in\mathbb{R}^{m\times d \times n}$}}
	(prh_p_in.north);
\draw [arrow] (ext_fpn_in.south) 
	--
	node[midway, right, text=textdark] {\rotatebox{\roti}{$\pmb{F}_{n}\in\mathbb{R}^{m\times n}$}} 
	(prh_fp_in.north);
\draw [arrow] (psh_p_out.south) 
	-- 
	node[midway, left, text=textdark] {\rotatebox{\roti}{{$\pmb{P}\in\mathbb{R}^{n\times d}$}}}
	(ext_p_out.north);
\draw [arrow] (psh_fp_out.south) 
	-- 
	node[midway, right, text=textdark] {\rotatebox{\roti}{{$\pmb{f}_{{p}}\in\mathbb{R}^{n}$}}}
	(ext_fp_out.north);


\begin{pgfonlayer}{background}
\draw [thick, dashed] (int_var.east) 
	-| ++(4.2cm,4.65cm) coordinate (agent_mid); 
\draw [arrow, dashed] (agent_mid) -| coordinate (agent_mid_aux)
	(psh_int_var.north);
\draw [arrow, dashed] (agent_mid_aux) -- (prh_int_var.south);
\draw [arrow, dashed] (agent_mid) |- (ssh_int_var.east);
\fill[darkGray] (agent_mid) circle (2pt);
\fill[darkGray] (agent_mid_aux) circle (2pt);

\draw[thick, rounded corners=5pt, color=gray!50] 
    ([xshift=0.0cm, yshift=0.5cm] ext_s_out.north) 
    rectangle
    ([xshift=0.5cm, yshift=0.25cm] ext_fp_out.south east);
\end{pgfonlayer}

\end{tikzpicture}%
    }
    \caption{%
    \nhu{long} (\nhu{short}) architecture comprising two main processes, Spiking Core and Selector; and three auxiliary processes, Spiking Handler, Receiver, and Sender, for receiving and sending information. 
    }\label{Fig:nhu}
\end{figure}
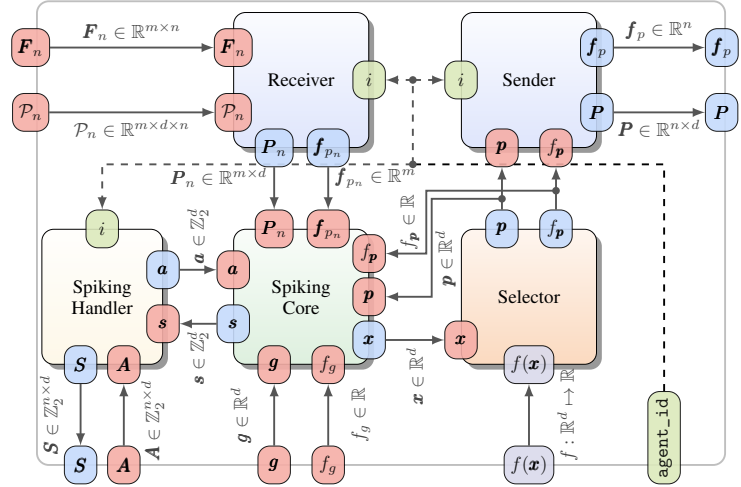

\noindent%
\nhu{}s are designed as self-contained modules responsible for encoding, evaluating, and updating candidate solutions during the optimisation process. 
\figurename~\ref{Fig:nhu} shows the \nhu{}'s internal architecture comprising two main processes, \ie Spiking Core and Selector. Besides, it also utilises three information management components serving as I/O peripherals, \ie a Spiking Handler, a Receiver, and a Sender. 
These internal processes are briefly described as follows:
\begin{itemize}
    \item The Spiking Core encodes and modifies candidate solutions using \(d\) spiking neurons. 
    \item The Selector evaluates a given position using the objective function \(f\), and maintains the local solution \((\pmb{p}, f_p)\) depending on the selection criterion.
    \item The Spiking Handler manages spiking communication by encoding local spike events into the global matrix \(\pmb{S}\) and decoding activation patterns \(\pmb{A}\).
    \item The Sender broadcasts the local information to the network by encoding \((p, f_p)\) into shared structures \((\pmb{P}, \pmb{f}_p)\).
    \item The Receiver gathers information from neighbours and extracts relevant data \((\pmb{P}_n, \pmb{f}_n)\) for local usage.
\end{itemize}
Moreover, the position index \(i\) (encoded in the \texttt{agent\_id}) is only shared with those processes interacting with external information. Likewise, the objective function is only accessed by the Selector. The subsequent paragraphs of this subsection detail these internal processes.

\noindent%

\subsubsection{Spiking Core}\label{Sec:SpikingCore}

\noindent%
This process is the central component of the \nhu{}. It comprises \(d\) spiking neurons that encode \(\pmb{x}\) and define the substance of our approach. 
To implement this process, the key components required include the transformation \(\mathcal{T}\), the spiking condition function \(\Phi\), and the dynamic updating \(h_d\) and spike-triggered \(h_s\) heuristics.
\algorithmname~\ref{Alg:SpikingCore} details this asynchronous process, which can be summarised in three main blocks. 
The first one (lines 1-3) corresponds to the initialisation for preparing and sending the first candidate solution. The second (lines 4-6) stands for waiting and reading the relevant information from ports.
Once information is available, the third block (7-14) performs the heuristic-based neuromorphic procedure detailed in \sectionname~\ref{Sec:Foundations}.

\begin{algorithm}[!htb]
\caption{Spiking Core asynchronous process.}
\label{Alg:SpikingCore}
\begin{algorithmic}[1]
\InPorts{%
    \( \pmb{a}, \pmb{p}, \pmb{g}, \pmb{P}_n, f_g,  f_p, \pmb{f}_{p_n}\) 
} \Comment{Activation, positions, and fitness}
\OutPorts{%
    \( \pmb{s} , \pmb{x}  \) 
} \Comment{Spiking signal and candidate solution}
    \State{\( \pmb{x}, \pmb{p},\pmb{g},f_p,f_g,\pmb{P}_n,\pmb{f}_{p_n}\gets\Call{Initialise}{ }\)}
    \State{\(\pmb{a}, \pmb{s}\gets\Call{InitialiseWith}{\pmb{0}_d}\)} 
    \State{\Call{Send}{\(\pmb{s}, \pmb{x}\)}}
\While{\Call{RunCondition}{ }}
    \Comment{Asynchronous execution}
\State{\Call{WaitForInputs}{ }}
    \Comment{Asynchronous event handling}
\State{\( \pmb{a}, \pmb{g}, \pmb{p}, f_g, f_p, \pmb{P}_n, \pmb{f}_{p_n} \gets \) \Call{Receive}{ }}  
\For{\textbf{each} \( j \in \N_d \)}
    \State \( \pmb{v}_j \gets \mathcal{T}(x_j) \) \Comment{Transform position to state}
    \State \( s_j \gets \varphi_s (\pmb{v}_j ) \) \Comment{Determine self-spiking condition} 
    
    \If{\( \Phi(\pmb{v}_j,\dots) \)} \Comment{Spiking condition function}
        \State \( \pmb{v}_j \gets h_s(\pmb{v}_j) \) \Comment{Apply spike-transition rule}
    \Else
        \State \( \pmb{v}_j \gets h_d(\pmb{v}_j) \) \Comment{Apply dynamical rule}
    \EndIf

    \State \( x_j \gets \mathcal{T}^{-1}(\pmb{v}_j) \) \Comment{Transform state back to position}
\EndFor
\State{%
    \Call{Send}{\( \pmb{s}, \pmb{x} \)}
}
\EndWhile
\end{algorithmic}
\end{algorithm}

\subsubsection{Selector}\label{Sec:Selector}

\noindent%
This process evaluates a candidate solution \(\pmb{x}\) using \( f \) and updates the local reference \(\pmb{p}, f_p\) throughout the evolutionary process. \algorithmname~\ref{Alg:Selector} presents its internal process, although simple, it implements a key mechanism in the behaviour of the \nhu{}, mainly because the objective function evaluation may involve high computational cost, delayed response, black-box conditions, and surrogate implementations in practical applications \cite{Henkes2024Approx}. 
This process employs a greedy selection strategy, which is crucial for ensuring the propagation of high-quality solutions throughout the search process.

\begin{algorithm}[!htb]
\caption{Selector asynchronous process.}
\label{Alg:Selector}
\begin{algorithmic}[1]
\InPorts{%
    \( \pmb{x} \) 
}
    \Comment{Candidate position}
\OutPorts{%
    \( \pmb{p}, f_p \) 
    \Comment{Best known position and its fitness}
}
\While{\Call{RunCondition}{ }}
    \Comment{Asynchronous execution}
    \State{\Call{WaitForInputs}{ }}
        \Comment{Asynchronous event handling}
    \State{\(\pmb{x}\gets\)\Call{Receive}{ }}
    \State{\( f_x \gets f(\pmb{x}) \)} 
        \Comment{Evaluate the candidate solution}

    \If{\( f_x < f_p \) \textbf{or} \(\lnot\texttt{is\_init}\)} 
        \Comment{Greedy selection}
        \State{\( \pmb{p} \gets \pmb{x},\, f_p \gets f_x \)} 
        \State{\( \texttt{is\_init} \gets \texttt{True} \)} 
    \EndIf
    \State{\Call{Send}{\(\pmb{p}, f_p\)}} 
\EndWhile
\end{algorithmic}
\end{algorithm}

\subsubsection{Spiking Handler}\label{Sec:SpikingHandler}

\noindent%
This process plays a critical role as a communication interface, linking each \nhu{} to the collective spike-driven dynamics of the SNN, despite spiking neurons being unaware of this global structure from a local perspective. This biologically inspired process emulates the presynaptic function of neurons by receiving information from neighbours and interpreting it as activation events. 
\algorithmname~\ref{Alg:SpikingHandler} details the procedure of this component that, contrary to those presented above, additional information about the \nhu{} identity (index \(i\)) is needed for proper encoding and decoding.
Specifically, the handler receives a spike vector \( \pmb{s} \) from the core and an activation matrix \( \pmb{A} \) from the SNN, updates the corresponding row of the global spike matrix \( \pmb{S} \), and extracts the relevant activation signals \( \pmb{a}\) for the unit. 

\begin{algorithm}[!htb]
\caption{Spiking Handler asynchronous process.}
\label{Alg:SpikingHandler}
\begin{algorithmic}[1]
\InPorts{%
    \( \pmb{A}, \pmb{s} \)
}
    \Comment{Activation matrix and local spike vector}
\OutPorts{%
    \( \pmb{S}, \pmb{a} \)
}
    \Comment{Spike matrix and activation vector}
\State \( \pmb{s} \gets \pmb{0}_d, \, \pmb{A} \gets \pmb{0}_{n \times d} \)
    \Comment{Zero initialisation}
\While{\Call{RunCondition}{ }}
    \Comment{Asynchronous execution}
    \State \Call{WaitForInputs}{ }
        \Comment{Handle asynchronous inputs}
    \State \( \pmb{s}, \pmb{A} \gets \Call{Receive}{ } \)

    \State \( (\pmb{S})_{i,:} \gets \pmb{s} \)
        \Comment{Encode spikes}
    \State \( \pmb{a} \gets (\pmb{A})_{i,:} \)
        \Comment{Decode activations}
    \State \Call{Send}{\( \pmb{S}, \pmb{a} \)}
\EndWhile
\end{algorithmic}
\end{algorithm}

\subsubsection{Sender and Receiver}\label{Sec:PositionSender}

\noindent%
These two processes perform the encoding and decoding tasks, similar to the Spike Handler process. 
However, we split them into dedicated processes because the information they handle is far more complex than binary arrays. 
On the one hand, the Sender broadcasts the local \( \pmb{p} \) and \( f_p\) encoded into the $i$\textsuperscript{th} row of the position matrix $\pmb{P} \in \R^{n \times d}$ and the fitness vector $\pmb{f}_p \in \R^{n}$, respectively.
On the other hand, the Receiver extracts the \( i \)\textsuperscript{th} neighbourhood's positions \( \pmb{P} \) and fitness \(\pmb{f}_{p_n}\) from \( \mathcal{P}_n \) and \( \pmb{F}_{n} \) shared externally.
This data is communicated with the rest of \nhu{}s through the position and fitness communication channels and depends on the floating point precision defined for the implementation.
Lastly, \algorithmname~\ref{Alg:PositionSender} and \ref{Alg:PositionReceiver} detail the Sender and Receiver.

\begin{algorithm}[!htb]
\caption{Sender asynchronous process.}
\label{Alg:PositionSender}
\begin{algorithmic}[1]
\InPorts{%
    \( \pmb{p}, f_p \)
}
    \Comment{Candidate position and fitness}
\OutPorts{%
    \( \pmb{P}, \pmb{f}_p \)
}
    \Comment{Position matrix and fitness vector}
\State \( \pmb{P} \gets \pmb{0}_{n\times d},\, \pmb{f}_p \gets \pmb{0}_n \)
    \Comment{Zero initialisation}
\While{\Call{RunCondition}{ }}
    \Comment{Asynchronous execution}
    \State \Call{WaitForInputs}{ }
        \Comment{Handle asynchronous inputs}
    \State \( \pmb{p}, f_p \gets \Call{Receive}{ } \)
    \State {\( (\pmb{P})_{i,:} \gets \pmb{p} \), 
        \( (\pmb{f}_{p})_{i} \gets f_p \)
    }
        \Comment{Encode position and fitness}
    \State \Call{Send}{\( \pmb{P}, \pmb{f}_p \)}
\EndWhile
\end{algorithmic}
\end{algorithm}

\begin{algorithm}[!htb]
\caption{Receiver asynchronous process.}
\label{Alg:PositionReceiver}
\begin{algorithmic}[1]
\InPorts{%
    \( \mathcal{P}_n, \pmb{F}_{n}\)
}
    \Comment{Neighbourhood's positions and fitness}
\OutPorts{%
    \( \pmb{P}_n, \pmb{f}_{p_n} \)
}
    \Comment{Positions and fitness of the \(i\)\textsuperscript{th} neighbourhood}
\State \( \pmb{P}_n \gets \pmb{0}_{m\times d},\, \pmb{f}_{p_n} \gets \pmb{0}_m \)
    \Comment{Zero initialisation}
\While{\Call{RunCondition}{ }}
    \Comment{Asynchronous execution}
    \State \Call{WaitForInputs}{ }
        \Comment{Handle asynchronous inputs}
    \State \( \mathcal{P}_n, \pmb{F}_{n} \gets \Call{Receive}{ } \)
    \State {\( \pmb{P}_n \gets (\mathcal{P}_n)_{:, :, i} \), 
        \( \pmb{f}_{p_n} \gets (\pmb{F}_{n})_{:, i} \)
    }
        \Comment{Decode information}
    \State \Call{Send}{\( \pmb{P}_n, \pmb{f}_{p_n} \)}
\EndWhile
\end{algorithmic}
\end{algorithm}

\subsection{Coordination Processes}

\noindent%
This section completes the \nha{}'s architectural description by detailing the coordination processes responsible for structuring information flow among \nhu{}s. 

\subsubsection{High-Level Selector}\label{Sec:HighLevelSelector}

\noindent%
This crucial coordination mechanism identifies and propagates the global best candidate \(\pmb{g}\) solution and its fitness \(f_p\) across all \nhu{}s. Compared to the (low-level) Selector, the substantial difference between them lies in the number of solutions being analysed. Apart from that, both implement a greedy selection that, unlike the selector processes, must be preserved in the high-level selection, at least for single-objective optimisation problems. \algorithmname~\ref{Alg:HighLevelSelector} details the High-Level Selector process.

\begin{algorithm}[!htb]
\caption{High-Level Selector asynchronous process.}
\label{Alg:HighLevelSelector}
\begin{algorithmic}[1]
\InPorts{%
    \( \pmb{P}, \pmb{f}_p \) 
        \Comment{Candidate positions and their fitness values}
}
\OutPorts{%
    \( \pmb{g}, f_g \) 
        \Comment{Best candidate position and its fitness}
}
\While{\Call{RunCondition}{ }}
    \Comment{Asynchronous execution}
    \State{\Call{WaitForInputs}{ }}
        \Comment{Synchronise incoming candidates}
    \State{\( \pmb{P}, \pmb{f}_p \gets \Call{Receive}{ } \)}
    \State{\( i_* \gets {\operatorname{argmin}}_{\forall i\in \N_n} \{ (\pmb{f}_p)_i \} \)} 
    \If{\( (\pmb{f}_p)_{i_*}< f_g \) \textbf{or} \(\lnot\texttt{is\_init} \)}
            \Comment{Greedy selection}
        \State{\( \pmb{g} \gets (\pmb{P})_{i_*},\, f_g \gets (\pmb{f}_p)_{i_*} \)} 
            \Comment{Update the best}
        \State{\( \texttt{is\_init} \gets \texttt{True} \)} 
    \EndIf
    \State{\Call{Send}{\(\pmb{g}, f_g\)}}
\EndWhile
\end{algorithmic}
\end{algorithm}

\subsubsection{Tensor Contraction Layer}\label{Sec:TensorContractionLayer}

\noindent%
This component transforms local spike emissions \(\pmb{S}\) into distributed activation \(\pmb{A}\) patterns across the SNN. This process emulates synaptic propagation by mapping the spike outputs of each \nhu{} onto its predefined neighbourhood, which implements a structured interaction aligned with the spiking topology. Thus, \(\pmb{A}\) is determined via the logical contraction operation adapted from \cite{Kossaifi2017tensor}, such as
\begin{equation}\label{Eq:logicalcontraction}
    (A)_{i,j} = \textstyle\bigvee_{k=1}^{n} \left((\mathcal{W})_{i,k,j}  \wedge (\pmb{S})_{k,j} \right), \, \forall i \in \N_n,\, j \in \N_d,
\end{equation}
where \(\mathcal{W} \in\Z_2^{n\times n\times d}\) is the weight tensor, which can be built simply by filling all the \( d \) slices using the adjacency matrix \(\pmb{W}_s\).
After that, the process broadcasts the spiking influence without explicit synchronisation, which permits each \nhu{} to read binary activations from its connected peers. 
\algorithmname~\ref{Alg:TensorContractionLayer} formalises the Tensor Contraction Layer process.

\begin{algorithm}[!htb]
\caption{Tensor Contraction Layer asynchronous process.}
\label{Alg:TensorContractionLayer}
\begin{algorithmic}[1]
\InPorts{%
    \( \pmb{S} \)
        \Comment{Spike matrix}
}
\OutPorts{%
    \( \pmb{A} \)
        \Comment{Activation matrix}
}
\State \( (\mathcal{W})_{:,:,j} \gets \pmb{W}_s,\,\forall j\in\N_d \)
    \Comment{Initialise contraction weights}
\While{\Call{RunCondition}{ }}
    \Comment{Asynchronous execution}
    \State \Call{WaitForInputs}{ }
        \Comment{Handle asynchronous inputs}
    \State \( \pmb{S} \gets \Call{Receive}{ } \)
    \State \( \pmb{A} \gets \Call{Contract}{\mathcal{W},\pmb{S}} \)
        \Comment{Compute activation using \eqref{Eq:logicalcontraction}}    
    \State \Call{Send}{\( \pmb{A} \)}
\EndWhile
\end{algorithmic}
\end{algorithm}

\subsubsection{Neighbour Manager}\label{Sec:NeighbourhoodManager}

\noindent%
This process systematically aggregates positional information from neighbouring candidate solutions, which permits the structuring of local interactions within \nha{short}.
\algorithmname~\ref{Alg:NeighbourhoodManager} specifies the procedure, which receives the particular best positions \( \pmb{P} \) and their fitness \( \pmb{f}_p \), merged in the communication channel from those messages sent by \nhu{}s, and processes them with an adjacency matrix \( \pmb{W}_x \) describing the connections between \nhu{}s. 
This matrix differs from \( \pmb{W}_s\), because it can be a real-valued matrix representing a weighted topology graph. However, we use a fixed `binary' version of \( \pmb{W}_x \) for simplicity.

\begin{algorithm}[!htb]
\caption{Neighbour Manager asynchronous process.}
\label{Alg:NeighbourhoodManager}
\begin{algorithmic}[1]
\InPorts{%
    \( \pmb{P}, \pmb{f}_p \)
}
    \Comment{Positions matrix and fitness vector}
\OutPorts{%
    \( \mathcal{P}_n, \pmb{F}_{n} \)
}
    \Comment{Neighbourhood positions and fitness values}
\State \( \pmb{N}_i \triangleq \left\{ j \in \N_n \mid (\pmb{W}_x)_{i,j} = 1 \right\},\,\forall i\in\N_n \)
    \Comment{Get neighbours}
\State \( m \gets \max_{\forall i\in\N_n} \left\{ \#\pmb{N}_i\right\} \)
    \Comment{Max. number of neighbours}
\State \( \mathcal{P}_n \gets \pmb{0}_{m\times d\times n},\, \pmb{F}_n\gets\pmb{0}_{m\times n} \)
    \Comment{Initialise arrays}
\While{\Call{RunCondition}{ }}
    \Comment{Asynchronous execution}
    \State \Call{WaitForInputs}{ }
        \Comment{Wait for position and fitness matrices}
    \State \( \pmb{P}, \pmb{f}_{p} \gets \Call{Receive}{ } \)
    \For{\textbf{each} \( i\in\N_n \)}
        \State \(  (\mathcal{P}_n)_{\ell,:,i} \gets (\pmb{P})_{k_\ell,:},\,\forall k_\ell\in\pmb{N}_i,\,\ell\in\N_{m} \)
            \Comment{Get positions}
        \State \( (\pmb{F}_n)_{\ell,i} \gets (\pmb{f}_p)_{k_\ell},\,\forall k_\ell\in\pmb{N}_i,\,\ell\in\N_{m} \)   
            \Comment{Get fitness}
    \EndFor
    \State \Call{Send}{\( \mathcal{P}_n, \pmb{f}_{p_n} \)}
\EndWhile
\end{algorithmic}
\end{algorithm}

\section{Methodology}\label{Sec:Methodology}

\noindent%
This work followed a methodology accordingly to the novelty of the proposed approach.
We designed and implemented the \nha{}, publicly available at %
\hyperlink{https://neuroptimiser.github.io}{neuroptimiser.github.io}. 
Plus, all scripts, configuration files, and experimental results are freely available in \cite{Cruz2025neurodataset}.
The framework was developed based on Intel's Lava NC framework~\cite{lava2023} for CPU simulation and neuromorphic deployment targeting the Loihi 2 chip. 
All the experiments were conducted in Python v3.10 on a 48-core AMD EPYC 7642 CPU with 512~GB RAM, under Grid'5000 using Debian 5.10 x86-64 GNU/Linux. Grid'5000 is a testbed supported by a scientific interest group hosted by Inria, CNRS, RENATER, and several Universities, as well as other organisations (visit \hyperlink{https://www.grid5000.fr}{grid5000.fr} for further information).

We adopted a three-fold experimental approach using the noiseless BBOB test suite from COCO and IOH~\cite{bbob2019, hansen2021coco, IOHexperimenter}, comprising 24 problems, six dimensionalities (2, 3, 5, 10, 20, and 40), and 15 instances per dimension. These are grouped as separable (\texttt{separ}), \textbf{f1}--\textbf{f5}; unimodal functions with low (\texttt{lcond}) and high (\texttt{hcond}) conditioning, \textbf{f6}--\textbf{f9} and \textbf{f10}--\textbf{f14}; and multimodal functions with either adequate (\texttt{multi}) or weak (\texttt{mult2}) global structure, \textbf{f15}--\textbf{f19} and \textbf{f20}--\textbf{f24}.

The first experiment characterised the principal behaviours of \nha{} implementations using the BBOB suite via IOH, with two dimensions, 30 \nhu{}s, and \(1000\times d\) steps. The dynamic heuristic \(h_d\) was instantiated with the Linear and Izhikevich neuron models in \eqref{Eq:2dsystem-diff} and \eqref{Eq:Izhikevich-NUM-heu} with random coefficients. 
All cases used 30 \nhu{}s and the 4\textsuperscript{th}-order Runge-Kutta method with \(\Delta t=0.01\). 
The spike-triggered heuristic \(h_s\) was set as Fixed~\eqref{Eq:hs:fixed} and DE/\emph{current-to-rand}/1~\eqref{suEq:de:cu2ra}. Moreover, \(\vartheta_j\) employed \eqref{subEq:vt:pg}; \(\pmb{W}_s\) represented a bidirectional ring; \(\pmb{W}_x\) was set random with \(m = 10\) neighbours; \(\pmb{x}_\text{ref}\) was defined as in \eqref{subEq:xref:pg} with weights equal to 0.5; and \(\varphi_s\) used the \(\ell_2\)-norm based on \eqref{subEq:vp:lp}. 

Subsequently, we carried out a confirmatory evaluation using the BBOB test suite from COCO. We selected the two best-performing \nha{short} variants from the preliminary analysis. An additional hybrid variant was created by blending these two models in a 50/50 proportion, producing a heterogeneous distribution of spiking models. The performance of these three \nha{}s was systematically compared against baseline algorithm results from the literature.

The third experiment assessed the computational efficiency and scalability of the framework. We measured average runtime per step and \nhu{} as a function of \(n\) and \(d\) using the same environment. In such a case, a hybrid-like \nha{} was implemented with the most computationally demanding setting, \ie, fully connected \(\pmb{W}_s\) and \(\pmb{W}_x\), \(\pmb{x}_\text{ref}\) as in \eqref{subEq:xref:pgn}, and randomly assigned neuron models. Seven instances per function were considered, evaluated across 2, 10, 20, and 40 dimensions, using configurations with 30, 60, and 90 \nhu{}s.

\section{Results and Discussion}\label{Sec:Results}


\noindent
The initial set of experiments examines the internal behaviour of several \nha{} implementations. It isolates the effect of core dynamics and spike-triggered heuristic selection on population trajectories, state evolution, and spiking activity. 
All configurations were evaluated on the BBOB suite with one representative two-dimensional function per category.

\begin{figure*}[!htb]\centering
    \def\leRaiser{9mm}
\begin{tabular}{@{}r*{4}{@{}c}@{}}
    \raisebox{\leRaiser}{\footnotesize\textbf{f1}} &
        \includegraphics[width=0.245\linewidth]{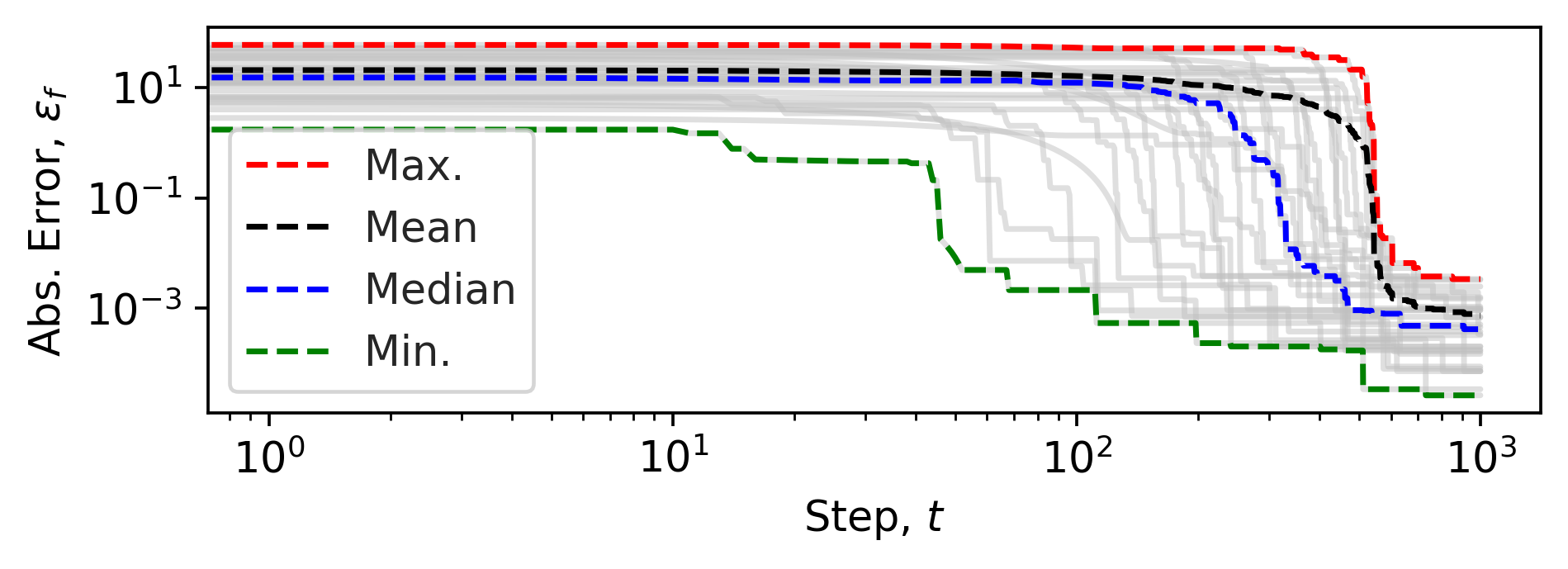} &
        \includegraphics[width=0.245\linewidth]{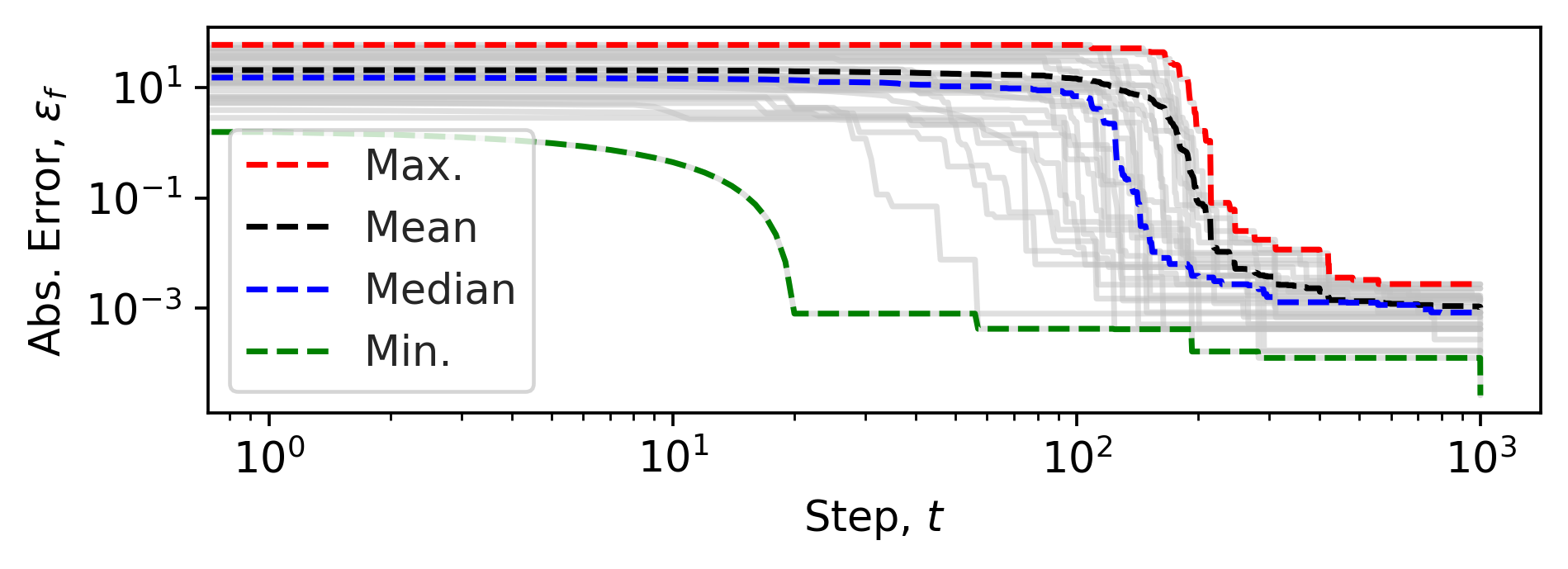} &
        \includegraphics[width=0.245\linewidth]{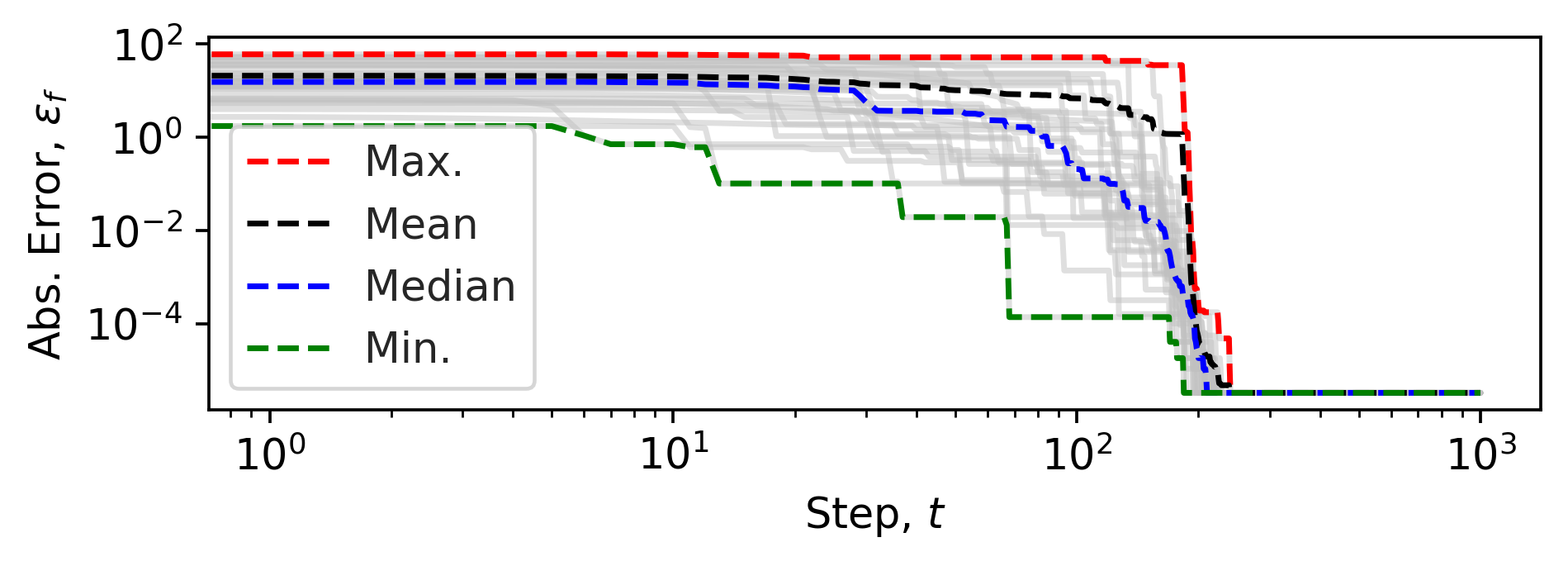} &
        \includegraphics[width=0.245\linewidth]{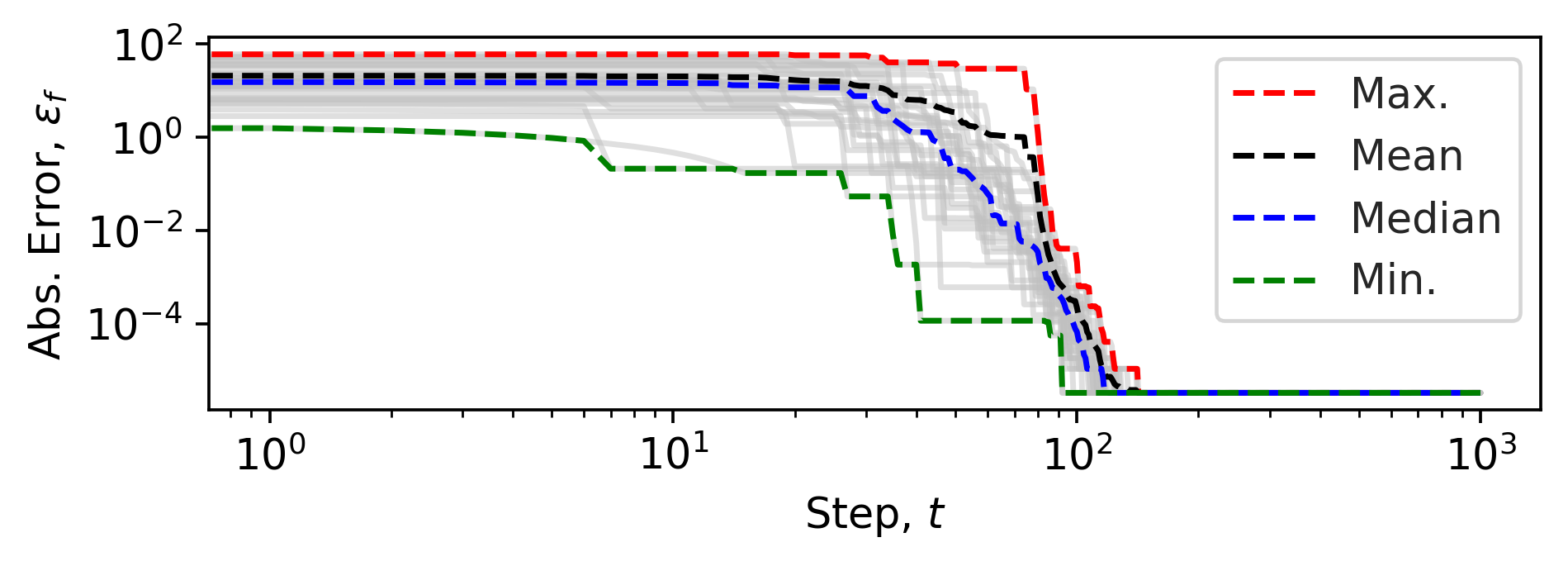} \\[0pt]
    \raisebox{\leRaiser}{\footnotesize\textbf{f6}} &
        \includegraphics[width=0.245\linewidth]{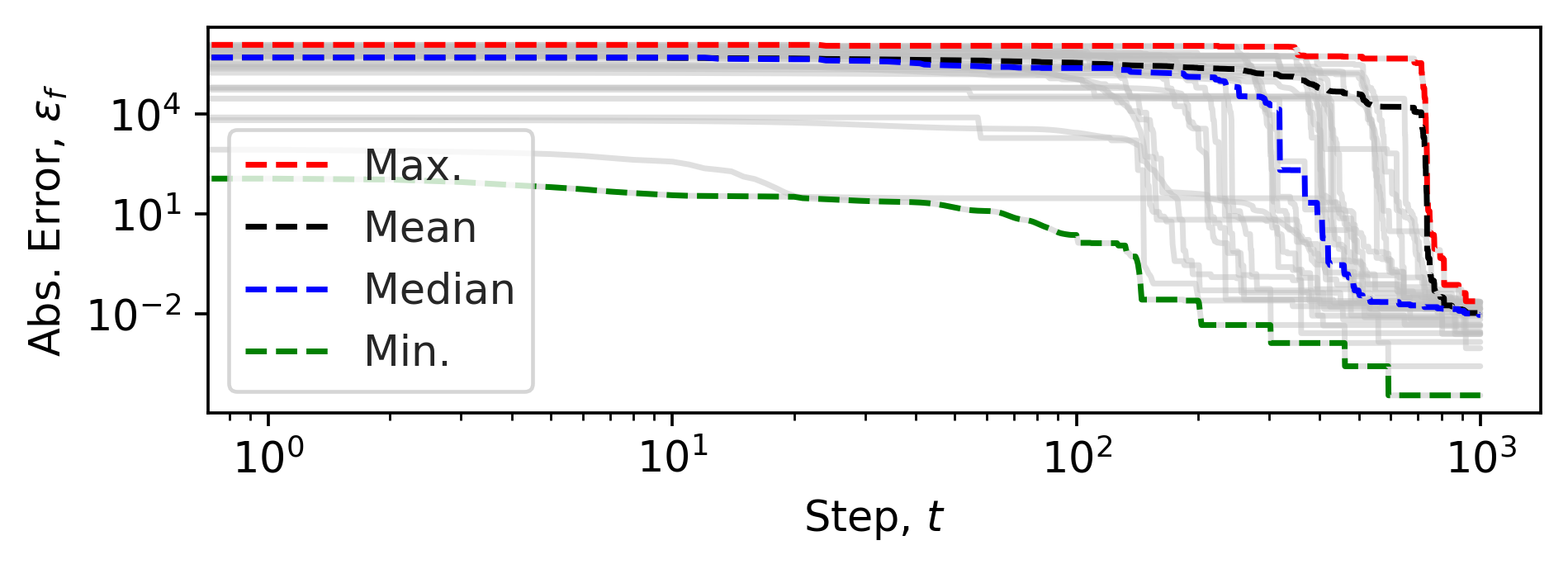} &
        \includegraphics[width=0.245\linewidth]{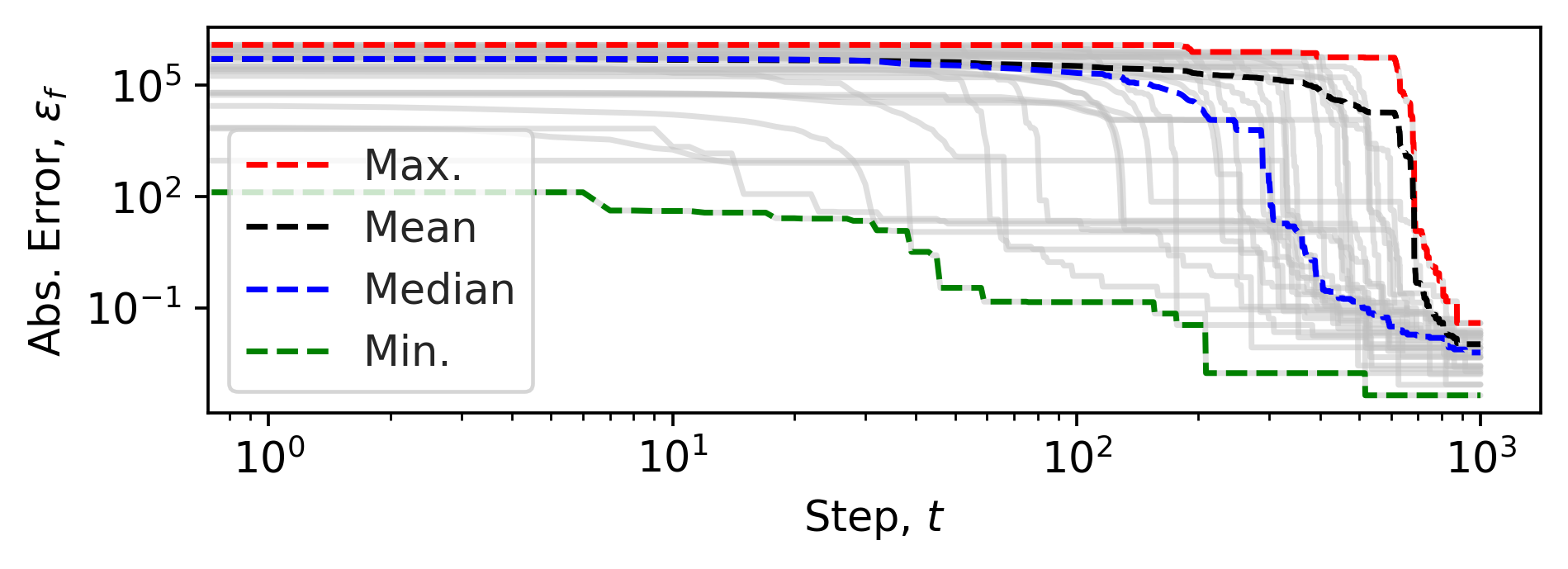} &
        \includegraphics[width=0.245\linewidth]{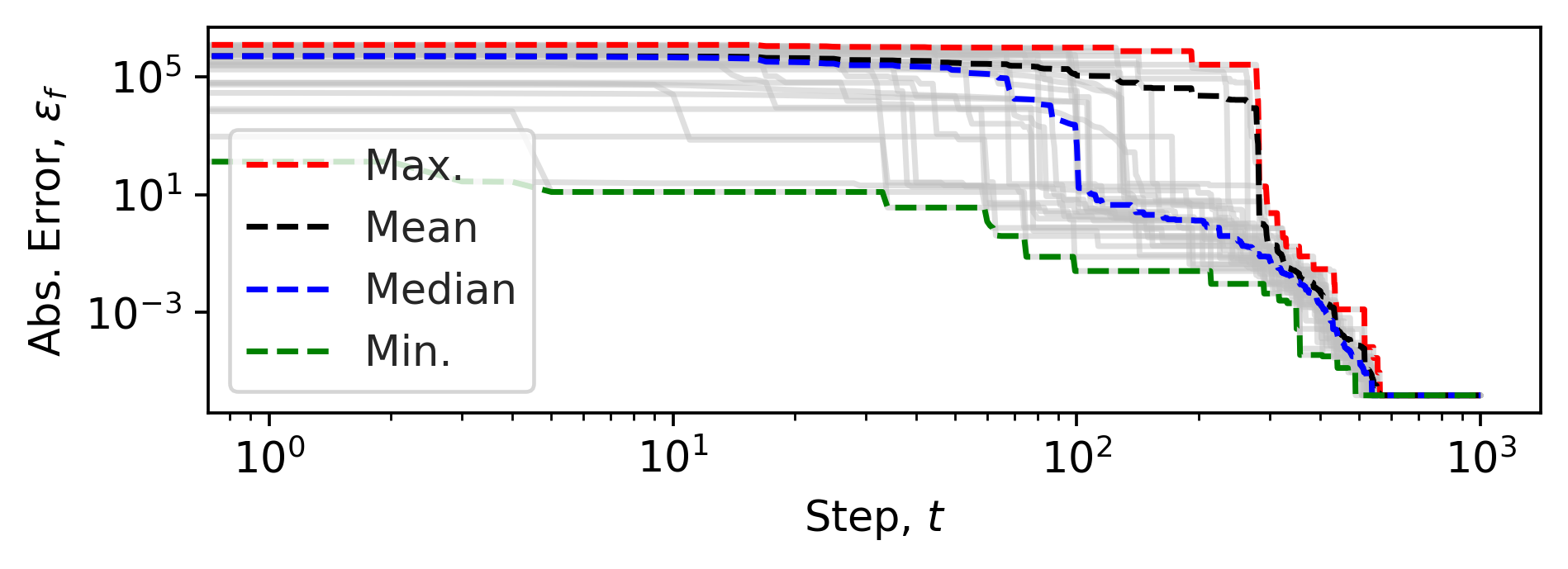} &
        \includegraphics[width=0.245\linewidth]{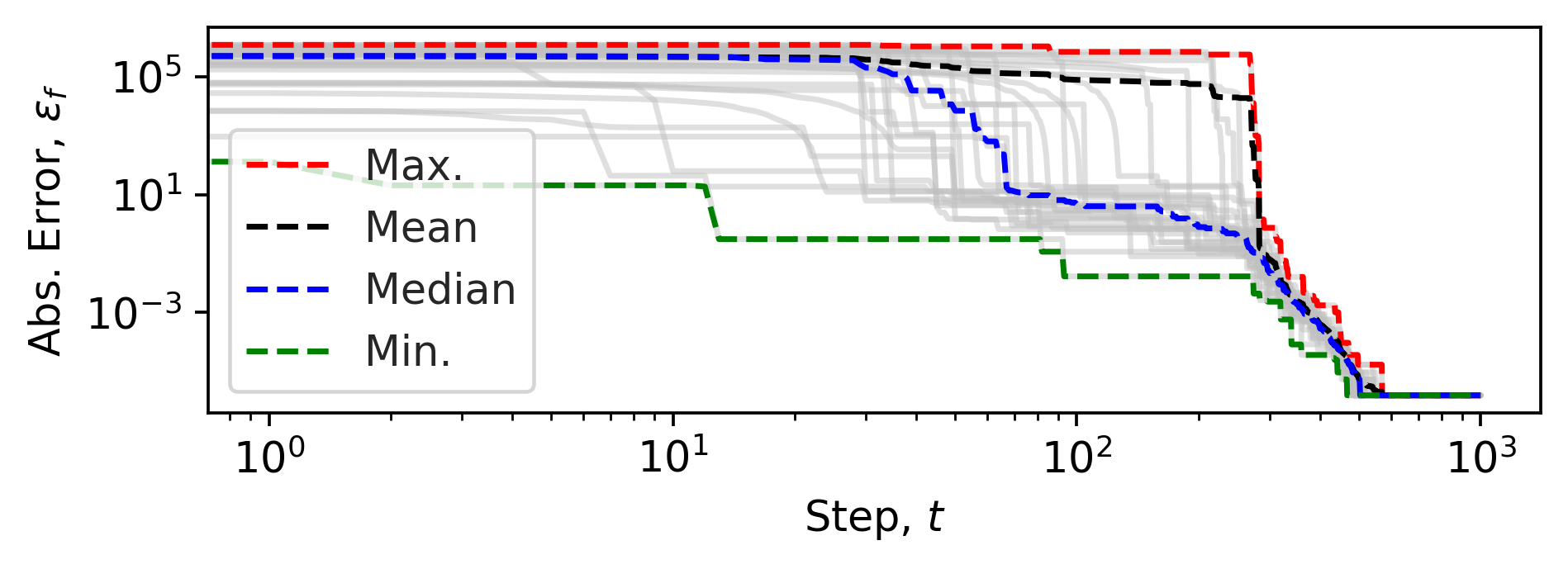} \\[0pt]
    \raisebox{\leRaiser}{\footnotesize\textbf{f10}} &
        \includegraphics[width=0.245\linewidth]{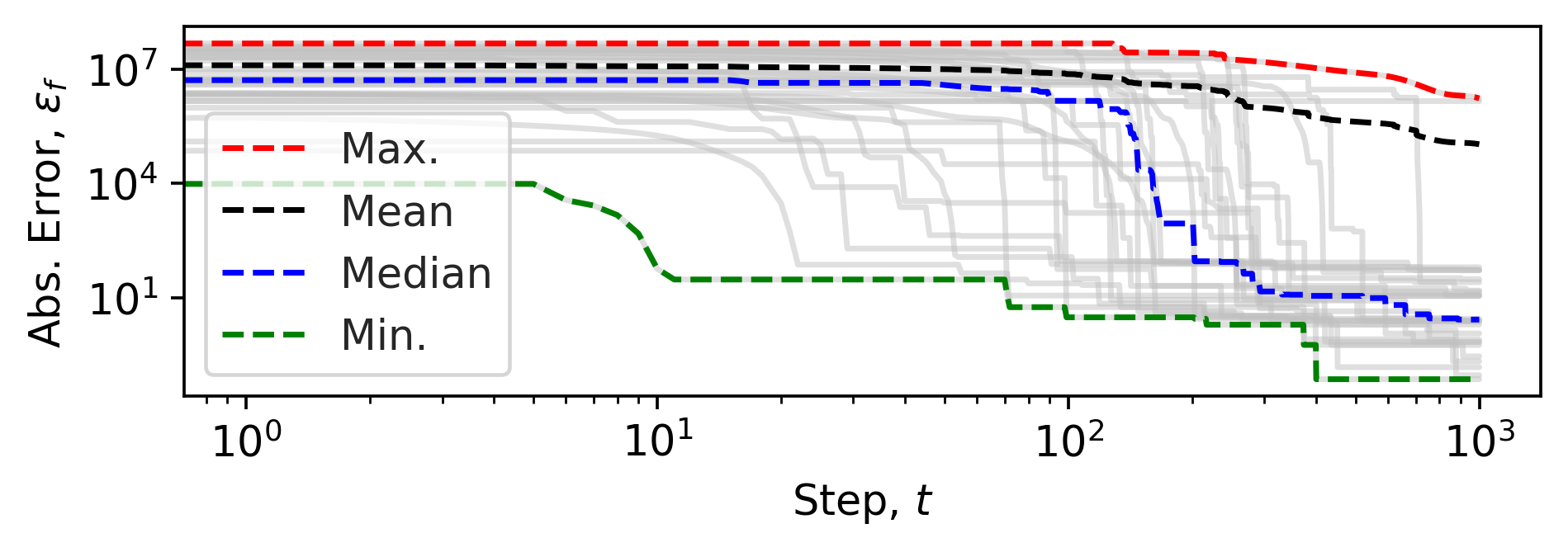} &
        \includegraphics[width=0.245\linewidth]{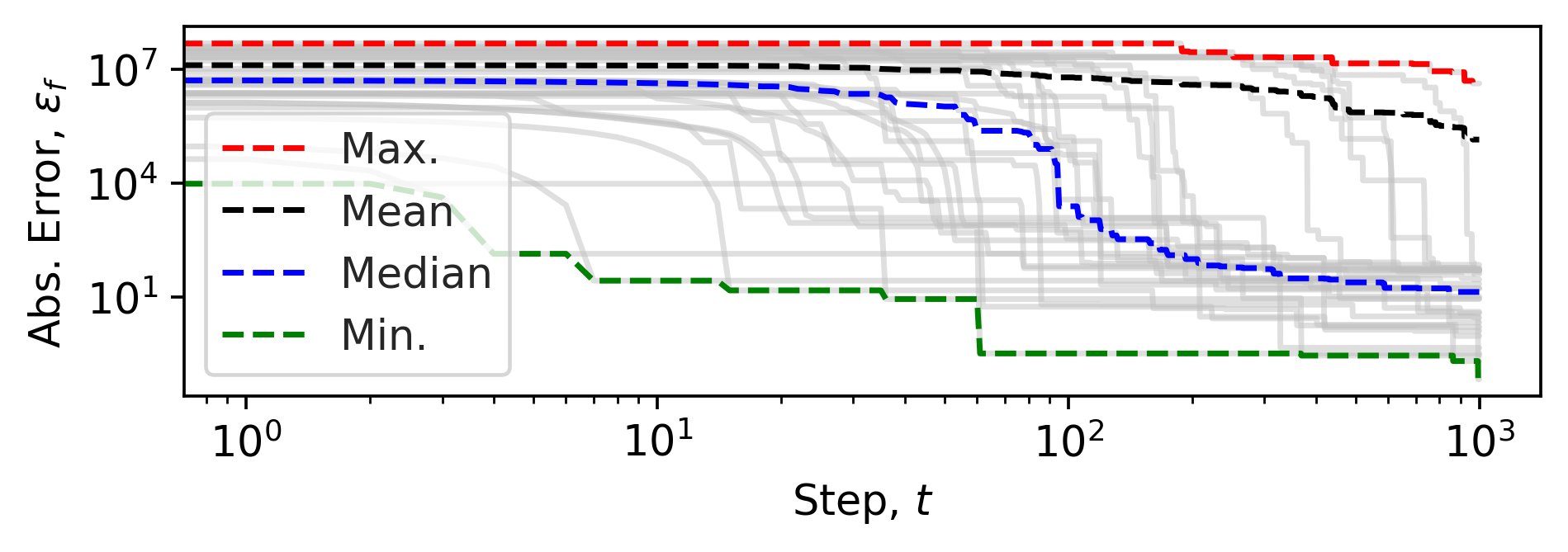} &
        \includegraphics[width=0.245\linewidth]{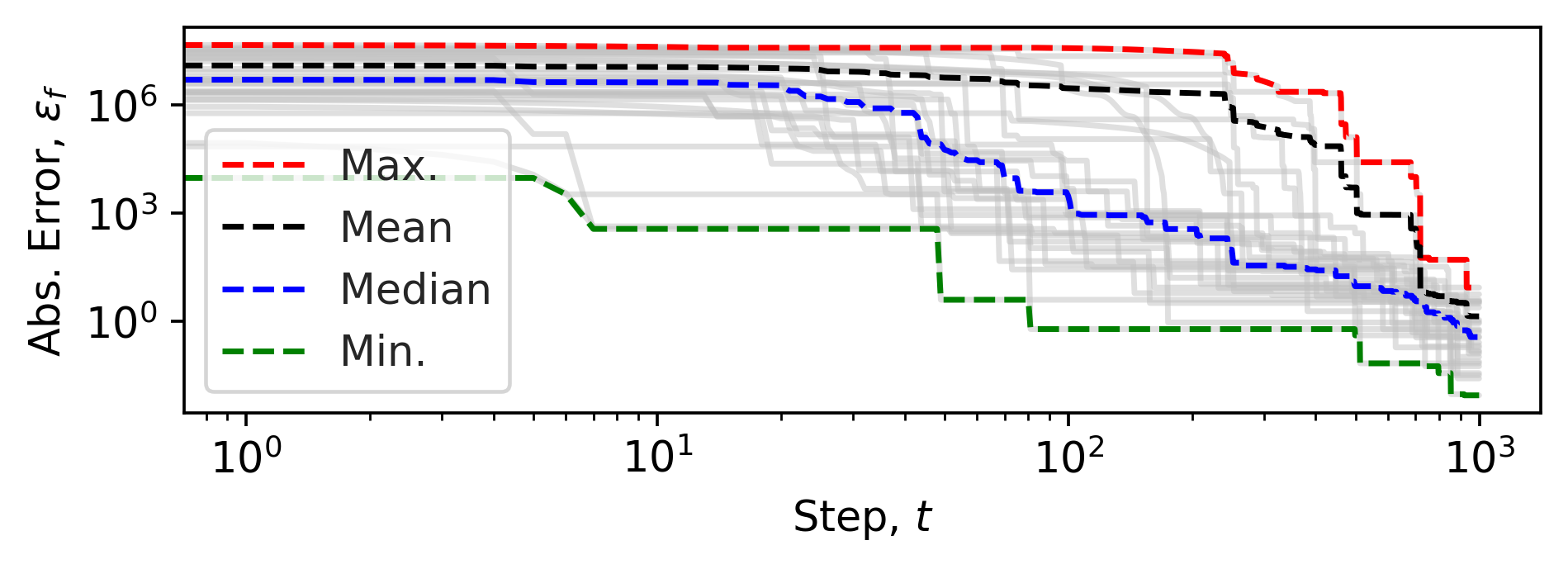} &
        \includegraphics[width=0.245\linewidth]{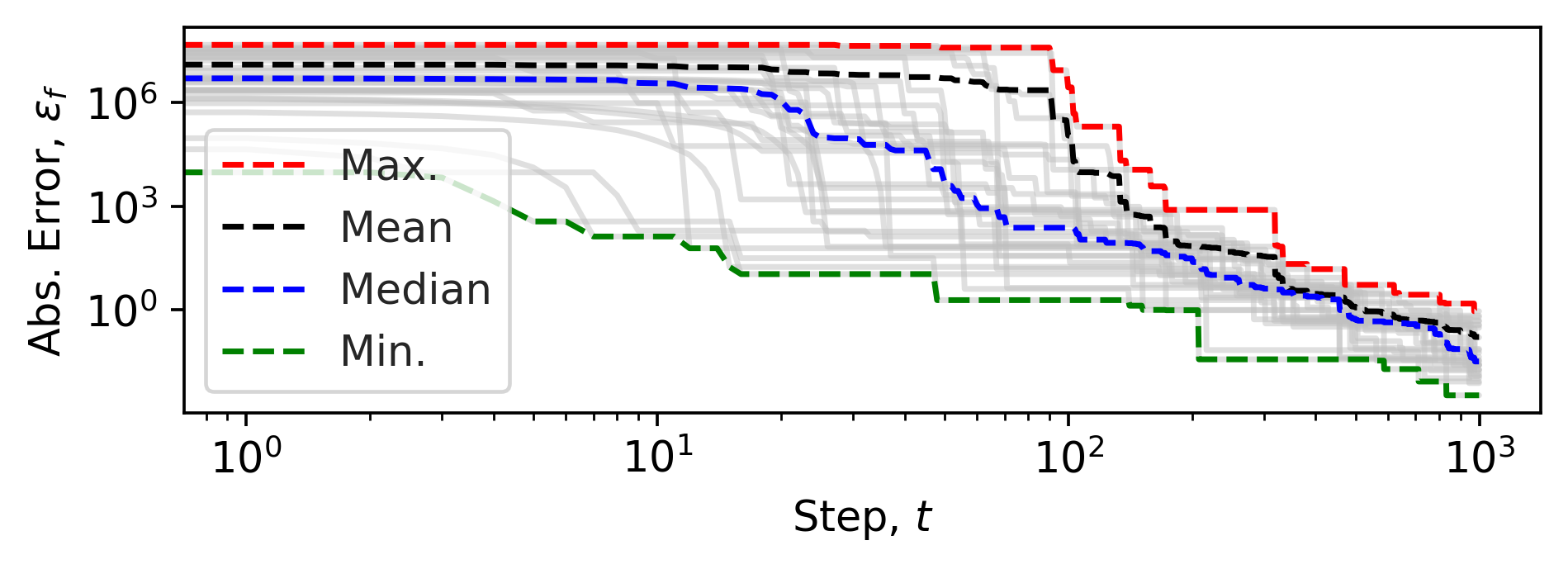} \\[0pt]
    \raisebox{\leRaiser}{\footnotesize\textbf{f15}} &
        \includegraphics[width=0.245\linewidth]{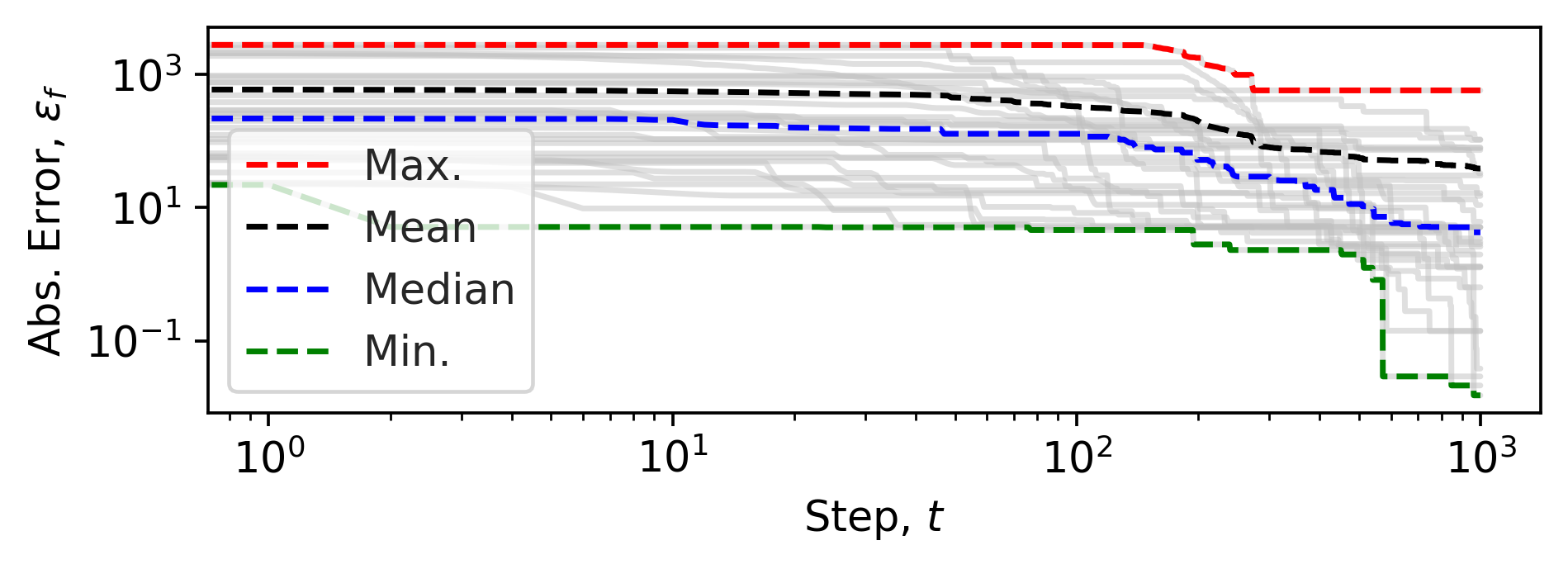} &
        \includegraphics[width=0.245\linewidth]{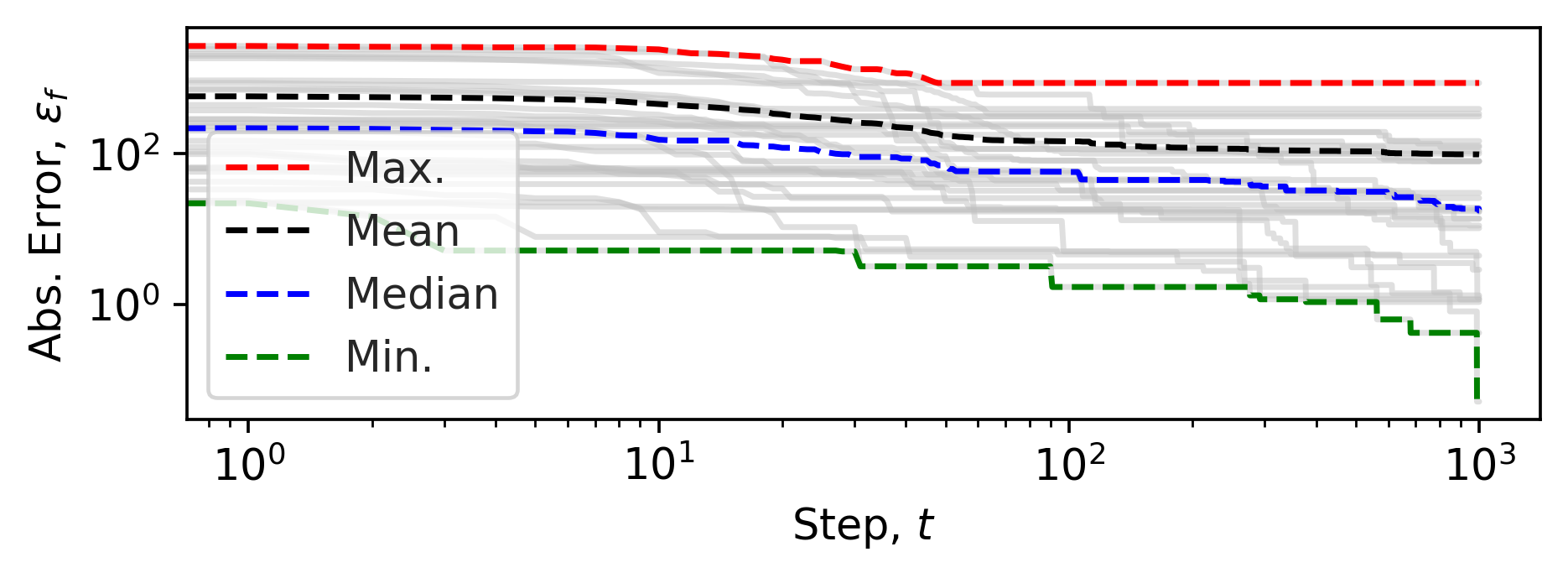} &
        \includegraphics[width=0.245\linewidth]{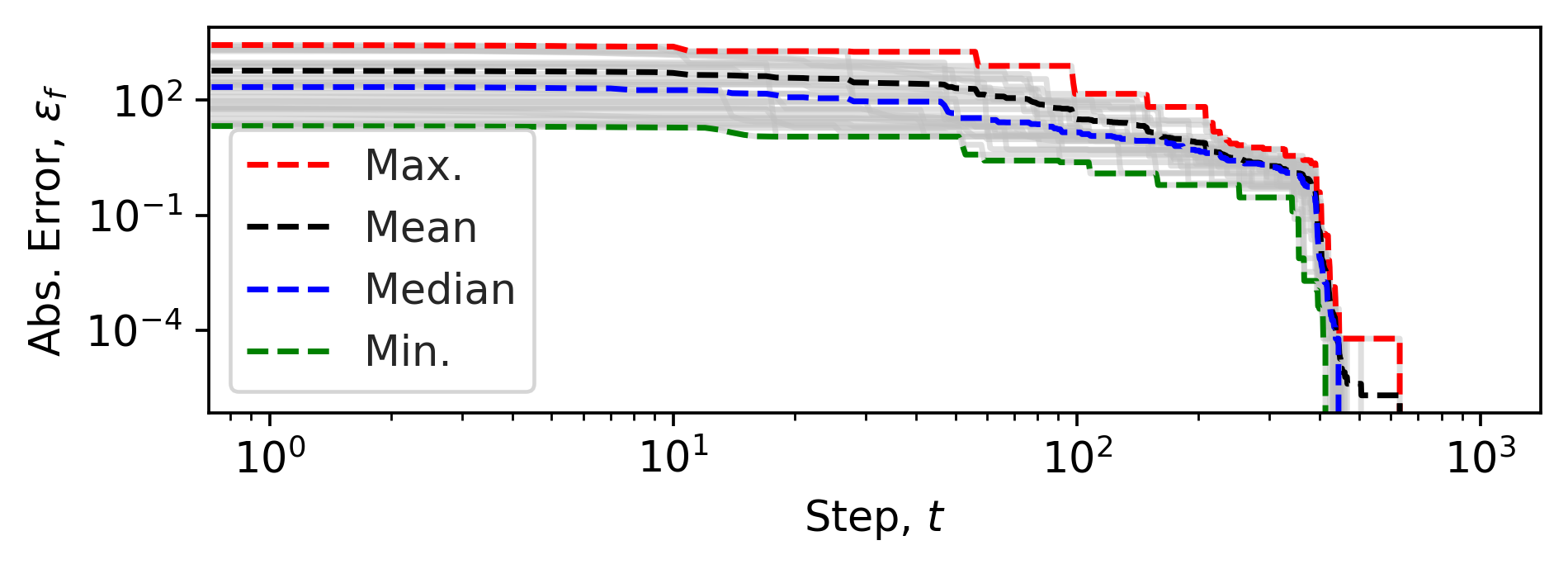} &
        \includegraphics[width=0.245\linewidth]{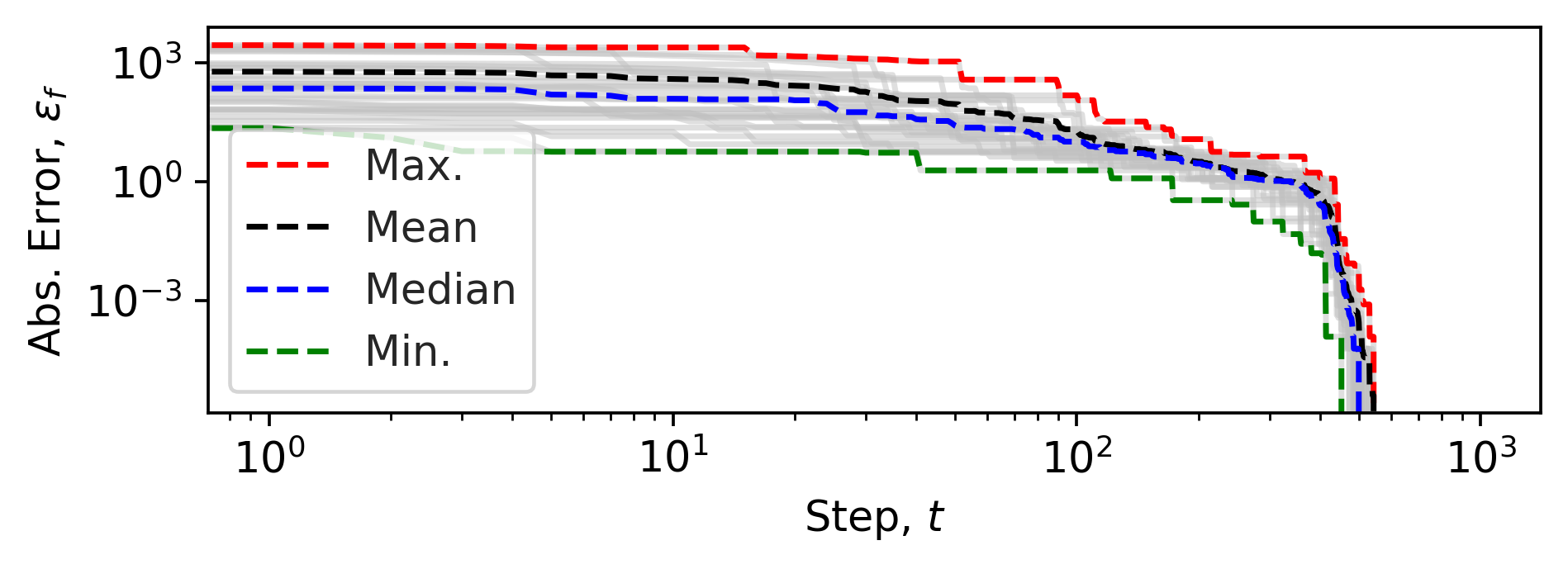} \\[0pt]
    \raisebox{\leRaiser}{\footnotesize\textbf{f20}} &
        \includegraphics[width=0.245\linewidth]{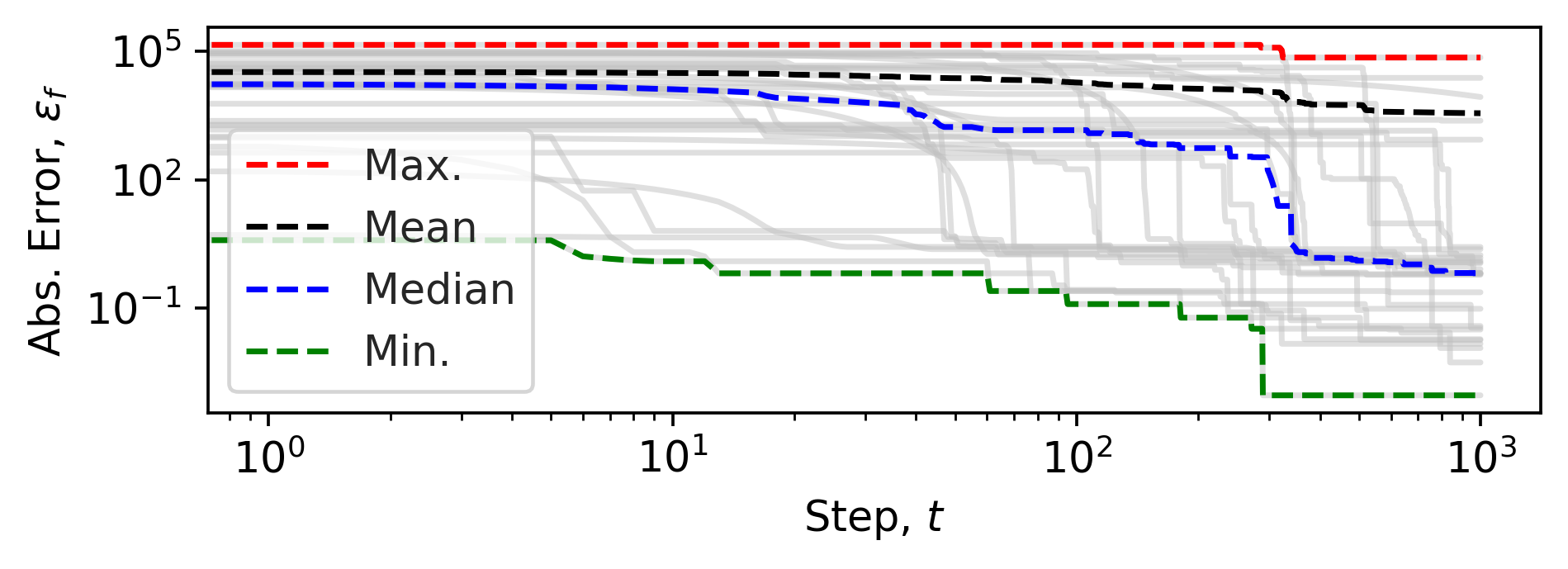} &
        \includegraphics[width=0.245\linewidth]{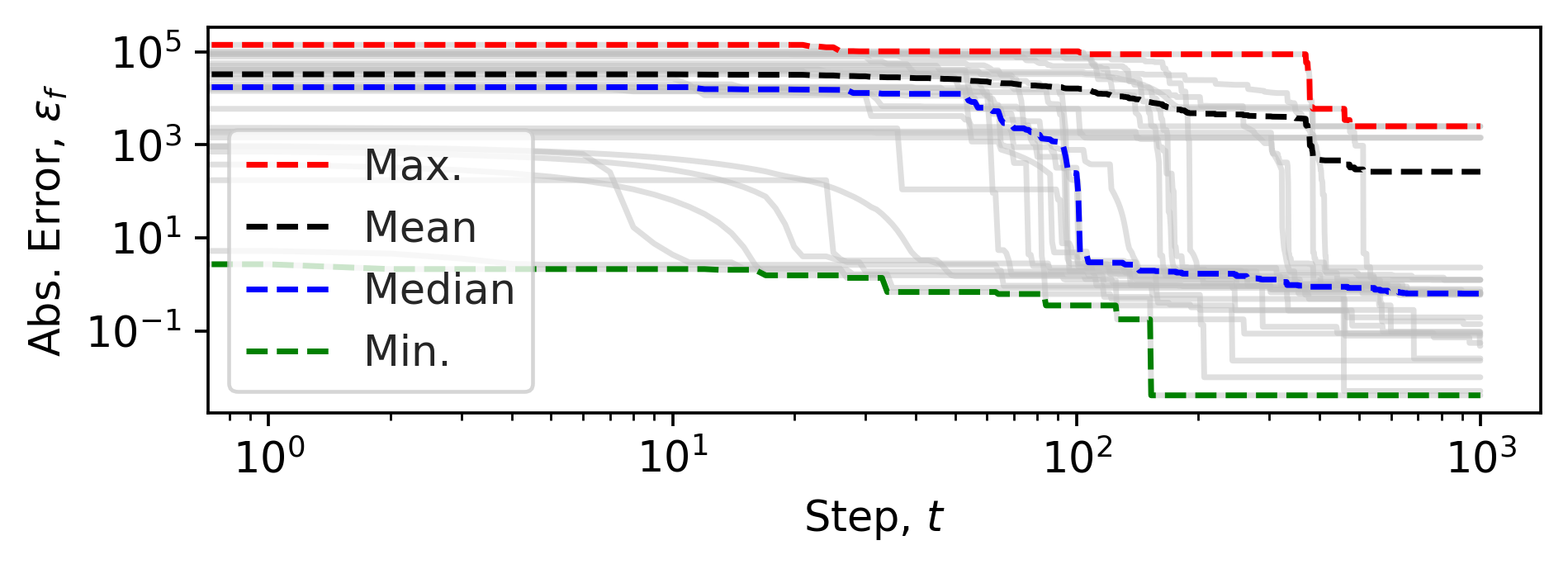} &
        \includegraphics[width=0.245\linewidth]{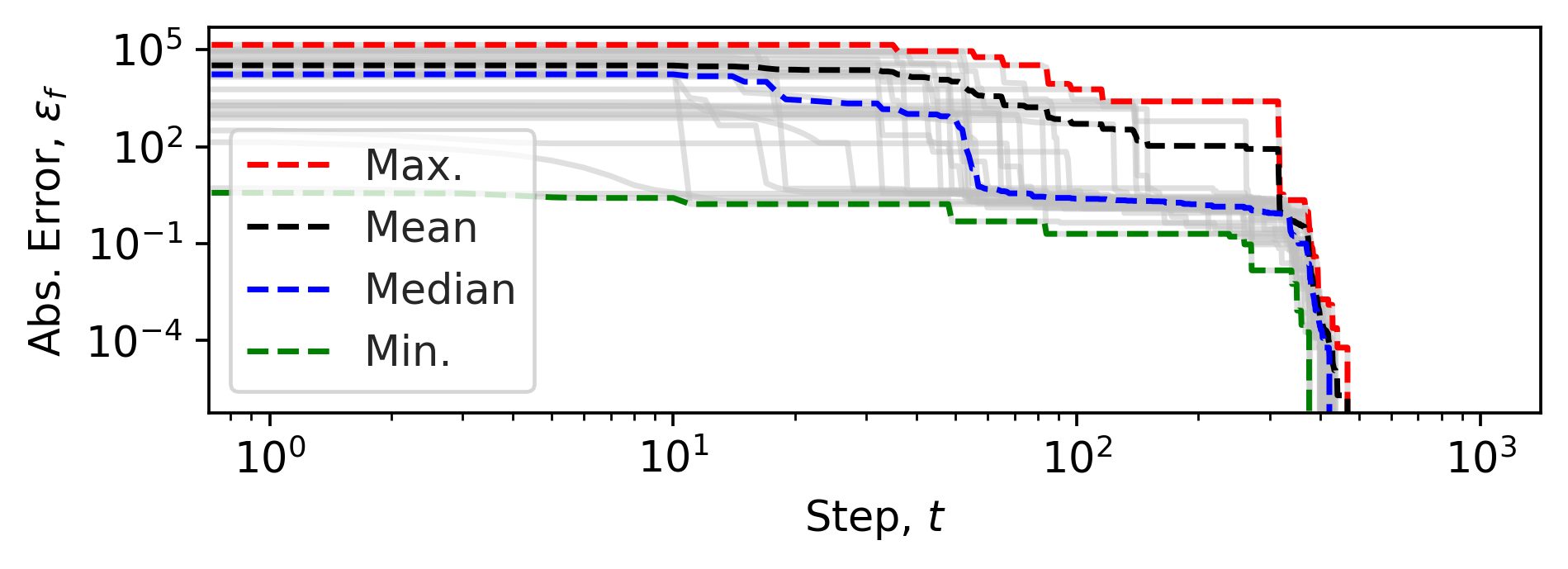} &
        \includegraphics[width=0.245\linewidth]{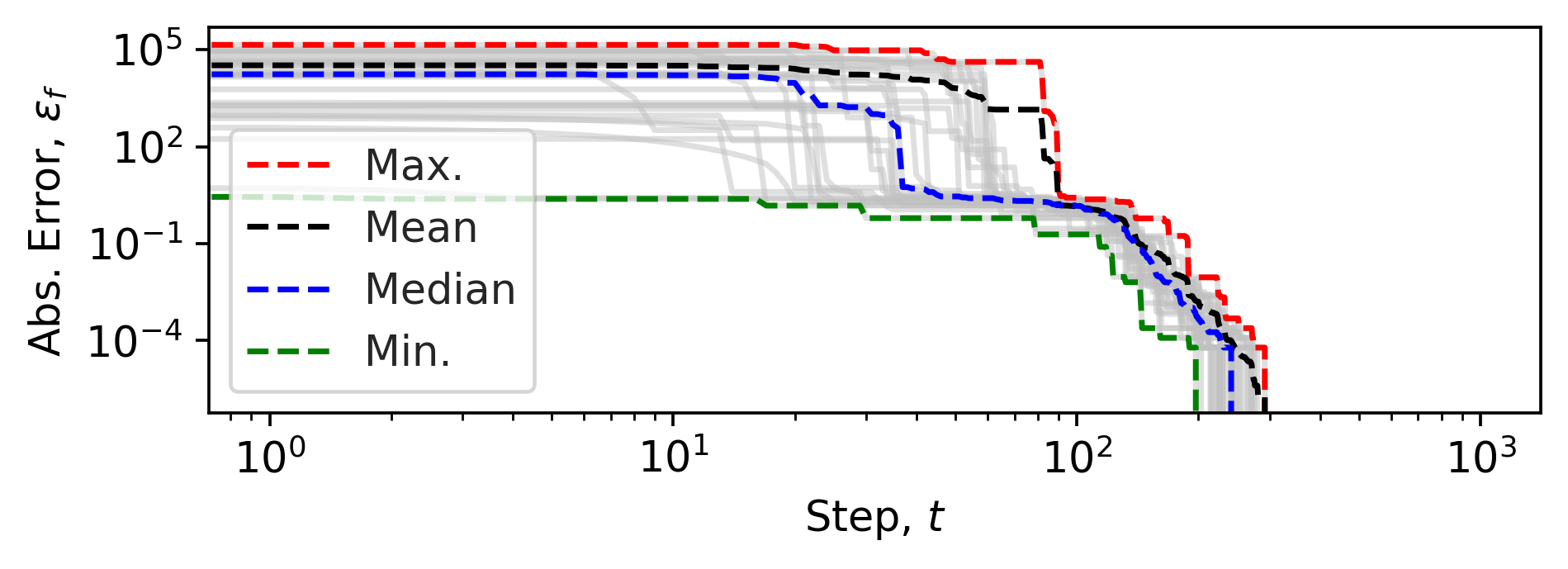} \\[0pt]
    & Linear+Fixed & Izhikevich+Fixed & Linear+DE/\emph{current-to-rand}/1 & Izhikevich+DE/\emph{current-to-rand}/1
\end{tabular}
\caption{\label{Fig:ExPre:Errors}%
    Absolute error \(\varepsilon_f = |f(\pmb{p}_i) - f(\pmb{x}_\ast)|\) over 1000 steps for the first 2D BBOB function of each different problem subgroup; \ie\texttt{separ}, \texttt{lcond}, \texttt{hcond}, \texttt{multi}, and \texttt{mult2}. 
    Each row stands for a function \(f_k\) from the BBOB suite.
    Each column corresponds to \nha{}s (\(h_d\)+\(h_s\)) implemented using the Linear and Izhikevich models as \(h_d\), and the Fixed and DE/\emph{current-to-rand}/1 rules as \(h_s\).
}
\end{figure*}

\figurename{}~\ref{Fig:ExPre:Errors} depicts the evolution of the absolute error \(\varepsilon_r\) for these representative 2D problems under certain combinations of \(h_d\) and \(h_s\) heuristics. 
We chose to present the most descriptive ones for brevity, although the entire set of results can be consulted on \cite{Cruz2025neurodataset}.
In both the linear and Izhikevich models with the fixed spike-triggered rule, the mean and median error curves descend in discrete steps, interrupted by plateaus of limited progress.
Considering that each grey curve corresponds to an individual \nhu{}'s best error, most do not remain flat and instead exhibit gradual improvement throughout the search.
This behaviour can be attributed to asynchronous progress and spike-driven information sharing.
For both cores, employing the DE/\emph{current-to-rand}/1 heuristic leads to an immediate reduction in population error variance, with statistical traces converging more rapidly.
Under Izhikevich's variant, the descent is sharper and the final error floor lower across all problem classes, particularly in \textbf{f15} and \textbf{f20}.
This contrast is evident when compared to the curves obtained with the fixed \(h_s\).

\begin{figure}[!ht]\centering
    \def\leRaiser{20mm}
    \def\leWidth{0.18\linewidth}
    \begin{tabular}{@{}c*{4}{@{ }c}@{}}
        \raisebox{\leRaiser}{\footnotesize\textbf{f1}} &
        \includegraphics[trim={3em 0em 0.6em 0em}, clip, width=\leWidth]{
                    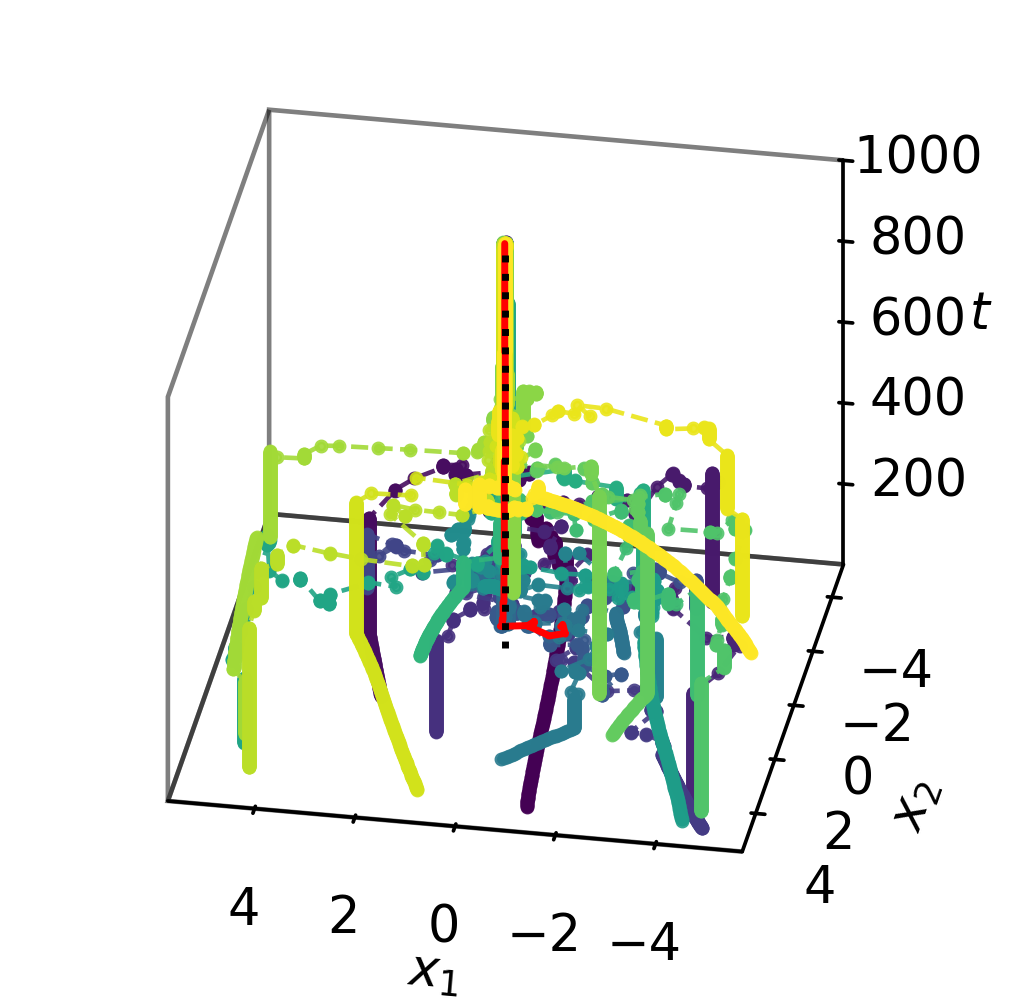} &
        \includegraphics[trim={3em 0em 0.6em 0em}, clip, width=\leWidth]{
                    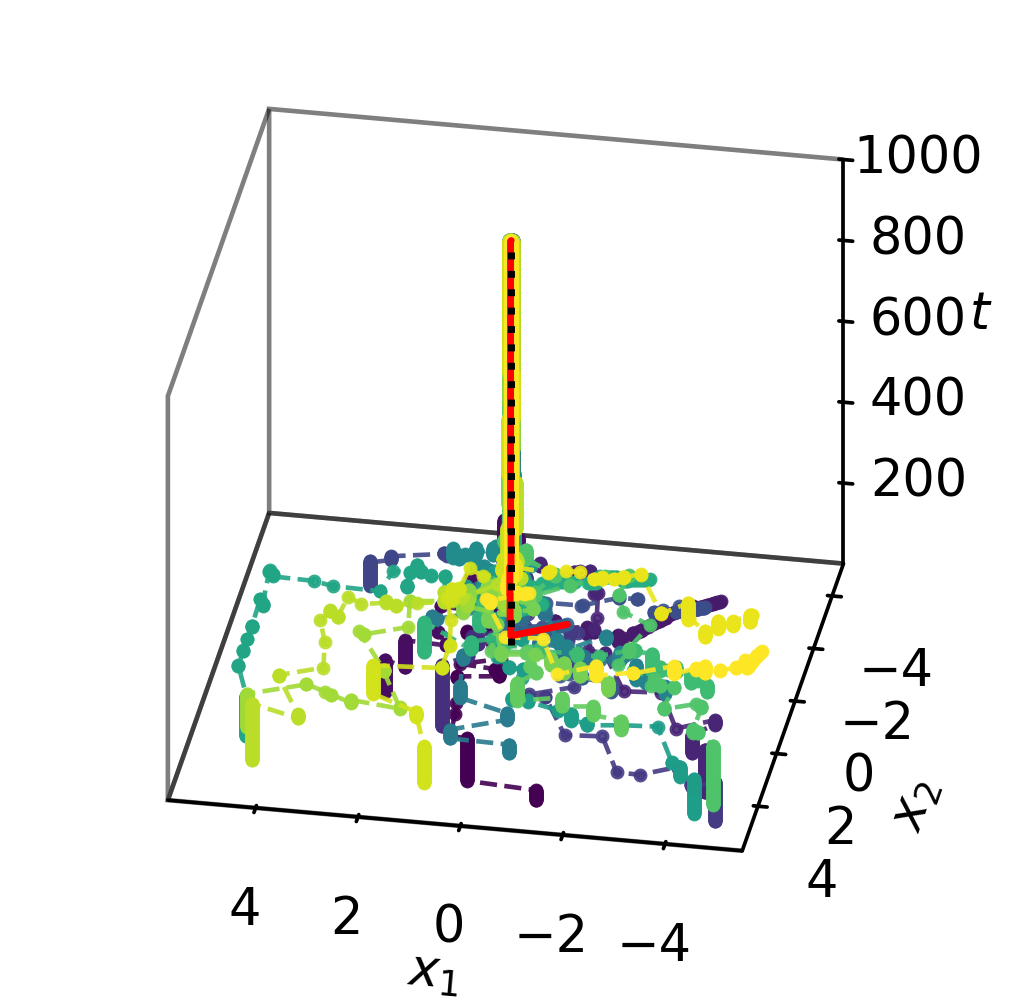} &
        \includegraphics[trim={3em 0em 0.6em 0em}, clip, width=\leWidth]{
                    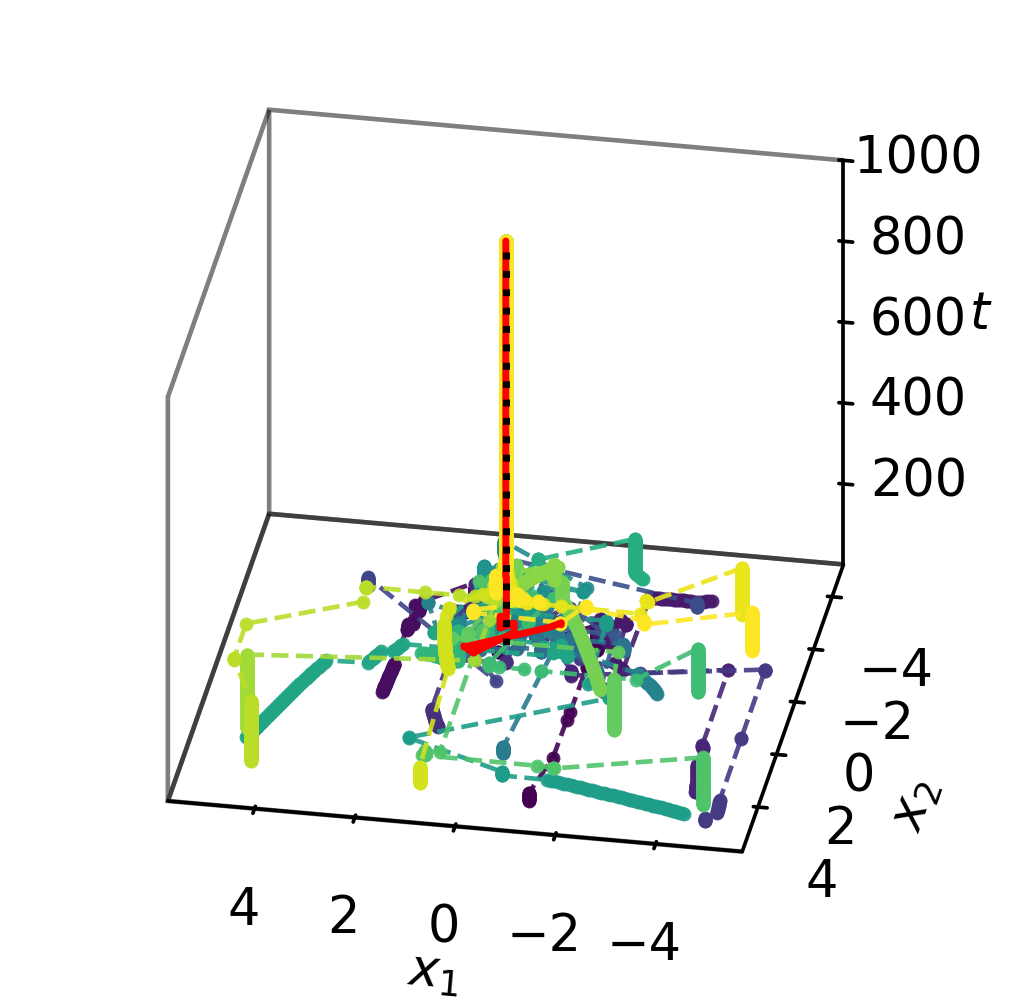} &
        \includegraphics[trim={3em 0em 0.6em 0em}, clip, width=\leWidth]{
                    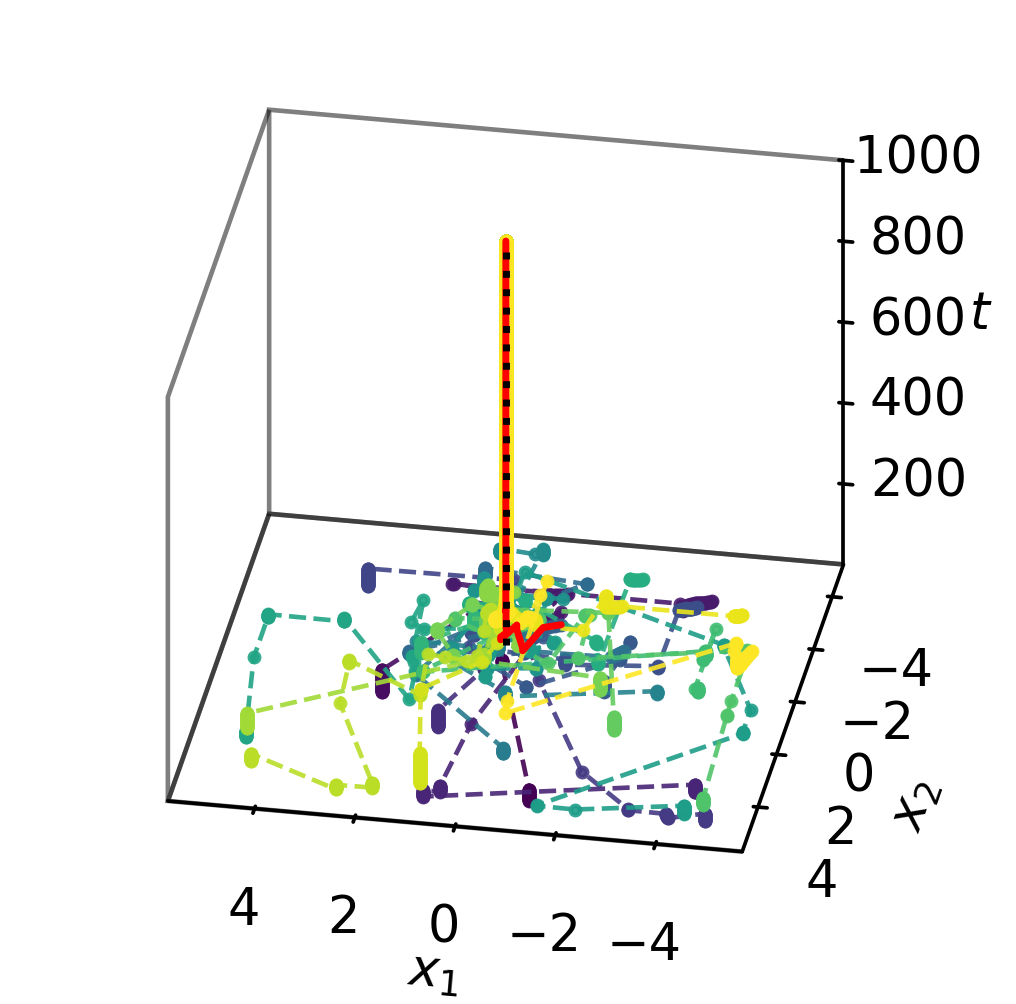} \\[0pt]
        \raisebox{\leRaiser}{\footnotesize\textbf{f6}} &
        \includegraphics[trim={3em 0em 0.6em 0em}, clip, width=\leWidth]{
                    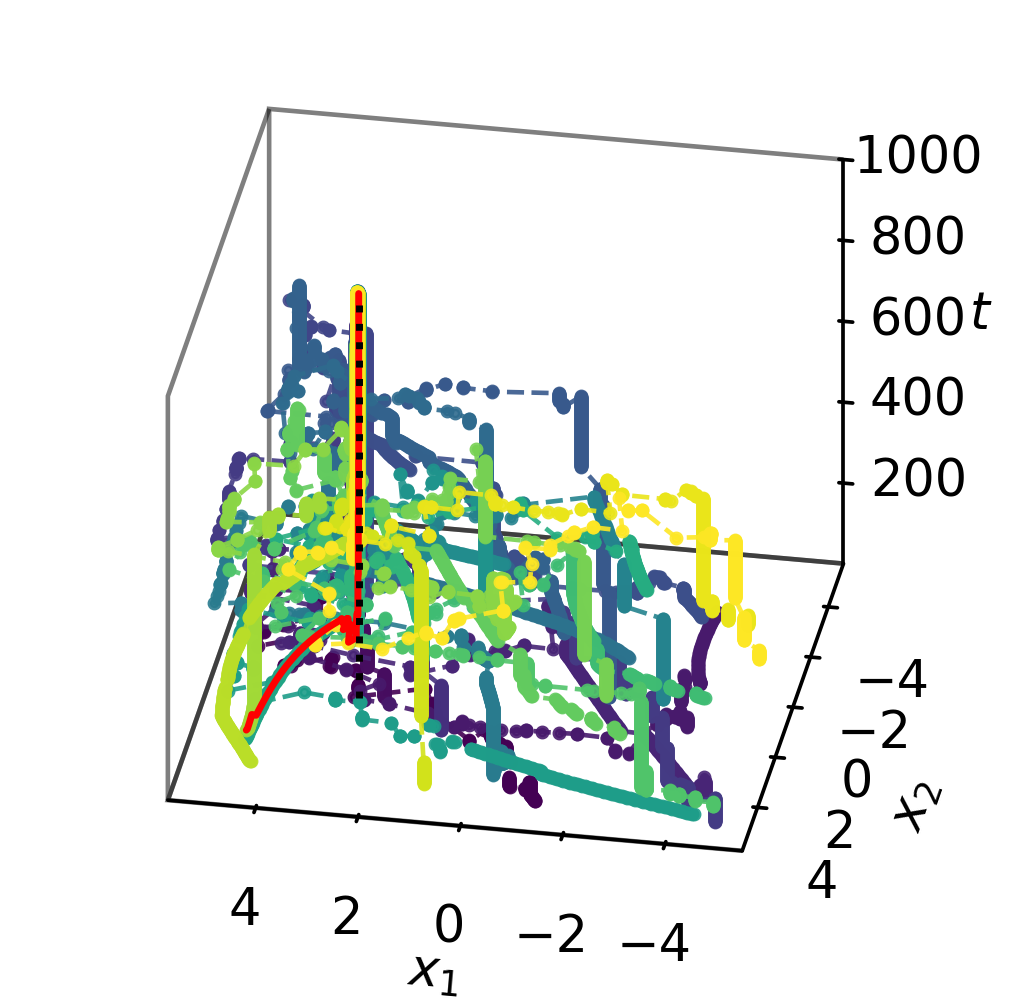} &
        \includegraphics[trim={3em 0em 0.6em 0em}, clip, width=\leWidth]{
                    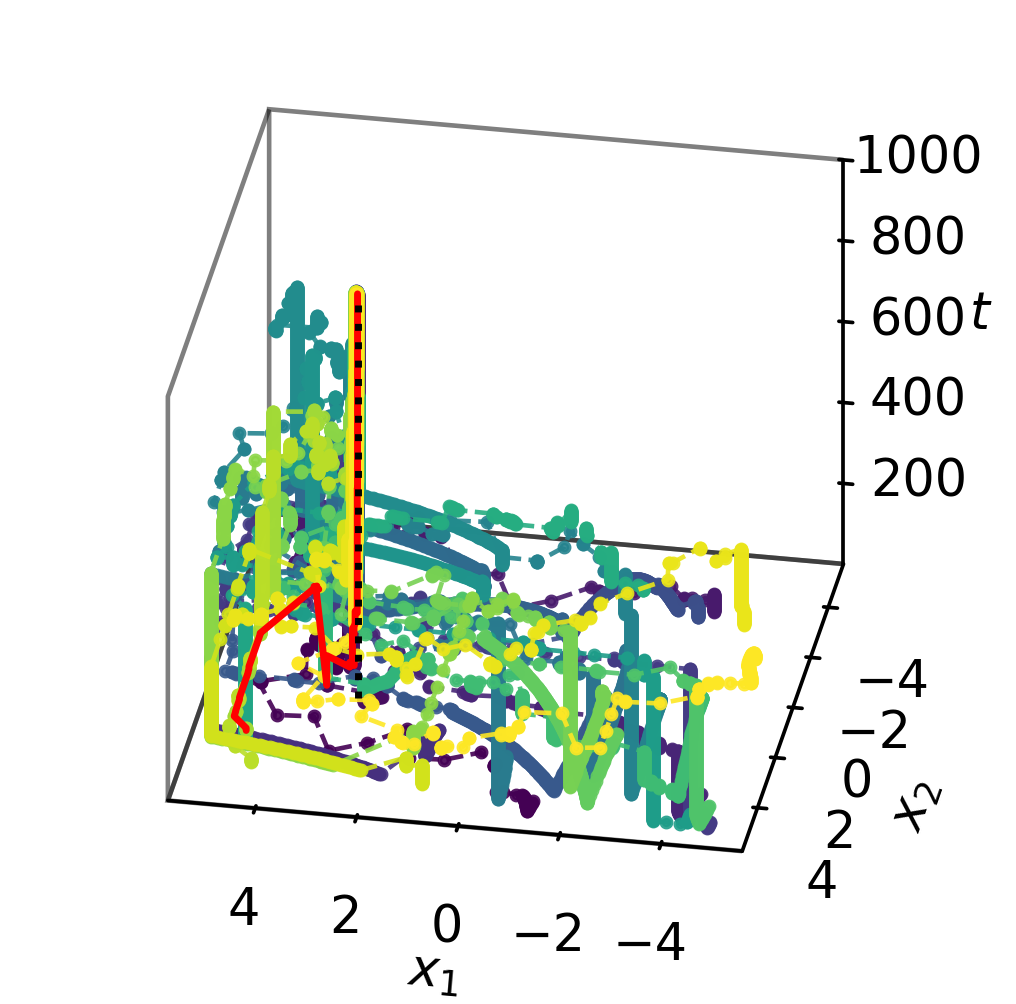} &
        \includegraphics[trim={3em 0em 0.6em 0em}, clip, width=\leWidth]{
                    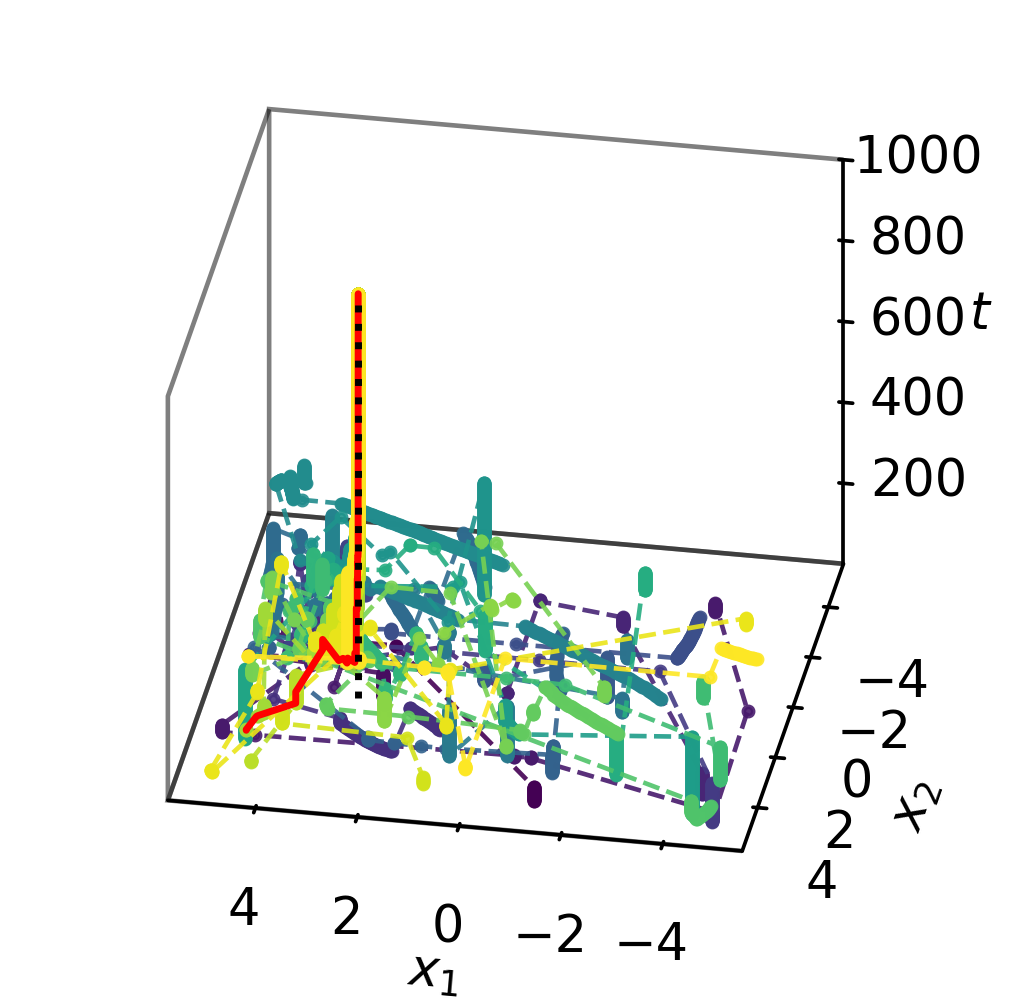} &
        \includegraphics[trim={3em 0em 0.6em 0em}, clip, width=\leWidth]{
                    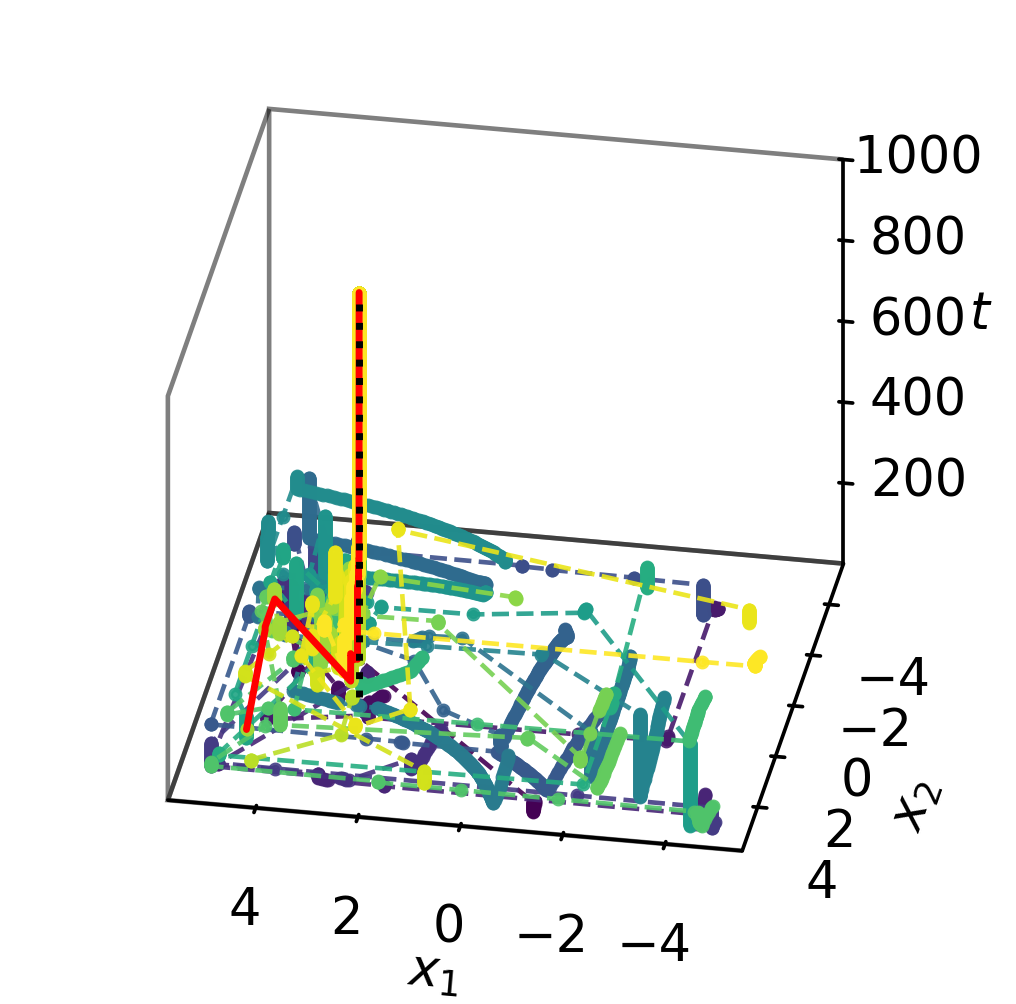} \\[0pt]
    \raisebox{\leRaiser}{\footnotesize\textbf{f10}} &
        \includegraphics[trim={3em 0em 0.6em 0em}, clip, width=\leWidth]{
                    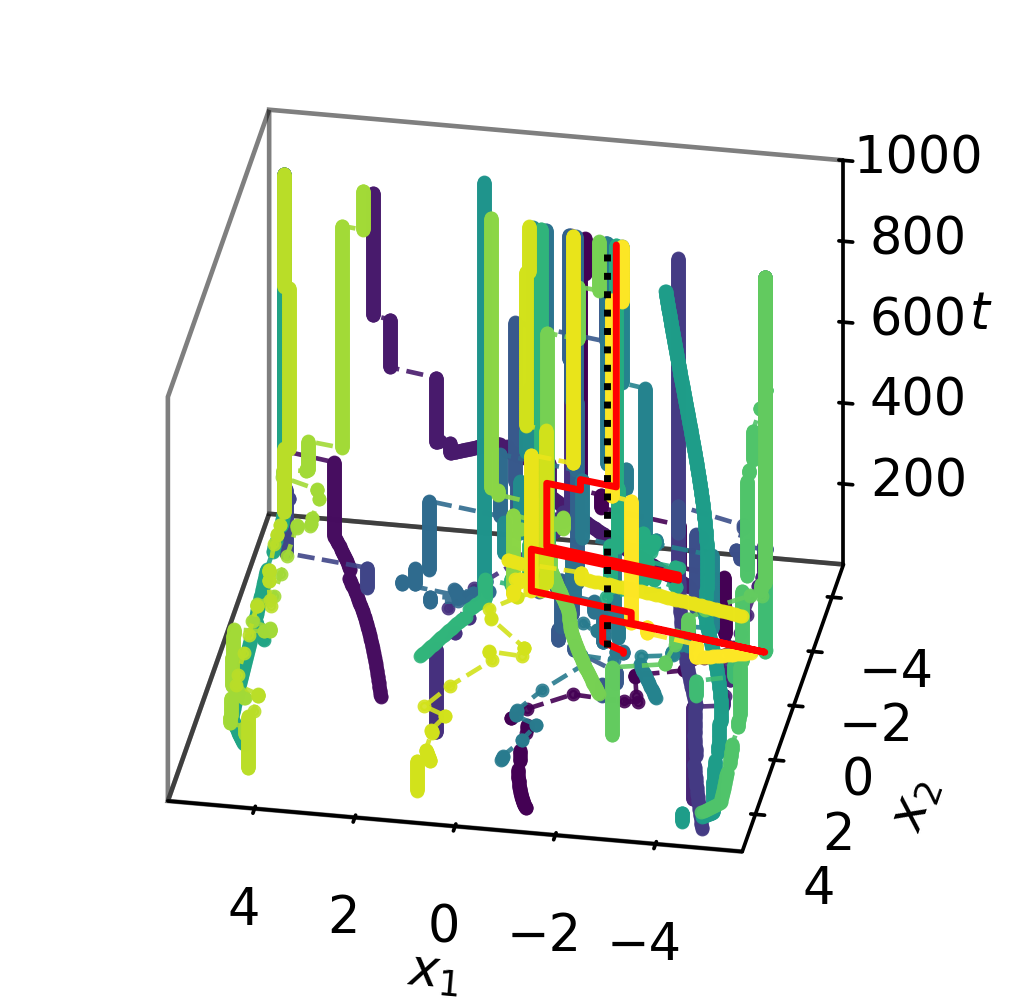} &
        \includegraphics[trim={3em 0em 0.6em 0em}, clip, width=\leWidth]{
                    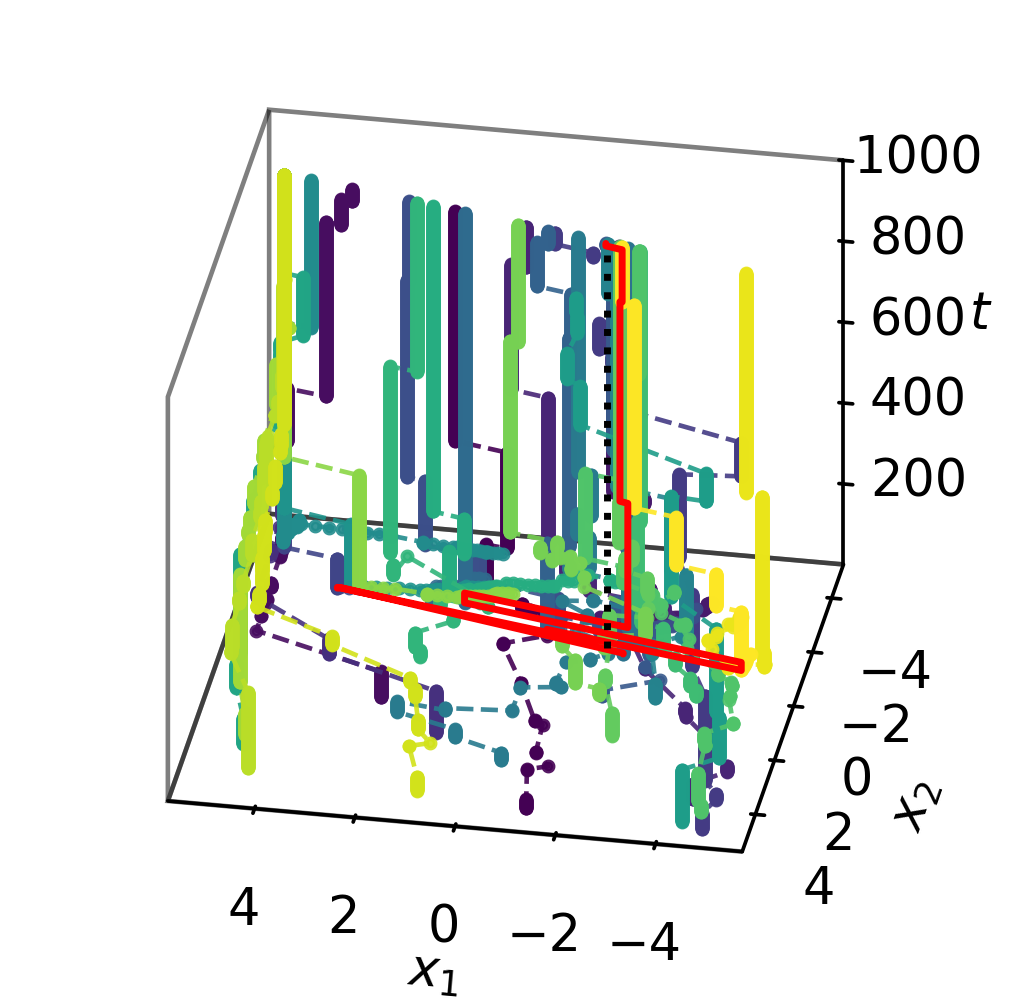} &
        \includegraphics[trim={3em 0em 0.6em 0em}, clip, width=\leWidth]{
                    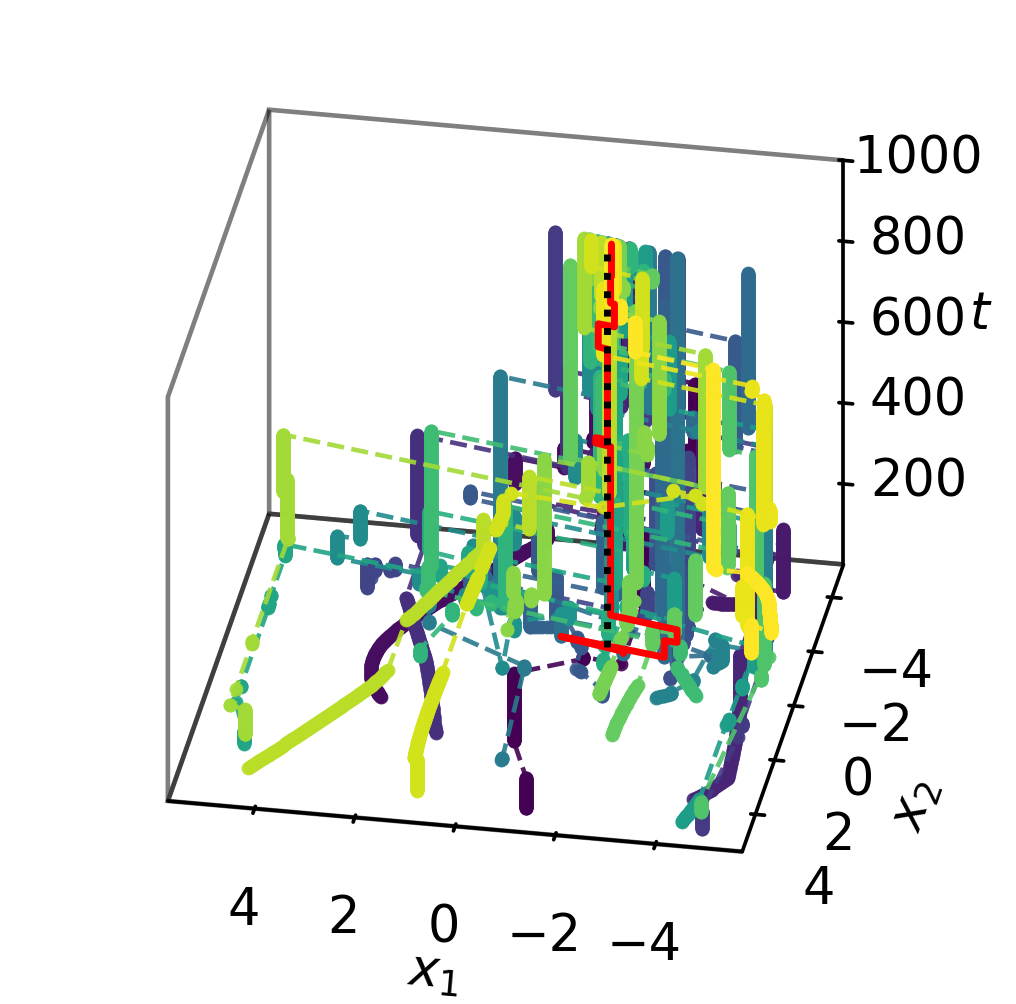} &
        \includegraphics[trim={3em 0em 0.6em 0em}, clip, width=\leWidth]{
                    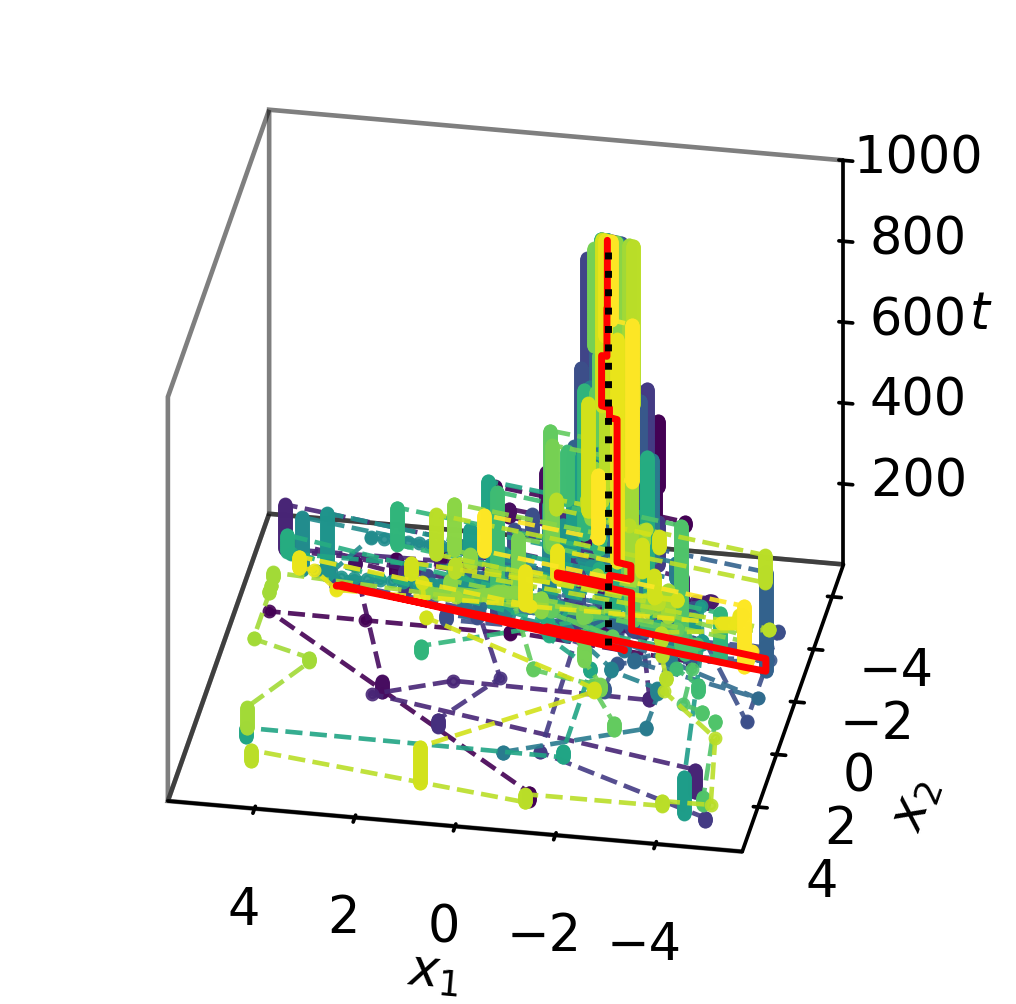} \\[0pt]
    \raisebox{\leRaiser}{\footnotesize\textbf{f15}} &
        \includegraphics[trim={3em 0em 0.6em 0em}, clip, width=\leWidth]{
                    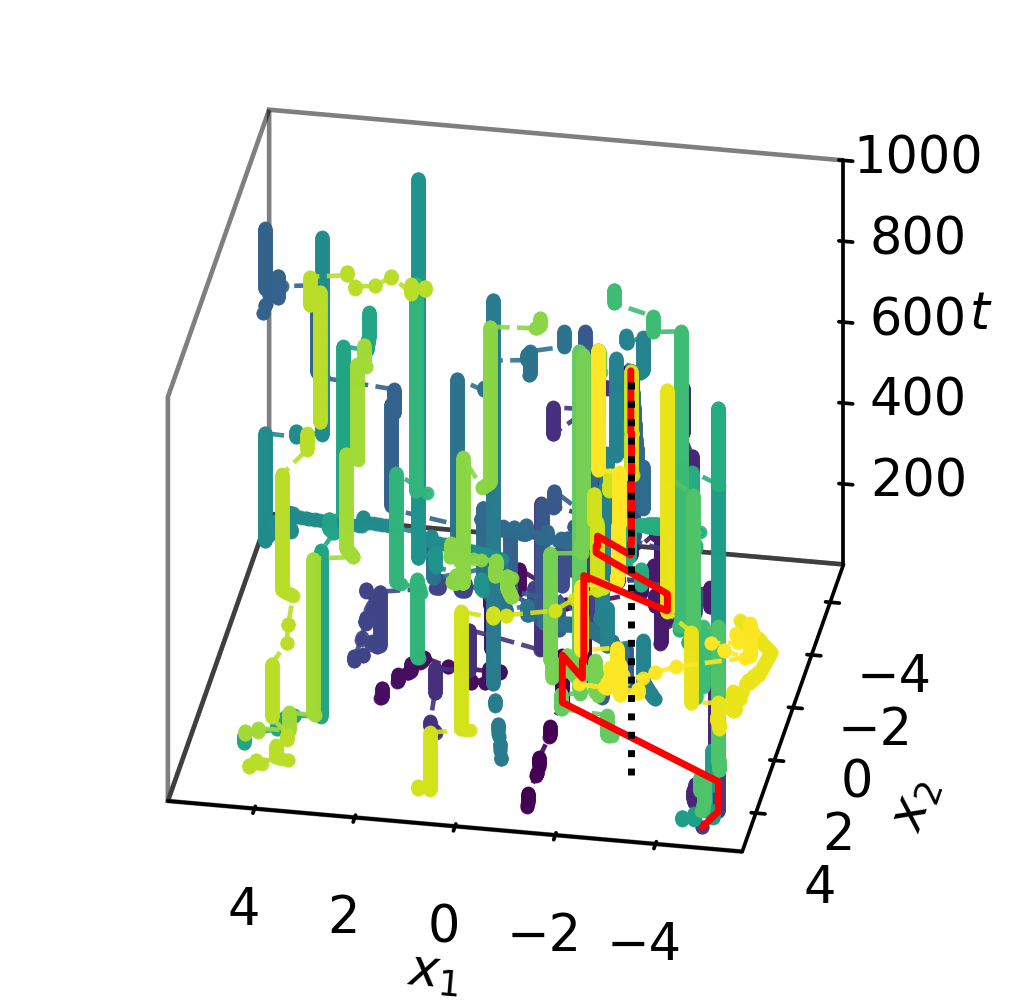} &
        \includegraphics[trim={3em 0em 0.6em 0em}, clip, width=\leWidth]{
                    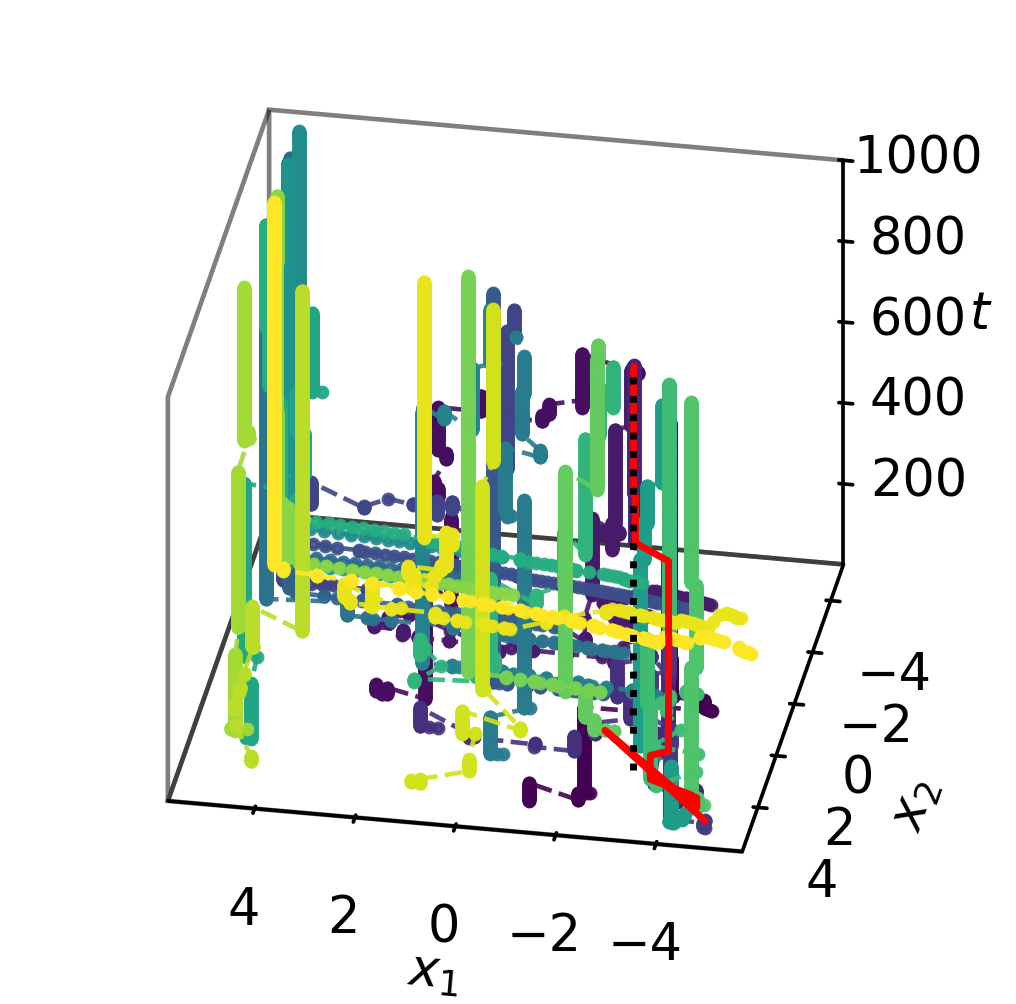} &
        \includegraphics[trim={3em 0em 0.6em 0em}, clip, width=\leWidth]{
                    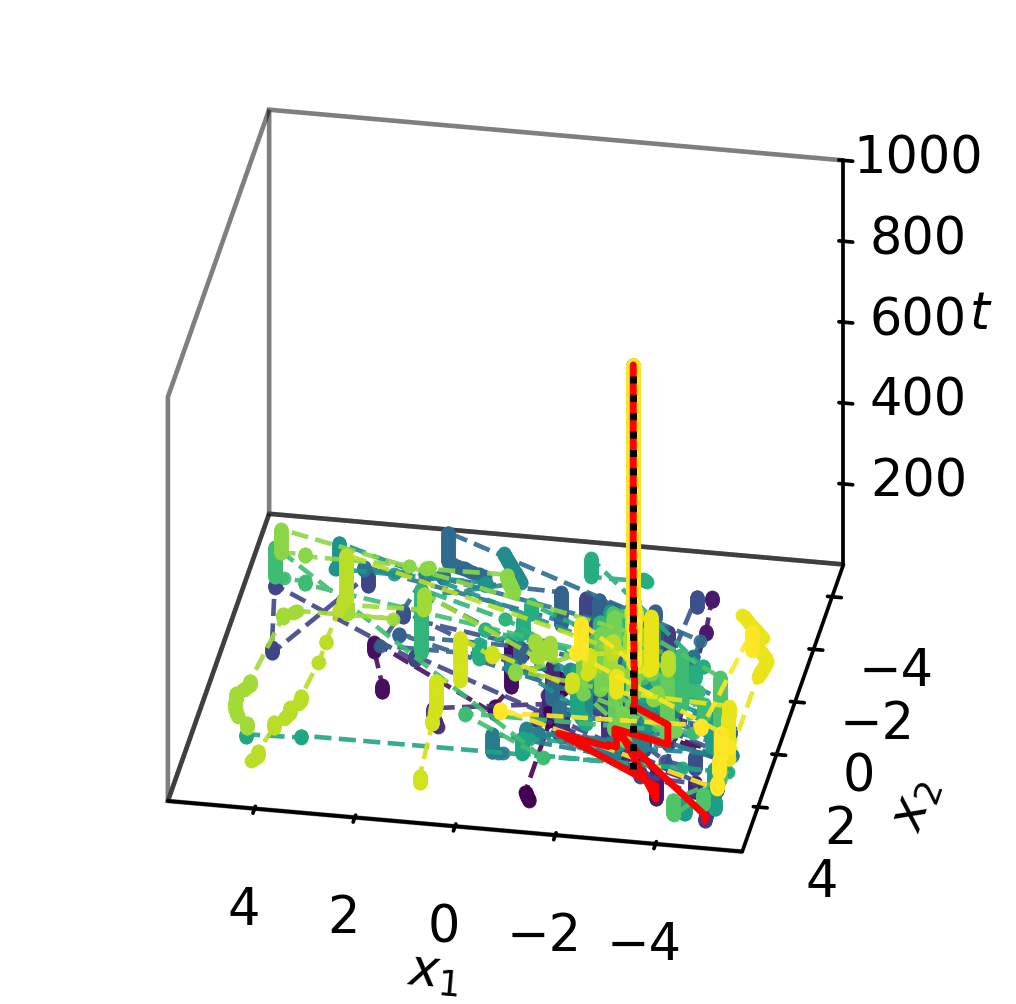} &
        \includegraphics[trim={3em 0em 0.6em 0em}, clip, width=\leWidth]{
                    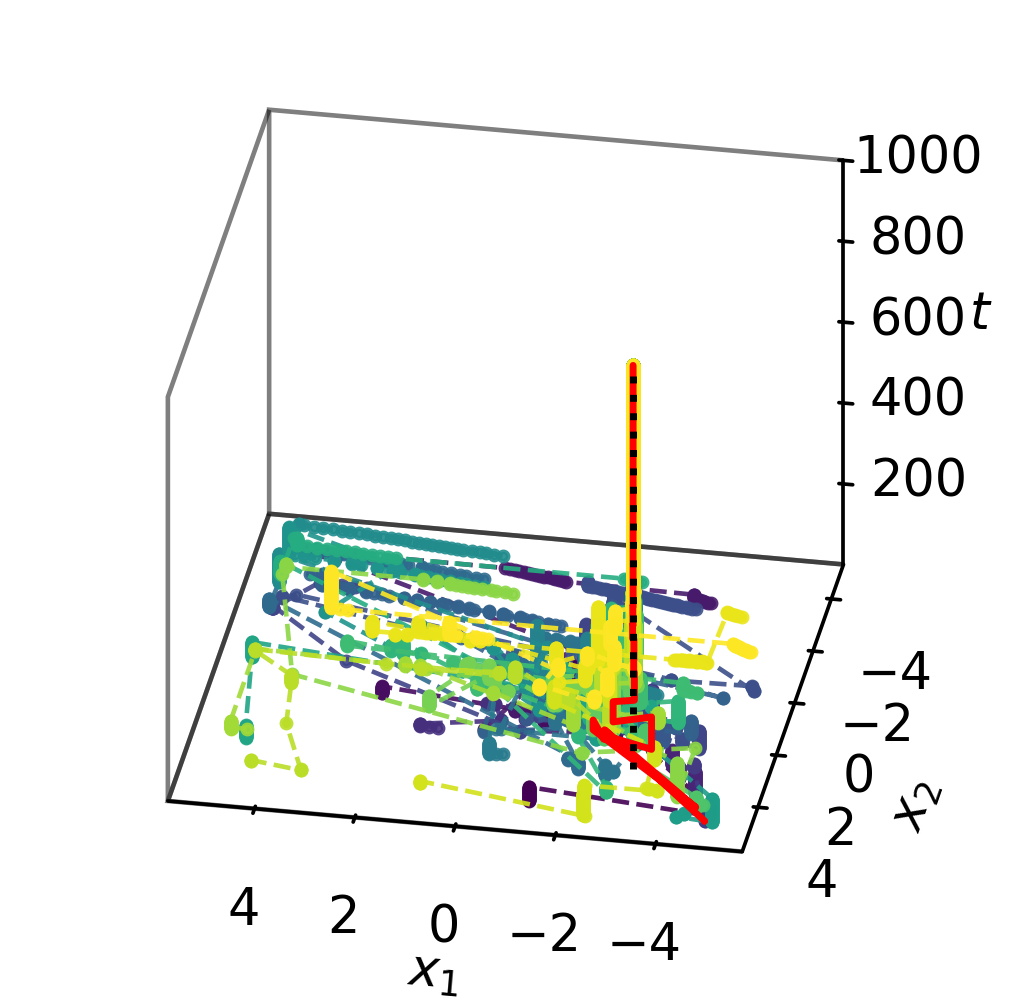} \\[0pt]
    \raisebox{\leRaiser}{\footnotesize\textbf{f20}} &
        \includegraphics[trim={3em 0em 0.6em 0em}, clip, width=\leWidth]{
                    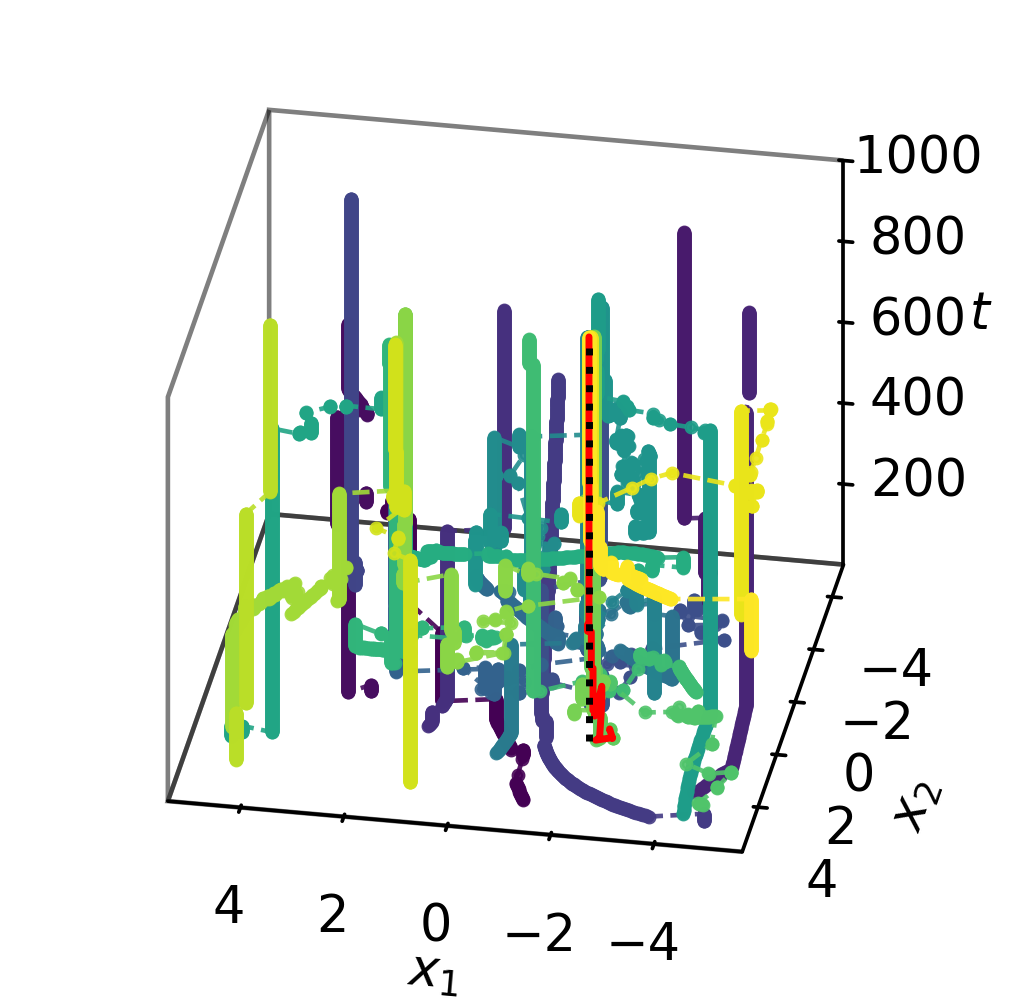} &
        \includegraphics[trim={3em 0em 0.6em 0em}, clip, width=\leWidth]{
                    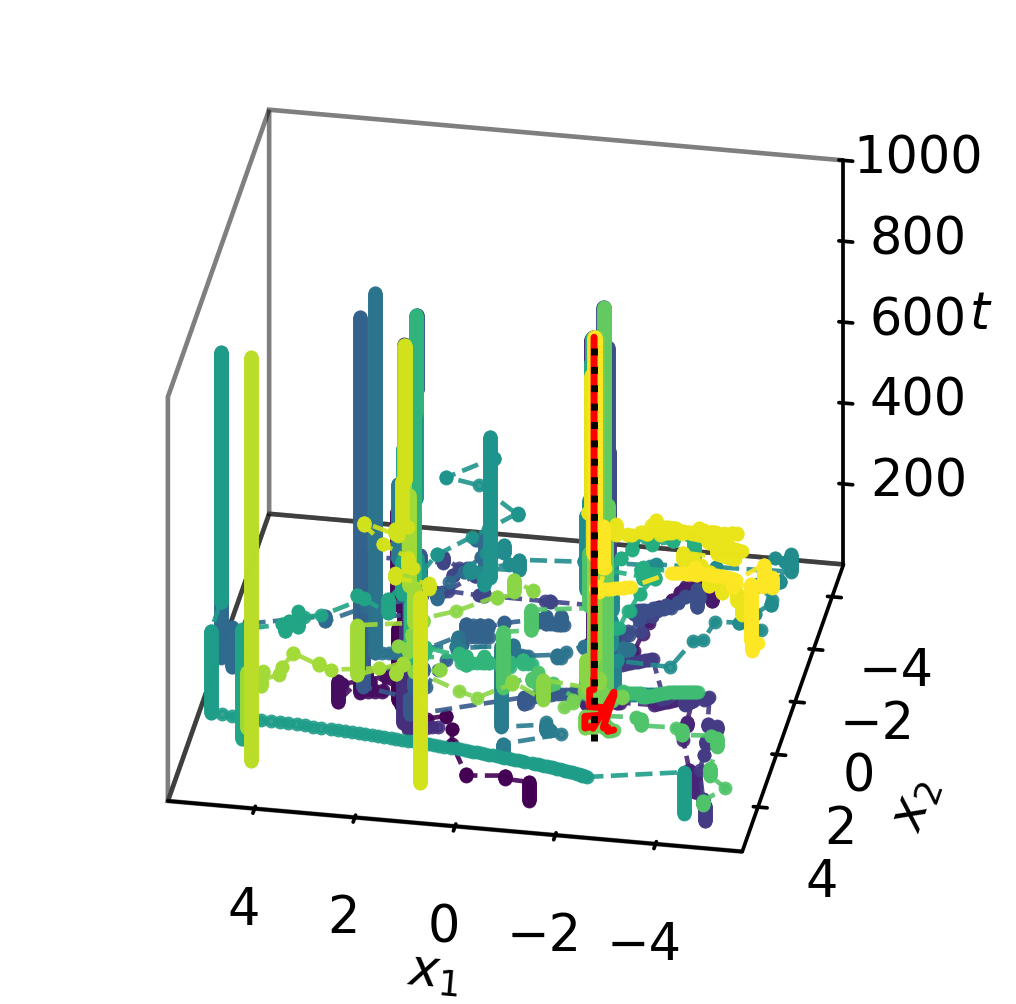} &
        \includegraphics[trim={3em 0em 0.6em 0em}, clip, width=\leWidth]{
                    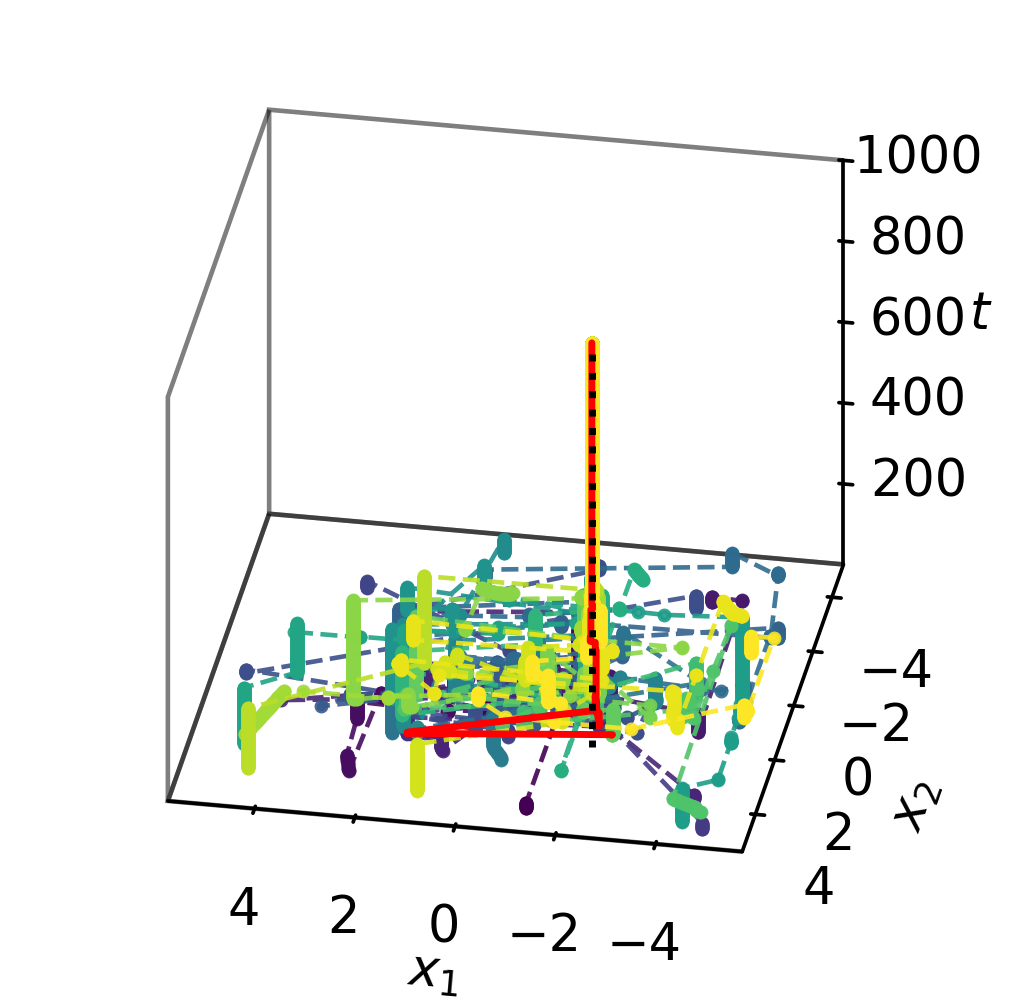} &
        \includegraphics[trim={3em 0em 0.6em 0em}, clip, width=\leWidth]{
                    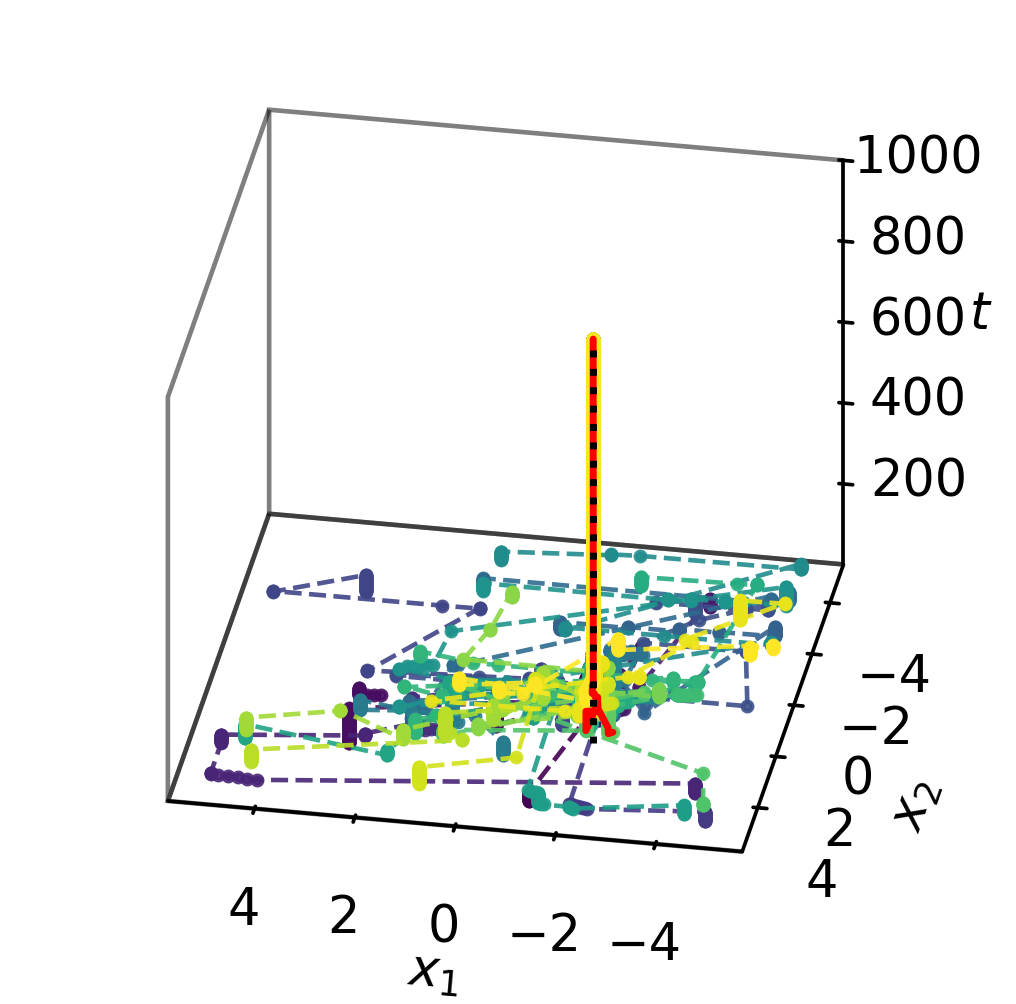} \\[0pt]
    & Linear+Fixed & Izhikevich+Fixed & Linear+DE/... & Izhikevich+DE/...
    \end{tabular}
    \caption{\label{Fig:ExPre:Positions}%
        Particular best positions (\( x_1,x_2\in\pmb{p}_i^t \)) of a 1000-step simulation with two \nha{short}s with 30 \nhu{}s searching on 2D problem domains.
        Each row stands for a function \(f_k\) from the BBOB suite.
        Each column corresponds to \nha{}s (\(h_d\)+\(h_s\)) implemented using the Linear and Izhikevich models as \(h_d\), and the Fixed and DE/\emph{current-to-rand}/1 rules as \(h_s\).
        Dashed black and red lines mark the target and global best.
    }
\end{figure}

Subsequently, \figurename{}~\ref{Fig:ExPre:Positions} presents the temporal evolution of best-so-far positions $(x_1, x_2 \in \pmb{p}^{t}_i)$ for each \nha{} configuration. 
Each vertical segment corresponds to an \nhu{} whose \(\pmb{p}_i^t\) remains unchanged for several steps.
Considering that the dashed black and red lines represent the target \((\pmb{x}_*)\) and the global best \((\pmb{g}^t)\) position evolution, respectively, we then verify that all the implementations approached the target position closely.
From these plots, we notice that the \nha{}s using fixed \(h_s\) exhibit gradual and staircase-like positional shifts, accompanied by extended static segments. 
These patterns are most pronounced in the Linear model and for ill-conditioned (\textbf{f10}) or multimodal (\textbf{f15}) landscapes. 
Besides, Izhikevich dynamics produce abrupt transitions and more localised position clustering, especially during steps of collective firing.
This is explained by the nature of linear systems, which tend to move smoothly in phase space, even when exhibiting swift dynamics dictated by nodal equilibrium points (cf.~\figurename{}~\ref{Fig:2ndSys}). Such a dynamic represents an excellent mechanism for exploitation courses. 
It is observed as static best positions, mainly on fixed rules, coinciding with those slow error plateaux in \figurename{}~\ref{Fig:ExPre:Errors}.
In contrast, the Izhikevich model mostly sweeps only the membrane potential variable, while the auxiliary variable remains invariant (cf.~\figurename{}~\ref{Fig:Izh-UV}). 
Moreover, DE/\emph{current-to-rand}/1 as \(h_s\) induces frequent and spatially coordinated advances. 
This rule drives the entire population towards the target position with vertical transitions and early trajectory collapse. 
With these implementations, we notice that \nha{} using traditional spiking neuron models, such as LIF and Izhikevich in \eqref{Eq:LIF-NUM} and \eqref{Eq:Izhikevich-NUM}, may be insufficient for effectively searching within complex problem landscapes.
While the problems are simple, these results highlight the \nha{}'s capacity to generate distinct behaviours and suggest that certain \(h_d\)+\(h_s\) combinations may require finer tuning to escape stagnation. 
It is worth mentioning that regarding the evolution of \(\pmb{g}^t\) towards \(\pmb{x}_\ast\) in \figurename{}~\ref{Fig:ExPre:Positions}, we can infer that those flatlines in \figurename{}~\ref{Fig:ExPre:Errors}, at least corresponding to the minimum value, can be primarily due to floating point precision issues.

\begin{figure}[!ht]\centering
    \def\leRaiser{20mm}
    \def\leWidth{0.46\linewidth}
    \begin{tabular}{@{}c*{2}{@{ }c}@{}}
    \raisebox{\leRaiser}{\footnotesize\textbf{f1}} &
        \includegraphics[width=\leWidth]{
                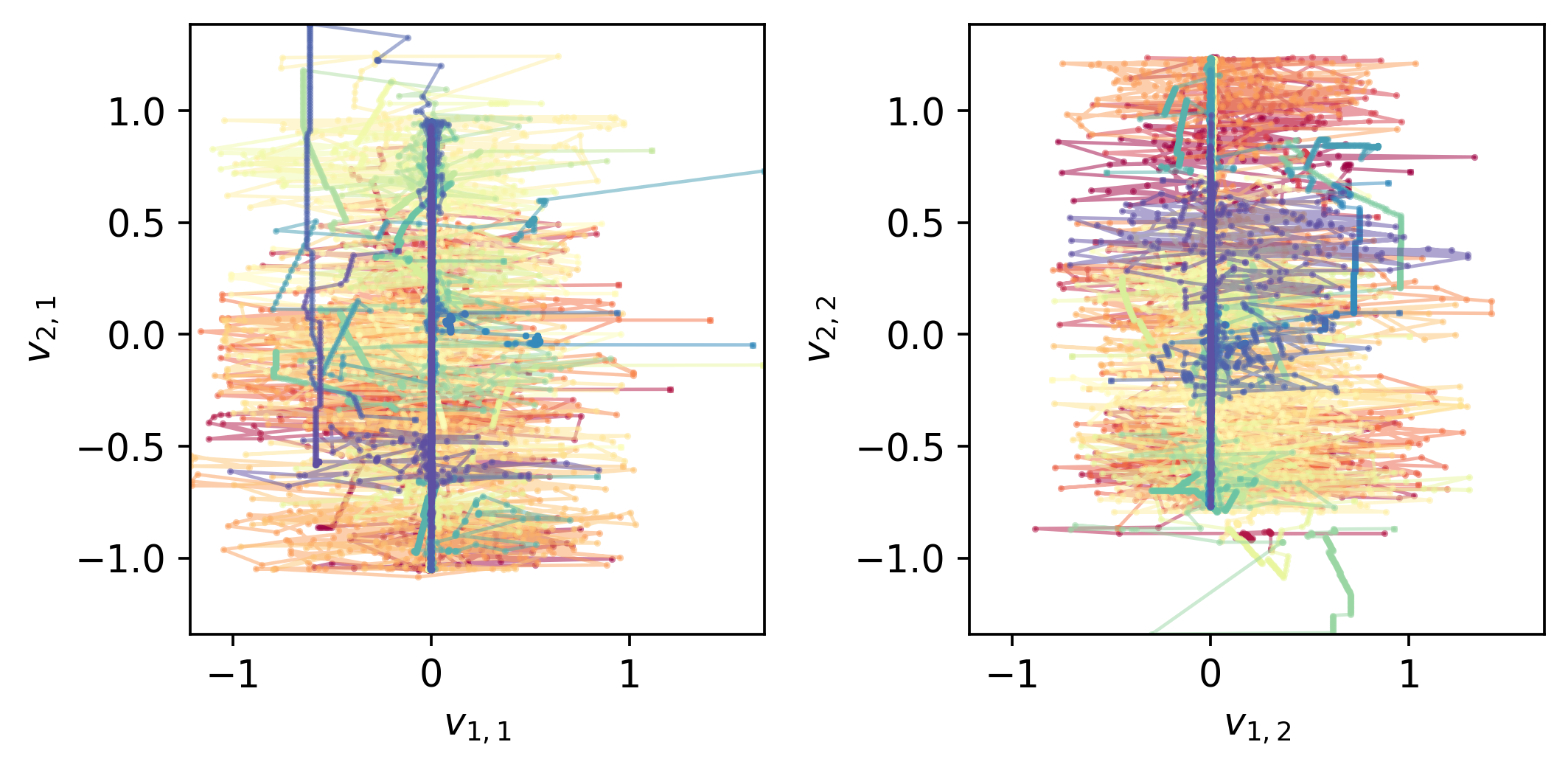} &
        \includegraphics[width=\leWidth]{
                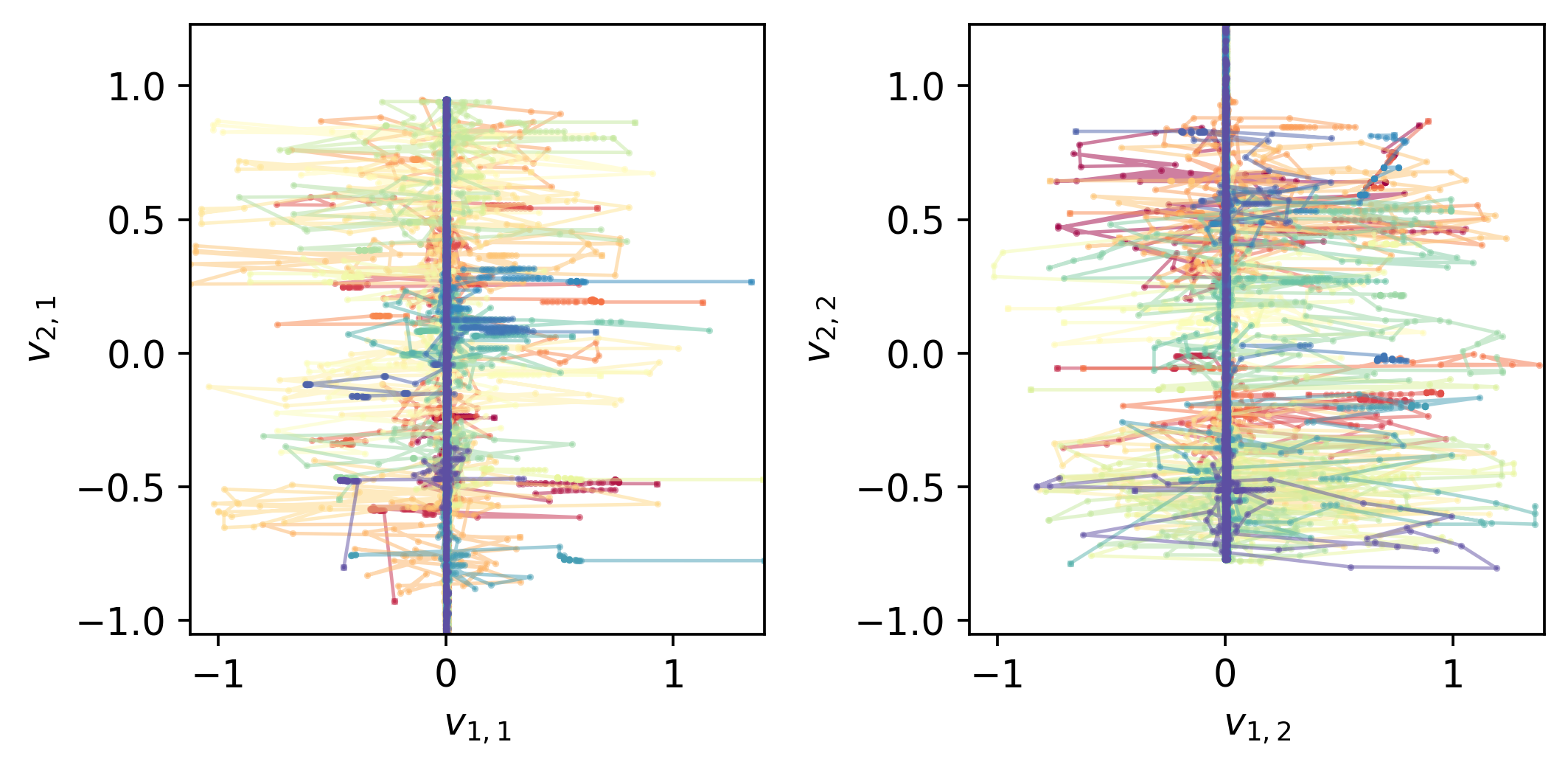} \\
    \raisebox{\leRaiser}{\footnotesize\textbf{f6}} &
        \includegraphics[width=\leWidth]{
                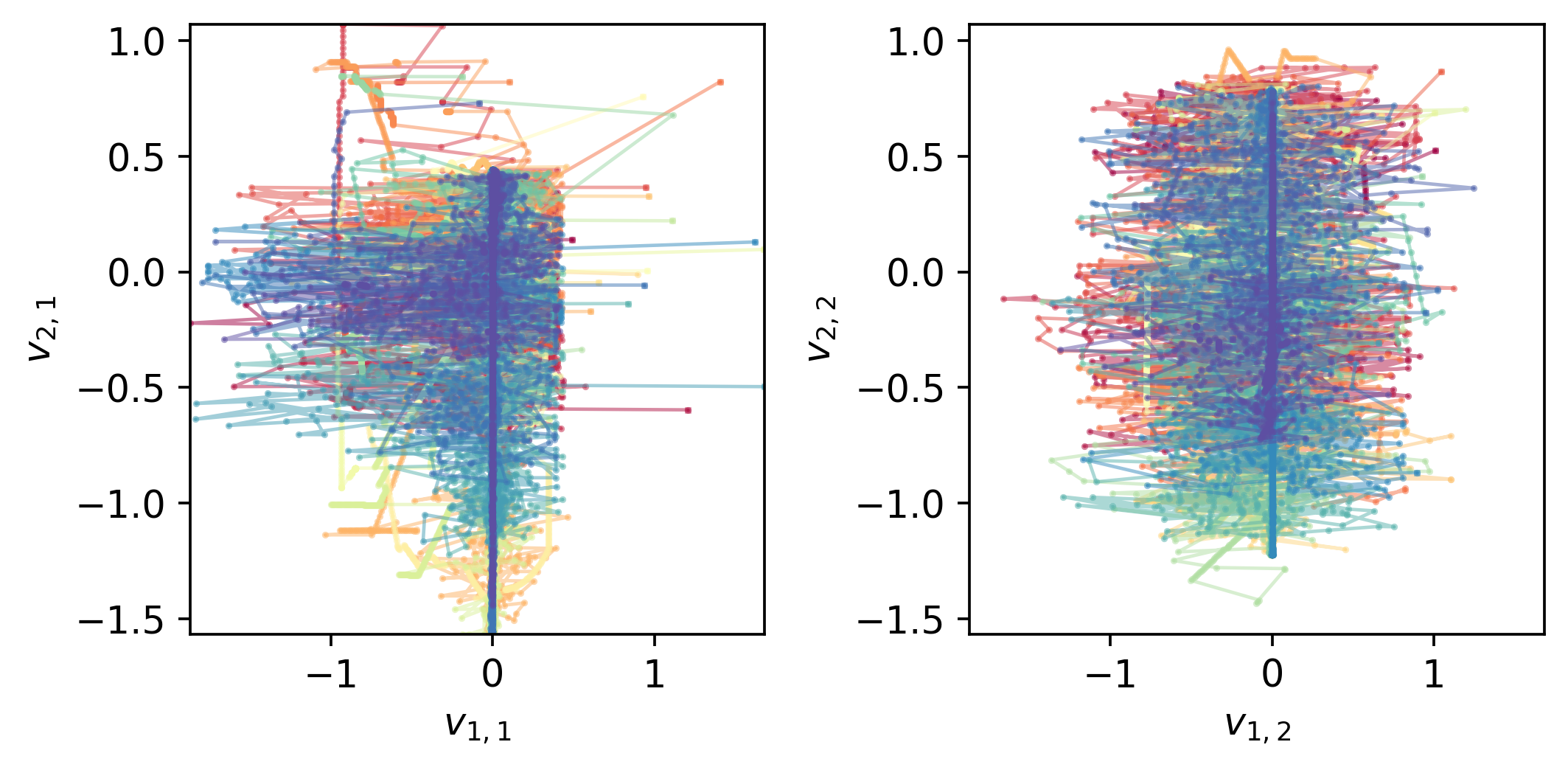} &
        \includegraphics[width=\leWidth]{
                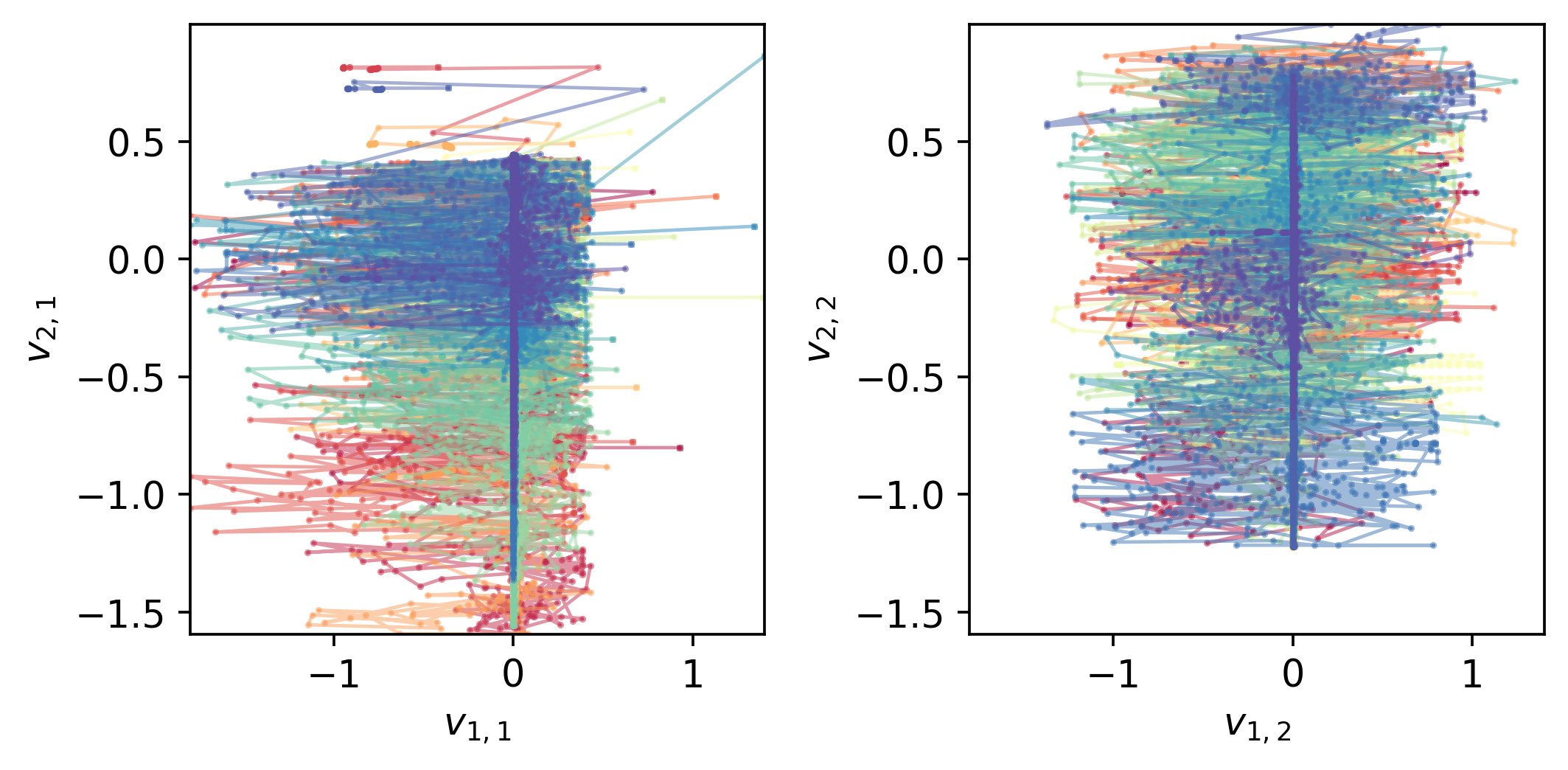} \\
    \raisebox{\leRaiser}{\footnotesize\textbf{f10}} &
        \includegraphics[width=\leWidth]{
                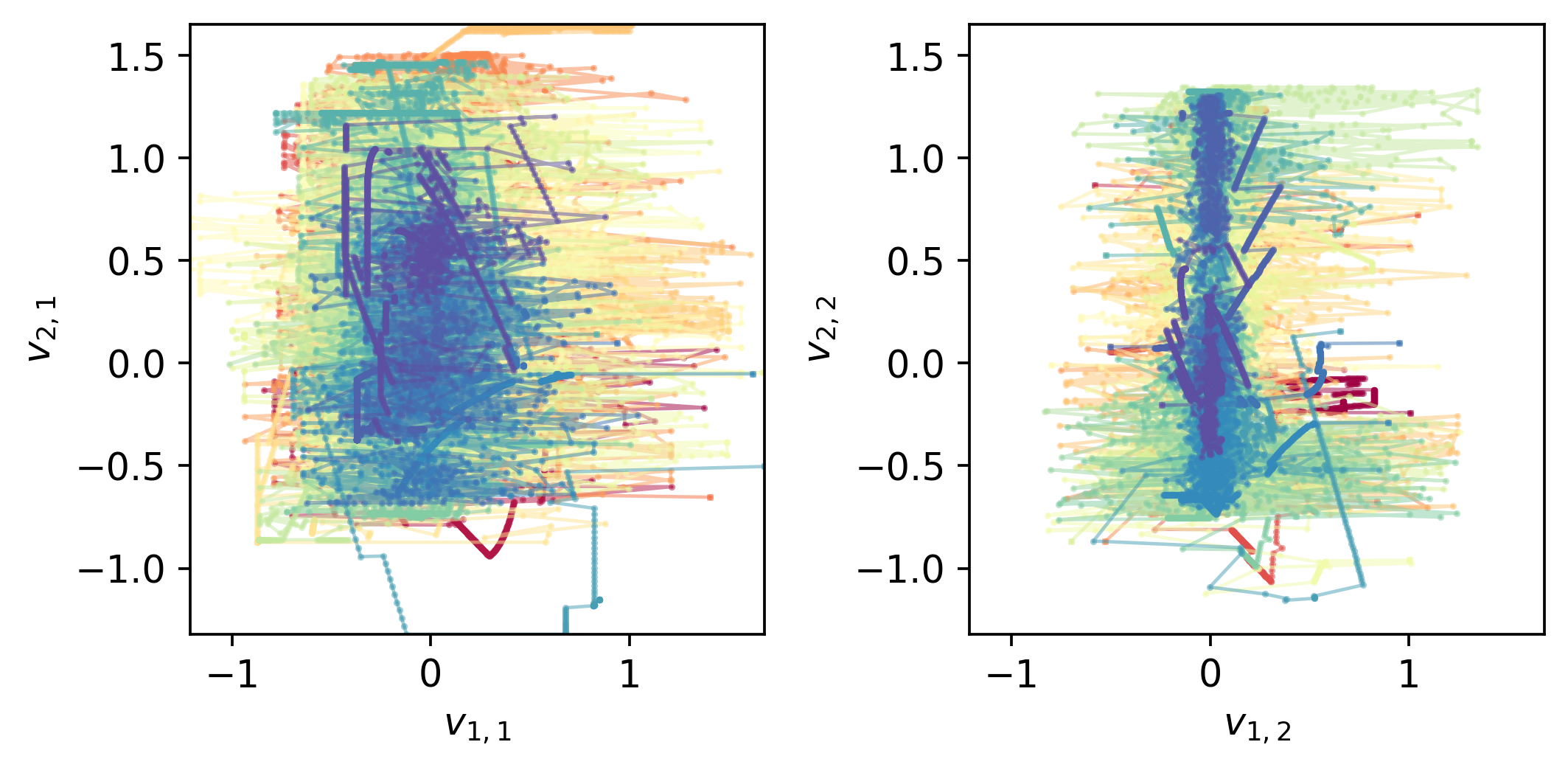} &
        \includegraphics[width=\leWidth]{
                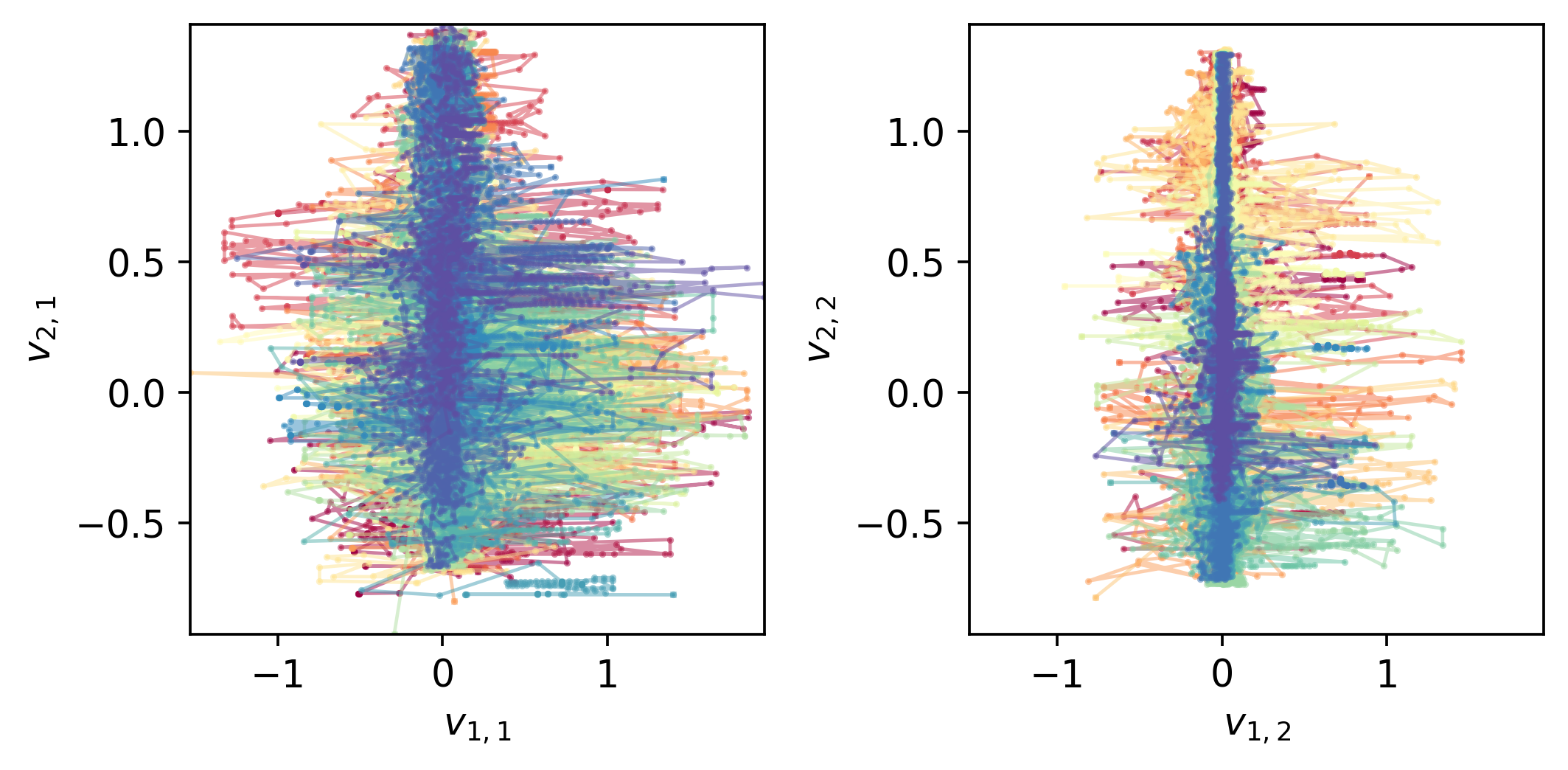} \\
    \raisebox{\leRaiser}{\footnotesize\textbf{f15}} &
        \includegraphics[width=\leWidth]{
                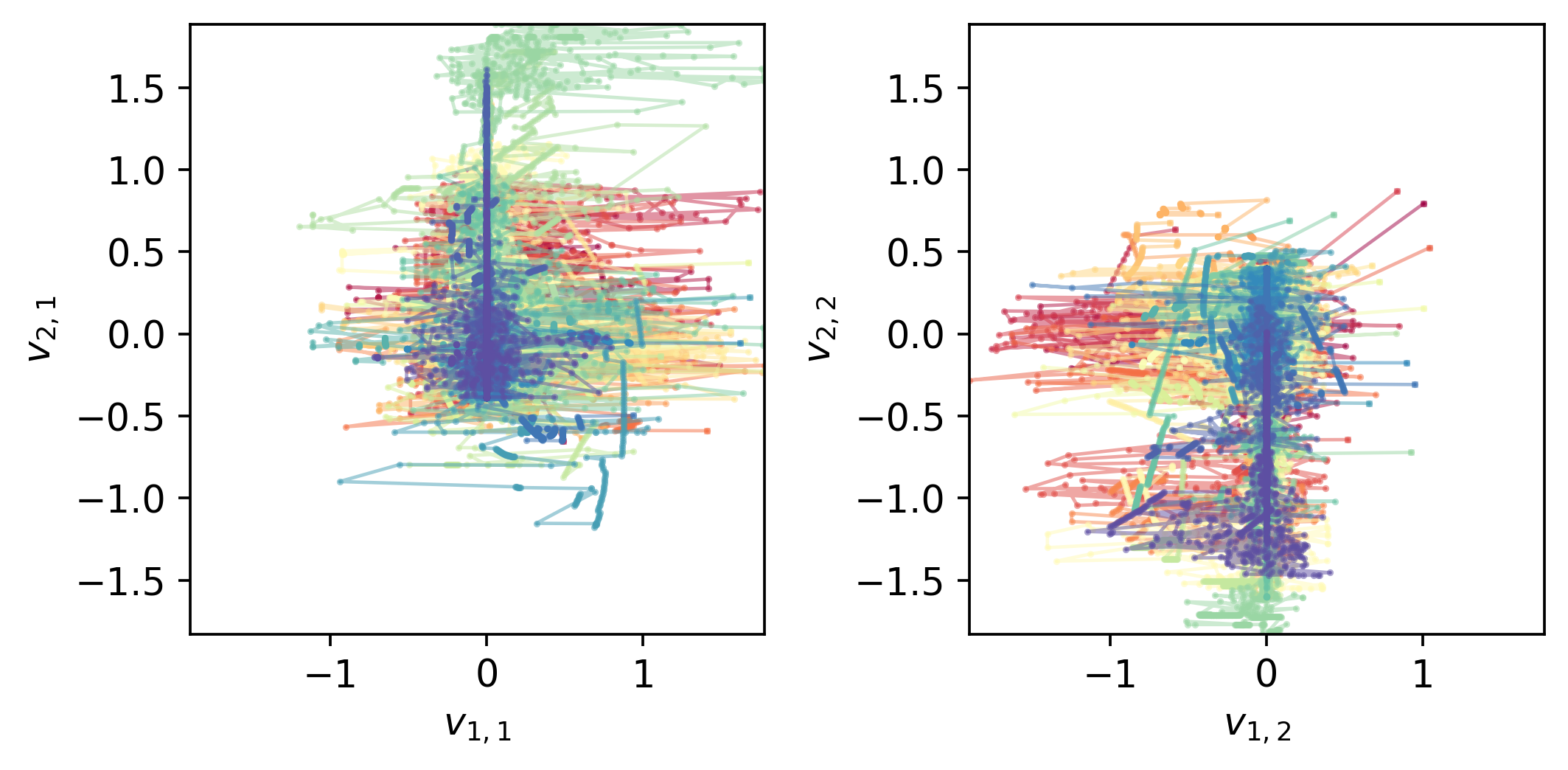} &
        \includegraphics[width=\leWidth]{
                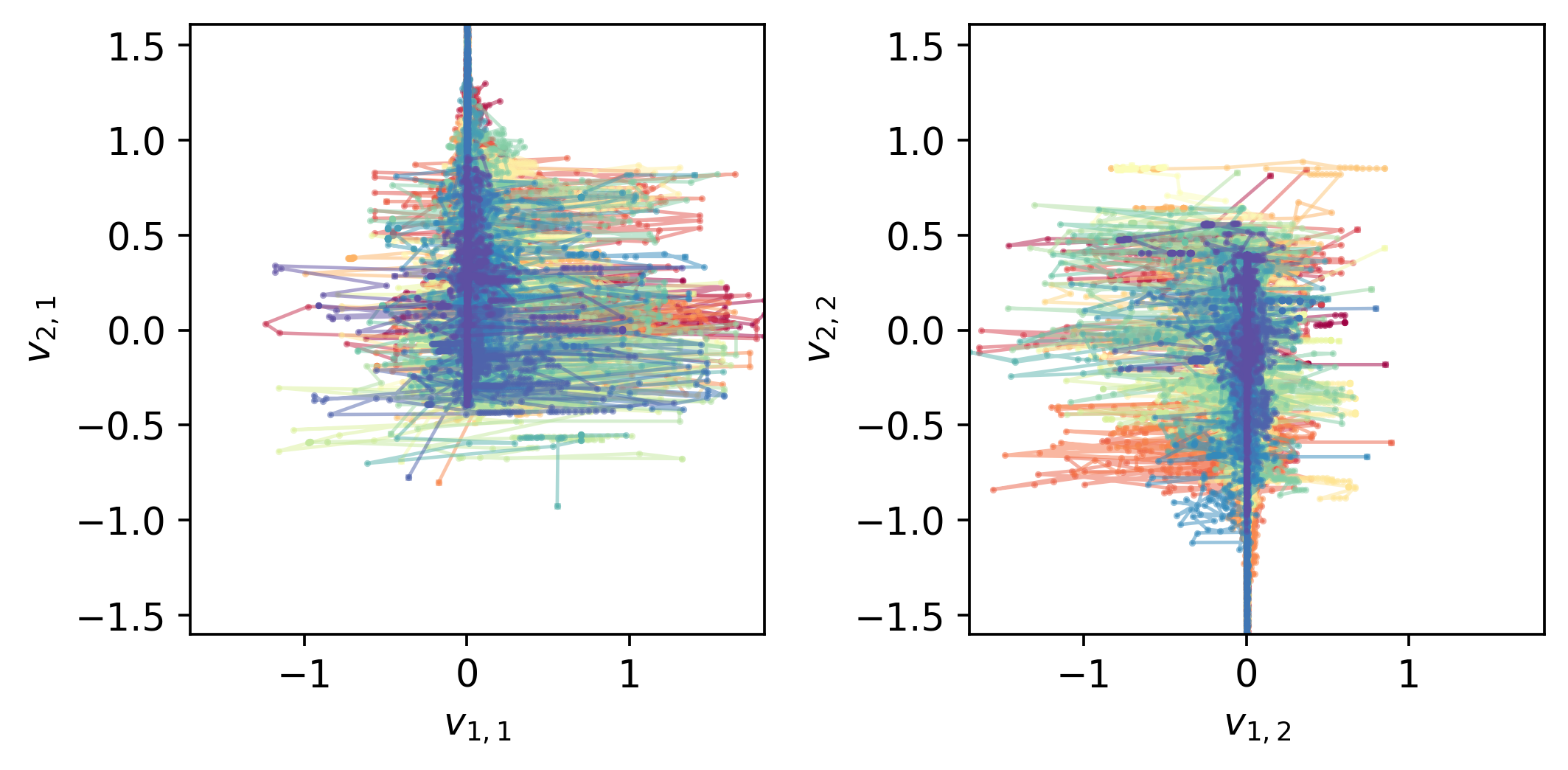} \\
    \raisebox{\leRaiser}{\footnotesize\textbf{f20}} &
        \includegraphics[width=\leWidth]{
                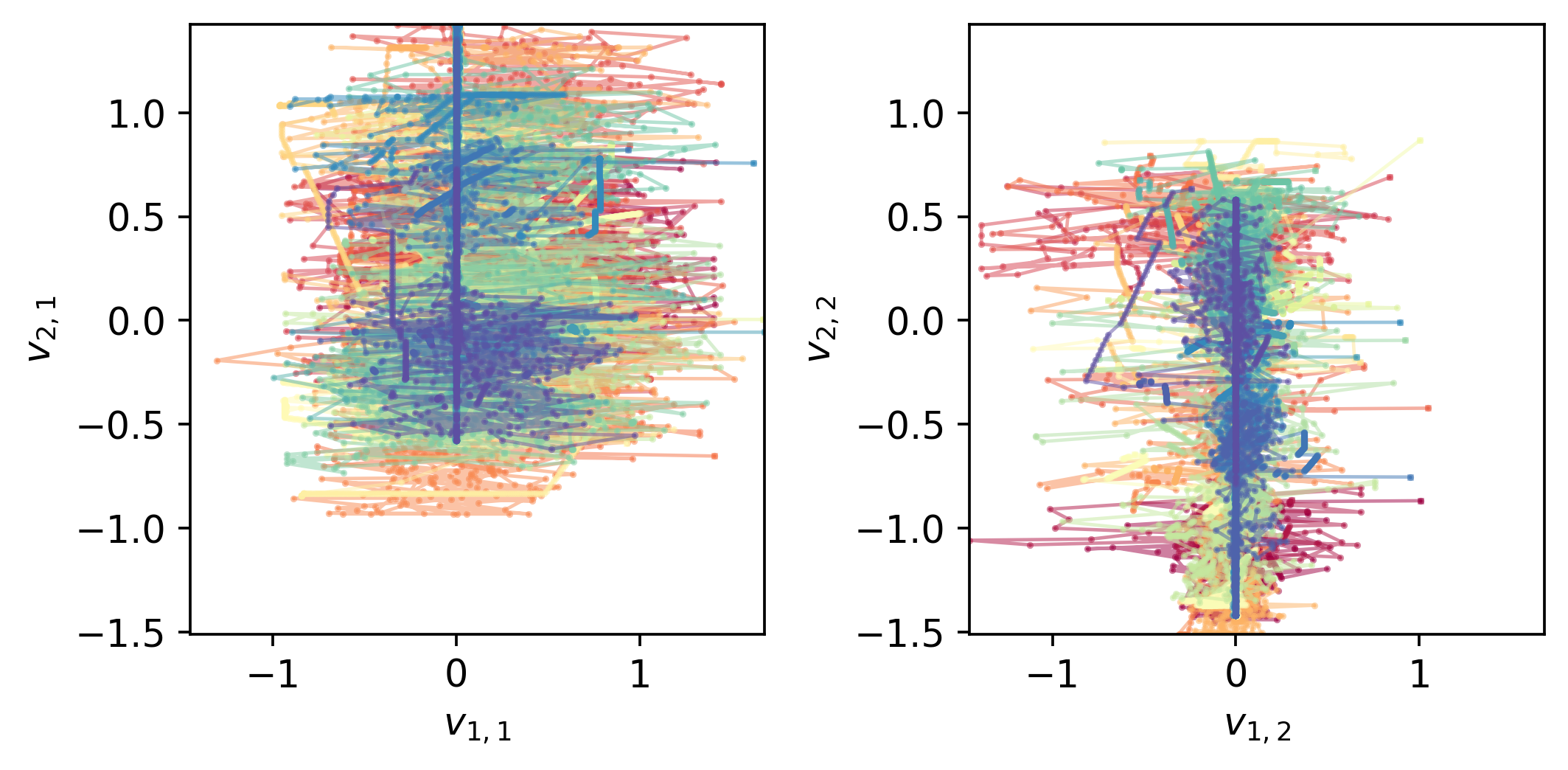} &
        \includegraphics[width=\leWidth]{
                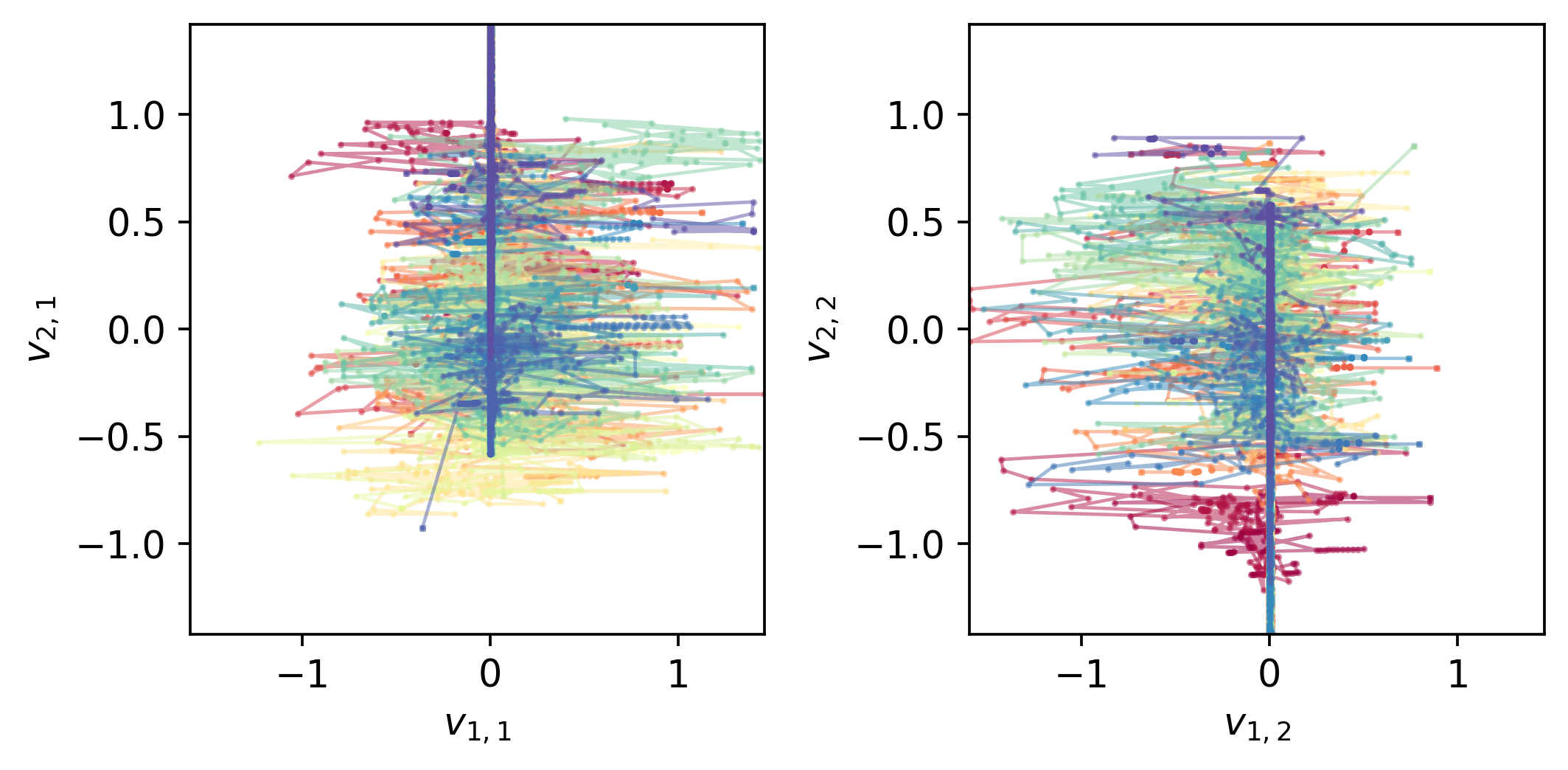} \\
    &   Linear+DE/\emph{current-to-rand}/1  & Izhikevich+DE/\emph{current-to-rand}/1
    \end{tabular}
    \caption{\label{Fig:ExPre:Portraits}%
        Portraits of neuromorphic state variables (\( v_{1,j},v_{2,j}\in\pmb{v}_j^t,\,\forall\ j\)) of a 1000-step simulation with two \nha{short}s with 30 \nhu{}s searching on 2D problem domains.
        Each row stands for a function \(f_k\) from the BBOB suite. Each column corresponds to \nha{}s (\(h_d\)+\(h_s\)) implemented with the Linear and Izhikevich models as \(h_d\), and DE/\emph{current-to-rand}/1 as \(h_s\).
    }
\end{figure}

Although the best positions provide interesting information about the evolutionary process, they are inherently limited by the greedy nature of this feature. For that reason, we analyse the portraits of state variables \((v_{1,j}, v_{2,j} \in \pmb{v}_j)\) for all steps and all \nhu{}s, as \figurename~\ref{Fig:ExPre:Portraits} displays. In this case, we only chose the DE/\emph{current-to-rand}/1 variant for both dynamic models to illustrate the different behaviours controlled mainly by $h_d$. 
While the linear model searches using two degrees of freedom, which is evident in trajectories describing perceptible spiral dynamics, the Izhikevich model focuses mainly on sweeping the membrane potential variables horizontally. In these portraits, the chosen \(\pmb{x}_\text{ref}\) makes \(v_{1,1}\!=\!v_{1,2}\!=\!0\) corresponds to a centring of the neuromorphic representation in the phase plane for each component.
When we compare the state transitions to the positions shown in \figurename~\ref{Fig:ExPre:Positions}, a direct correspondence is evident. In the linear case, the broad exploration of the state space produces scattered and irregular position updates. In the Izhikevich case, the horizontal banding of state trajectories leads to periods where many positions change abruptly at the same step. This results in tight position clustering and collective shifts in the population.

\begin{figure}[!ht]\centering
\def\leRaiser{20mm}
\def\leWidth{0.45\linewidth}
\begin{tabular}{@{}c*{2}{@{ }c}@{}}
    \raisebox{\leRaiser}{\footnotesize\textbf{f1}} &
        \includegraphics[width=\leWidth]{
            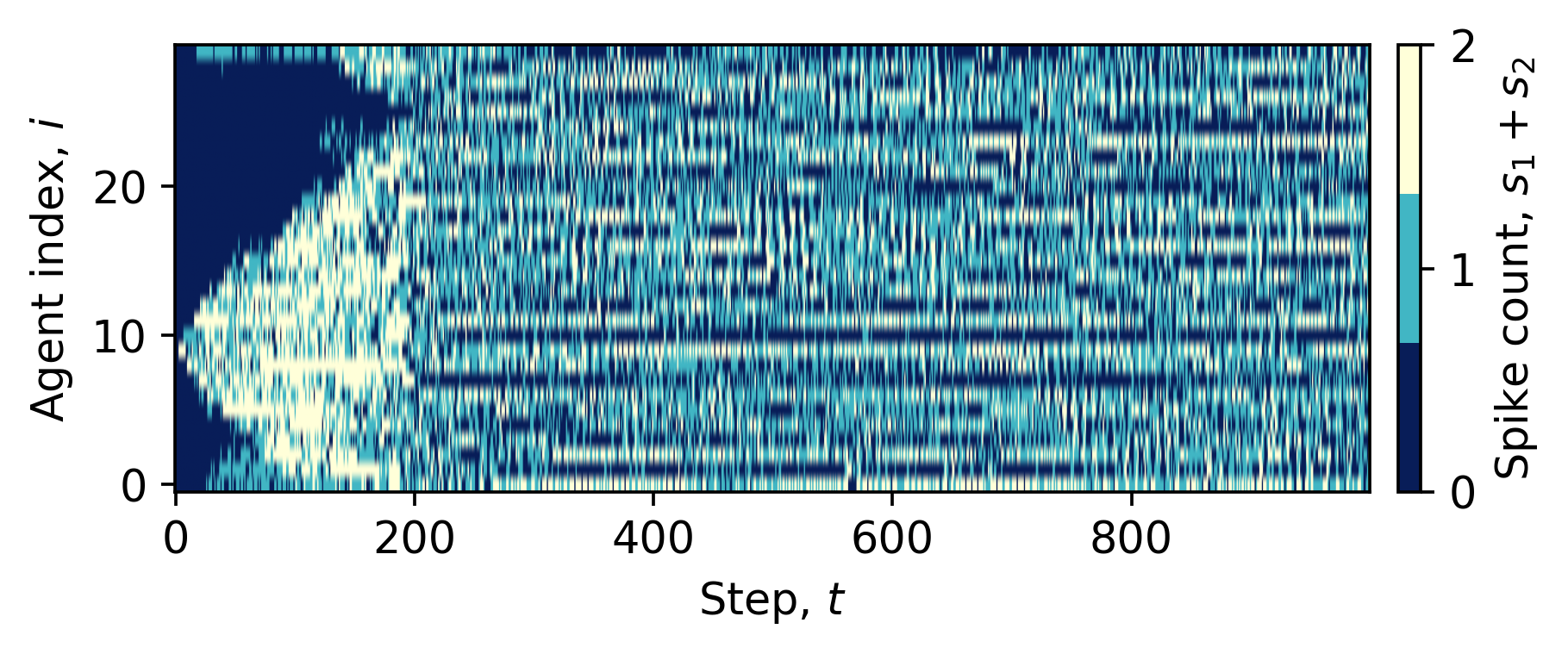} &
        \includegraphics[width=\leWidth]{
            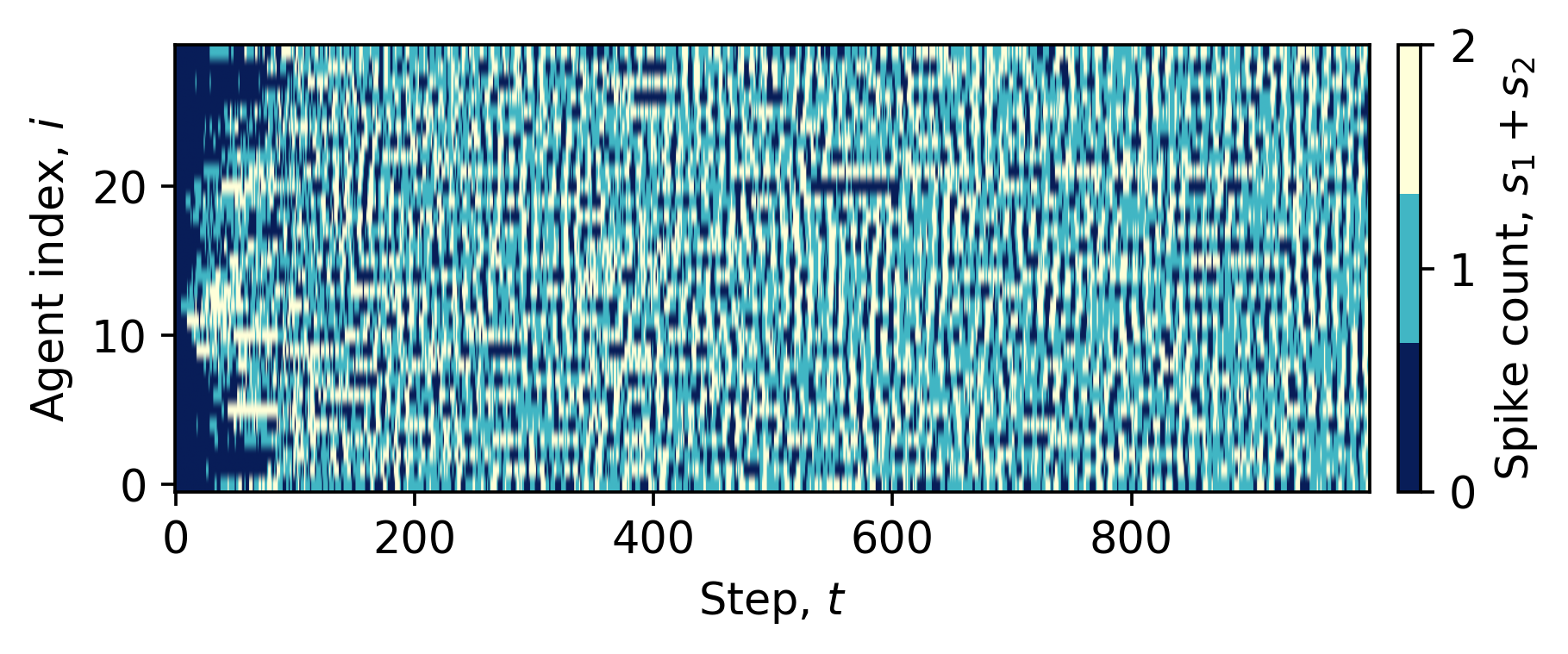} \\
    \raisebox{\leRaiser}{\footnotesize\textbf{f6}} &
        \includegraphics[width=\leWidth]{
            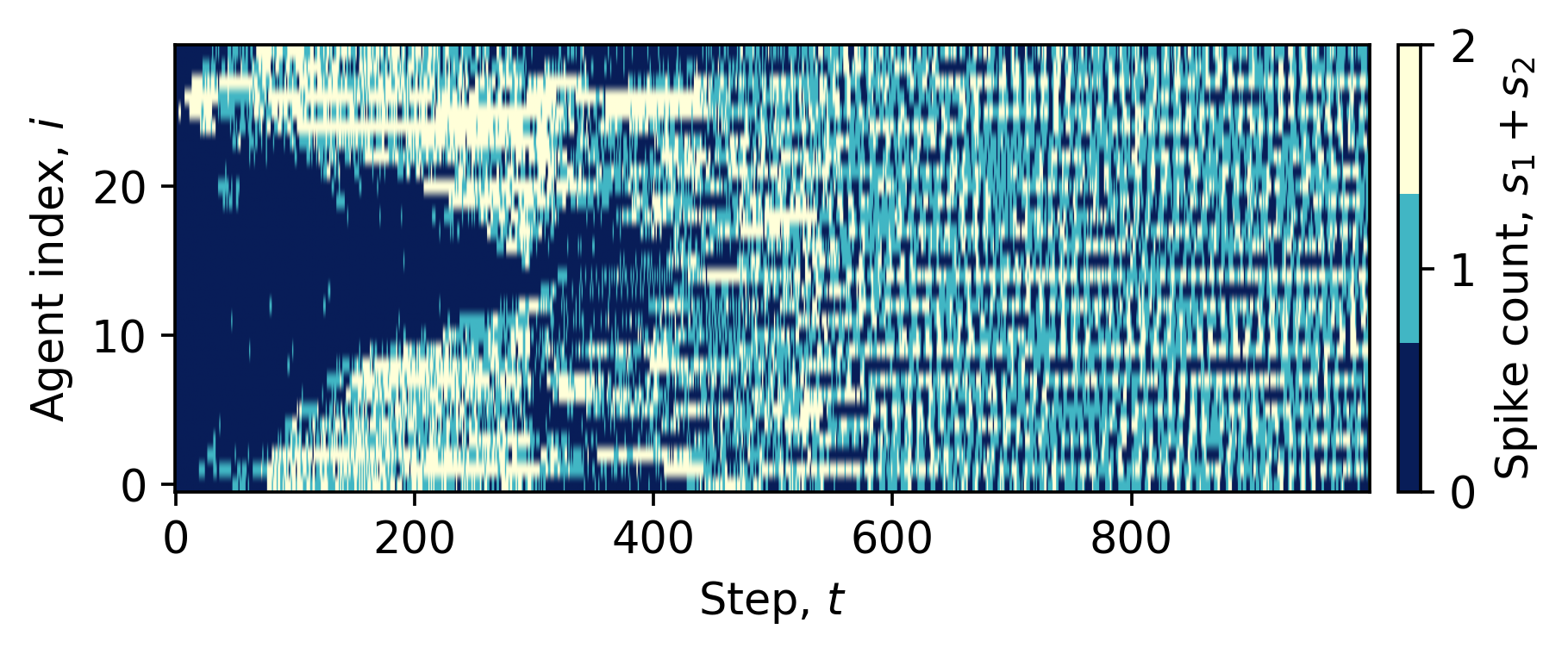} &
        \includegraphics[width=\leWidth]{
            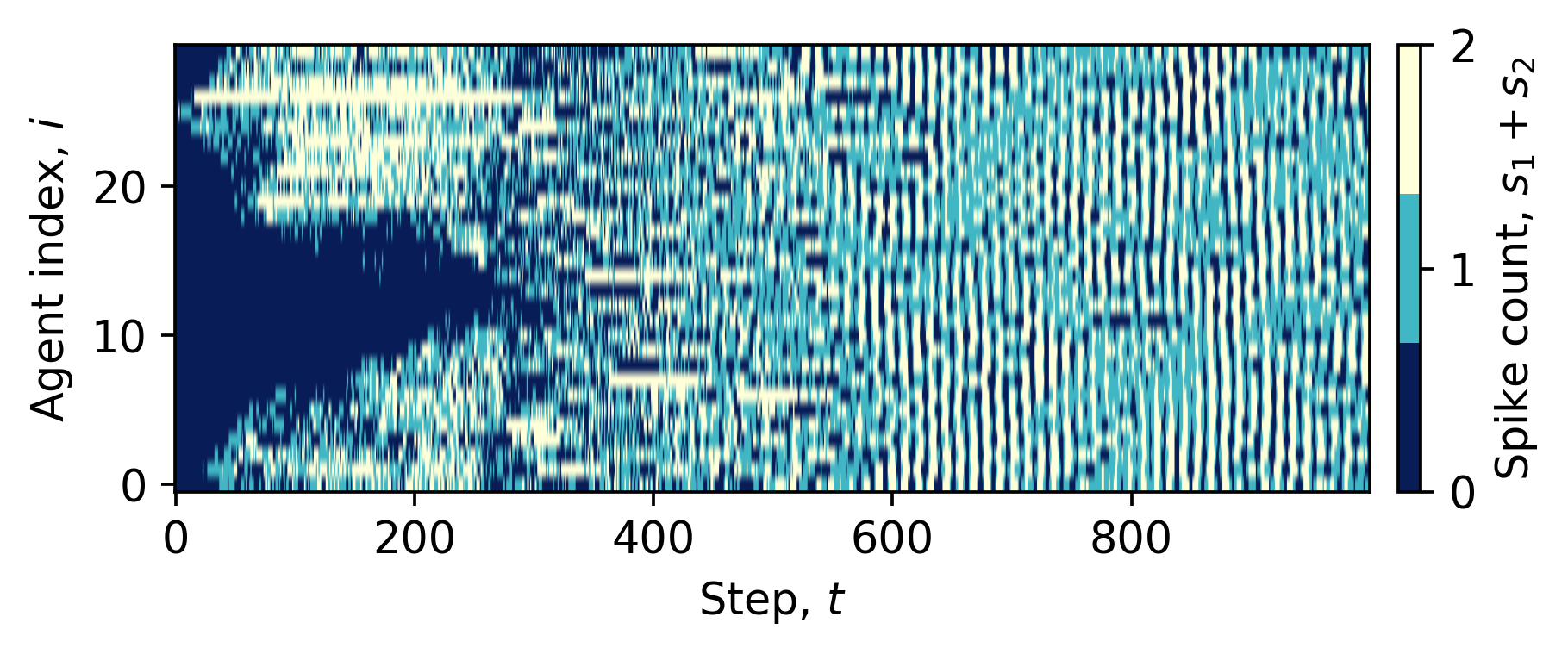} \\
    \raisebox{\leRaiser}{\footnotesize\textbf{f10}} &
        \includegraphics[width=\leWidth]{
            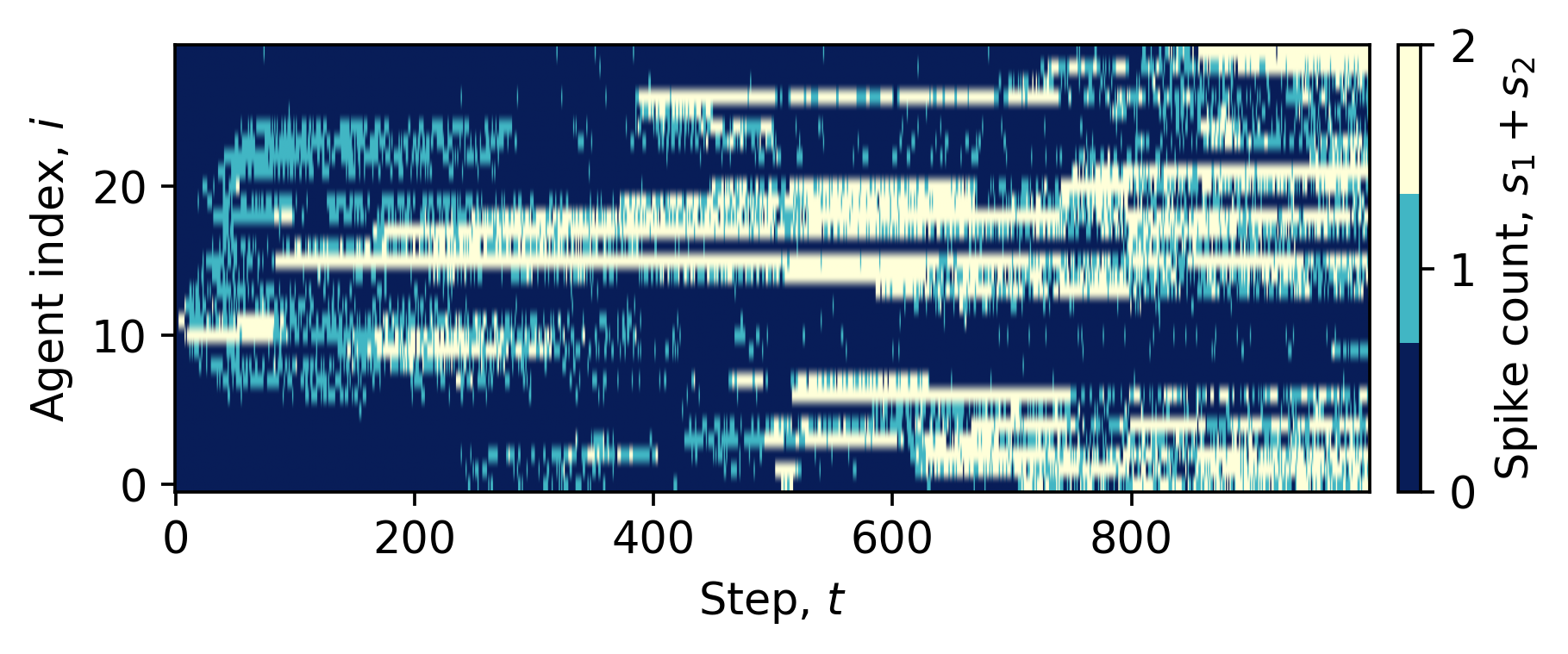} &
        \includegraphics[width=\leWidth]{
            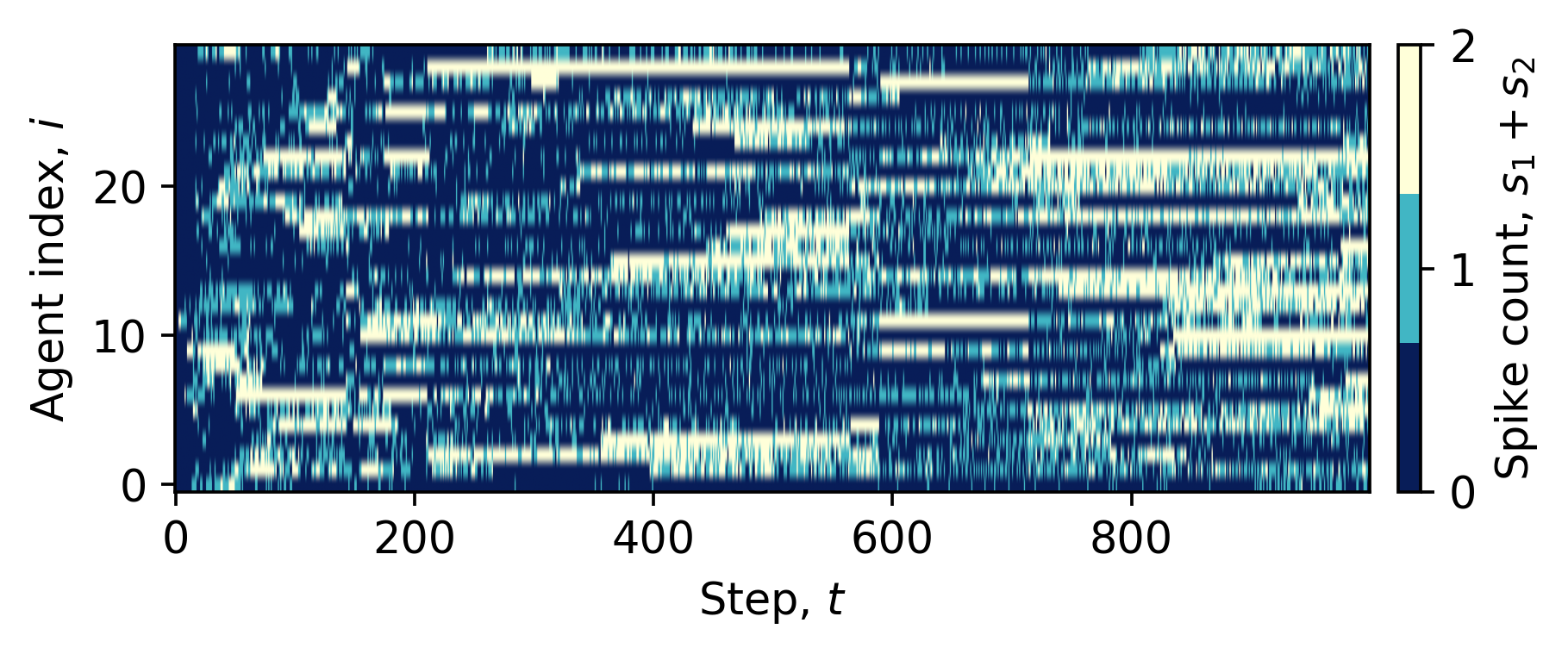} \\
    \raisebox{\leRaiser}{\footnotesize\textbf{f15}} &
        \includegraphics[width=\leWidth]{
            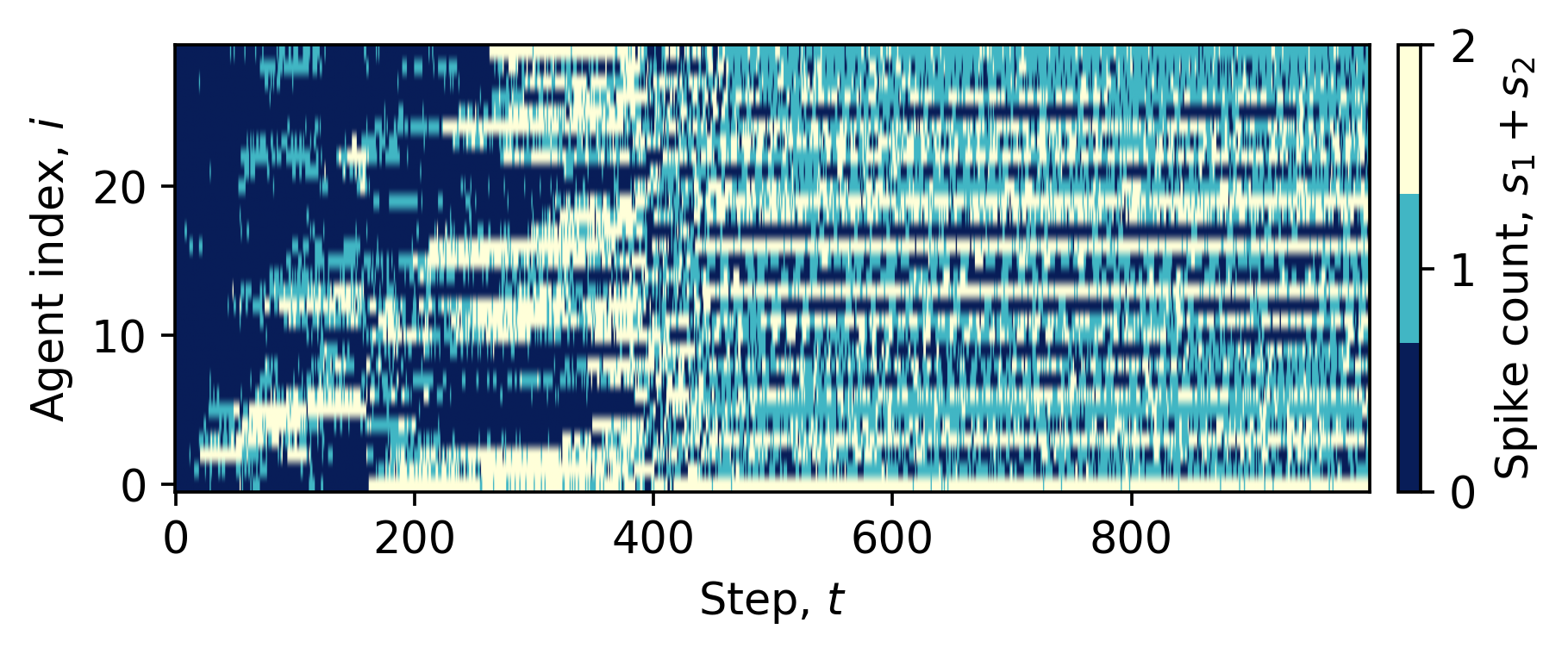} &
        \includegraphics[width=\leWidth]{
            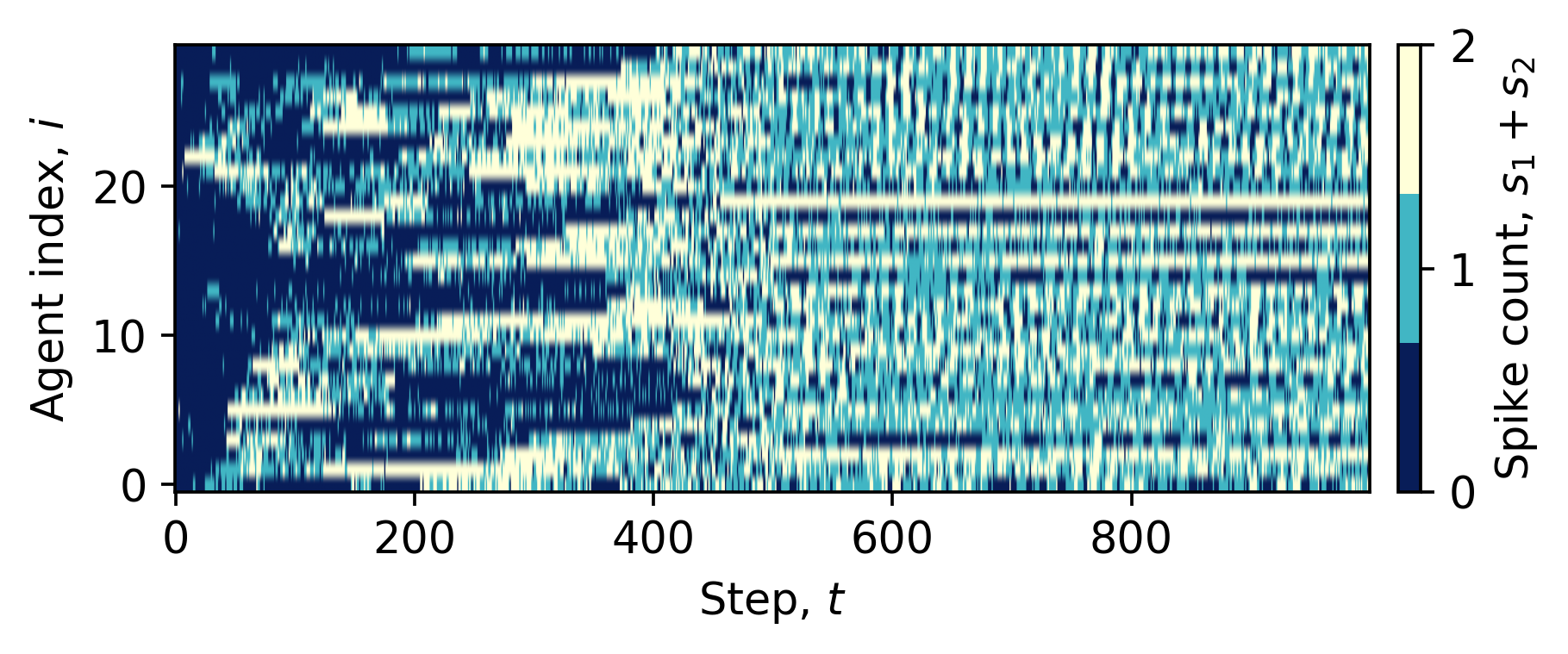} \\
    \raisebox{\leRaiser}{\footnotesize\textbf{f20}} &
        \includegraphics[width=\leWidth]{
            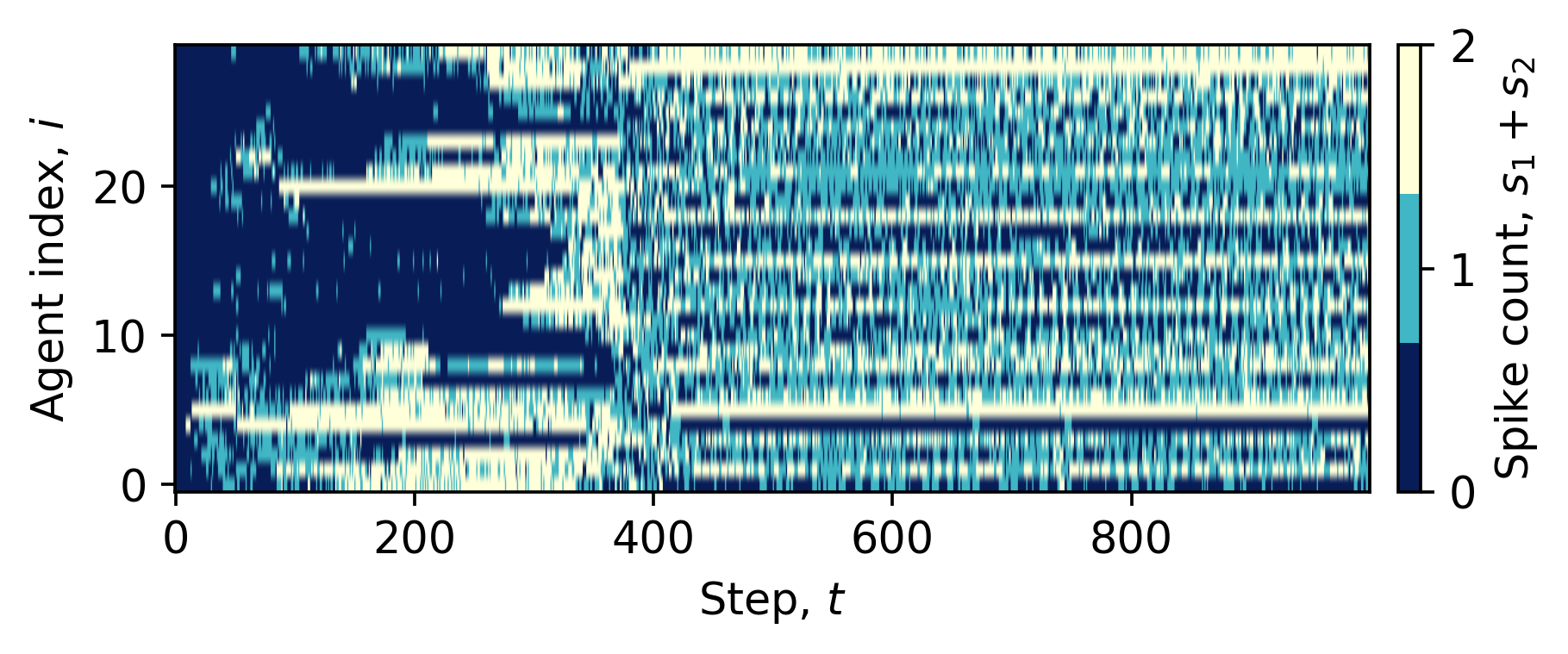} &
        \includegraphics[width=\leWidth]{
            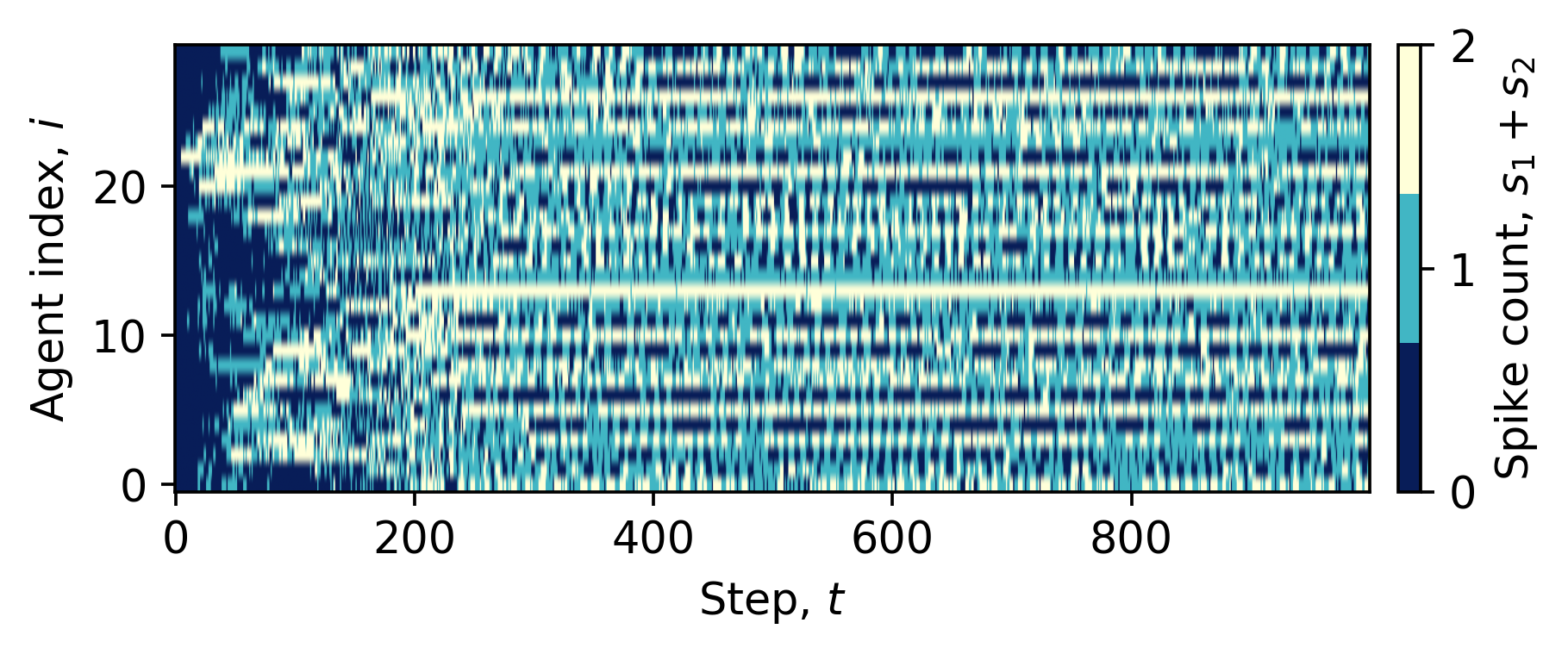} \\
    &   Linear+DE/\emph{current-to-rand}/1 & Izhikevich+DE/\emph{current-to-rand}/1
\end{tabular}
\caption{\label{Fig:ExPre:Spikes}%
    Spiking signal count (\( s_1 + s_2 \)) of a 1000-step simulation with two \nha{short}s with 30 \nhu{}s, each one with an index \(i\), searching on 2D problem domains.
    Each row stands for a function \(f_k\) from the BBOB suite.
    Each column corresponds to \nha{}s (\(h_d\)+\(h_s\)) implemented with the Linear and Izhikevich models as \(h_d\), and DE/\emph{current-to-rand}/1 as \(h_s\).
}
\end{figure}

Lastly, we complement the analysis of the internal \nha{} dynamics by examining the spike activity generated during the search. \figurename~\ref{Fig:ExPre:Spikes} displays the temporal distribution of spike counts per \nhu{} for the previously analysed configurations.
In all the spiking maps, we notice that slight triangular patterns arise on the first steps from the bidirectional ring topology implemented for the SNN via \(\pmb{W}_s\).
This spike event propagation is more evident on \nhu{}s using the linear model because of their smooth dynamic in the space state discussed in \figurename{}~\ref{Fig:ExPre:Portraits}.
For the Izhikevich configuration, frequent vertical bands reflect near-simultaneous firing of multiple units, caused by the nonlinear dynamics. 
In the case of \textbf{f1}, the Izhikevich model reaches the shrunk spiking condition set $\mathcal{S}$ earlier than the linear variant, reflecting a faster contraction of the state variables and an earlier onset of collective spiking.
Besides, for \textbf{f10} from the ill-conditioned functions, we observe that the spike activity is more fragmented in both configurations, with fewer instances of broad synchrony across the population. The presence of persistent diagonal, but narrow, bands in the linear case and irregular vertical patterns in the Izhikevich case indicates that units reach the spiking condition at different times. This is consistent with slower and less coordinated progress in both state and position trajectories for this challenging landscape (cf.~\figurename{}~\ref{Fig:ExPre:Positions} and \ref{Fig:ExPre:Portraits}).
From this analysis, we corroborate that the spiking condition set directly adapts the behaviour of each \nhu{} during the search by controlling when rules are triggered.

The subsequent experiment comprised three \nha{} variants across all BBOB function classes and dimensionalities.
In this experiment, Neuropt-Lin and Neuropt-Izh correspond to those implementations using the Linear and Izhikevich models as \(h_d\) and DE/\emph{current-to-rand}/1 as \(h_s\), as examined in the previous section. 
Neuropt-Hyb denotes a heterogeneous variant comprising equal proportions of both core models within the population. 

\begin{figure}[!ht]
    \centering\hfill%
    \subfloat[Neuropt-Lin]{%
        \includegraphics[width=0.33\linewidth]{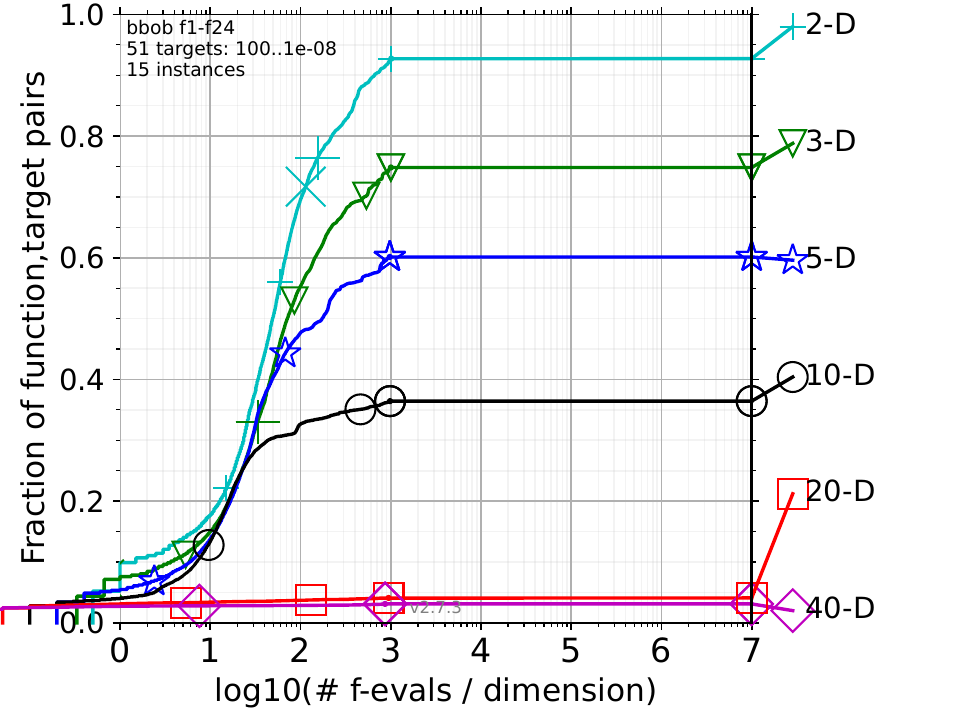}%
        }\hfill%
    \subfloat[Neuropt-Izh]{%
        \includegraphics[width=0.33\linewidth]{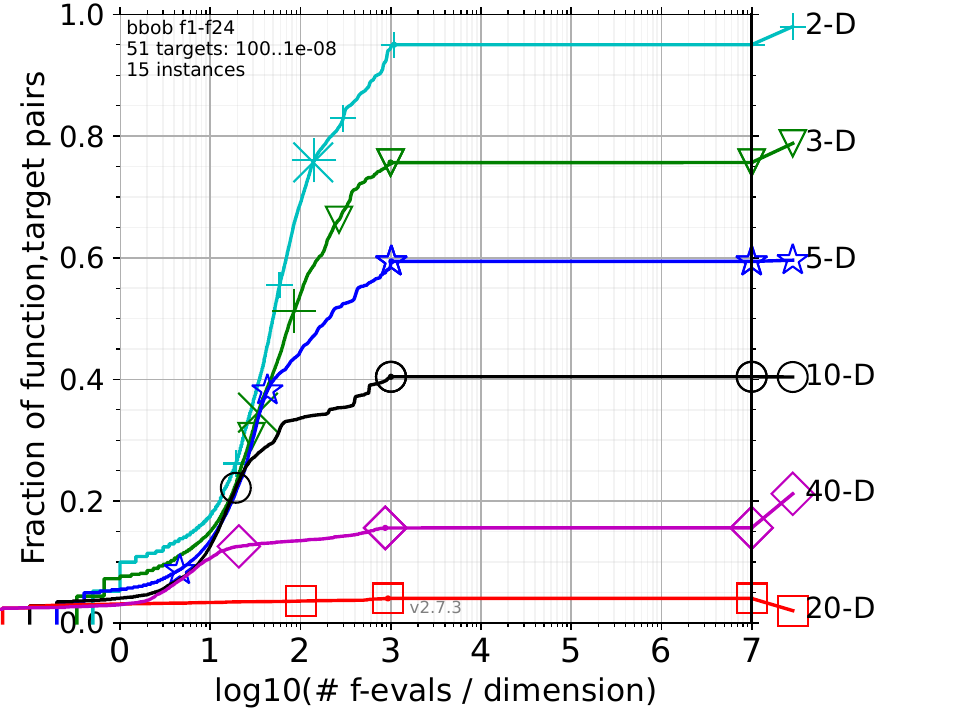}%
        }\hfill%
    \subfloat[Neuropt-Hyb]{%
        \includegraphics[width=0.33\linewidth]{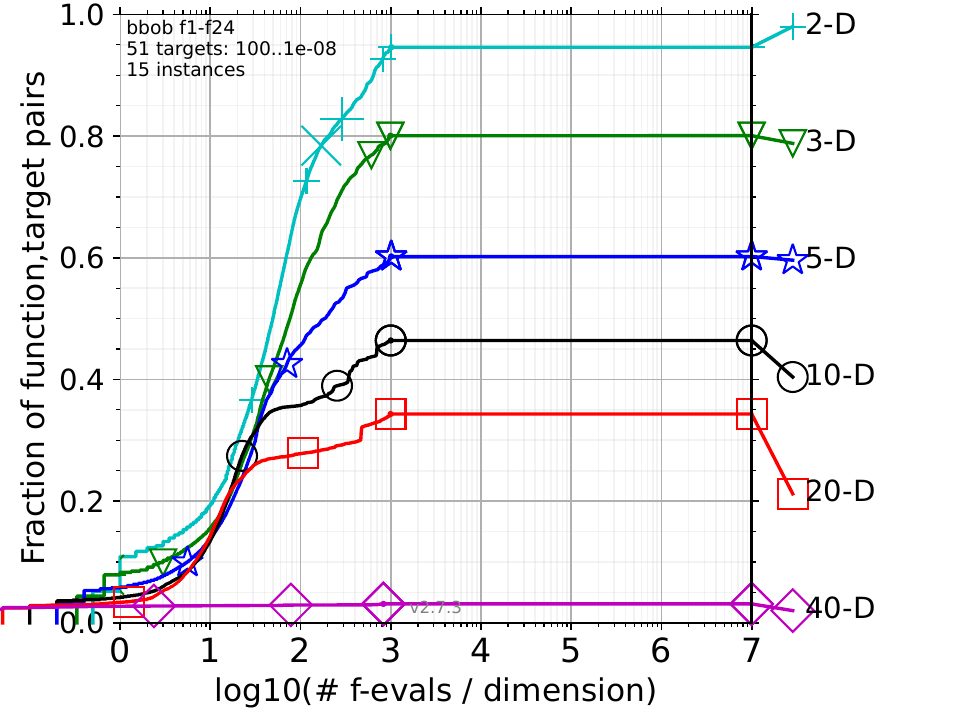}%
        }\hfill%
    \caption{\label{Fig:ExConf:ECD}%
    Empirical Cumulative Distribution Functions (ECDFs) of normalised runtimes for the three \nha{} implementations on the BBOB test suite. Each curve shows the fraction of function-target pairs solved as a function of \(f\)-evaluations per dimension, for 51 targets \(10^{[-8..2]}\) over all functions, instances, and dimensions. Higher curves indicate more efficient convergence.
    \cocoversion{}
    }
\end{figure}

\figurename~\ref{Fig:ExConf:ECD} displays the Empirical Cumulative Distribution Functions (ECDFs) of normalised runtimes for all function groups and all dimensions.
In general, we notice that all variants achieve a high fraction of solved function–target pairs for low-dimensional cases (\ie up to 5D) with a reduced evaluation budget.
In the intermediate dimensionality, Neuropt-Izh consistently exhibits faster convergence than Neuropt-Lin. At the same time, Neuropt-Hyb maintains performance comparable to the best homogeneous variant and reveals increased robustness as dimensionality grows.
For 20D and 40D, all approaches show reduced success, with the fraction of solved pairs flattening well below one. 
The plateaux observed in the results arise from the constrained evaluation budget employed in the simulations rather than from any intrinsic limitation of the \nha{} variants. This effect becomes more evident in the 20D and 40D settings. Despite this, the overall trend across dimensions reveals robust and scalable convergence, confirming the potential of the proposed spike-driven asynchronous mechanisms for optimisation problems.

\begin{figure}[!ht]\centering
\def\leRaiser{15mm}
\def\leWidth{0.3\linewidth}
\begin{tabular}{@{}c*{3}{@{ }c}@{}}
    \raisebox{\leRaiser}{\footnotesize\rotatebox{90}{\(d=5\)}} &
        \includegraphics[width=\leWidth]{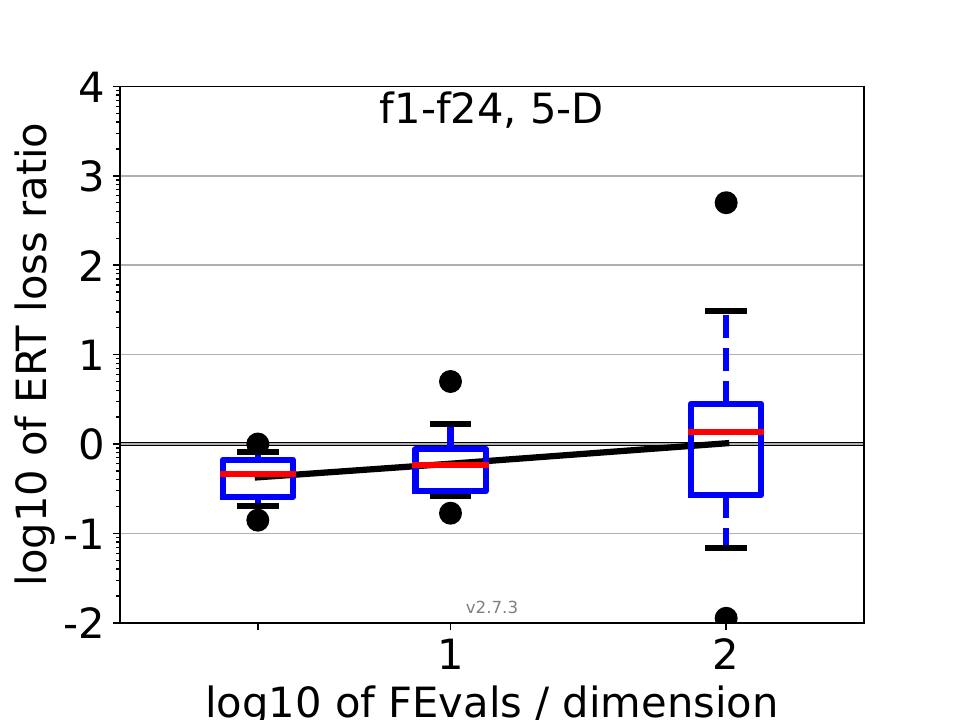} &
        \includegraphics[width=\leWidth]{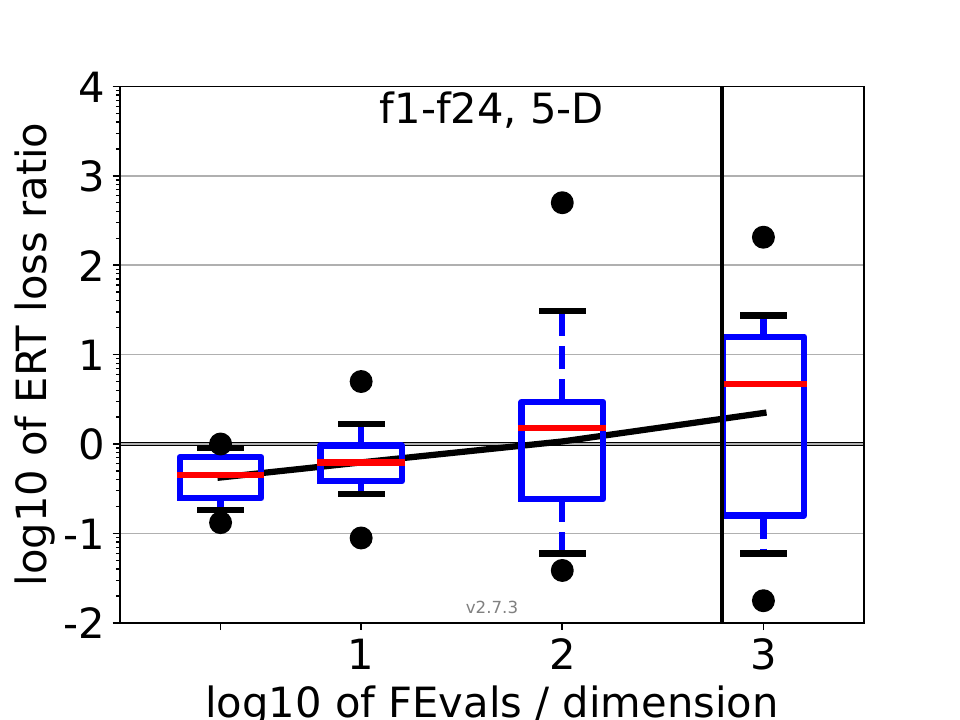} &
        \includegraphics[width=\leWidth]{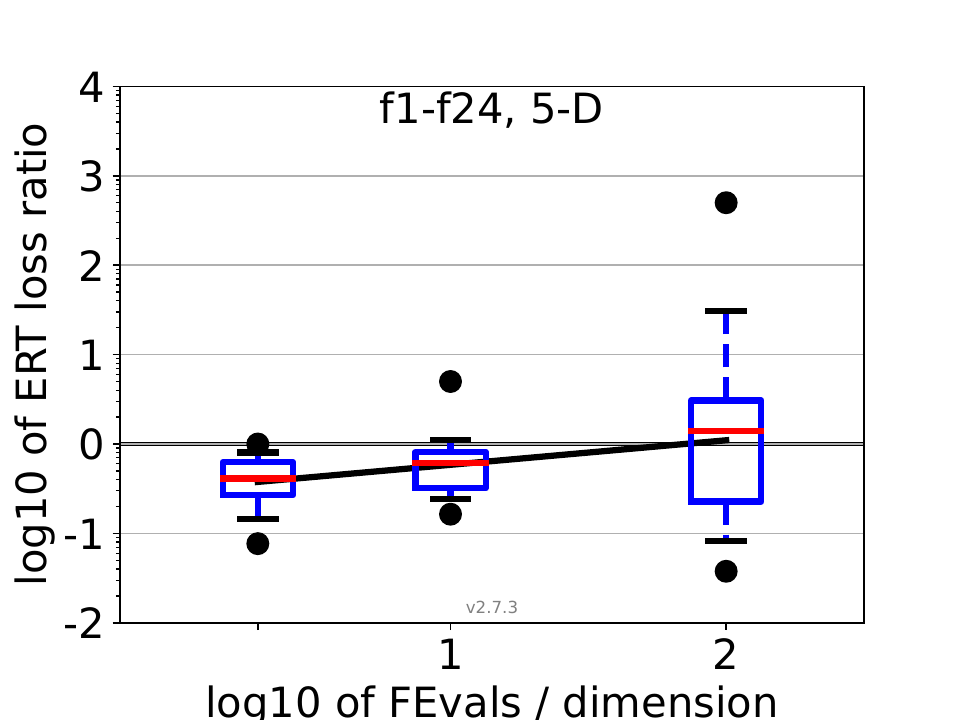} \\
    \raisebox{\leRaiser}{\footnotesize\rotatebox{90}{\(d=20\)}} &
        \includegraphics[width=\leWidth]{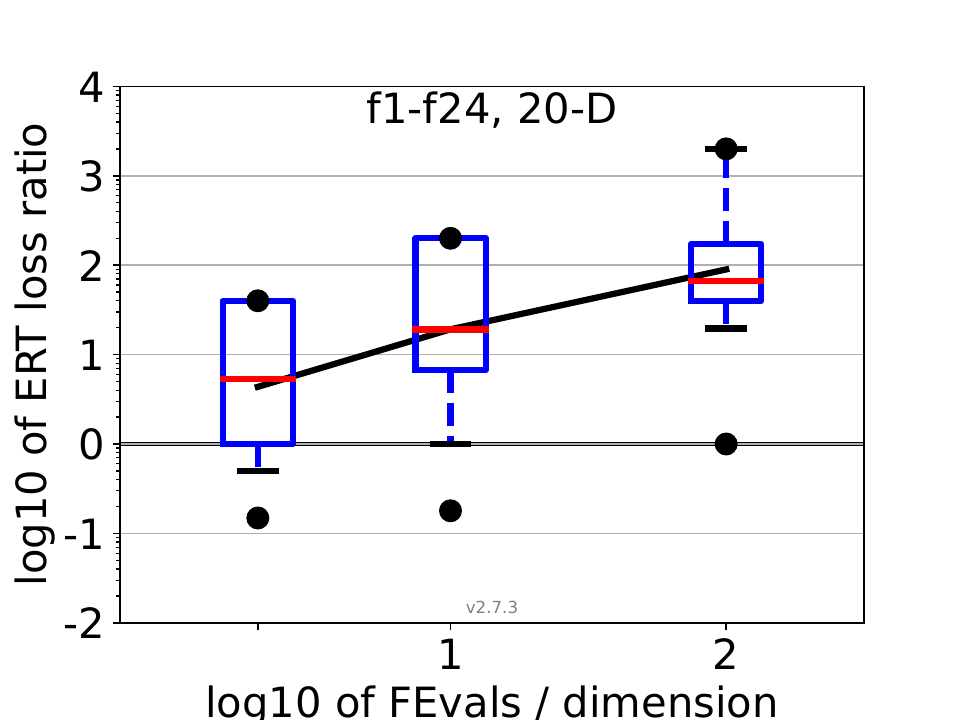} &
        \includegraphics[width=\leWidth]{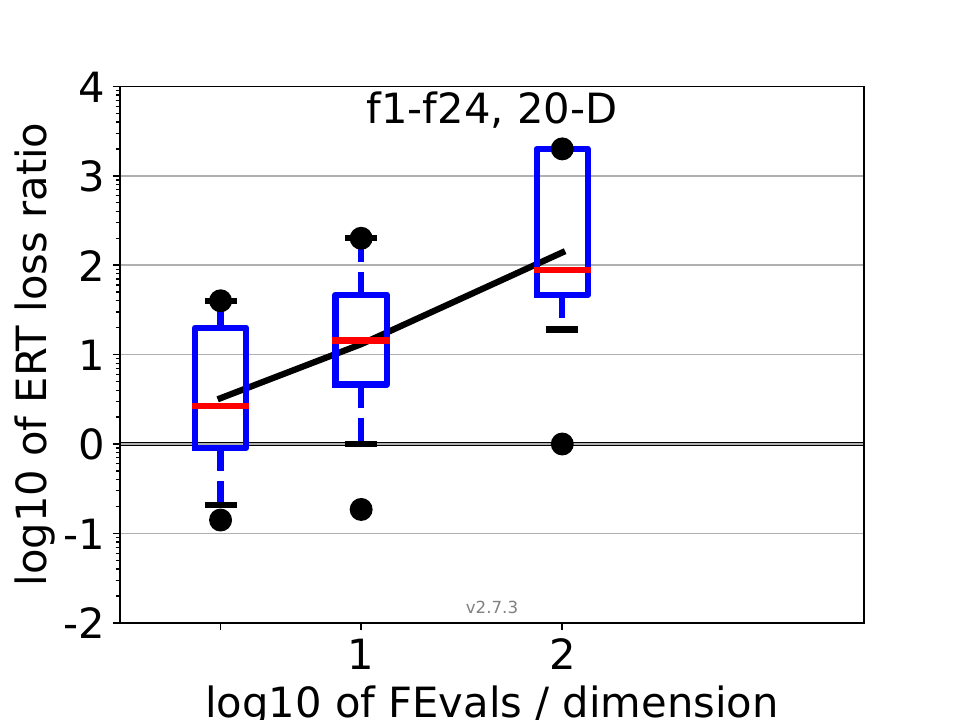} &
        \includegraphics[width=\leWidth]{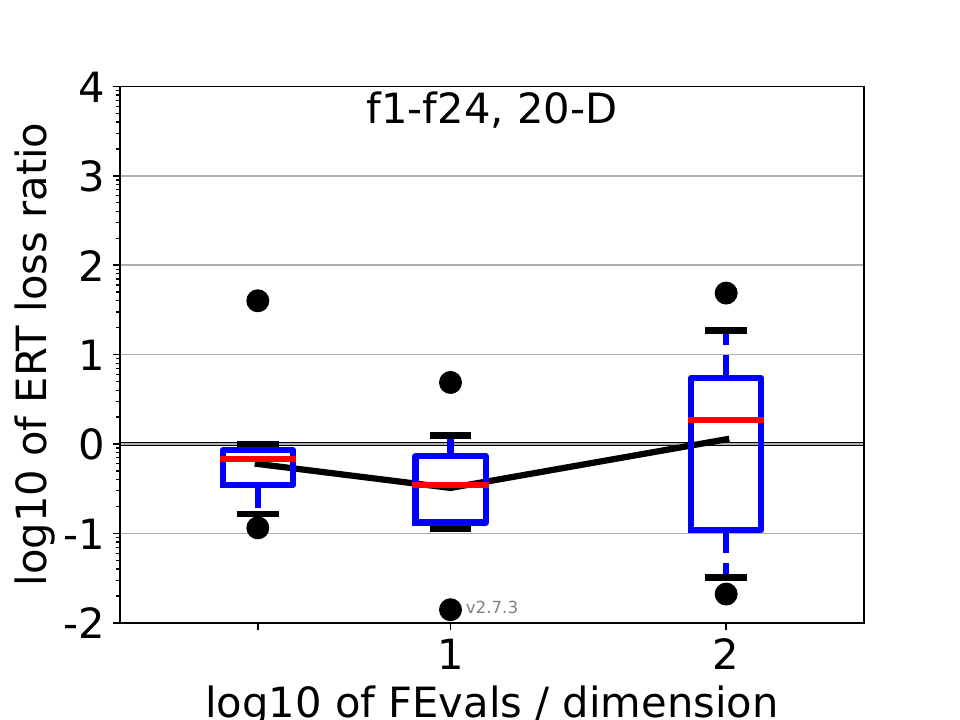} \\
    & {\footnotesize Neuropt-Lin}
    & {\footnotesize Neuropt-Izh}
    & {\footnotesize Neuropt-Hyb}
    %
\end{tabular}
\caption{\label{Fig:ExConf:logloss}%
    Expected Running Time (\ERT) loss ratio vs. the budget in number of \(f\)-evaluations per dimension. For each budget, the \ERT\ to reach the best target \(f\)-value is divided by the \ERT\ of the best BBOB 2009 method for the same target. Lines show the geometric mean; box-whiskers indicate 25-75\% (box), 10-90\% (caps), and min-max (points). The vertical line marks the evaluation budget limit for this subset.
    \cocoversion{}}
    
\end{figure}

This trend is further clarified in \figurename~\ref{Fig:ExConf:logloss}, which quantifies the Expected Running Time (\ERT) loss ratio of each variant relative to the BBOB 2009 reference. 
Recall that \ERT\ quantifies the expected number of function evaluations required by an algorithm to reach a given target function value \cite{hansen2021coco}. 
In 5D, all \nha{} variants maintain loss ratios near or below unity across most budgets. 
This verifies that the observed performance in low dimensions matches established benchmarks. 
As dimensionality increases to 20D, the dispersion in loss ratios grows and median values shift above one, consistent with the plateaux seen previously. 
This increase is attributable to the constrained evaluation budget rather than algorithmic stagnation, as hard instances remain unsolved within the allocated steps. 
Occasionally, high outliers appear when challenging instances are not solved within the budget, causing the \ERT\ loss ratio to spike for those cases \cite{hansen2021coco}. 
Hence, these results support that the spike-driven approach retains competitive efficiency in moderate dimensions and remains robust even as problem complexity increases.

\begin{figure*}[!ht]\centering
\def\leRaiser{10mm}
\def\leWidth{0.19\linewidth}
\begin{tabular}{@{}c*{5}{@{ }c}@{}}
\raisebox{\leRaiser}{\footnotesize\rotatebox{90}{\texttt{separ}}} &
      \includegraphics[width=\leWidth]{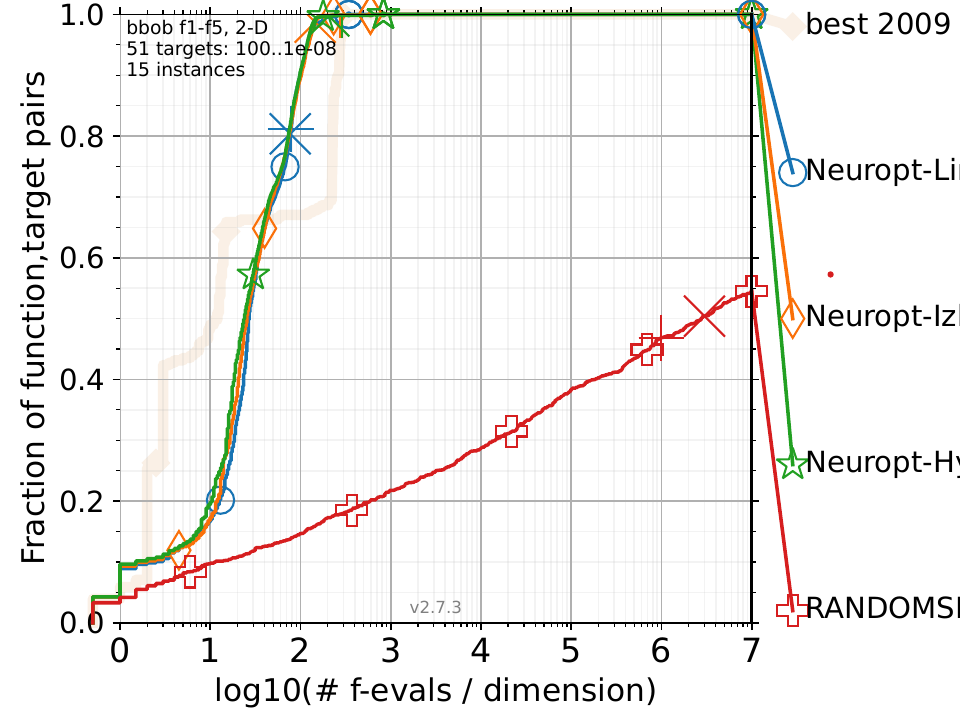} &
      \includegraphics[width=\leWidth]{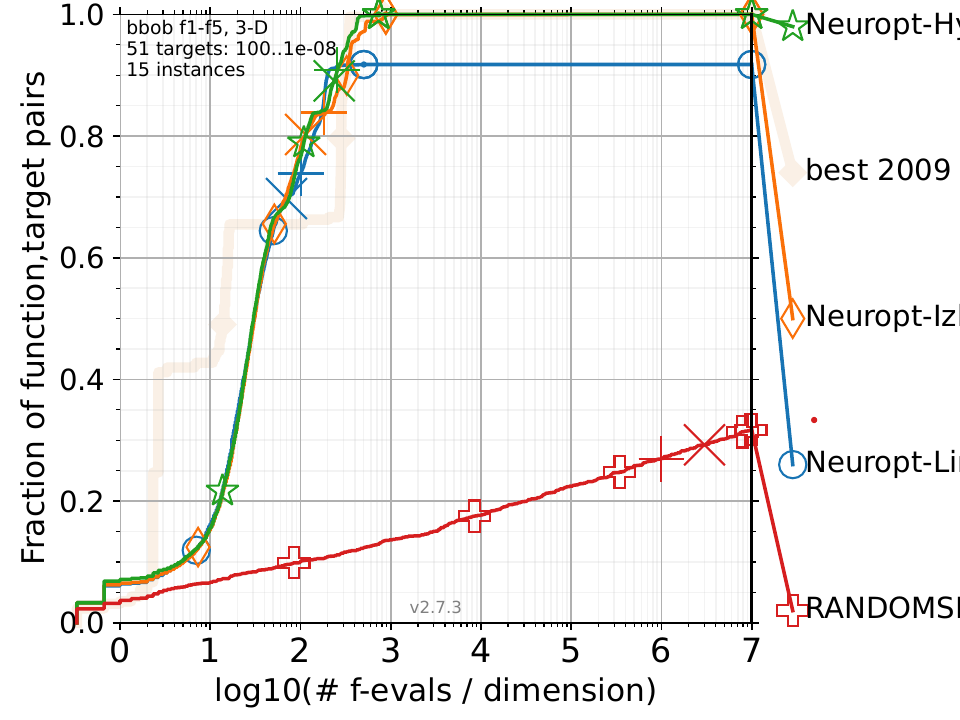} &
      \includegraphics[width=\leWidth]{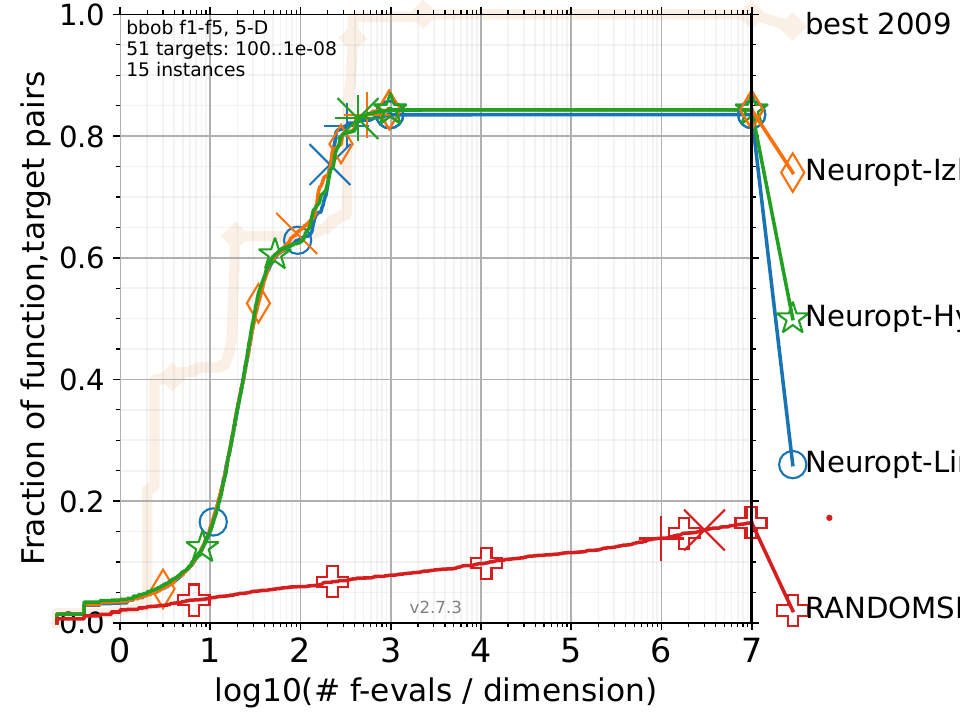} &
      \includegraphics[width=\leWidth]{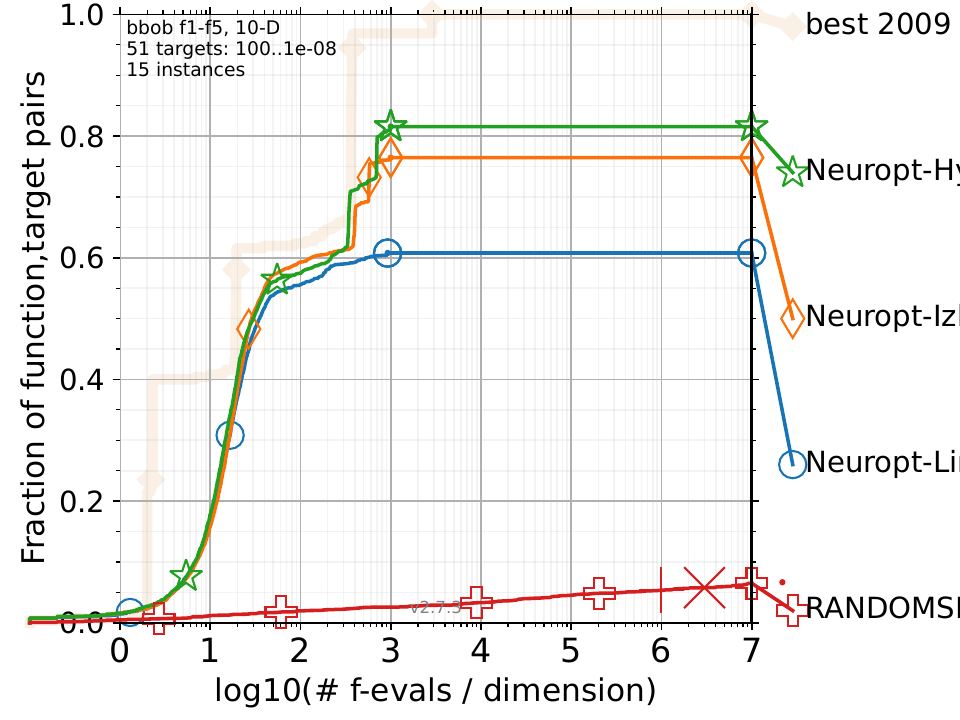} &
      \includegraphics[width=\leWidth]{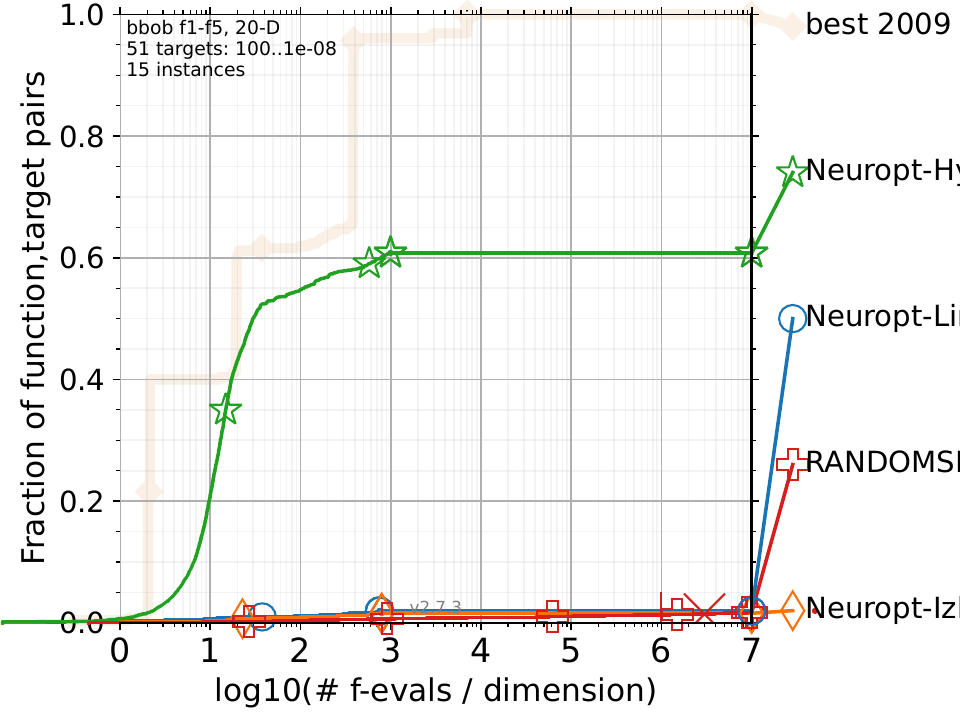} \\
\raisebox{\leRaiser}{\footnotesize\rotatebox{90}{\texttt{lcond}}} &
      \includegraphics[width=\leWidth]{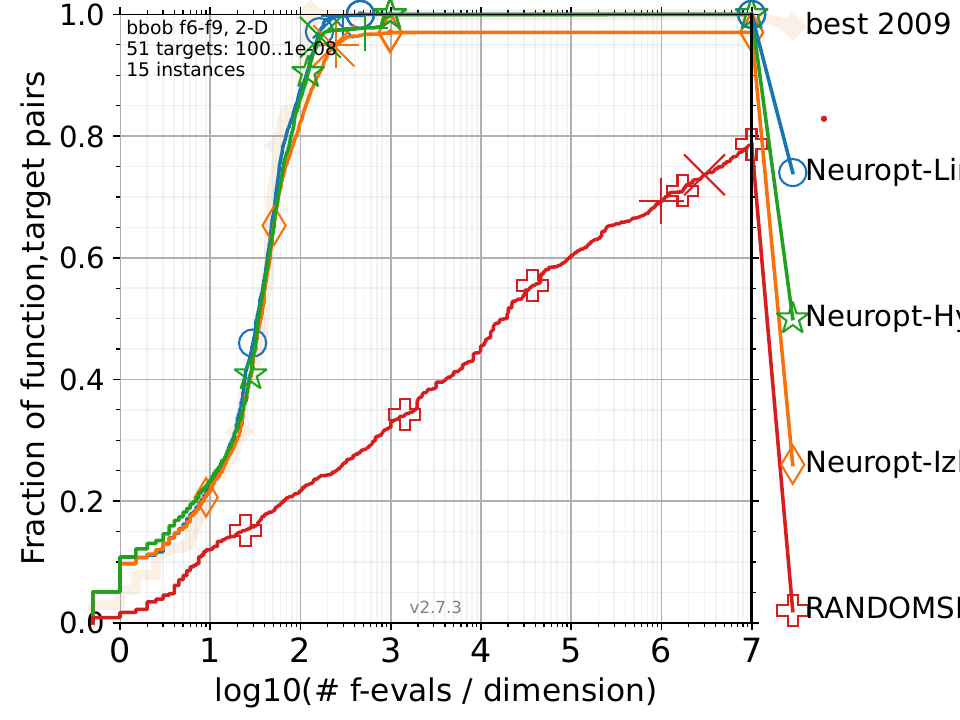} &
      \includegraphics[width=\leWidth]{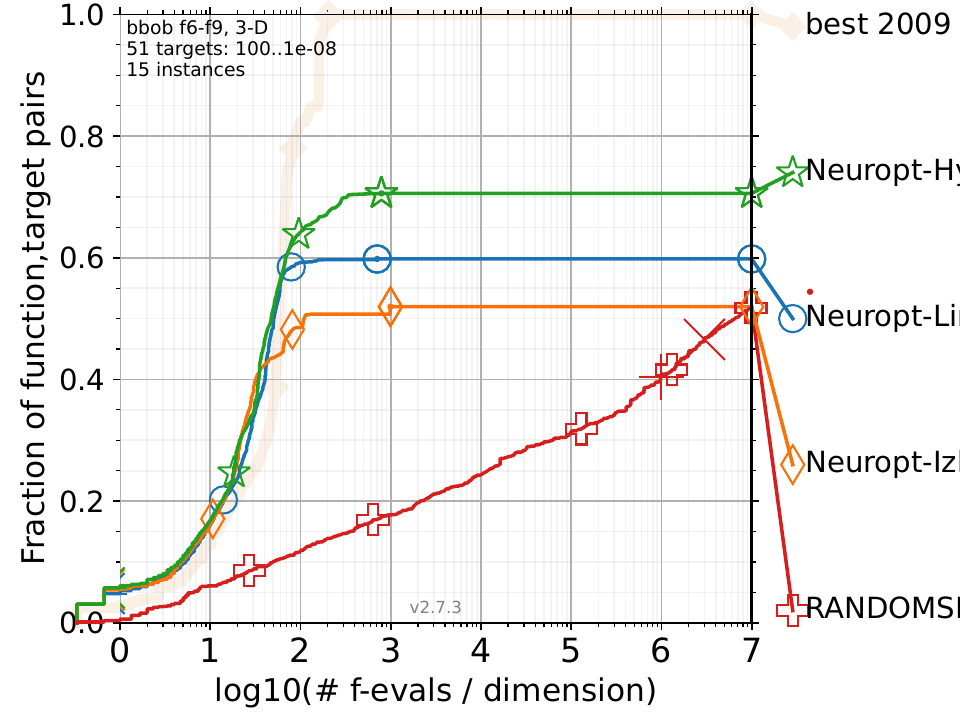} &
      \includegraphics[width=\leWidth]{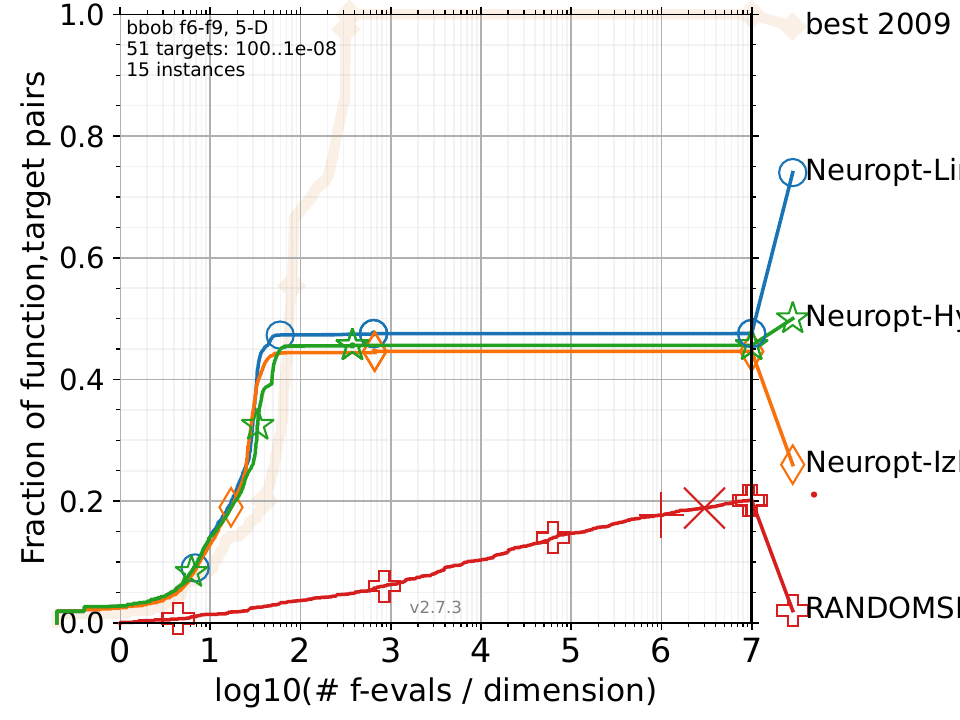} &
      \includegraphics[width=\leWidth]{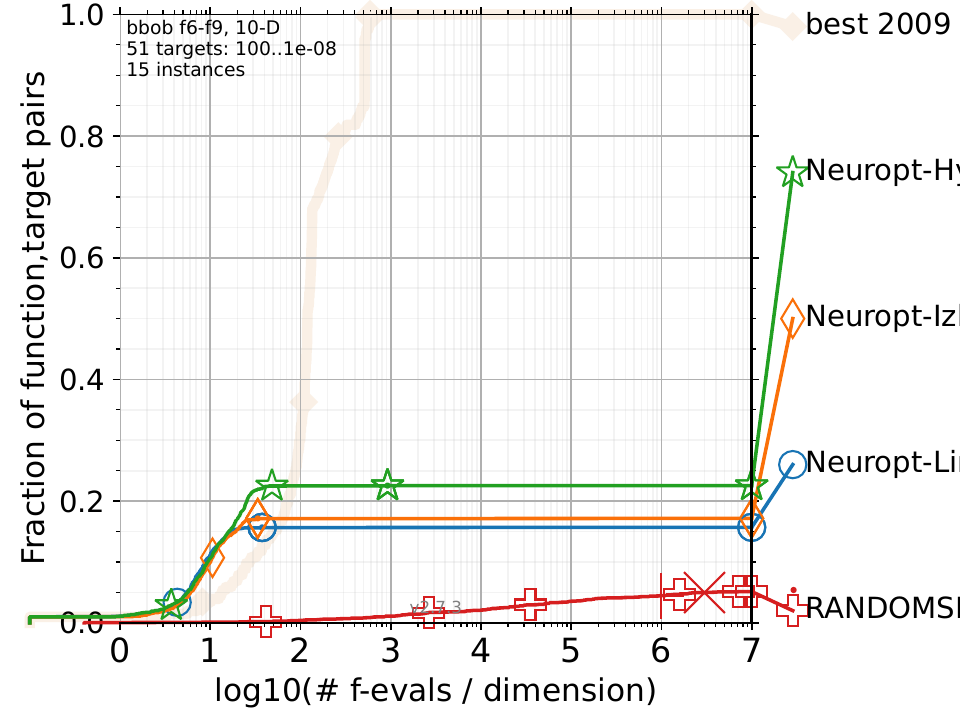} &
      \includegraphics[width=\leWidth]{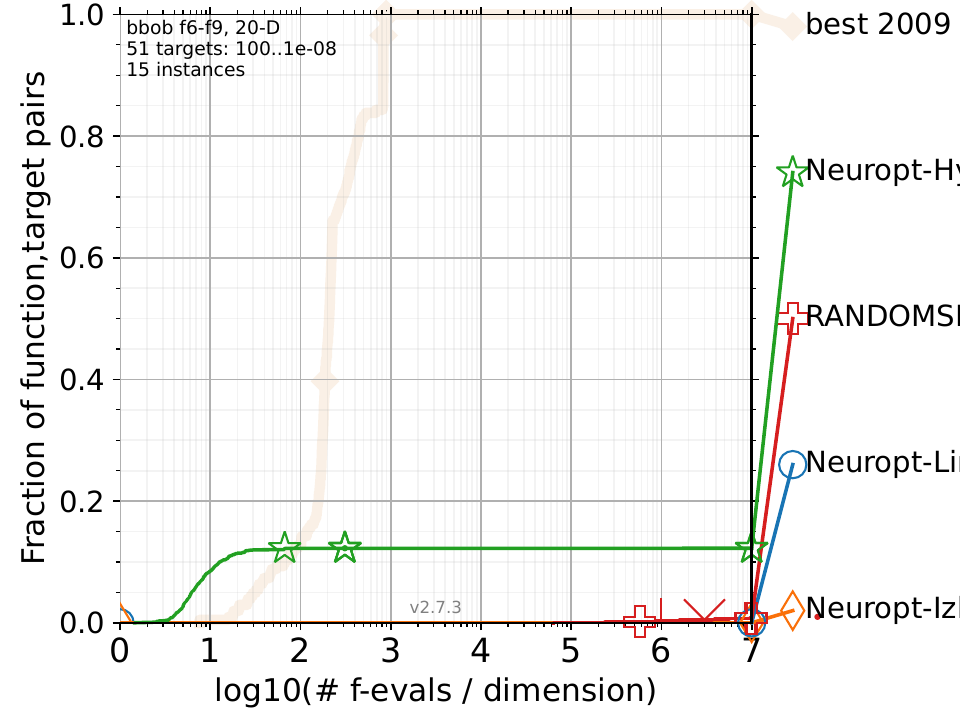} \\
\raisebox{\leRaiser}{\footnotesize\rotatebox{90}{\texttt{hcond}}} &
      \includegraphics[width=\leWidth]{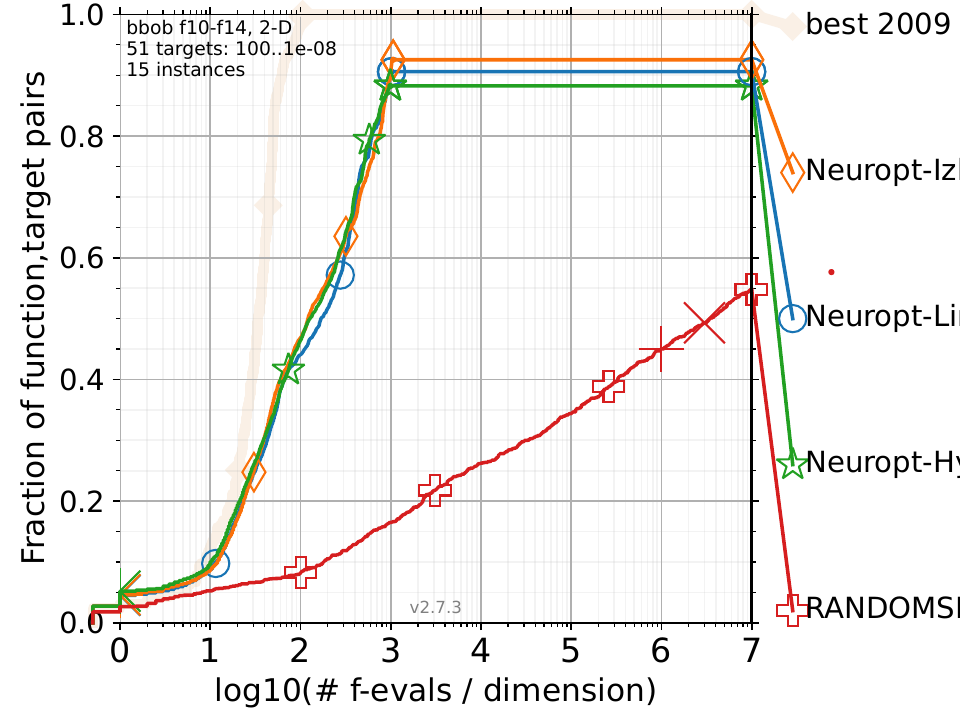} &
      \includegraphics[width=\leWidth]{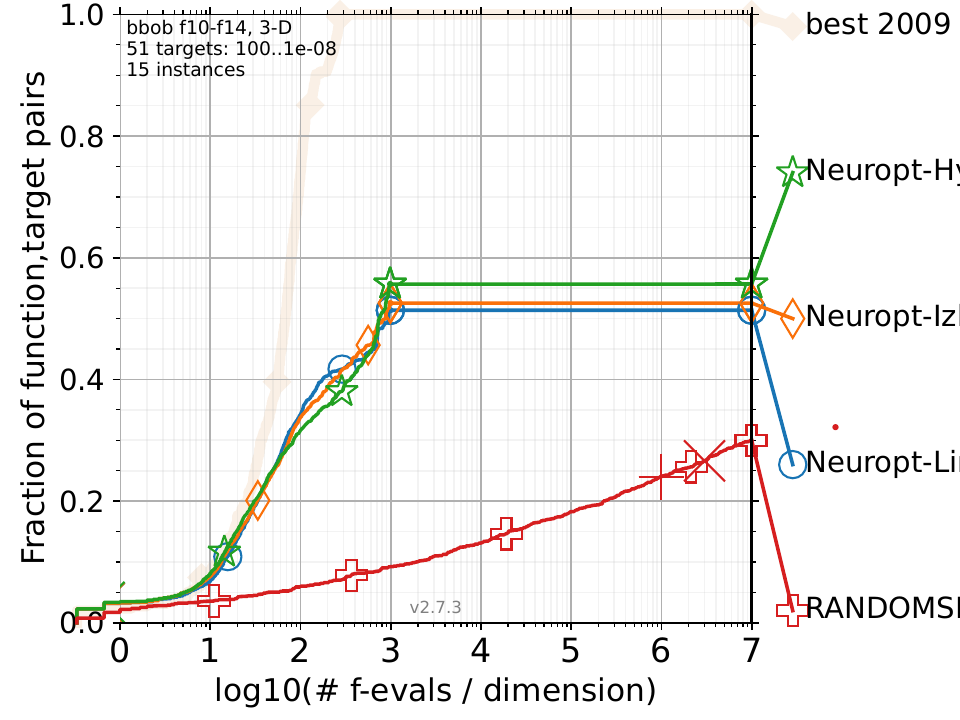} &
      \includegraphics[width=\leWidth]{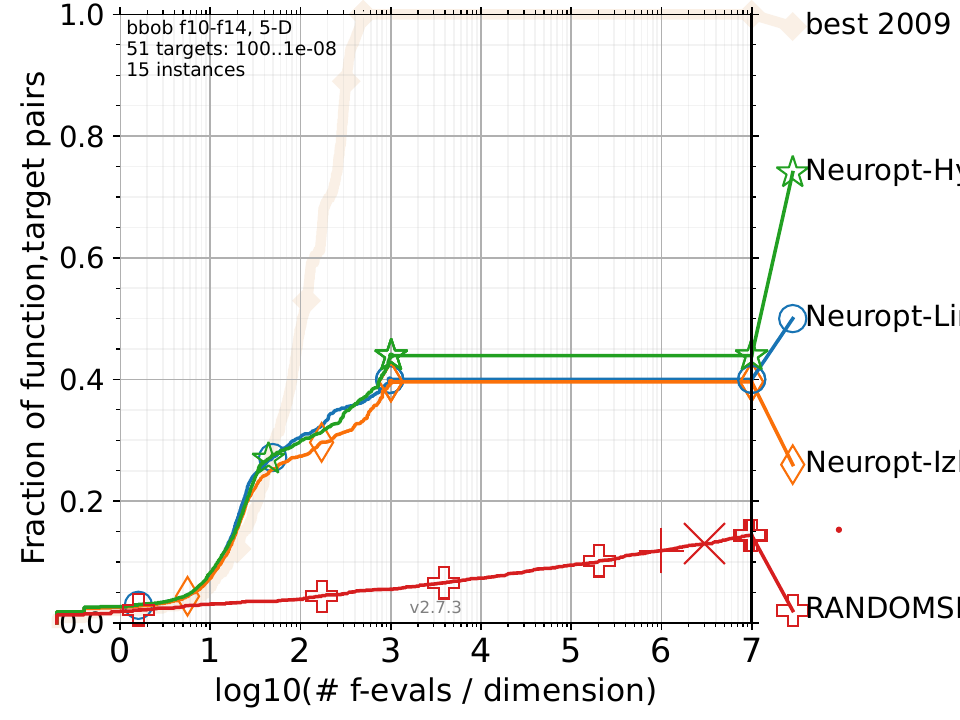} &
      \includegraphics[width=\leWidth]{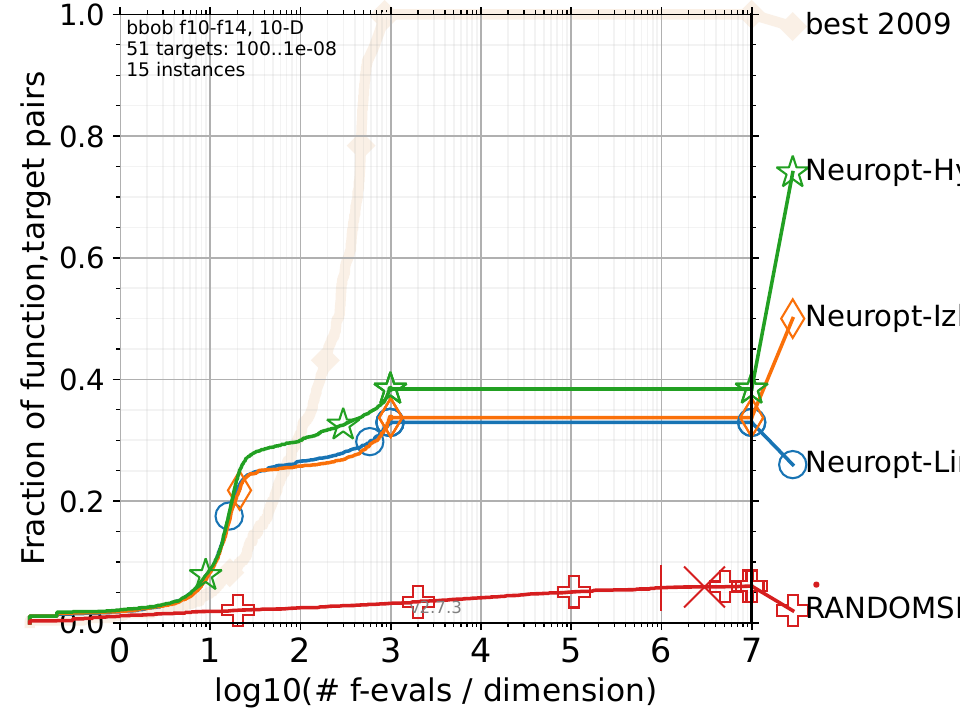} &
      \includegraphics[width=\leWidth]{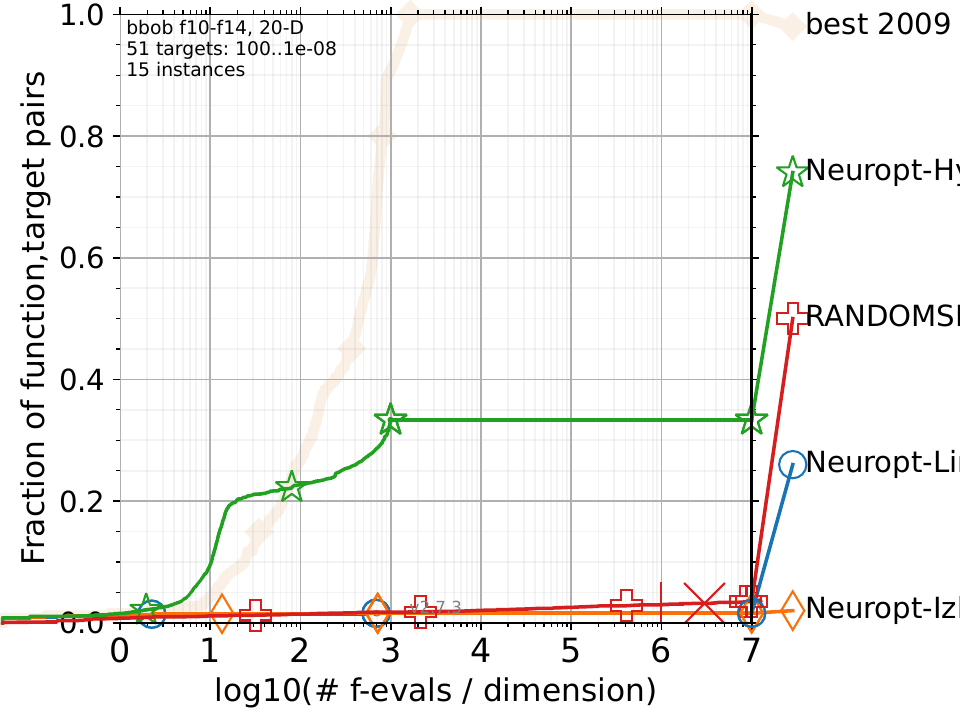} \\
\raisebox{\leRaiser}{\footnotesize\rotatebox{90}{\texttt{multi}}} &
      \includegraphics[width=\leWidth]{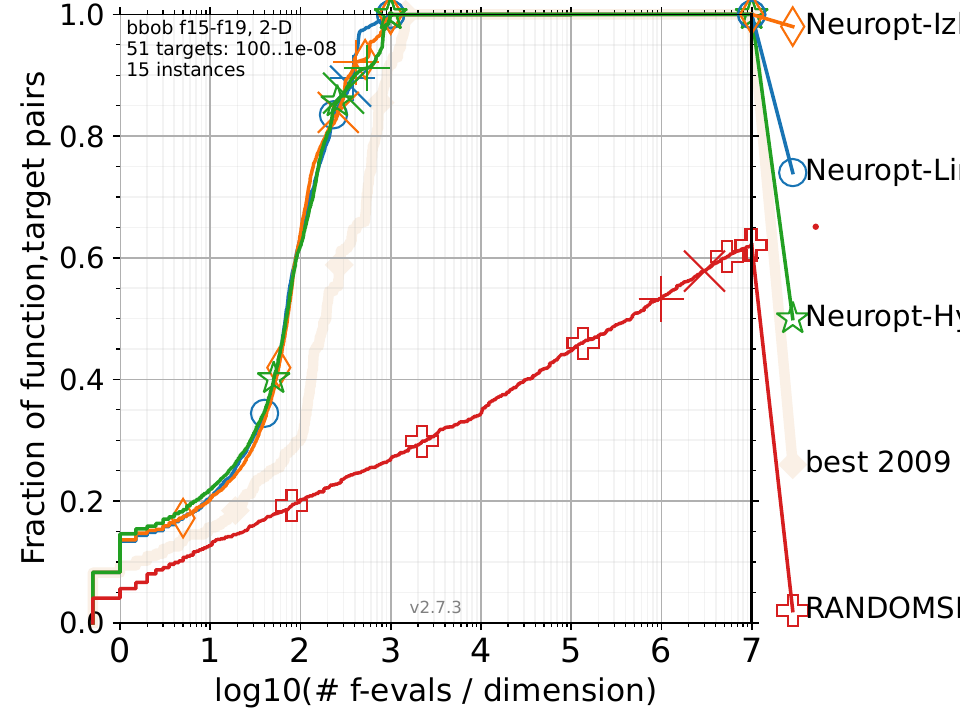} &
      \includegraphics[width=\leWidth]{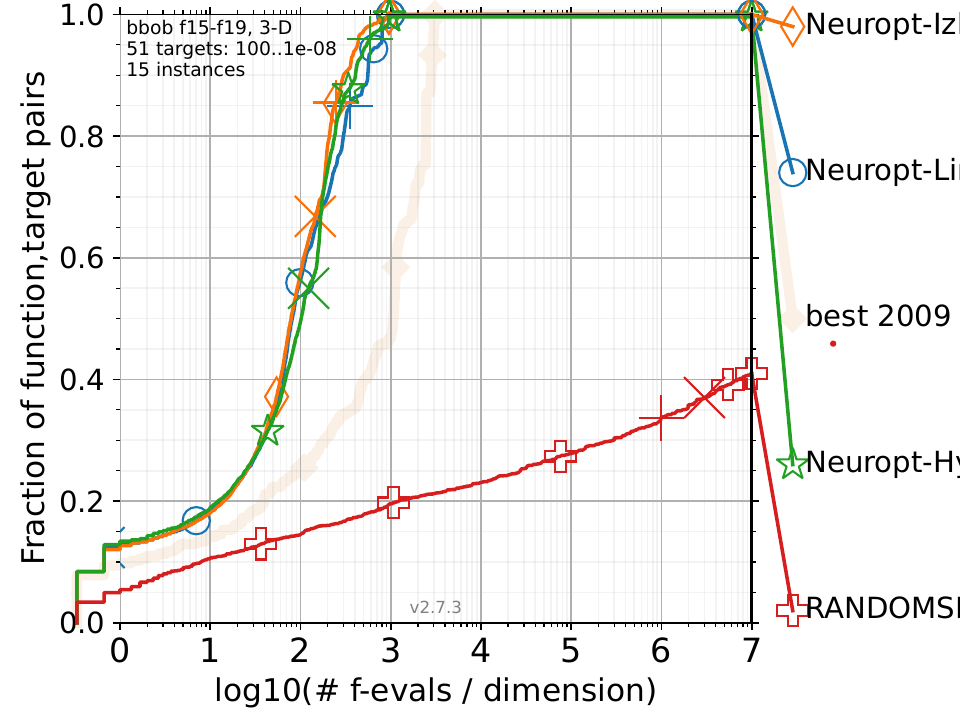} &
      \includegraphics[width=\leWidth]{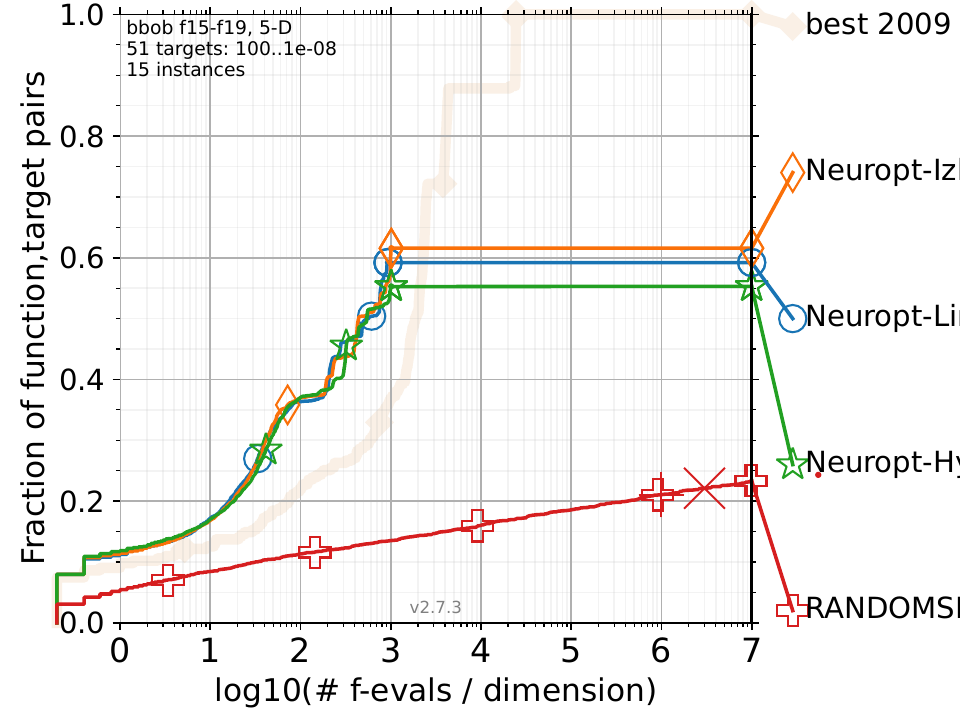} &
      \includegraphics[width=\leWidth]{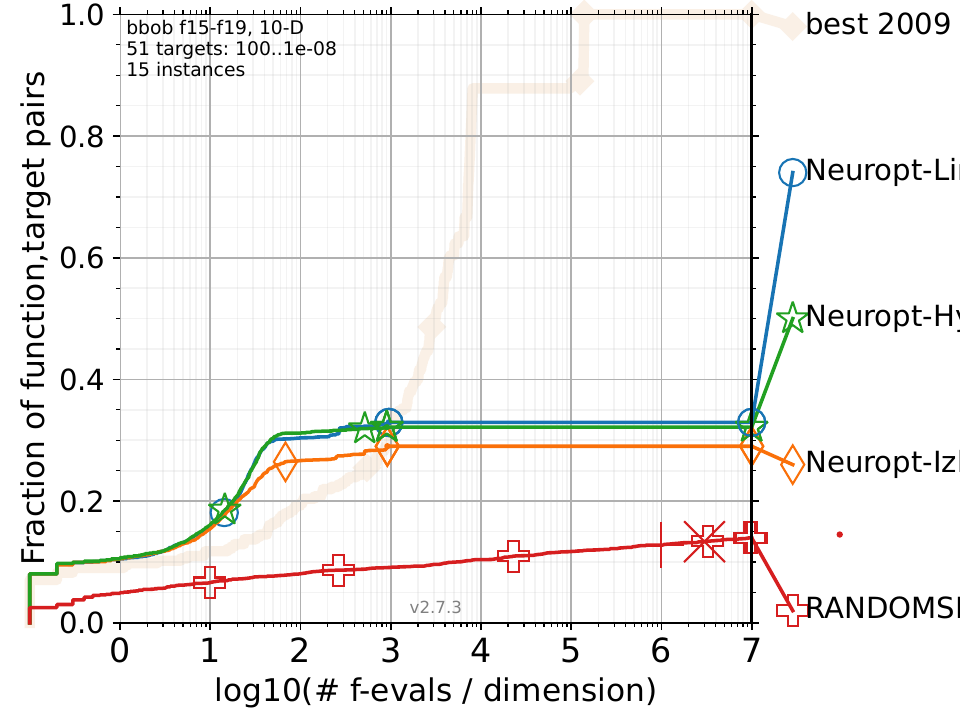} &
      \includegraphics[width=\leWidth]{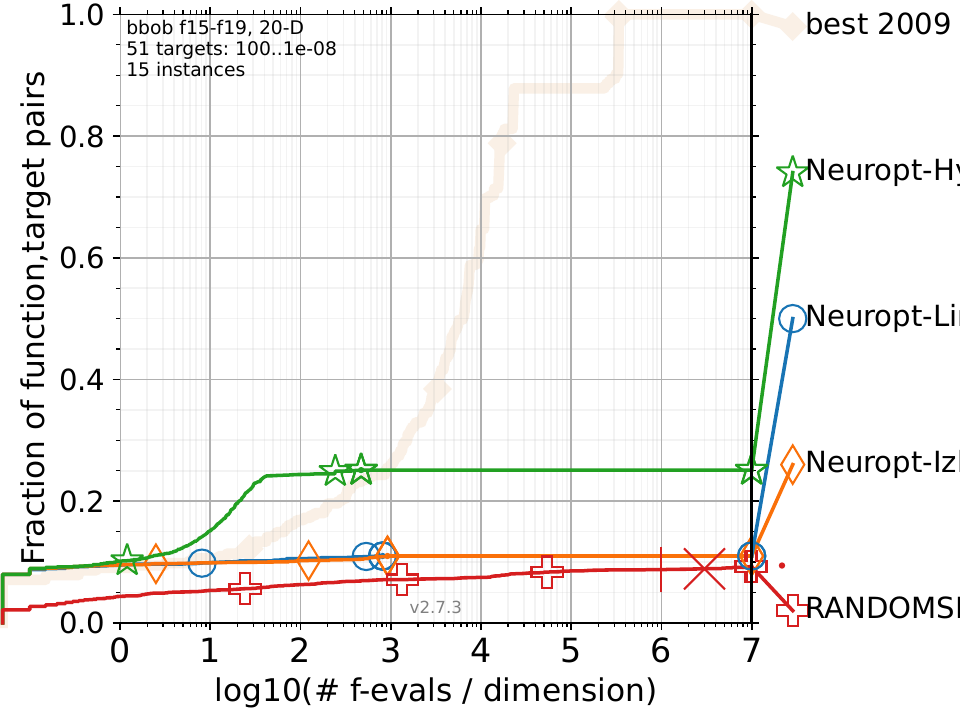} \\
\raisebox{\leRaiser}{\footnotesize\rotatebox{90}{\texttt{mult2}}} &
      \includegraphics[width=\leWidth]{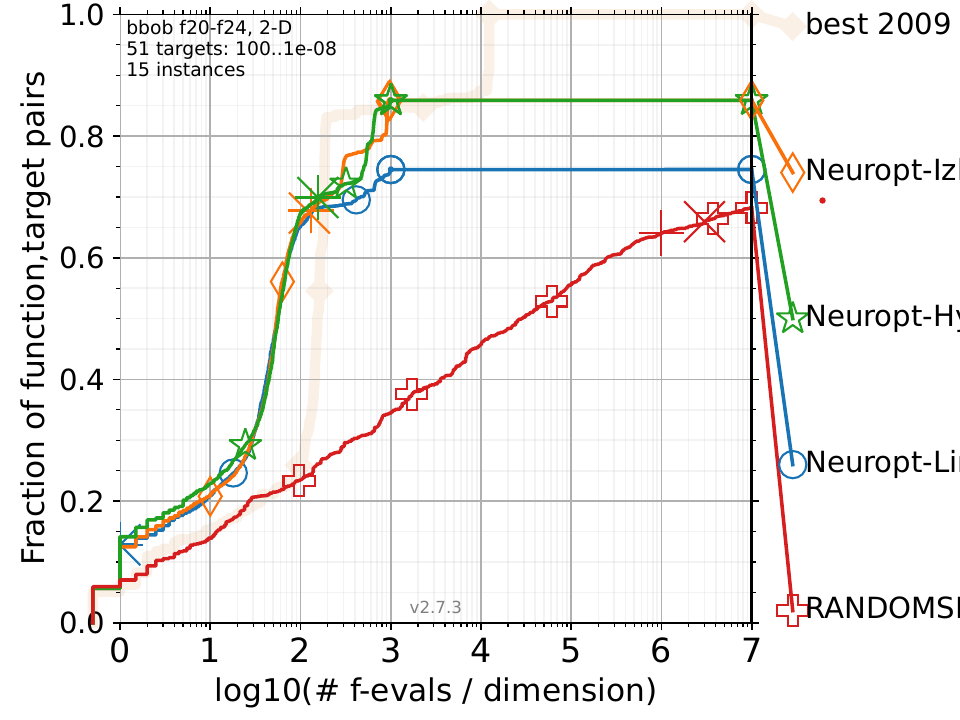} &
      \includegraphics[width=\leWidth]{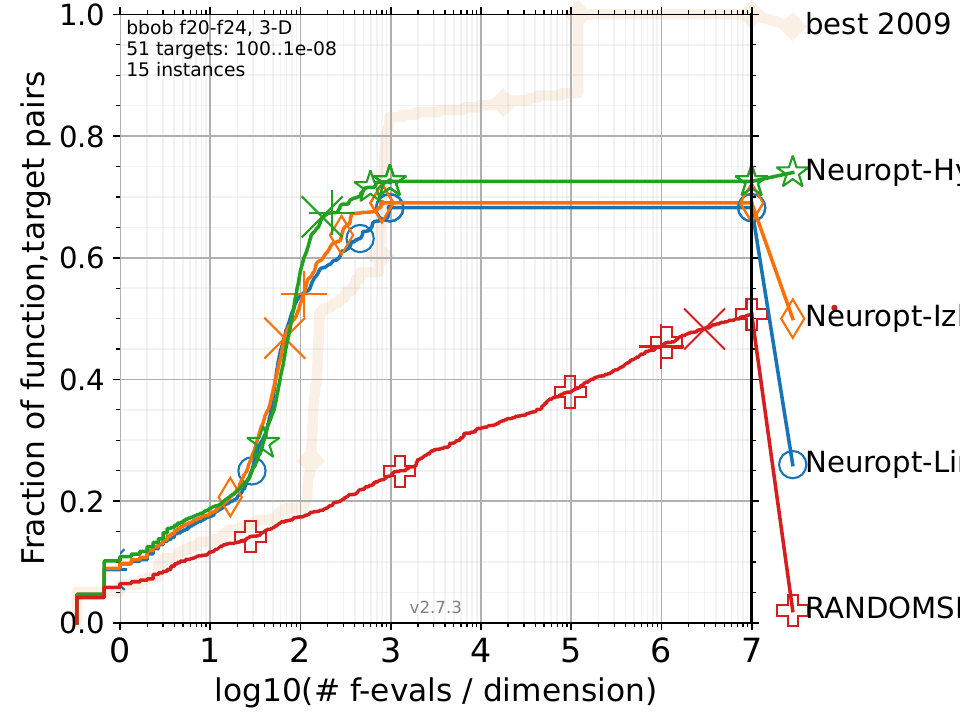} &
      \includegraphics[width=\leWidth]{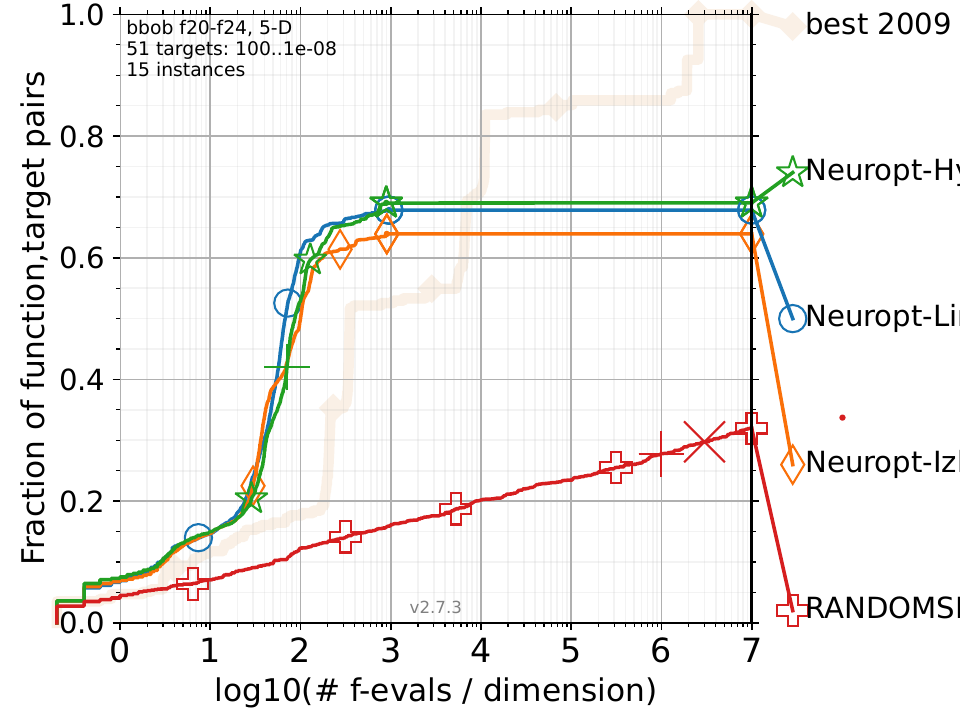} &
      \includegraphics[width=\leWidth]{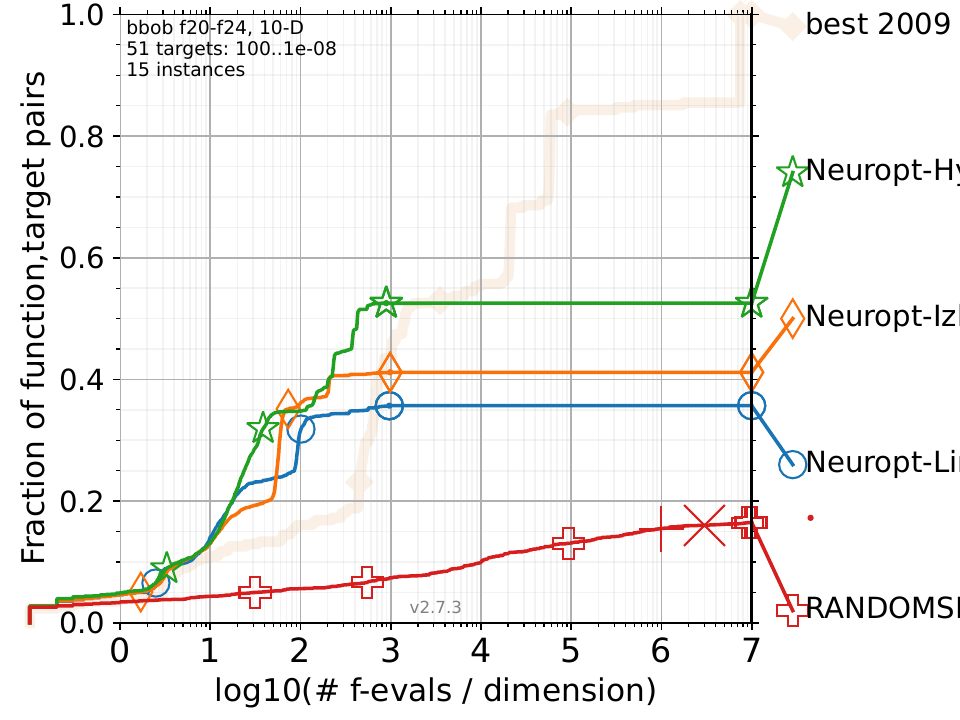} &
      \includegraphics[width=\leWidth]{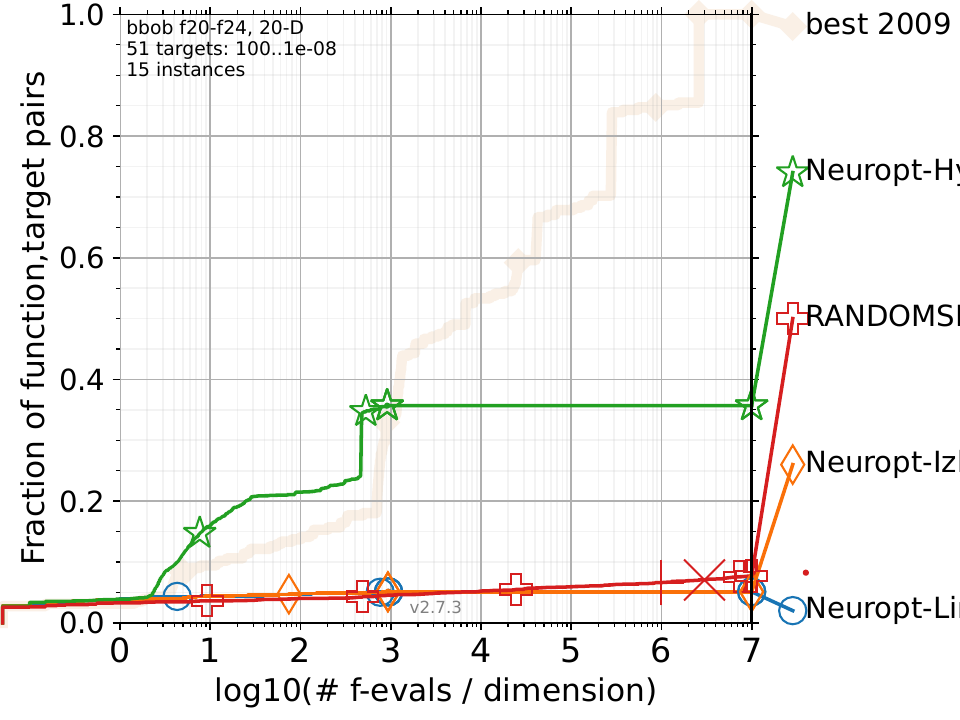} \\
\raisebox{\leRaiser}{\footnotesize\rotatebox{90}{\texttt{all}}} &
      \includegraphics[width=\leWidth]{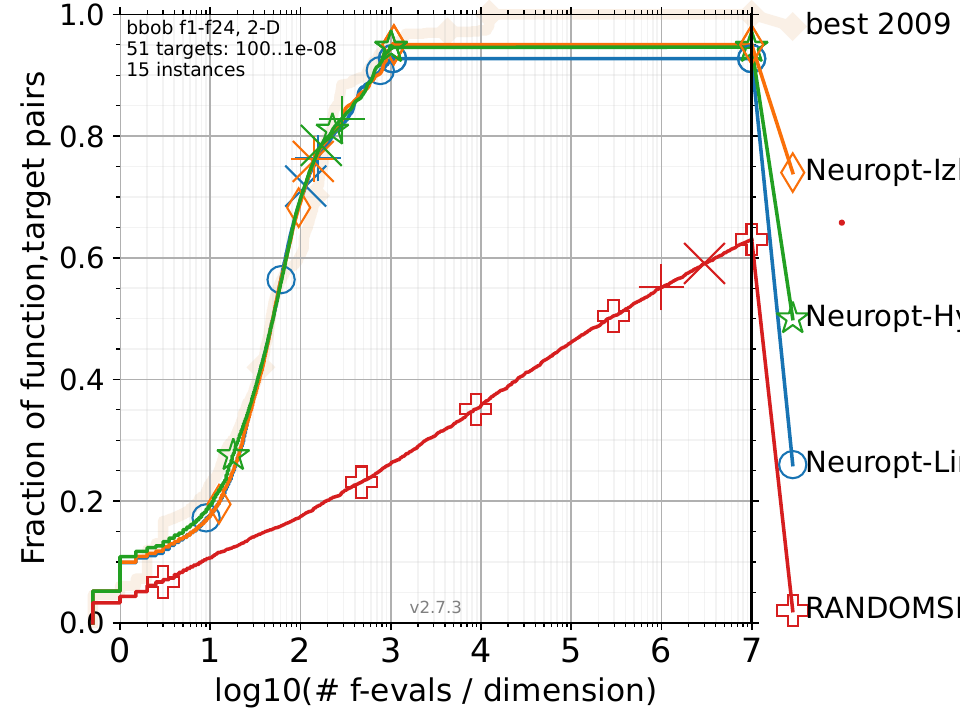} &
      \includegraphics[width=\leWidth]{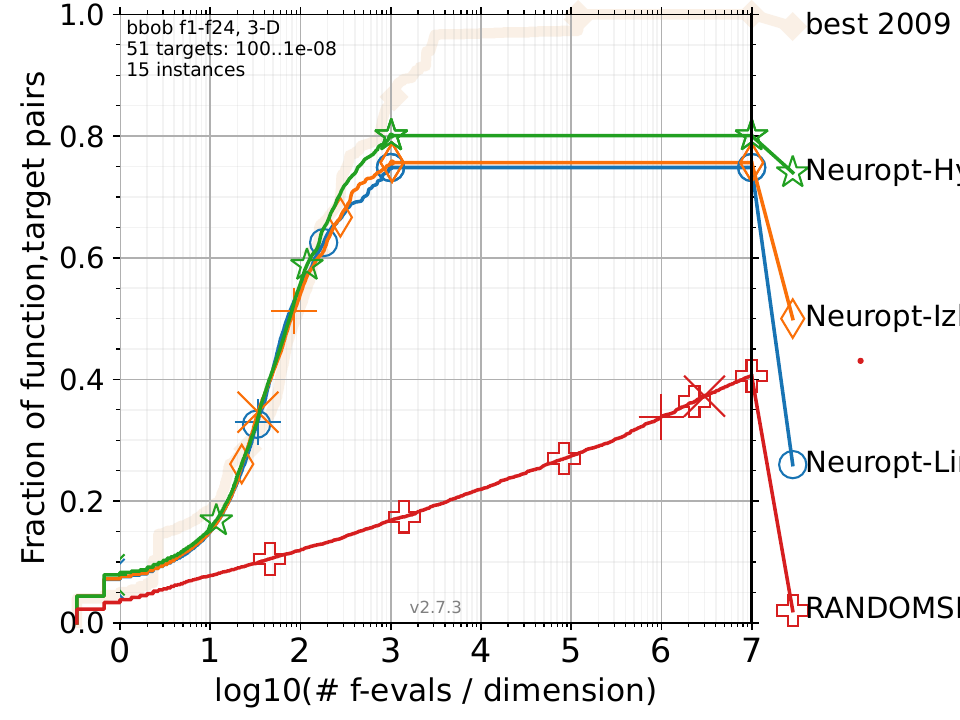} &
      \includegraphics[width=\leWidth]{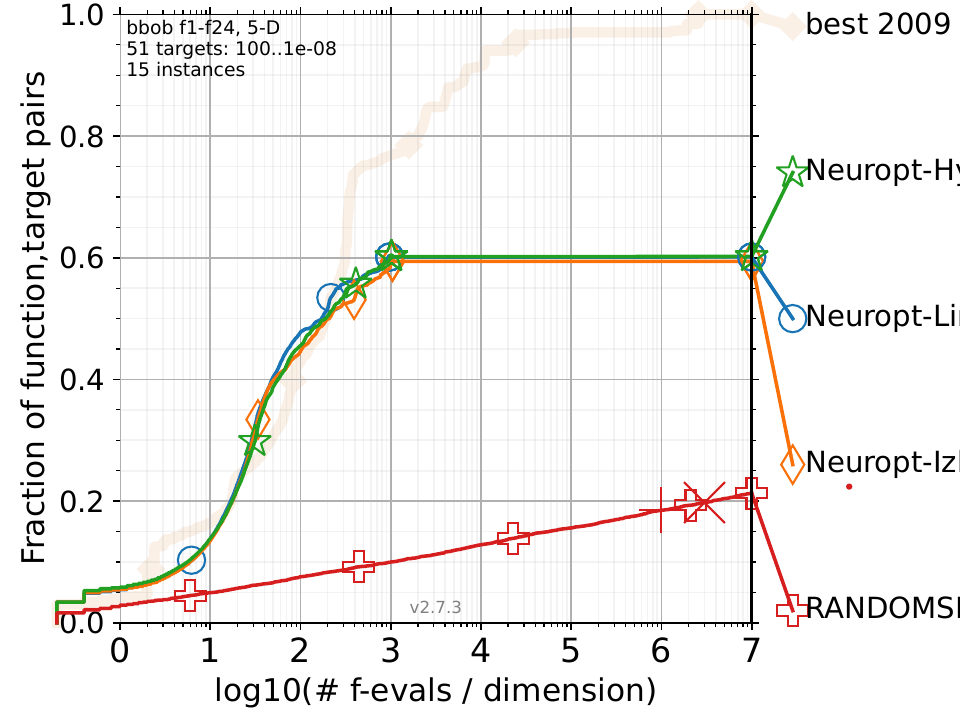} &
      \includegraphics[width=\leWidth]{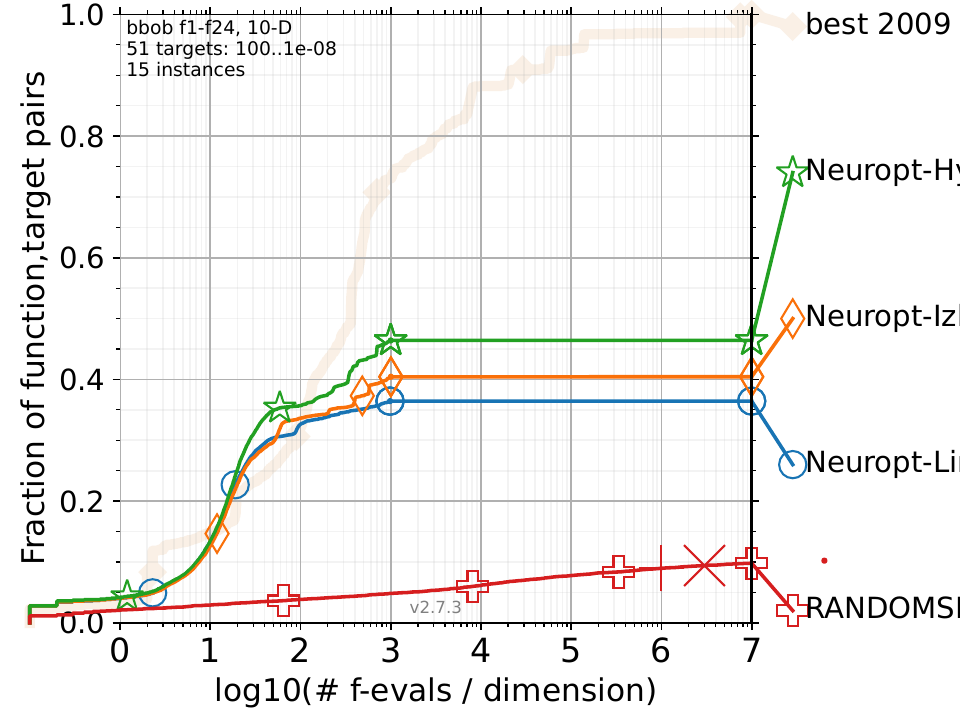} &
      \includegraphics[width=\leWidth]{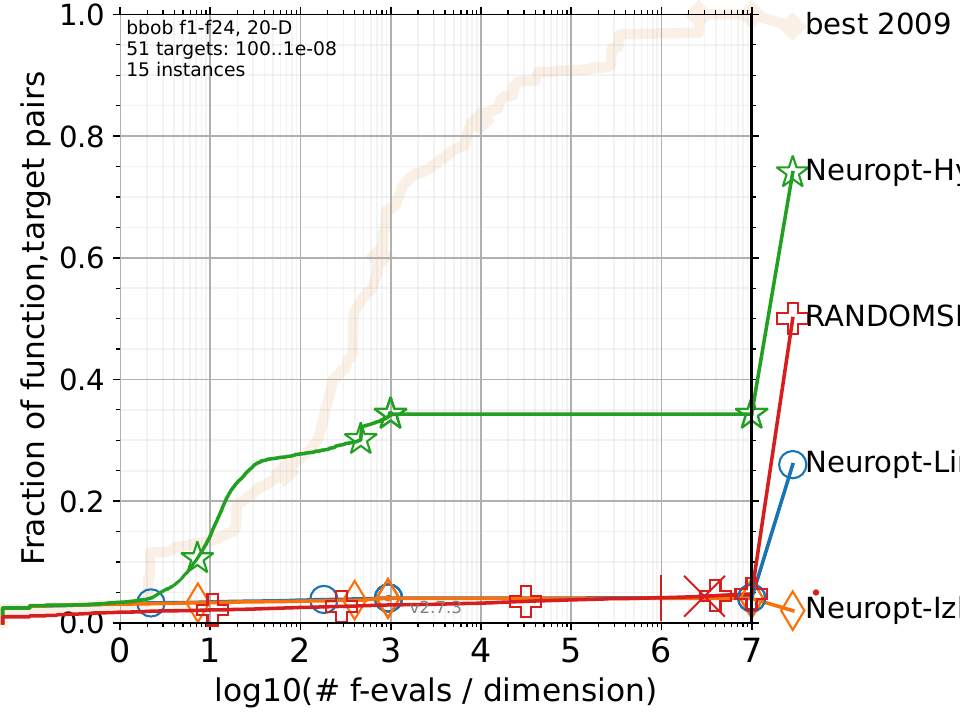} \\
& \(d=2\) & \(d=3\) & \(d=5\) & \(d=10\) & \(d=20\) 
\end{tabular}
\caption{\label{Fig:ExConf:ECDdet}%
    Bootstrapped Empirical Cumulative Distribution Function (ECDF) of the number of \(f\)-evaluations divided by dimension for 51 targets in \(10^{[-8..2]}\) across all function subgroups, with 15 instances per dimension, shown for dimensions 2, 5, 10, and 20. The best algorithm from BBOB 2009 is shown as a light, thick reference line. 
    Legend:
    {\color{CornflowerBlue}$\bigcirc$}:~\algorithmA
    , {\color{Orange}$\diamondsuit$}:~\algorithmB
    , {\color{Green}$\hollowstar$}:~\algorithmC
    , {\(\filledplus[red]\)}:~\algorithmD \cite{auger2009benchmarking}. 
    \cocoversion{}
    }
\end{figure*}

\figurename~\ref{Fig:ExConf:ECDdet} details the ECDFs of normalised runtimes for all \nha{} variants, RANDOMSEARCH auger noiseless~\cite{auger2009benchmarking}, and the BBOB 2009 reference, split down by function subgroup and dimension. 
In line with the global trends reported previously, both Neuropt-Izh and Neuropt-Hyb achieve rapid convergence and high fractions of solved pairs for separable (\texttt{separ}) and low-conditioned (\texttt{lcond}) functions up to 10D, closely tracking the BBOB 2009 reference and significantly outperforming RANDOMSEARCH. 
On high-conditioned (\texttt{hcond}) and multimodal (\texttt{multi} and \texttt{mult2}) problems, the fraction of solved pairs decreases as dimension increases, with the difference becoming most pronounced in 20D. 
For these tough groups, Neuropt-Hyb exhibits increased robustness, often maintaining higher solved fractions than either of the homogeneous variants, consistent with the expected benefits of population diversity in challenging landscapes. 
Across all scenarios, RANDOMSEARCH displays a steady and gradual increase, but rarely reaches the highest fractions of solved pairs within the available budget, higher than the one used by our approach.

It is worth noting that the objective of this study is to demonstrate the viability of spike-driven heuristics as a general optimisation framework, rather than to achieve state-of-the-art performance.
Instead, the comparative analysis is intended to confirm that the \nha{} framework achieves competitive results even when operating under a limited evaluation budget, reduced population sizes, and without extensive parameter fine-tuning. 
There is a massive room for improvement and future research. 
For this reason, we employed results from a random search implementation and the best BBOB 2009 algorithm as the lower and upper benchmarks to establish the performance bounds. 
As previously noticed, the flat lines observed in the ECDFs, especially for high-dimensional and complex functions, primarily reflect the restrictive evaluation budget rather than an intrinsic limitation of the neuroptimiser approach. 
These results corroborate the analysis in \figurename~\ref{Fig:ExConf:ECD} and \ref{Fig:ExConf:logloss}, supporting the observation that performance boundaries are shaped by resource allocation and problem complexity.

Furthermore, \tablename{}~\ref{Tab:ERTsDetails} complements the convergence analysis by providing detailed \ERT\ for the first function of each BBOB subgroup in 5D. (The complete dataset can be found in \cite{Cruz2025neurodataset}.)
For \textbf{f1}, all \nha{} configurations reach the majority of targets in most trials, with Neuropt-Lin and Neuropt-Hyb closely matching or exceeding the BBOB 2009 baseline for all attainable targets. 
However, it is clear that none of the \nha{} variants consistently achieves the final targets.
On \textbf{f6} (\texttt{lcond}), only the first two targets are reliably reached by all \nha{}s. 
For \textbf{f10} (\texttt{hcond}), all configurations stagnate at coarse targets, consistent with the previous ECDF trends.
In the weakly structured subgroup, both Neuropt-Lin and Neuropt-Hyb report the best ERTs on \textbf{f15} for all targets except the final one, with \ERT\ ratios below one and a clear improvement over RANDOMSEARCH for that target.
For \textbf{f20}, Neuropt-Lin exhibits good scalability, solving the initial targets in all trials, surpassing RANDOMSEARCH, with Neuropt-Lin showing stable and consistently lower ERTs. 
These results corroborate the ECDF and ERT analyses and confirm that while homogeneous \nha{}s can excel in some instances, Neuropt-Hyb provides more robust performance across problem types.

\def\dimpp{05}
\begin{table}[!ht]\centering
\caption{\label{Tab:ERTsDetails}%
    Expected runtime (ERT, in $f$-evaluations) for each algorithm and target, normalised by the best BBOB 2009 ERT in 5D. Each cell depicts the ERT ratio, along with the half-difference between the 10\textsuperscript{th} and 90\textsuperscript{th} percentiles (bootstrapped) in parentheses. Reference ERTs are in the first row. \#succ is the number of successful trials for the final target $f_\text{opt} + 10^{-8}$. Italics indicate medians where the target was not reached. Stars mark statistically best entries; $\uparrow$ flags some worse than the reference. Bold indicates best results.
    \cocoversion%
    }
\setlength{\tabcolsep}{1pt}
\renewcommand{\arraystretch}{1.05}
\vspace{0pt}\centering
\resizebox{\columnwidth}{!}{%
\begin{tabular}{@{}l@{ }*{7}{@{}r@{}l@{}}|@{}r@{}@{}l@{}}%
     \hline
      $\pmb{\Delta f_\text{opt}}$%
      & \multicolumn{2}{@{}c@{}}{\(10^1\)}%
      & \multicolumn{2}{@{}c@{}}{\(10^0\)}%
      & \multicolumn{2}{@{}c@{}}{\(10^{-1}\)}%
      & \multicolumn{2}{@{}c@{}}{\(10^{-2}\)}%
      & \multicolumn{2}{@{}c@{}}{\(10^{-3}\)}%
      & \multicolumn{2}{@{}c@{}}{\(10^{-5}\)}%
      & \multicolumn{2}{@{}c@{}|}{\(10^{-7}\)}%
      & \multicolumn{2}{@{}c@{}}{\#succ}\\
    \hline
    \textbf{f1} & \multicolumn{2}{@{}c@{}}{11 \quad} & \multicolumn{2}{@{}c@{}}{12 \quad} & \multicolumn{2}{@{}c@{}}{12 \quad} & \multicolumn{2}{@{}c@{}}{12 \quad} & \multicolumn{2}{@{}c@{}}{12 \quad} & \multicolumn{2}{@{}c@{}}{12 \quad} & \multicolumn{2}{@{}c@{}|}{12} & 15 & /15\\\hline
\algAtables\hspace*{\fill} & 0 & .66\mbox{\tiny (0.4)} & 2 & .1\mbox{\tiny (0.2)} & 3 & .4\mbox{\tiny (0.4)} & 4 & .2\mbox{\tiny (0.5)} & 5 & .1\mbox{\tiny (0.6)} & 6 & .8\mbox{\tiny (0.8)} & 68 & \mbox{\tiny (171)} & 4 & /15\\
\algBtables\hspace*{\fill} & 0 & .79\mbox{\tiny (0.6)} & \textbf{2} & \textbf{.1}\mbox{\tiny (0.5)} & \textbf{3} & \textbf{.2}\mbox{\tiny (0.5)} & \textbf{4} & \textbf{.1}\mbox{\tiny (0.5)} & \textbf{5} & \textbf{.0}\mbox{\tiny (0.6)} & \textbf{6} & \textbf{.6}\mbox{\tiny (0.8)} & 62 & \mbox{\tiny (116)} & 5 & /15\\
\algCtables\hspace*{\fill} & \textbf{0} & \textbf{.57}\mbox{\tiny (0.5)} & 2 & .2\mbox{\tiny (0.5)} & 3 & .4\mbox{\tiny (0.5)} & 4 & .4\mbox{\tiny (0.7)} & 5 & .2\mbox{\tiny (0.8)} & 6 & .7\mbox{\tiny (0.8)} & \textbf{29} & \textbf{}\mbox{\tiny (8)} & 5 & /15\\
\algDtables\hspace*{\fill} & 7 & .5\mbox{\tiny (10)} & 1698 & \mbox{\tiny (1877)} & \multicolumn{2}{@{\,}l@{\,}}{6.9e5\mbox{\tiny (1e6)}} & \multicolumn{2}{@{\,}l@{\,}}{$\infty$} & \multicolumn{2}{@{\,}l@{\,}}{$\infty$} & \multicolumn{2}{@{\,}l@{\,}}{$\infty$} & \multicolumn{2}{@{\,}l@{\,}|}{$\infty$\,\textit{5e6}} & 0 & /15\\ 
    \textbf{f6} & \multicolumn{2}{@{}c@{}}{114 \quad} & \multicolumn{2}{@{}c@{}}{214 \quad} & \multicolumn{2}{@{}c@{}}{281 \quad} & \multicolumn{2}{@{}c@{}}{404 \quad} & \multicolumn{2}{@{}c@{}}{580 \quad} & \multicolumn{2}{@{}c@{}}{1038 \quad} & \multicolumn{2}{@{}c@{}|}{1332} & 15 & /15\\\hline
\algAtables\hspace*{\fill} & \textbf{0} & \textbf{.34}\mbox{\tiny (0.1)}$_{\uparrow0}$ & 0 & .47\mbox{\tiny (0.2)} & 0 & .69\mbox{\tiny (0.1)} & \multicolumn{2}{@{\,}l@{\,}}{\textbf{$\infty$}} & \multicolumn{2}{@{\,}l@{\,}}{\textbf{$\infty$}} & \multicolumn{2}{@{\,}l@{\,}}{\textbf{$\infty$}} & \multicolumn{2}{@{\,}l@{\,}|}{$\infty$\,\textit{2}} & 0 & /15\\
\algBtables\hspace*{\fill} & 0 & .41\mbox{\tiny (0.1)}$_{\uparrow0}$ & \textbf{0} & \textbf{.43}\mbox{\tiny (0.1)} & \multicolumn{2}{@{\,}l@{\,}}{$\infty$} & \multicolumn{2}{@{\,}l@{\,}}{\textbf{$\infty$}} & \multicolumn{2}{@{\,}l@{\,}}{\textbf{$\infty$}} & \multicolumn{2}{@{\,}l@{\,}}{\textbf{$\infty$}} & \multicolumn{2}{@{\,}l@{\,}|}{$\infty$\,\textit{2}} & 0 & /15\\
\algCtables\hspace*{\fill} & 0 & .34\mbox{\tiny (0.1)}$_{\uparrow0}$ & 0 & .45\mbox{\tiny (0.2)}$_{\uparrow1}$ & \textbf{0} & \textbf{.46}\mbox{\tiny (0.1)} & \multicolumn{2}{@{\,}l@{\,}}{\textbf{$\infty$}} & \multicolumn{2}{@{\,}l@{\,}}{\textbf{$\infty$}} & \multicolumn{2}{@{\,}l@{\,}}{\textbf{$\infty$}} & \multicolumn{2}{@{\,}l@{\,}|}{$\infty$\,\textit{2}} & 0 & /15\\
\algDtables\hspace*{\fill} & 303 & \mbox{\tiny (425)} & \multicolumn{2}{@{\,}l@{\,}}{5.7e4\mbox{\tiny (7e4)}} & \multicolumn{2}{@{\,}l@{\,}}{$\infty$} & \multicolumn{2}{@{\,}l@{\,}}{\textbf{$\infty$}} & \multicolumn{2}{@{\,}l@{\,}}{\textbf{$\infty$}} & \multicolumn{2}{@{\,}l@{\,}}{\textbf{$\infty$}} & \multicolumn{2}{@{\,}l@{\,}|}{$\infty$\,\textit{5e6}} & 0 & /15\\
    \textbf{f10} & \multicolumn{2}{@{}c@{}}{349 \quad} & \multicolumn{2}{@{}c@{}}{500 \quad} & \multicolumn{2}{@{}c@{}}{574 \quad} & \multicolumn{2}{@{}c@{}}{607 \quad} & \multicolumn{2}{@{}c@{}}{626 \quad} & \multicolumn{2}{@{}c@{}}{829 \quad} & \multicolumn{2}{@{}c@{}|}{880} & 15 & /15\\\hline
\algAtables\hspace*{\fill} & \textbf{2} & \textbf{.1}\mbox{\tiny (2)} & \multicolumn{2}{@{\,}l@{\,}}{\textbf{$\infty$}} & \multicolumn{2}{@{\,}l@{\,}}{\textbf{$\infty$}} & \multicolumn{2}{@{\,}l@{\,}}{\textbf{$\infty$}} & \multicolumn{2}{@{\,}l@{\,}}{\textbf{$\infty$}} & \multicolumn{2}{@{\,}l@{\,}}{\textbf{$\infty$}} & \multicolumn{2}{@{\,}l@{\,}|}{$\infty$\,\textit{2}} & 0 & /15\\
\algBtables\hspace*{\fill} & \multicolumn{2}{@{\,}l@{\,}}{$\infty$} & \multicolumn{2}{@{\,}l@{\,}}{\textbf{$\infty$}} & \multicolumn{2}{@{\,}l@{\,}}{\textbf{$\infty$}} & \multicolumn{2}{@{\,}l@{\,}}{\textbf{$\infty$}} & \multicolumn{2}{@{\,}l@{\,}}{\textbf{$\infty$}} & \multicolumn{2}{@{\,}l@{\,}}{\textbf{$\infty$}} & \multicolumn{2}{@{\,}l@{\,}|}{$\infty$\,\textit{2}} & 0 & /15\\
\algCtables\hspace*{\fill} & 3 & .3\mbox{\tiny (2)} & \multicolumn{2}{@{\,}l@{\,}}{\textbf{$\infty$}} & \multicolumn{2}{@{\,}l@{\,}}{\textbf{$\infty$}} & \multicolumn{2}{@{\,}l@{\,}}{\textbf{$\infty$}} & \multicolumn{2}{@{\,}l@{\,}}{\textbf{$\infty$}} & \multicolumn{2}{@{\,}l@{\,}}{\textbf{$\infty$}} & \multicolumn{2}{@{\,}l@{\,}|}{$\infty$\,\textit{2}} & 0 & /15\\
\algDtables\hspace*{\fill} & \multicolumn{2}{@{\,}l@{\,}}{$\infty$} & \multicolumn{2}{@{\,}l@{\,}}{\textbf{$\infty$}} & \multicolumn{2}{@{\,}l@{\,}}{\textbf{$\infty$}} & \multicolumn{2}{@{\,}l@{\,}}{\textbf{$\infty$}} & \multicolumn{2}{@{\,}l@{\,}}{\textbf{$\infty$}} & \multicolumn{2}{@{\,}l@{\,}}{\textbf{$\infty$}} & \multicolumn{2}{@{\,}l@{\,}|}{$\infty$\,\textit{5e6}} & 0 & /15\\
%
   \textbf{f15} & \multicolumn{2}{@{}c@{}}{511 \quad} & \multicolumn{2}{@{}c@{}}{9310 \quad} & \multicolumn{2}{@{}c@{}}{19369 \quad} & \multicolumn{2}{@{}c@{}}{19743 \quad} & \multicolumn{2}{@{}c@{}}{20073 \quad} & \multicolumn{2}{@{}c@{}}{20769 \quad} & \multicolumn{2}{@{}c@{}|}{21359} & 14 & /15\\\hline
\algAtables\hspace*{\fill} & 0 & .21\mbox{\tiny (0.1)}$_{\uparrow0}$ & \textbf{0} & \textbf{.19}\mbox{\tiny (0.1)} & \textbf{0} & \textbf{.10}\mbox{\tiny (0.1)} & \textbf{0} & \textbf{.10}\mbox{\tiny (0.1)} & \textbf{0} & \textbf{.10}\mbox{\tiny (0.1)} & 0 & .09\mbox{\tiny (0.1)} & 0 & .18\mbox{\tiny (2e-3)} & 0 & /15\\
\algBtables\hspace*{\fill} & 0 & .23\mbox{\tiny (0.1)}$_{\uparrow0}$ & 0 & .20\mbox{\tiny (0.1)} & 0 & .11\mbox{\tiny (0.1)} & 0 & .11\mbox{\tiny (0.1)} & 0 & .11\mbox{\tiny (0.1)} & \textbf{0} & \textbf{.07}\mbox{\tiny (0.0)} & \multicolumn{2}{@{\,}l@{\,}|}{$\infty$\,\textit{2}} & 0 & /15\\
\algCtables\hspace*{\fill} & \textbf{0} & \textbf{.18}\mbox{\tiny (0.1)}$_{\uparrow0}$ & 0 & .26\mbox{\tiny (0.2)} & 0 & .12\mbox{\tiny (0.1)} & 0 & .12\mbox{\tiny (0.1)} & 0 & .12\mbox{\tiny (0.1)} & 0 & .11\mbox{\tiny (0.0)} & \textbf{0} & \textbf{.08}\mbox{\tiny (1e-3)} & 0 & /15\\
\algDtables\hspace*{\fill} & 6853 & \mbox{\tiny (7246)} & \multicolumn{2}{@{\,}l@{\,}}{$\infty$} & \multicolumn{2}{@{\,}l@{\,}}{$\infty$} & \multicolumn{2}{@{\,}l@{\,}}{$\infty$} & \multicolumn{2}{@{\,}l@{\,}}{$\infty$} & \multicolumn{2}{@{\,}l@{\,}}{$\infty$} & \multicolumn{2}{@{\,}l@{\,}|}{$\infty$\,\textit{5e6}} & 0 & /15\\
   \textbf{f20} & \multicolumn{2}{@{}c@{}}{16 \quad} & \multicolumn{2}{@{}c@{}}{851 \quad} & \multicolumn{2}{@{}c@{}}{38111 \quad} & \multicolumn{2}{@{}c@{}}{51362 \quad} & \multicolumn{2}{@{}c@{}}{54470 \quad} & \multicolumn{2}{@{}c@{}}{54861 \quad} & \multicolumn{2}{@{}c@{}|}{55313} & 14 & /15\\\hline
\algAtables\hspace*{\fill} & \textbf{0} & \textbf{.98}\mbox{\tiny (0.3)} & \textbf{0} & \textbf{.29}\mbox{\tiny (0.2)}$_{\uparrow0}$ & \multicolumn{2}{@{\,}l@{\,}}{\textbf{9.8e-3}\mbox{\tiny (3e-3)}} & \multicolumn{2}{@{\,}l@{\,}}{\textbf{7.5e-3}\mbox{\tiny (2e-3)}} & \multicolumn{2}{@{\,}l@{\,}}{\textbf{7.3e-3}\mbox{\tiny (2e-3)}} & \multicolumn{2}{@{\,}l@{\,}}{\textbf{7.6e-3}\mbox{\tiny (2e-3)}} & 0 & .01\mbox{\tiny (7e-3)} & 4 & /15\\
\algBtables\hspace*{\fill} & 1 & .1\mbox{\tiny (0.3)} & 0 & .46\mbox{\tiny (0.2)} & 0 & .01\mbox{\tiny (7e-3)} & 0 & .01\mbox{\tiny (5e-3)} & 0 & .01\mbox{\tiny (5e-3)} & \multicolumn{2}{@{\,}l@{\,}}{9.4e-3\mbox{\tiny (2e-3)}} & \textbf{0} & \textbf{.01}\mbox{\tiny (1e-3)} & 3 & /15\\
\algCtables\hspace*{\fill} & 1 & .1\mbox{\tiny (0.2)} & 0 & .34\mbox{\tiny (0.1)}$_{\uparrow0}$ & 0 & .01\mbox{\tiny (6e-3)}$_{\uparrow1}$ & 0 & .01\mbox{\tiny (5e-3)}$_{\uparrow1}$ & 0 & .01\mbox{\tiny (4e-3)}$_{\uparrow1}$ & 0 & .01\mbox{\tiny (5e-3)}$_{\uparrow1}$ & 0 & .02\mbox{\tiny (0.0)} & 5 & /15\\
\algDtables\hspace*{\fill} & 29 & \mbox{\tiny (40)} & 9230 & \mbox{\tiny (1e4)} & \multicolumn{2}{@{\,}l@{\,}}{$\infty$} & \multicolumn{2}{@{\,}l@{\,}}{$\infty$} & \multicolumn{2}{@{\,}l@{\,}}{$\infty$} & \multicolumn{2}{@{\,}l@{\,}}{$\infty$} & \multicolumn{2}{@{\,}l@{\,}|}{$\infty$\,\textit{5e6}} & 0 & /15\\
\hline
\end{tabular}}
\end{table}

\begin{figure}[!ht]
    \centering
    \includegraphics[width=0.75\linewidth]{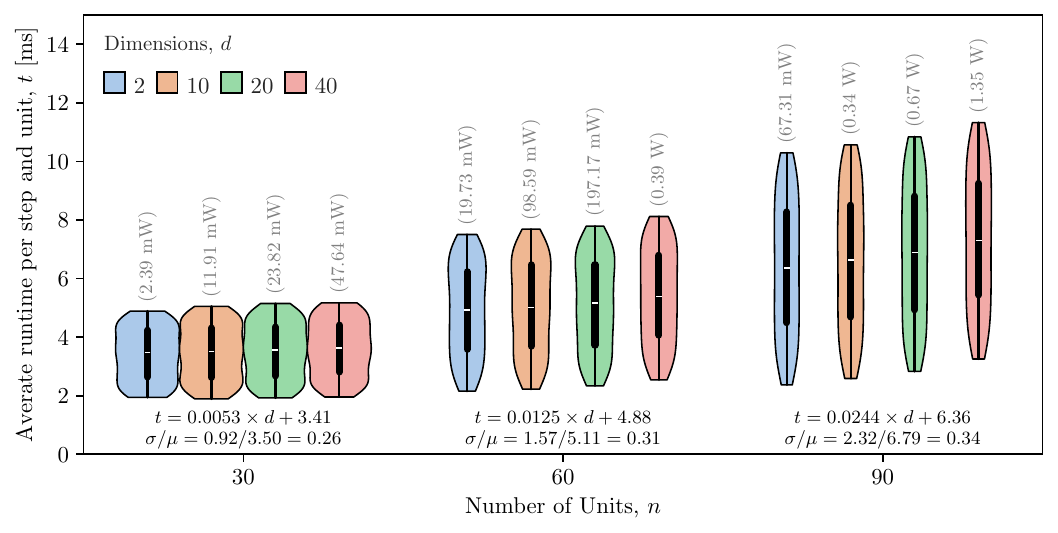}
    \vspace{-8pt}
    \caption{\label{Fig:ExTime:computing-time}%
    Average runtime per step and unit (\(t\)~[ms]) as a function of the number of \nhu{}s \(n\) and dimensionality \(d\). Problem instances were evaluated across 2, 10, 20, and 40 dimensions, using configurations with 30, 60, and 90 \nhu{}s. Violin plots display the distribution of per-unit runtimes across trials. Mean runtime (white dot), estimated power consumption (in parentheses), linear regression equations, and \(\sigma/\mu\) ratios are reported for each setting.
    }
\end{figure}

Last but not least, to evaluate the computational efficiency of the proposed framework, we measured the average runtime per step and \nhu{} across varying numbers of units and problem dimensionalities.
\figurename~\ref{Fig:ExTime:computing-time} displays these average runtimes varying the number of \nhu{}s \(n\) and problem dimensionality \(d\). 
We observe that the computational cost increases linearly with both parameters, confirming the expected scaling of asynchronous and event-driven architectures. 
Even for the most significant tested values, the per-unit runtime remains below 12~ms, with the slope increasing moderately as \(d\) grows due to the added cost of spike propagation and local updates. 
Across all experiments, we notice that the relative variability (\(\sigma/\mu\)) remains consistently low, indicating the regularity of the implementations. 
These results also validate the computational efficiency of the proposed framework and align with the behaviour reported in NC literature~\cite{Davies2018, Schuman2022}, where event-driven models are known to maintain low latency and power consumption, especially under sparse activity. 
As all experiments were conducted as CPU simulations with the most demanding connectivity settings, we anticipate further gains when deployed on dedicated neuromorphic hardware, where locality and event-driven computation can be fully leveraged.

Additionally, \figurename~\ref{Fig:ExTime:computing-time} also presents the power estimates for all evaluated settings in parentheses above each violin.
We assessed these values as upper bounds on the power consumption of our framework when deployed on Intel's Loihi chip~\cite{Davies2018}. 
This estimation accounts for the number of synaptic events (\(N_{\text{syn}} = n(n-1)md\)), neuron updates, and spike transmissions per step. 
Using Loihi's reported costs of \(23.6\)~pJ per synaptic event, \(81\)~pJ per neuron update, and \(8.7\)~pJ per spike, the total energy per step is given by \(E_{\text{step}} = 23.6\, N_{\text{syn}} + 89.7n\)~pJ.
With a simulation time step of \(\Delta t_\text{sim} = 0.5\)~ms, the average power is computed as \(P_{\text{avg}} = E_{\text{step}}/\Delta t_\text{sim}\). 
Therefore, for the worst-case configuration tested (\(n = 90\), \(d = 40\), \(m = 89\)), this yields \(E_{\text{step}} \approx 0.67\)~mJ and \(P_{\text{avg}} \approx 1.35\)~W. 
In practice, the actual power consumption will be lower due to sparse spike activity (\figurename~\ref{Fig:ExPre:Spikes}) and the event-driven design of Loihi, which ensures that inactive units consume negligible energy. 

\section{Conclusion}

\noindent
This work addressed the challenge of scalable, low-power optimisation by introducing \nha{short}, a framework that combines population-based metaheuristics and neuromorphic computing. 
This approach performs an evolutionary search using only asynchronous, spike-triggered dynamics, eliminating the need for central coordination or surrogate modelling.
We established its foundational components, comprising a population of \nhu{}s that encode candidate solutions and evolve their state through event-driven dynamics.
Each unit receives input from neighbour processes via spike events, with state transitions triggered by local conditions only using a dynamic \(h_d\) and spike-triggered \(h_s\) heuristics.
We specified how dynamic models, either discrete or continuous, can be defined as \(h_d\), as well as evolutionary mechanisms such as perturbation and selection are embedded as \(h_s\).
Moreover, we detailed all the other relevant components to bear in mind when implementing this approach, such as the spiking condition that triggers transitions, the bidirectional transformation for encoding and decoding candidate solutions, and the thresholding mechanism, among others.

We implemented and tested several \nha{short} variants, considering combinations mainly of two-dimensional dynamic systems such as linear and Izhikevich models as \(h_d\), and the fixed and DE/\emph{current-to-rand} heuristics as \(h_s\).
For testing, we employed Intel's Lava platform, targeting the Loihi 2 chip, for materialising our approach and the noiseless BBOB suite up to 40 dimensions.
Results showed that the \nha{} approaches reliably converged on separable, unimodal, and moderately conditioned problems, while matching reference methods in low and moderate dimensions.
Plus, we found that performance decreased and variance increased on high-dimensional, multimodal, and ill-conditioned functions under tight evaluation budgets.
Such an outcome highlights the limitations of fixed spike-triggered rules and uniform model assignment, as homogeneous configurations are often outperformed by the heterogeneous variant on more complex problems.
Recall that homogeneous variants used the same dynamic model and heuristic for all units, while the heterogeneous variant blended linear and Izhikevich cores within the \nhu{}s.
Furthermore, we observed a linear scaling of runtime and resource consumption with increasing population size and dimensionality, with per-unit runtimes under 12~ms and milliwatt-level power estimates. 
We confirmed the suitability of the architecture for practical NC hardware deployments.

This framework stands for the first complete and reproducible integration of neuromorphic heuristic-based optimisers. It constitutes a principled path toward low-energy, real-time, and fully decentralised optimisation.
For future work, further progress requires adaptive rule selection, increased model heterogeneity, and large-scale deployment on physical NC hardware. Additionally, we plan to integrate advanced dynamic models, extend support to multi-objective and large-scale optimisation, and enable real-time evaluation in embedded environments.

\section*{Acknowledgments}

\noindent
The authors utilised OpenAI's ChatGPT‑4o and Grammarly for language polishing under human supervision. No AI was used for scientific ideas, data analysis, or interpretation. The authors remain fully responsible for the manuscript.

\bibliography{IEEEabrv, references}
\bibliographystyle{IEEEtran}

\end{document}